\documentclass[10pt,twocolumn,letterpaper]{article}

\usepackage[pagenumbers]{wacv} % To force page numbers, e.g. for an arXiv version

\usepackage{algorithm}
\usepackage{listings}
\usepackage{graphicx}
\usepackage{amsmath}
\usepackage{amssymb}
\usepackage{booktabs}
\usepackage{tabularx}
\usepackage{makecell}
\usepackage{enumitem}
\usepackage{multirow}
\usepackage{indentfirst}

\usepackage[pagebackref,breaklinks,colorlinks]{hyperref}

\usepackage[capitalize]{cleveref}
\crefname{section}{Sec.}{Secs.}
\Crefname{section}{Section}{Sections}
\Crefname{table}{Table}{Tables}
\crefname{table}{Tab.}{Tabs.}

\def\wacvPaperID{*****} % *** Enter the WACV Paper ID here
\def\confName{WACV}
\def\confYear{2024}

\begin{document}

\title{SDXL-Lightning: Progressive Adversarial Diffusion Distillation}

\author{Shanchuan Lin \quad Anran Wang \quad Xiao Yang \\
ByteDance Inc.\\
{\tt\small \{peterlin, anran.wang, yangxiao.0\}@bytedance.com}
}

\newcommand\banner{%
    \vspace{30pt}
    \hspace*{-0.26\textwidth}
    \centering
    \footnotesize
    \setlength\tabcolsep{0pt}
    \renewcommand{\arraystretch}{0}
    \begin{tabular}{ c c c c c c }
        \includegraphics[width=0.25\textwidth]{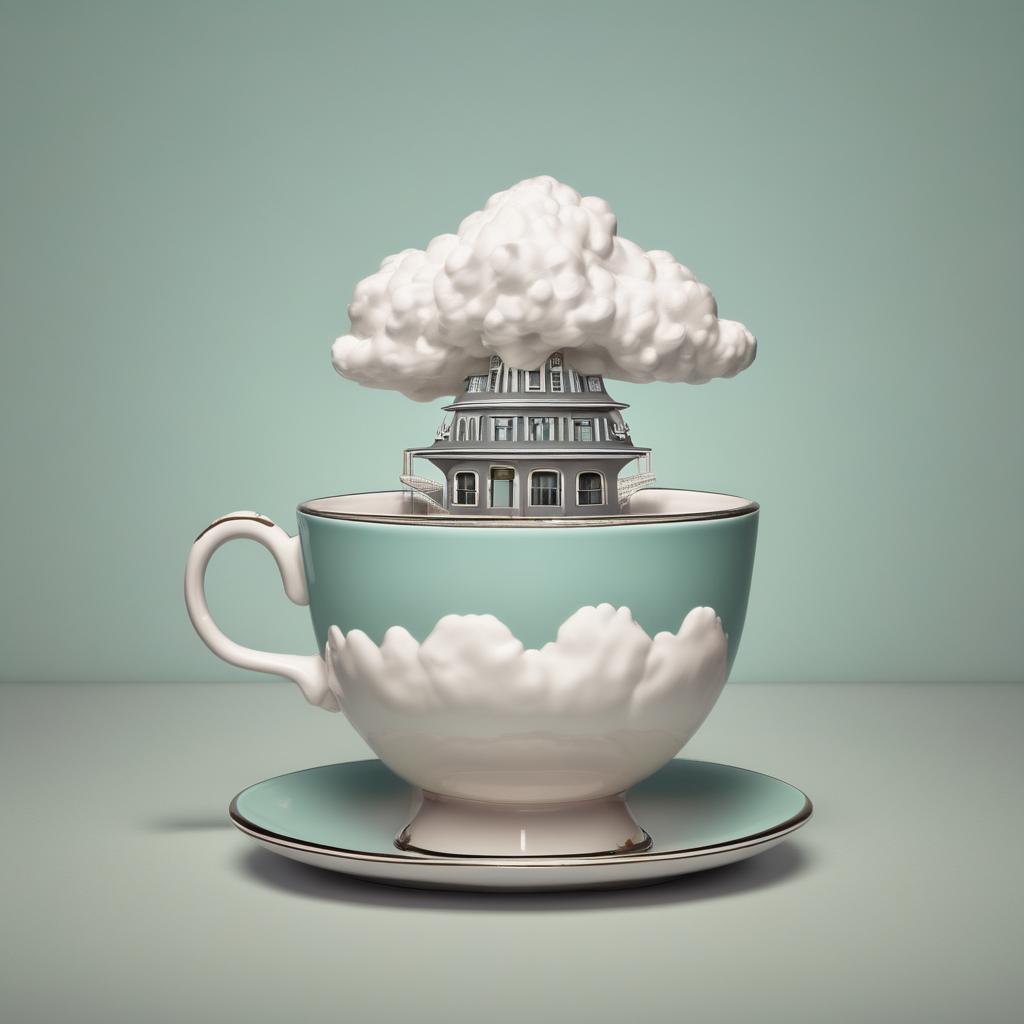} &
        \includegraphics[width=0.25\textwidth]{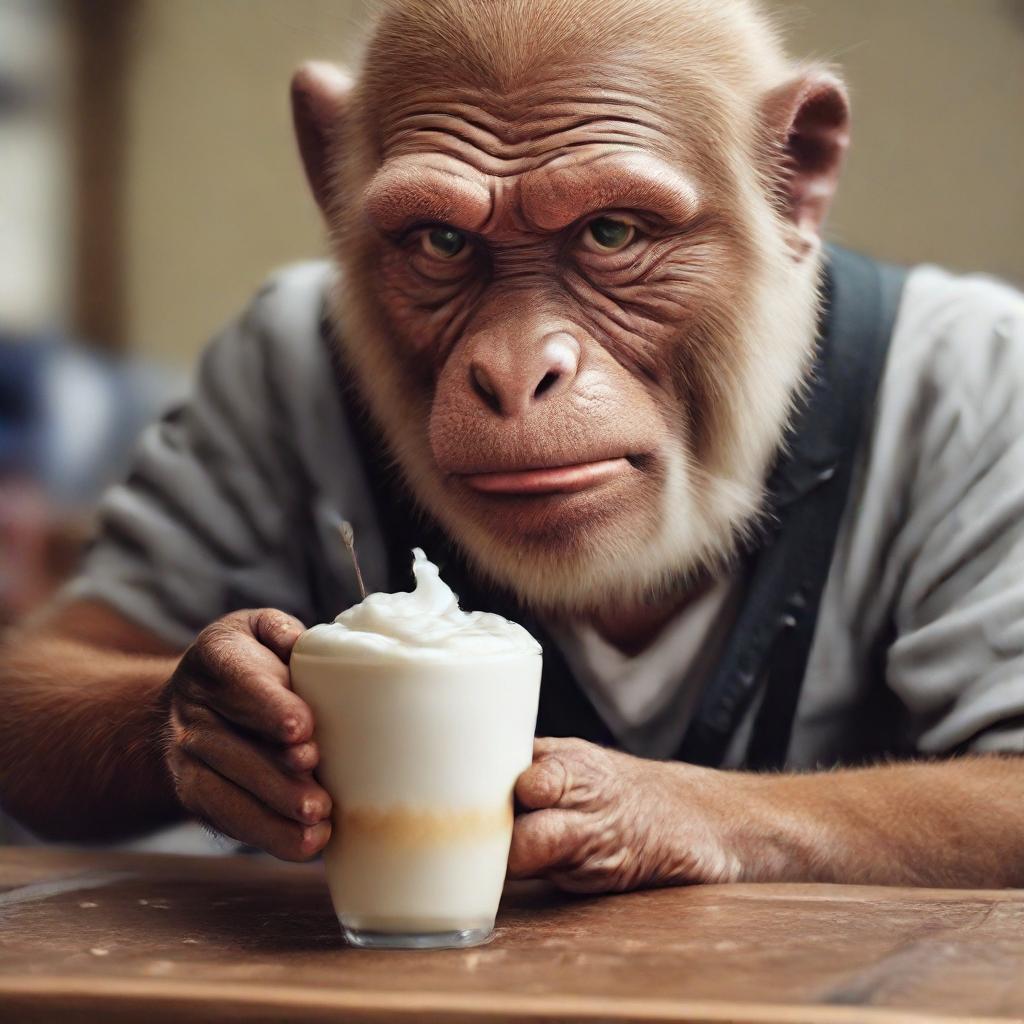} &
        \includegraphics[width=0.25\textwidth]{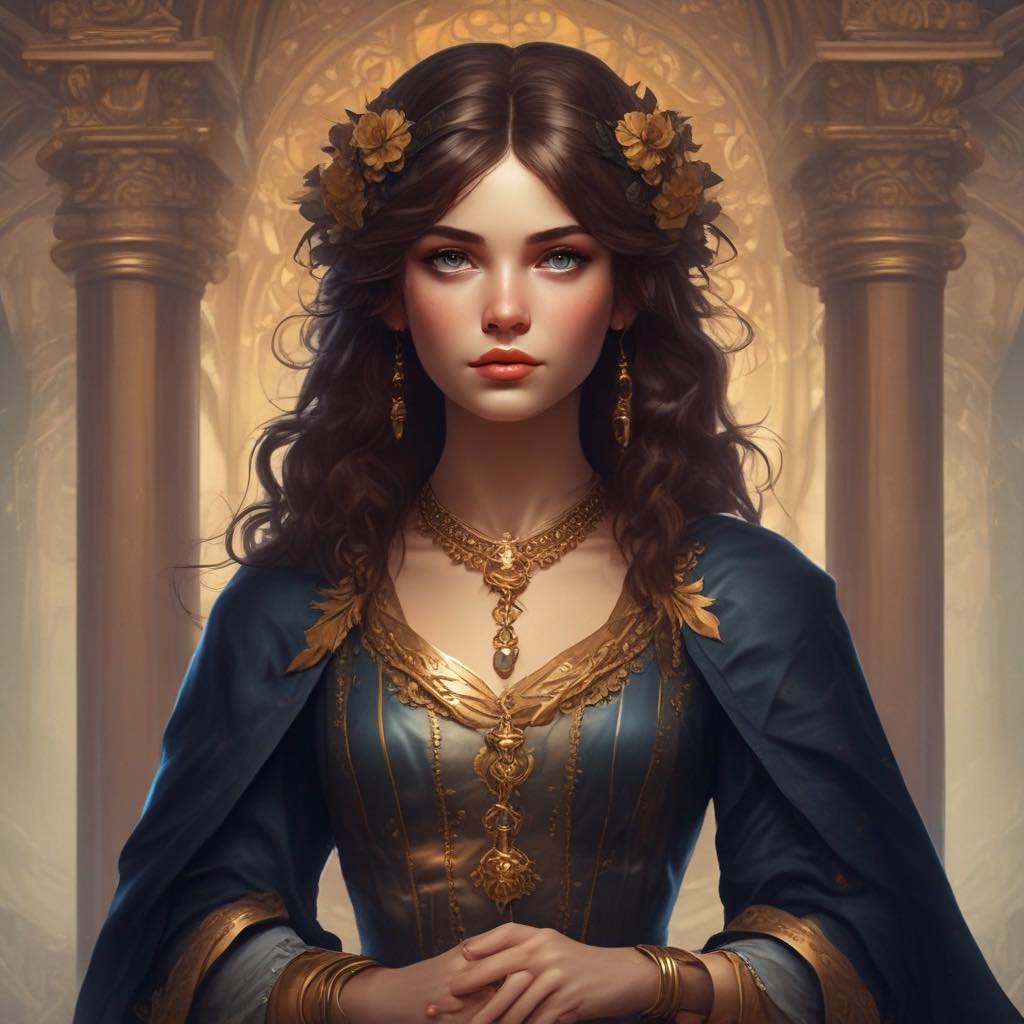} & 
        \includegraphics[width=0.25\textwidth]{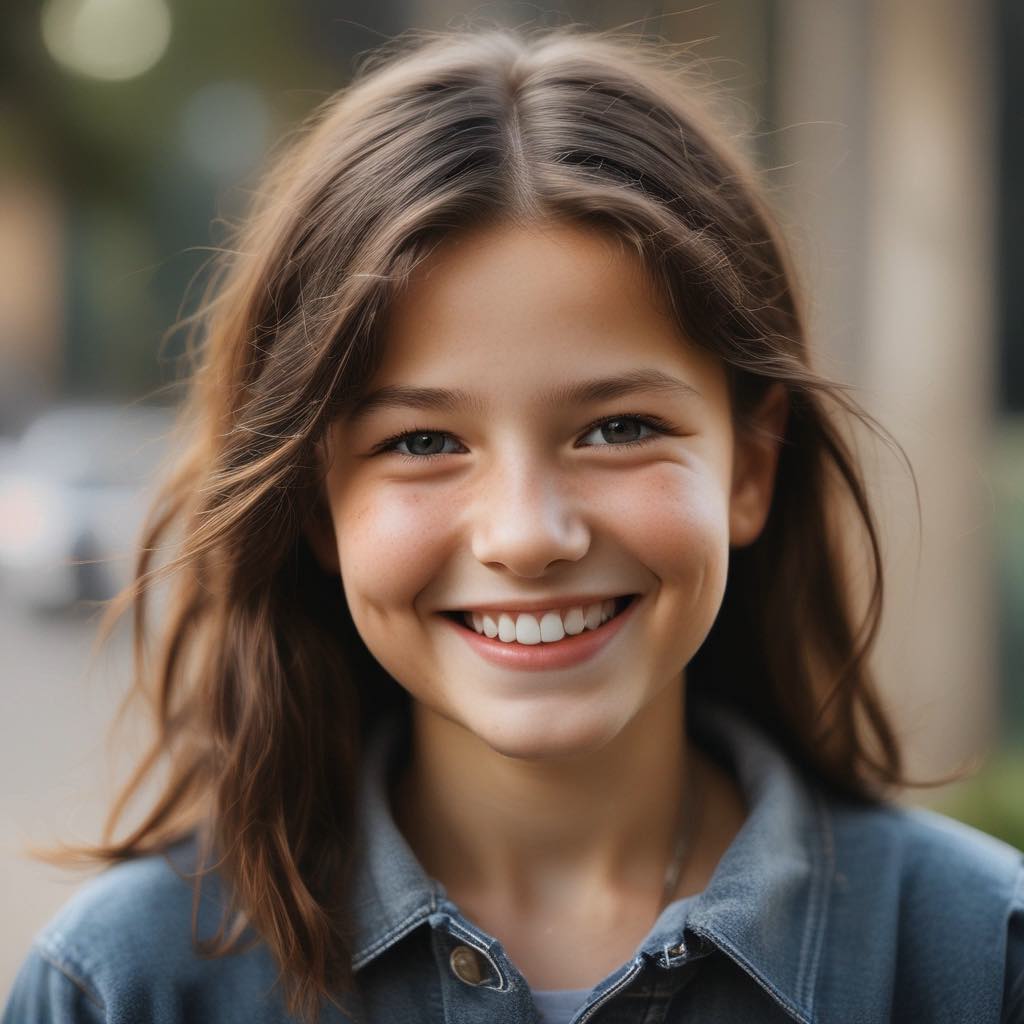} &
        \includegraphics[width=0.25\textwidth]{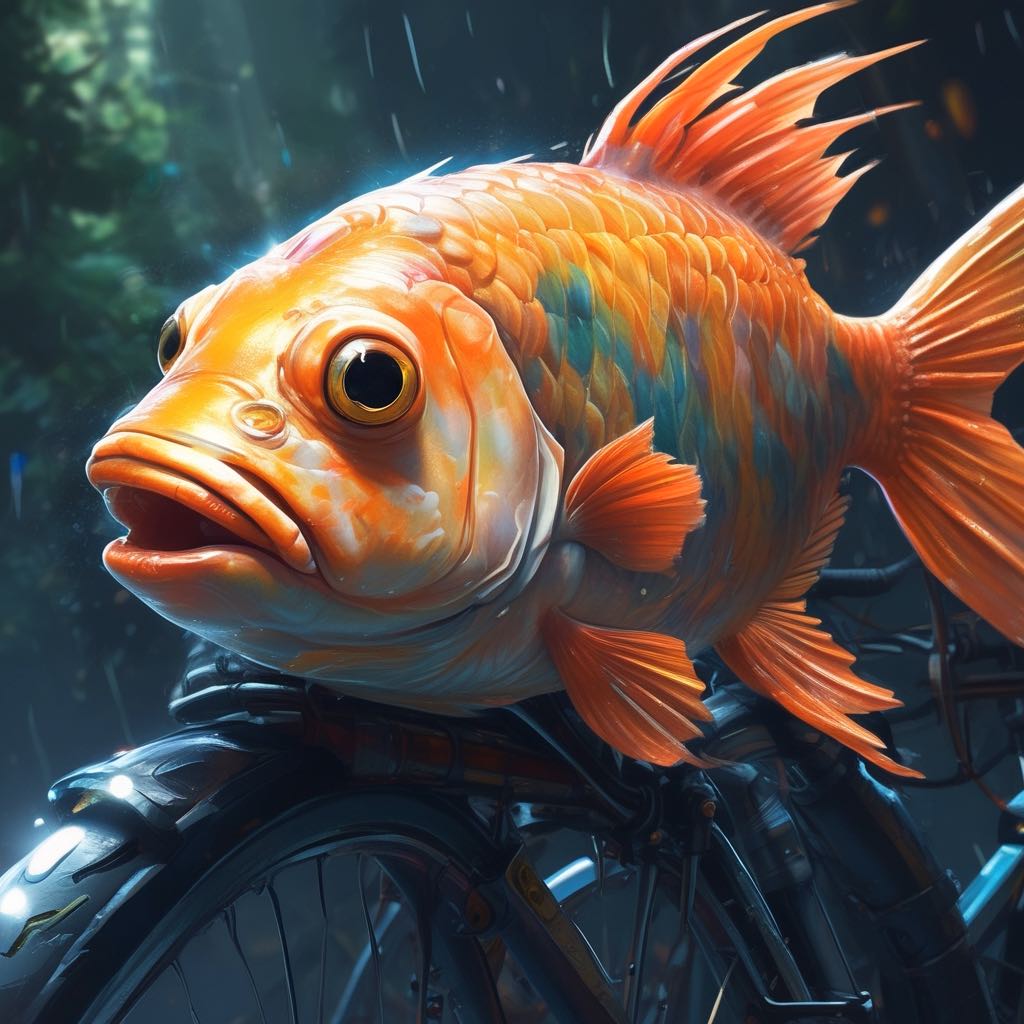} &
        \includegraphics[width=0.25\textwidth]{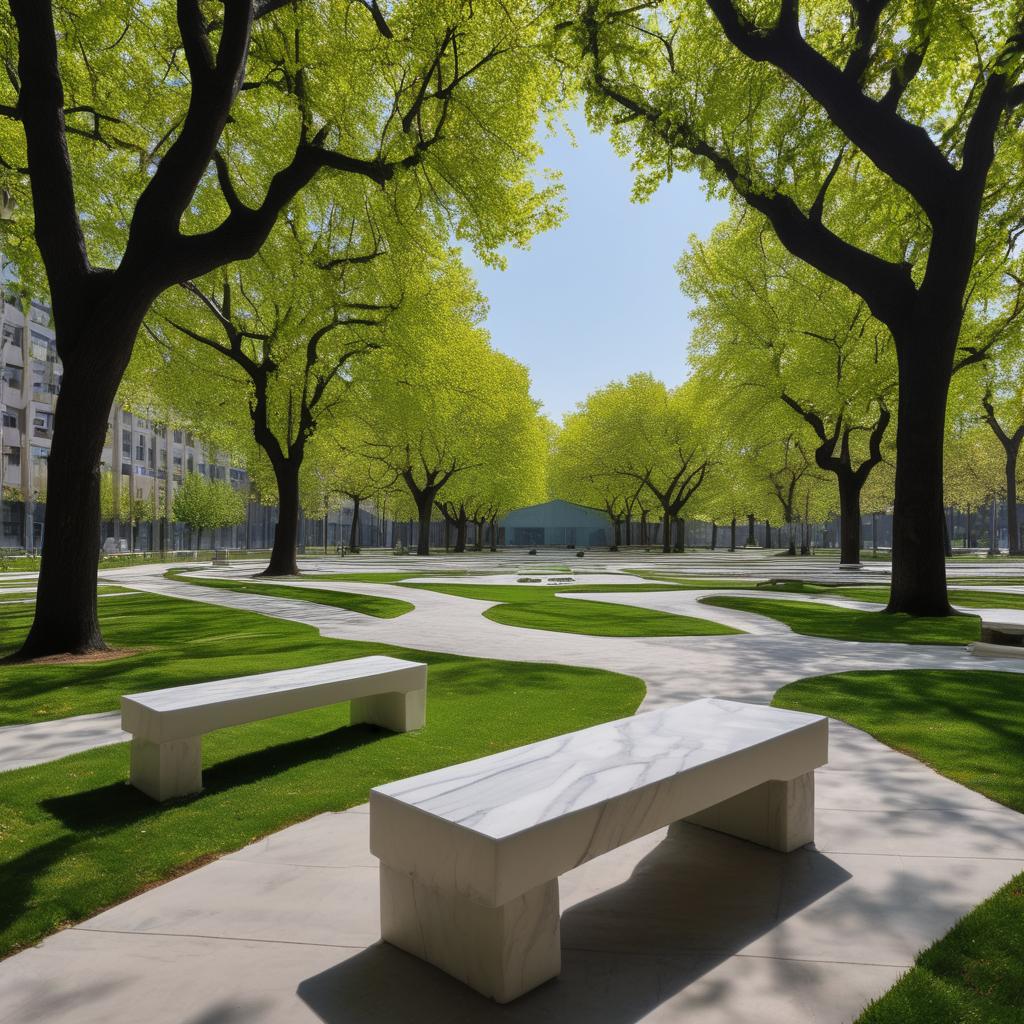} \\
        \includegraphics[width=0.25\textwidth]{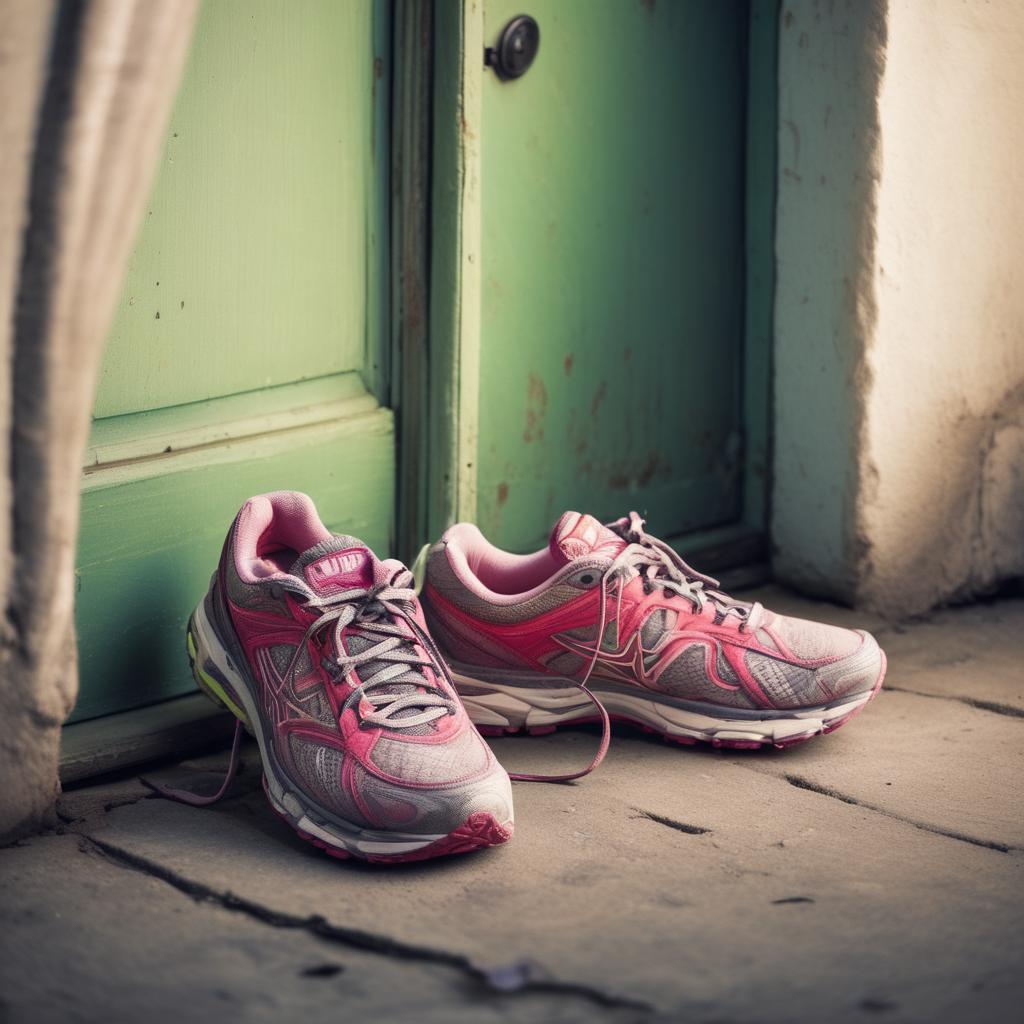} &
        \includegraphics[width=0.25\textwidth]{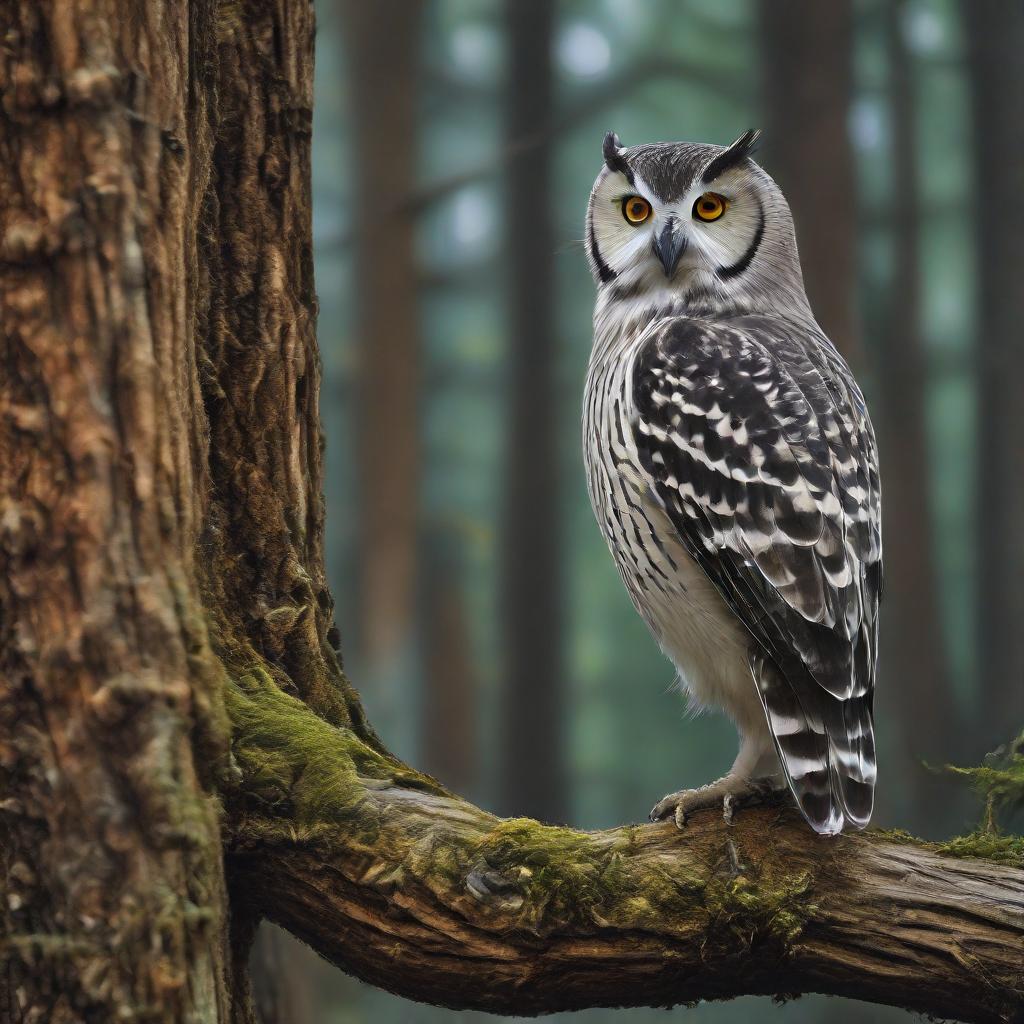} & 
        \includegraphics[width=0.25\textwidth]{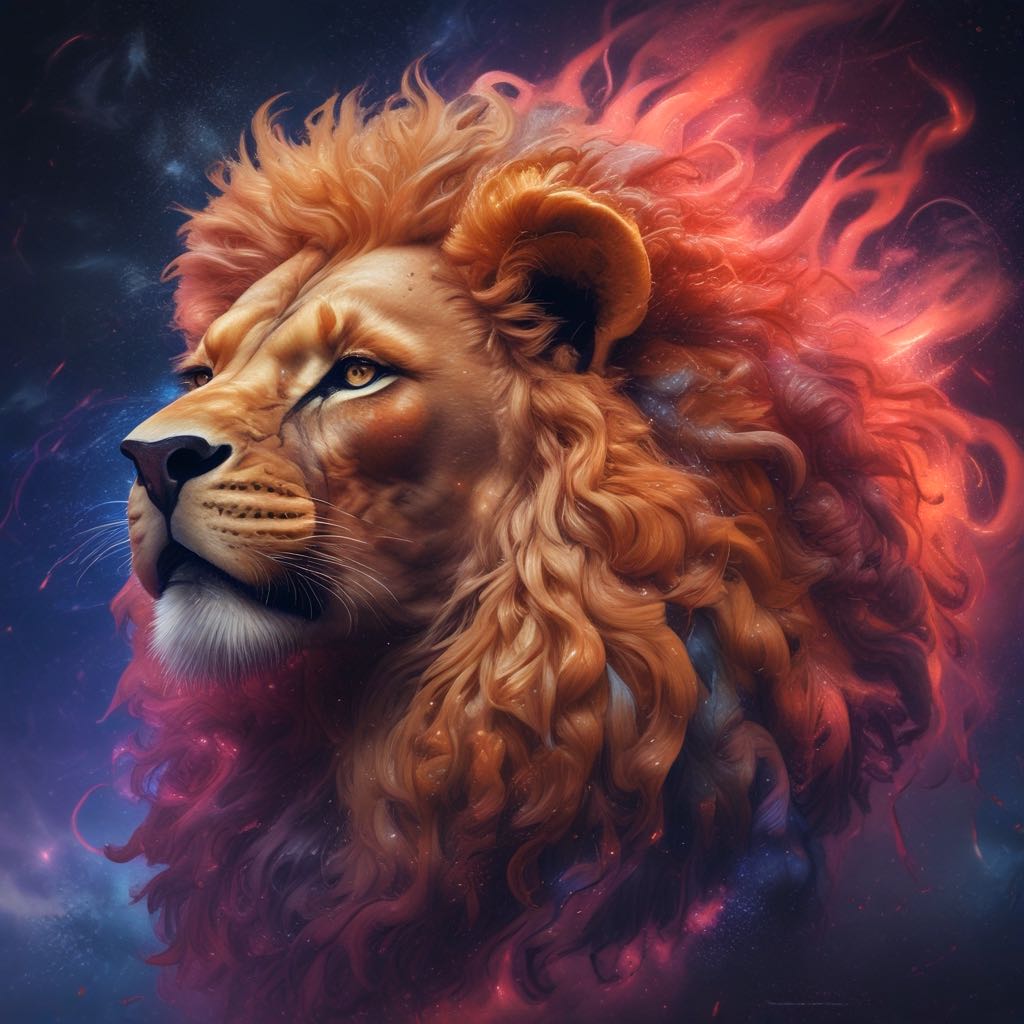} &
        \includegraphics[width=0.25\textwidth]{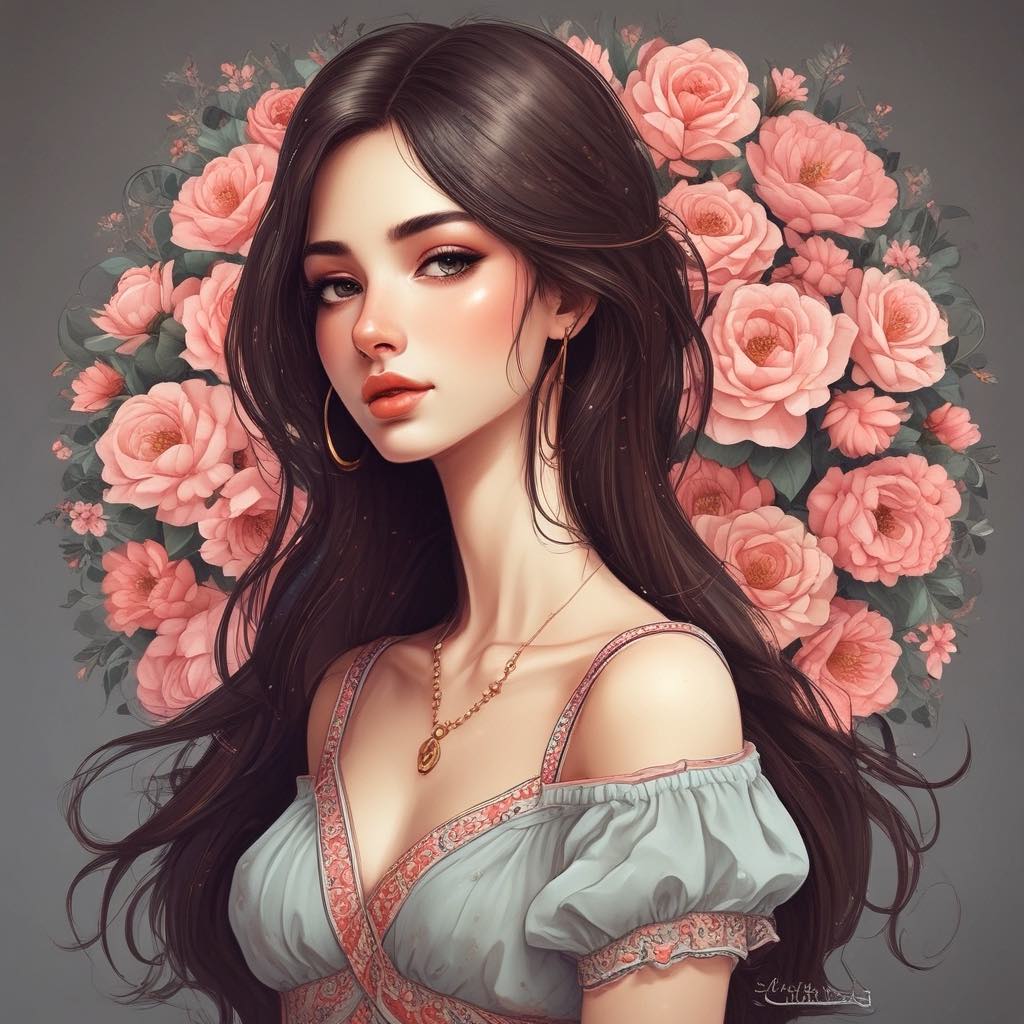} & 
        \includegraphics[width=0.25\textwidth]{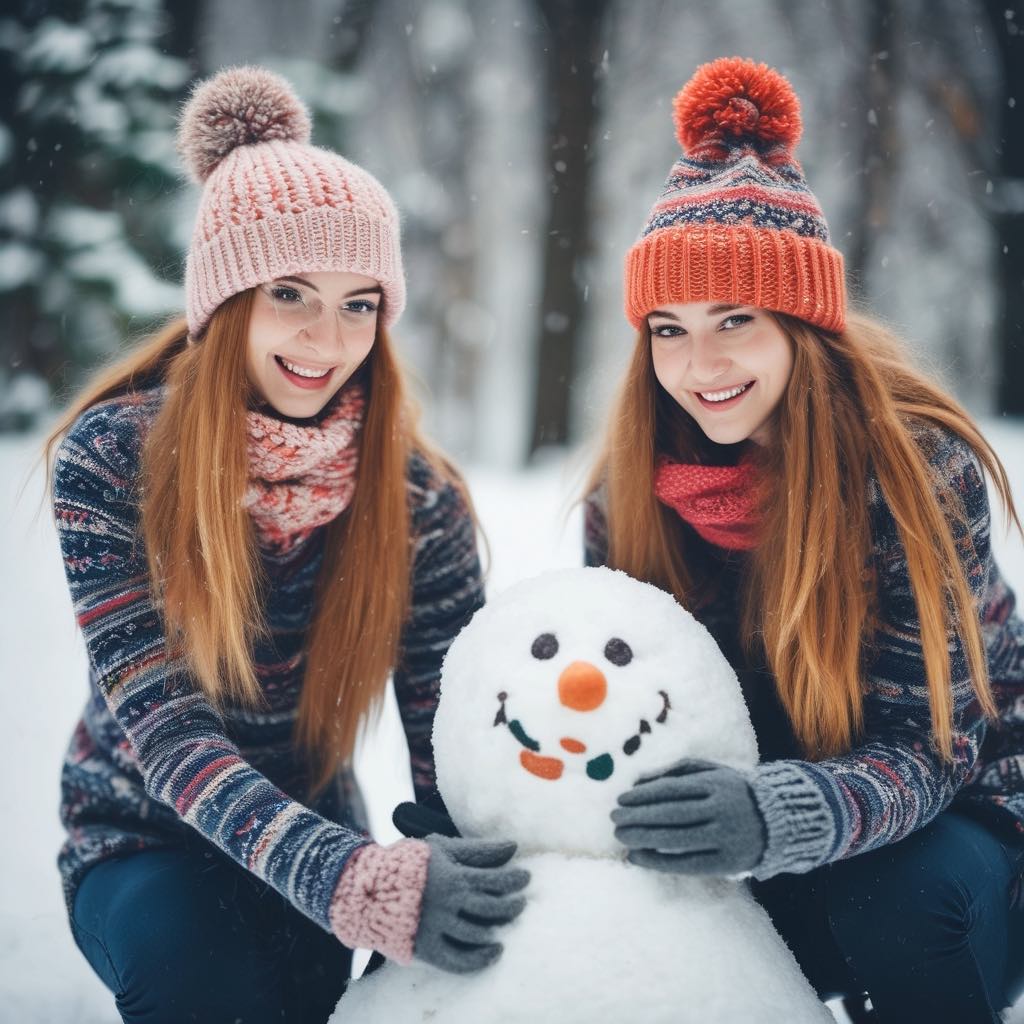} &
        \includegraphics[width=0.25\textwidth]{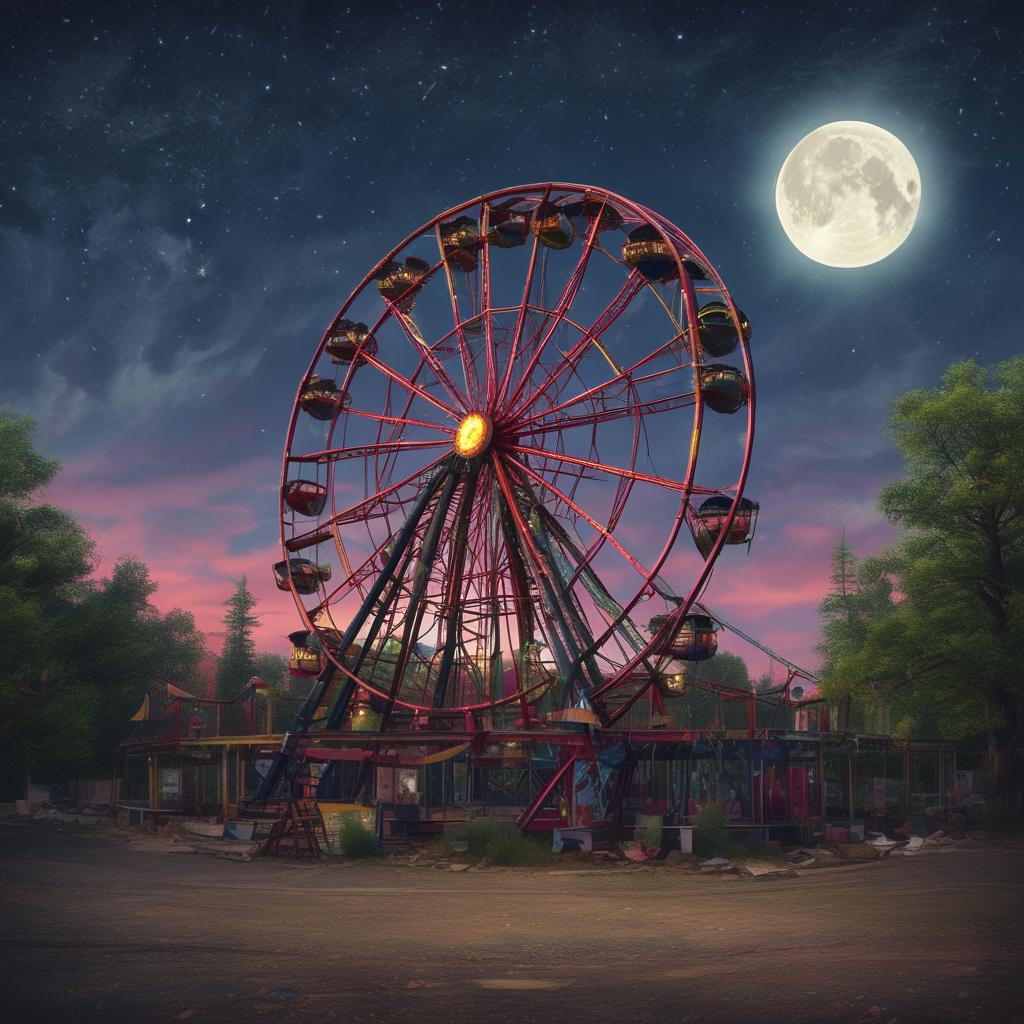} \\
         & \multicolumn{1}{c}{\rule{0pt}{10pt}1 Step} & \multicolumn{1}{|c}{2 Steps} & \multicolumn{1}{|c}{4 Steps} & \multicolumn{1}{|c}{8 Steps}
    \end{tabular}
    % \captionof{figure}{Introducing SDXL-Lightning, our new state-of-the-art one-step/few-step 1024px text-to-image generative model.}
    \vspace{25pt}
}

\date{}
\maketitle

%%%%%%%%% ABSTRACT

\begin{abstract}
We propose a diffusion distillation method that achieves new state-of-the-art in one-step/few-step 1024px text-to-image generation based on SDXL. Our method combines progressive and adversarial distillation to achieve a balance between quality and mode coverage. In this paper, we discuss the theoretical analysis, discriminator design, model formulation, and training techniques. We open-source our distilled SDXL-Lightning models both as LoRA and full UNet weights.
\end{abstract}

\let\thefootnote\relax\footnotetext{Model: \href{https://huggingface.co/ByteDance/SDXL-Lightning}{https://huggingface.co/ByteDance/SDXL-Lightning}}

\section{Introduction}

Diffusion models \cite{ho2020denoising,sohldickstein2015deep,song2021scorebased} are a rising class of generative models that has achieved state-of-the-art results in a wide range of applications, such as text-to-image \cite{podell2023sdxl,rombach2022highresolution,saharia2022photorealistic,chen2023pixartalpha,ramesh2022hierarchical,zheng2022movq,pernias2023wuerstchen}, text-to-video \cite{guo2023animatediff,blattmann2023align,zhou2023magicvideo,ho2022imagen,blattmann2023stable,singer2022makeavideo}, and image-to-video \cite{blattmann2023stable}, \etc. However, the iterative generation process of diffusion models is slow and computationally expansive. How to generate high-quality samples faster is an actively researched area and is the main focus of our work.

Conceptually, the generation involves a probability flow that gradually transports samples between the data and the noise probability distribution. The diffusion model learns to predict the gradient at any location of this flow. The generation is simply transporting samples from the noise distribution to the data distribution by following the predicted gradient through the flow. Because the flow is complex and curved, the generation must take a small step at a time. Formally, the flow can be expressed as an ordinary differential equation (ODE) \cite{song2021scorebased}. In practice, generating a high-quality data sample requires more than 50 inference steps.

Different approaches to reduce the number of inference steps have been researched. Prior works have proposed better ODE solvers to account for the curving nature of the flow \cite{song2022denoising,lu2022dpmsolver,lu2023dpmsolver,karras2022elucidating,liu2022pseudo,zhao2023unipc}. Others have proposed formulations to make the flow straighter \cite{liu2022flow,lipman2023flow}. Nonetheless, these approaches generally still require more than 20 inference steps.

Model distillation \cite{salimans2022progressive,song2023consistency,song2023improved,kim2023consistency,sauer2023adversarial,luo2023latent,luo2023lcmlora,yin2023onestep,liu2023instaflow,xu2023ufogen}, on the other hand, can achieve high-quality samples under 10 inference steps. Instead of predicting the gradient at the current flow location, it changes the model to directly predict the next flow location much farther ahead. Existing methods can achieve good results using 4 or 8 inference steps, but the quality is still not production-acceptable using 1 or 2 inference steps. Our method falls under the model distillation umbrella and achieves much superior quality compared to existing methods.

Our method combines the best of both worlds from progressive \cite{salimans2022progressive} and adversarial distillation \cite{sauer2023adversarial}. Progressive distillation ensures that the distilled model follows the same probability flow and has the same mode coverage as the original model. However, progressive distillation with mean squared error (MSE) loss produces blurry results under 8 inference steps and we provide theoretical analysis in our paper. To mitigate the issue, we use adversarial loss at every stage of the distillation to strike a balance between quality and mode coverage. Progressive distillation also brings an additional benefit, \ie, for multi-step sampling, our model predicts the next location on the ODE trajectory instead of jumping to the ODE trajectory endpoints every time by other distillation approaches \cite{song2023consistency,sauer2023adversarial,yin2023onestep}. This better preserves the original model behavior and facilitates better compatibility with LoRA modules \cite{hu2021lora} and control plugins \cite{zhang2023adding,guo2023animatediff,ye2023ipadapter}.

Furthermore, our paper proposes innovative discriminator design, loss objectives, and stable training techniques. Specifically, we use the pre-trained diffusion UNet encoder as the discriminator backbone and fully operate in latent space. We propose two adversarial loss objectives to trade off sample quality and mode coverage. We investigate the implication of diffusion schedules and output formulation. We discuss techniques to stabilize the adversarial training.

Our distillation method produces new state-of-the-art SDXL \cite{podell2023sdxl} models that support one-step/few-step generation at 1024px resolution. We open-source our distilled models as SDXL-Lightning.

\section{Background}

\subsection{Diffusion Model}

The forward diffusion process \cite{ho2020denoising} gradually transforms samples from the data distribution to the Gaussian noise distribution. Given a data sample $x_0$, noise $\epsilon \sim \mathcal{N}(0, \mathbf{I})$ and time $t \sim \mathcal{U}(1, T)$. The forward function is defined as the following, with $\bar\alpha_t$ as the manually defined schedule \cite{ho2020denoising}:
\begin{equation}
    x_t = \mathbf{forward}(x_0, \epsilon, t) \equiv \sqrt{\bar{\alpha}_t}x_0 + \sqrt{1 - \bar{\alpha}_t}\epsilon
\end{equation}

A neural network $f: \mathbb{R}^d \rightarrow \mathbb{R}^d$ is trained to predict the gradient field $u_t$ at any location $x_t$ of the flow. The network is conditioned on time $t$ and optionally other conditions $c$:
\begin{equation}
    \hat{u}_t = f(x_t, t, c)
\end{equation}

Many prior works formulate the network to perform noise prediction \cite{ho2020denoising}, \ie $\hat\epsilon = f(x_t, t, c)$. We can use the conversion function $\mathbf{x}$ to convert the prediction to $\hat{x}_0$ space:
\begin{equation}
    \hat{x}_0  = \mathbf{x}(x_t, \hat\epsilon, t) \equiv (x_t - \sqrt{1-\bar\alpha_t} \hat\epsilon) / \sqrt{\bar\alpha_t}
    \label{eq:conversion-eps-pred}
\end{equation}

Alternatively, the network can be formulated to perform data sample prediction \cite{ho2020denoising}, \ie $\hat{x}_0 = f(x_t, t, c)$. We can use the conversion function $\boldsymbol{\epsilon}$ to convert the prediction to $\hat{\epsilon}$ space:
\begin{equation}
    \hat\epsilon = \boldsymbol{\epsilon}(x_t, \hat{x}_0, t) \equiv (x_t - \sqrt{\bar{\alpha}_t} \hat{x}_0) / \sqrt{1 - \bar{\alpha}_t}
\end{equation}

Regardless of the formulation, the network in essence predicts the gradient $\hat{u}_t$. Given the gradient $u_t$ at any location $x_t$, we can move samples along the flow:
\begin{equation}
\begin{aligned}
    x_{t'} & = \mathbf{move}(x_t, u_t, t, t') \\
           & \equiv \mathbf{forward}(\mathbf{x}(x_t, u_t, t), \boldsymbol{\epsilon}(x_t, u_t, t), t')
\end{aligned}
\end{equation}

The generation process is simply moving sample $x_T \sim \mathcal{N}(0, \mathbf{I})$ from $t=T$ to $t=0$ a small step at a time.

\subsection{Latent Diffusion Model}

Instead of directly generating samples at the data space, latent diffusion models (LDMs) \cite{rombach2022highresolution} propose to first train a Variational Autoencoder (VAE) \cite{kingma2022autoencoding} that encodes the data to a more compact latent space. Diffusion models are trained to generate the latent codes, which are passed through the VAE decoder to generate the final data sample.

Latent diffusion models are widely adopted for high-resolution image and video generation due to their computational efficiency. SDXL \cite{podell2023sdxl} is the state-of-the-art text-to-image generation model that can generate 1024px resolution images from 128px latent space.

\subsection{Progressive Distillation}

Progressive distillation \cite{salimans2022progressive} trains the student to predict directions pointing to the next flow location as if the teacher has performed multiple steps.

Specifically, given data $x_0, c$ from the dataset, and noise $\epsilon \sim \mathcal{N}(0, \mathbf{I})$, we jump to arbitrary timestep $t$:
\begin{equation}
    x_t = \mathbf{forward}(x_0, \epsilon, t)
\end{equation}

We use the frozen teacher network $f_{\mathrm{teacher}}$ to perform $n$ inference steps to derive $x_{t-ns}$ ($t-ns$ clamped to $[0, T]$):
\small
\begin{align}
    u_t &= f_{\mathrm{teacher}}(x_t, t, c) \\
    x_{t-s} &= \mathbf{move}(x_t, u_t, t, t-s) \\
    u_{t-s} &= f_{\mathrm{teacher}}(x_{t-s}, t-s, c) \\
    x_{t-2s} &= \mathbf{move}(x_{t-s}, u_{t-s}, t-s, t-2s) \\
    & \dots \\
    u_{t-(n-1)s} &= f_{\mathrm{teacher}}(x_{t-(n-1)s}, t-(n-1)s, c) \\
    x_{t-ns} &= \mathbf{move}(x_{t\!-\!(\!n\!-\!1\!)\!s}, u_{t\!-\!(\!n\!-\!1\!)\!s}, t\!-\!(\!n\!-\!1\!)\!s, t\!-\!ns)
\end{align}
\normalsize

Then, we train the student network $f_{\mathrm{student}}$ to predict a direction field $\hat{u}_t$ that points from $x_t$ directly to $x_{t-ns}$:
\begin{align}
    \hat{u}_t &= f_{\mathrm{student}}(x_t, t, c) \\
    \hat{x}_{t-ns} &= \mathbf{move}(x_t, \hat{u}_t, t, t-ns)
\end{align}

The original work uses MSE loss \cite{salimans2022progressive}:
\begin{equation}
    \mathcal{L}_{\mathrm{mse}} = \|\hat{x}_{t-ns} - x_{t-ns}\|_2^2
    \label{eq:progressive-mse}
\end{equation}

Once the student model converges, it is used as the teacher model and the distillation process repeats. In theory, it can produce one-step generation models, but in practice, models produce blurry results. We analyze this issue in \Cref{sec:why-mse-fails}.

\subsection{Adversarial Distillation}

Adversarial training involves a minimax optimization between a discriminator network that aims to identify generated samples from real samples and a generator network that aims to fool the discriminator. It was originally proposed as Generative Adversarial Networks (GANs) \cite{goodfellow2014generative}, a standalone class of generative networks, but it suffers from issues such as mode collapse and instability. Recent studies have found that the adversarial objective can be incorporated in diffusion training \cite{xiao2022tackling} and distillation \cite{sauer2023adversarial,xu2023ufogen}.

SDXL-Turbo \cite{sauer2023adversarial} is the latest and the most popular open-source model using adversarial diffusion distillation. It follows prior works \cite{sauer2021projected,sauer2023stylegant} to use a pre-trained image encoder DINOv2 \cite{oquab2023dinov2} as the discriminator backbone to accelerate training. However, this brings several limitations. First, using an off-the-shelf vision encoder means it must operate in the pixel space instead of the latent space, which significantly increases computation, memory consumption, and training time, making high-resolution distillation impractical. This is likely the reason SDXL-Turbo only supports up to 512px resolution. Second, an off-the-shelf vision encoder only works at $t=0$. The distilled model has to be trained to jump to ODE trajectory endpoints $x_0$, but since the quality for one-step inference is not good enough, random noises are added again for multi-step inference. This way of multi-step inference significantly alters the model behavior, making it less compatible with existing LoRA modules \cite{hu2021lora} and control plugins \cite{zhang2023adding,guo2023animatediff,ye2023ipadapter}. Third, off-the-shelf encoders may be hard to find for other datasets (anime, line arts, \etc) and modalities (video, audio, \etc). This reduces the generalizability of the distillation method. Lastly, the adversarial objective alone does not force the model to follow the same probability flow, so mode coverage is not enforced.

Our method uses the diffusion model's U-Net encoder as the discriminator backbone. This allows us to efficiently distill in the latent space for high-resolution models, supports discrimination at all timesteps, and is generalizable to all datasets and modalities. Our method also allows control over the trade-off between quality and mode coverage, as later discussed in \Cref{sec:adversarial-objective,sec:mode-relax}.

\subsection{Other Distillation Methods}

We briefly discuss the advantages of our approach compared to other distillation methods.

Consistency Model (CM) \cite{song2023consistency,song2023improved} also requires jumping to the ODE trajectory endpoints at every inference step. This causes large model behavior changes for multi-step sampling which reduces compatibility with LoRA modules and plugins. This method has been applied to SDXL \cite{luo2023latent,luo2023lcmlora} but its generation quality is poor under 8 steps. Consistency Trajectory Model (CTM) \cite{kim2023consistency} adds adversarial loss and supports jumping to arbitrary flow locations, but the adversarial training is applied post-distillation, instead of during the distillation, and the method has not been applied to large-scale text-to-image models.

Rectified Flow (RF) \cite{liu2022flow,liu2023instaflow} straightens the flow by repeatedly training with deterministic data and noise pairs. However, its few-step generation quality is still poor. Additionally, since the model has only seen specific data and noise pairs during the distillation, it no longer supports data pairing with arbitrary noise. This impacts the ability for image editing such as SDEdit \cite{meng2022sdedit}. 

Score Distillation Sampling (SDS) \cite{poole2022dreamfusion} has been used in SDXL-Turbo \cite{sauer2023adversarial} to stabilize adversarial training, yet its effect is minimal and it cannot be used as a distillation method alone. Variation Score Distillation (VSD) \cite{wang2023prolificdreamer} has recently been used in diffusion distillation \cite{yin2023onestep}. However, it requires training an additional score model of the negative distribution during the distillation process, and like the discriminator in adversarial training, it also involves a dynamic training target that can negatively affect training stability. There is no open-source model for comparison, and our preliminary experiments find our method achieves better quality.

\subsection{LoRA}

Low-Rank Adaptation (LoRA) \cite{hu2021lora} is an efficient finetuning technique. It only trains a small number of additional parameters to the model and has become particularly popular for training stylization modules for existing text-to-image models.

LCM-LoRA \cite{luo2023lcmlora} is the first to show that model distillation can also be trained as a LoRA module. This ensures minimum parameter changes and can be conveniently plugged into the existing ecosystem.

Our work is inspired by this approach and we provide our distilled models both as LoRAs for convenient plug and play and as full models for even better quality.

\section{Method}

\subsection{Why Distillation with MSE Fails}
\label{sec:why-mse-fails}

\begin{figure}[h]
    \centering
    \vspace{-10pt}
    \includegraphics[width=\linewidth]{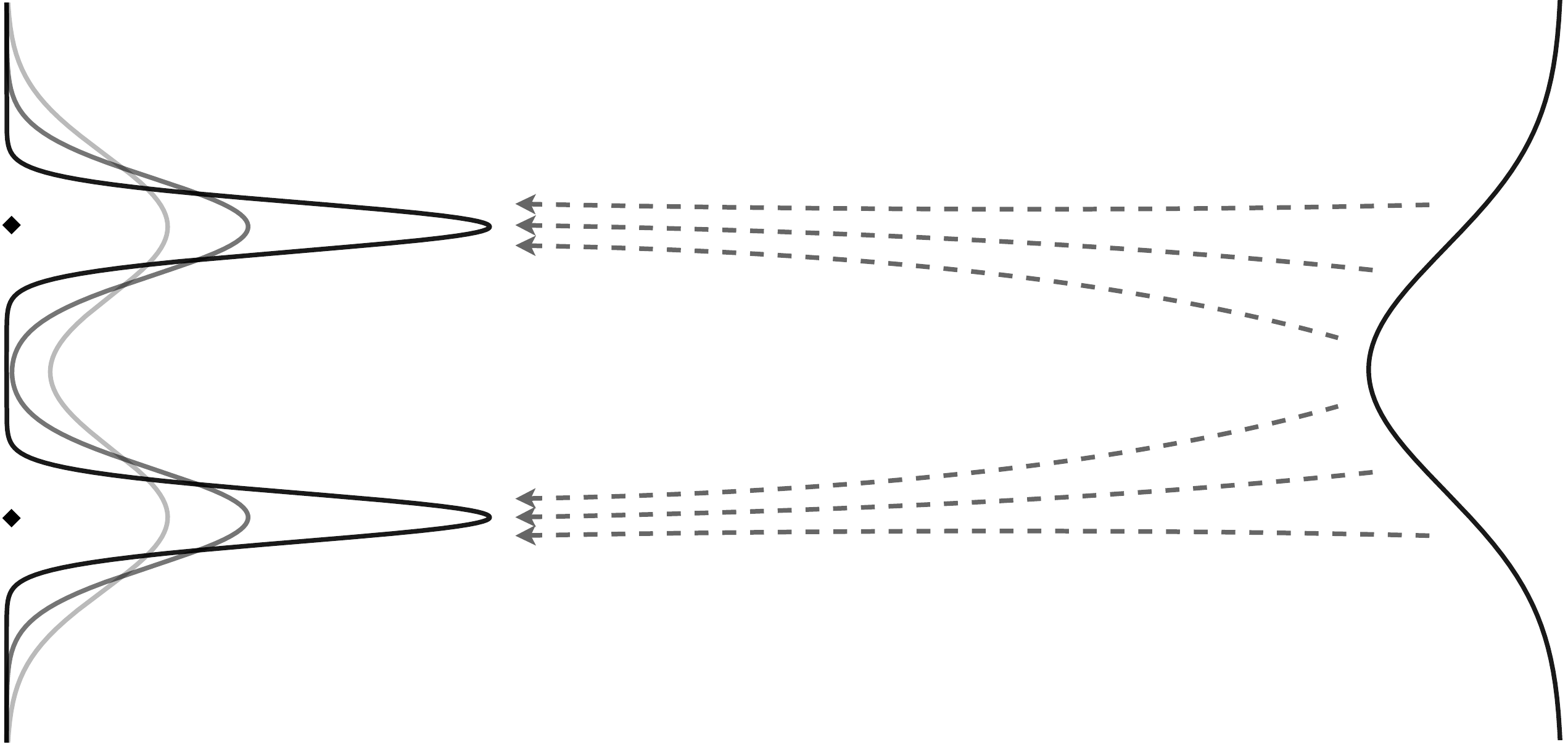}
    \caption{Illustration of multiple possible flows learned by models with different capacities. Distilled student models for few-step generations do not have the same capacity to match with the teacher models, leading to blurry results with MSE loss.}
    \label{fig:ode_ambiguity}
\end{figure}

The learned probability flow is determined by the dataset, the forward function \cite{lipman2023flow,liu2022flow}, the loss function \cite{lin2024diffusion}, and the model capacity. Given finite training samples, the underlying data distribution is ambiguous. The maximum likelihood estimation (MLE) is a distribution that assigns even probability only to the observed samples and zero everywhere else. If the model has infinite capacity, it will learn a flow of this maximum likelihood estimation and overfit to always produce observed samples and generate no new data. In practice, diffusion models can generate new data because neural networks are not exact learners.

When the model is used in multi-step generations, it is stacked and has a higher Lipschitz constant and more non-linearities to approximate a more complex distribution. But when the model is used in few-step generations, it no longer has the same amount of capacity to approximate well the same distribution. This is evidenced by diffusion models can have very sharp changes in results despite small changes in the initial noises \cite{guo2023smooth}, but the distilled models have much smoother latent traversal. This explains why distillation with MSE loss produces blurry results. The student model simply does not have the capacity to match the teacher.

Additionally, neural network parameter optimization involves a complex landscape. Even models with the same capacity can hardly match output exactly since parameters can get stuck at different local minima.

We find that other distance metrics, \eg L1 and perceptual loss \cite{zhang2018unreasonable,lin2024diffusion}, also produce undesirable results. On the other hand, we find adversarial objectives to be effective in mitigating this issue.

\subsection{Adversarial Objective}
\label{sec:adversarial-objective}

Instead of using the MSE loss between the student-predicted $\hat{x}_{t-ns}$ and teacher-predicted $x_{t-ns}$ as in \Cref{eq:progressive-mse}, we use an adversarial discriminator. Specifically, our discriminator $D: \mathbb{R}^d \rightarrow \mathbb{R} \in [0,1]$ computes the probability of $x_{t-ns}$ being generated from the teacher as opposed to the student, given condition $x_t$ and $c$.
\begin{equation}
    D(x_t, x_{t-ns}, t, t-ns, c)
\end{equation}

We use non-saturated adversarial loss \cite{goodfellow2014generative} and train the discriminator and the student model in alternating steps. This encourages the student prediction $\hat{x}_{t-ns}$ to be closer to the teacher prediction $x_{t-ns}$:
\begin{align}
    p &= D(x_t, x_{t-ns}, t, t-ns, c) \\
    \hat{p} &= D(x_t, \hat{x}_{t-ns}, t, t-ns, c) \\
    \mathcal{L}_{D} &= -\log(p) - \log(1 - \hat{p}) \\
    \mathcal{L}_{G} &= -\log(\hat{p})
\end{align}

The condition on $x_t$ is important for preserving the probability flow. This is because the teacher's generation of $x_{t-ns}$ is deterministic from $x_t$. By providing the discriminator both $x_{t-ns}$ and $x_t$, the discriminator learns the underlying probability flow and the student must also follow the same flow to fool the discriminator.

Our formulation is very similar to a prior work \cite{xiao2022tackling} except we use it for distillation instead of training from scratch. Note that this approach only preserves the probability flow and ensures mode coverage when used in distillation.

\subsection{Discriminator Design}

A prior work \cite{lin2024diffusion} has shown that a pre-trained diffusion model's U-Net \cite{ronneberger2015unet} encoder can be used as a vision backbone. Such a pre-trained backbone is very suitable for our discriminator because it has been pre-trained on the target dataset, directly operates in the latent space, supports noised input at all timesteps, and supports text condition.

We follow the approach and copy the encoder and midblock of the pre-trained SDXL model as our discriminator backbone $d$. We pass $x_{t-ns}$ and $x_t$ independently through the shared backbone $d$, concatenate the hidden features after the midblock in the channel dimension, and pass it to a prediction head. The prediction head consists of simple blocks of $4\times4$ convolution with a stride of 2, group normalization \cite{wu2018group} with 32 groups, and SiLU activation \cite{hendrycks2023gaussian,ramachandran2017searching} layers to further reduce the spatial dimension. The output is projected to a single value and clamped to $[0, 1]$ range with sigmoid $\sigma(\cdot)$. Together they form the complete discriminator $D$:

\begin{equation}
\begin{aligned}
    &D(x_t, x_{t-ns}, t, t-ns, c) \\
    & \quad \equiv \sigma\bigg(\mathrm{head}\Big( d(x_{t-ns}, t-ns, c), d(x_t, t, c)\Big)\bigg)
\end{aligned}
\end{equation}

Note that the backbone is initialized with the pre-trained weights and we train the entire discriminator without freezing the backbone. We find our training stable without the need for expansive R1 regularization \cite{mescheder2018training} nor switching to L2 attention \cite{kim2021lipschitz,kang2023scaling}. Additional stabilization techniques are discussed in \Cref{sec:stable-techniques}.

\subsection{Relax the Mode Coverage}
\label{sec:mode-relax}

\begin{figure}[h]
    \centering
    \begin{tabular}{@{}c@{}c@{}c@{}c@{}}
        \includegraphics[width=0.25\linewidth]{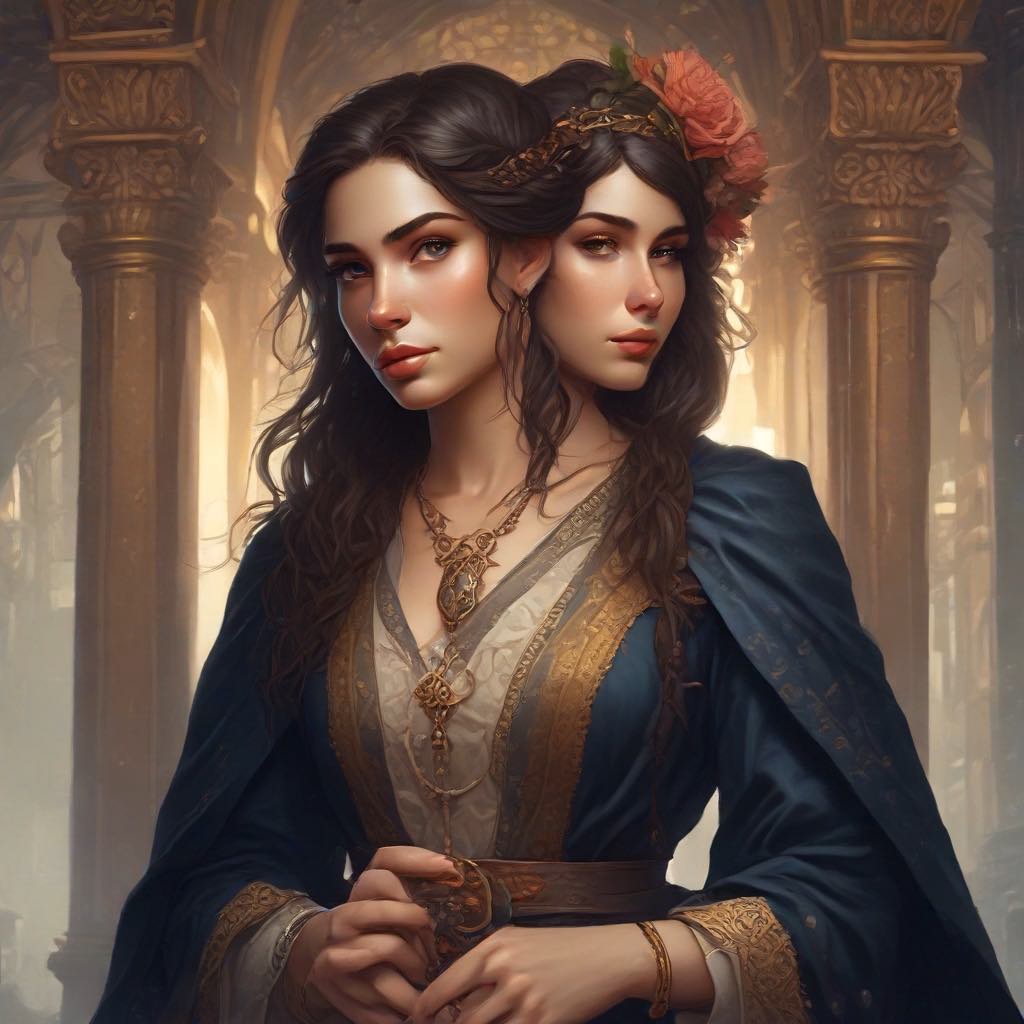} &
        \includegraphics[width=0.25\linewidth]{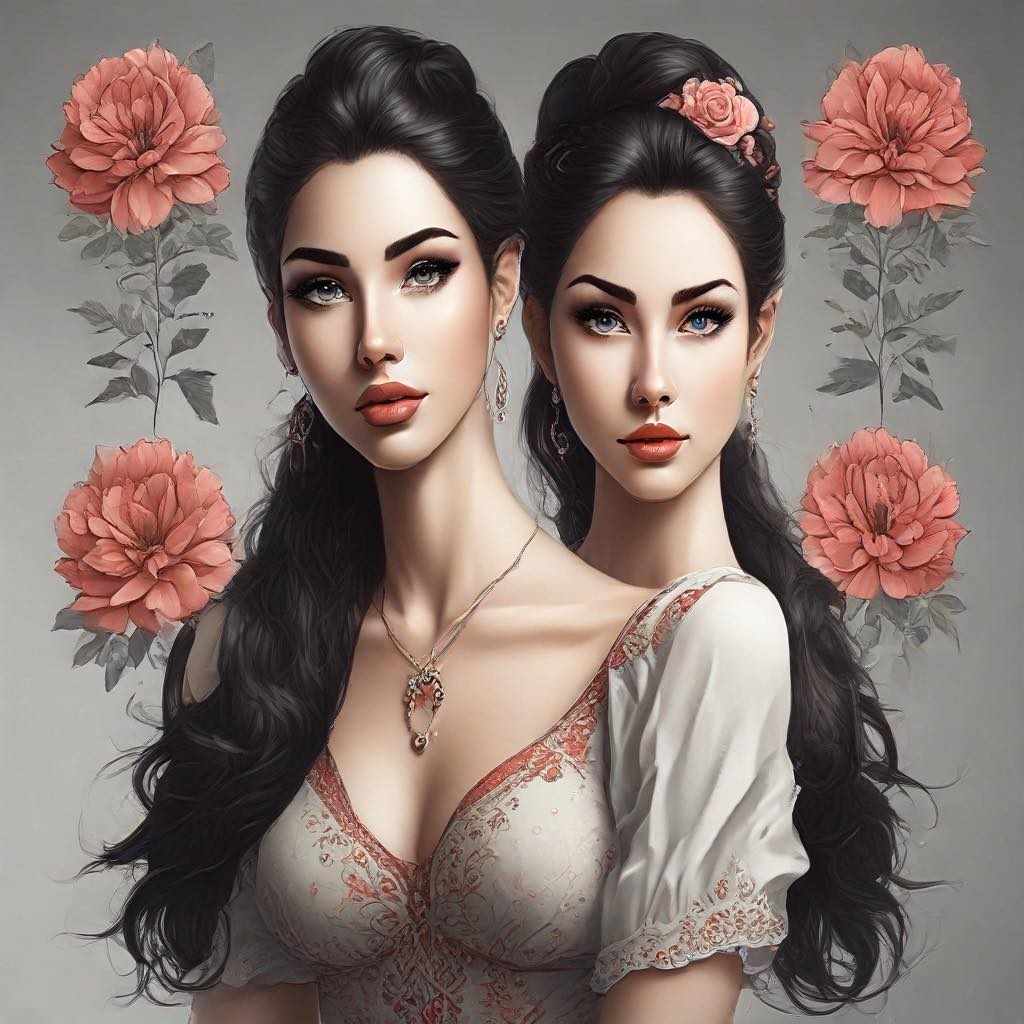} &
        \includegraphics[width=0.25\linewidth]{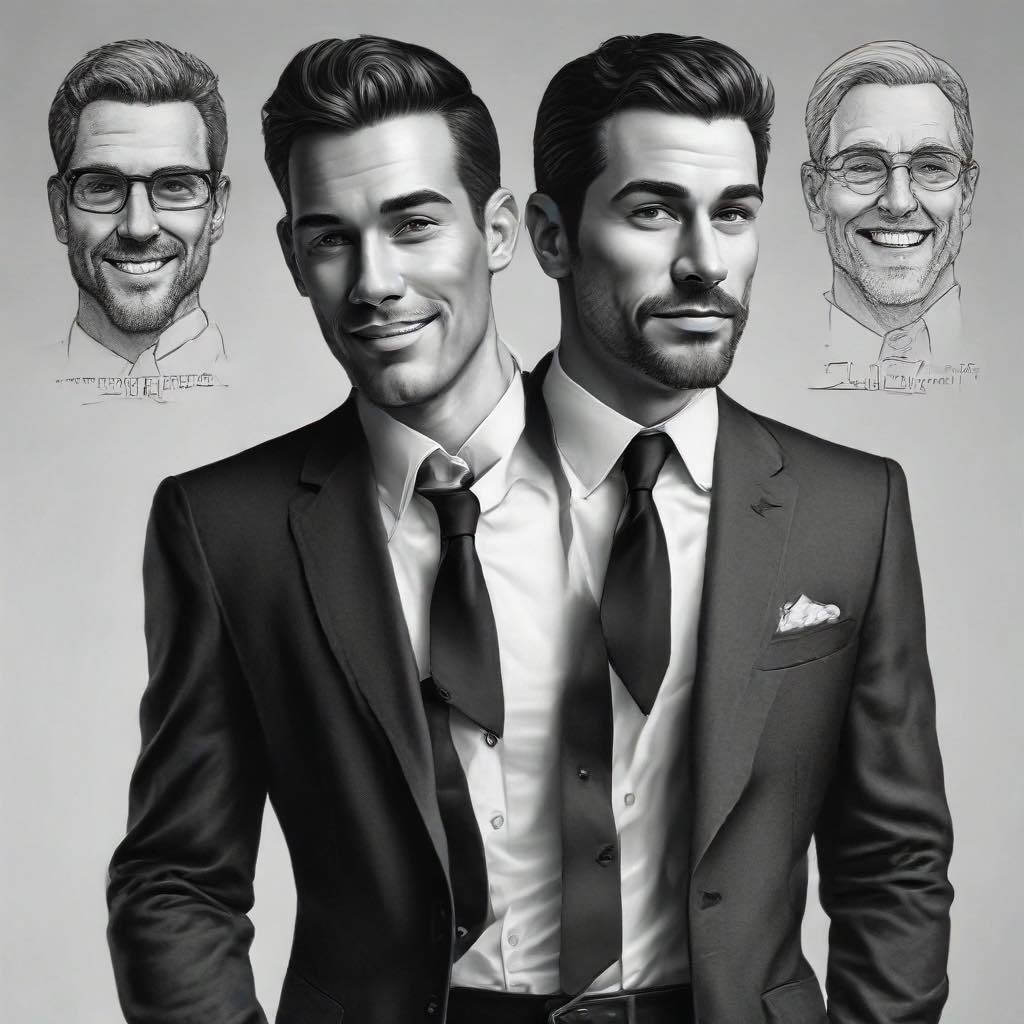} &
        \includegraphics[width=0.25\linewidth]{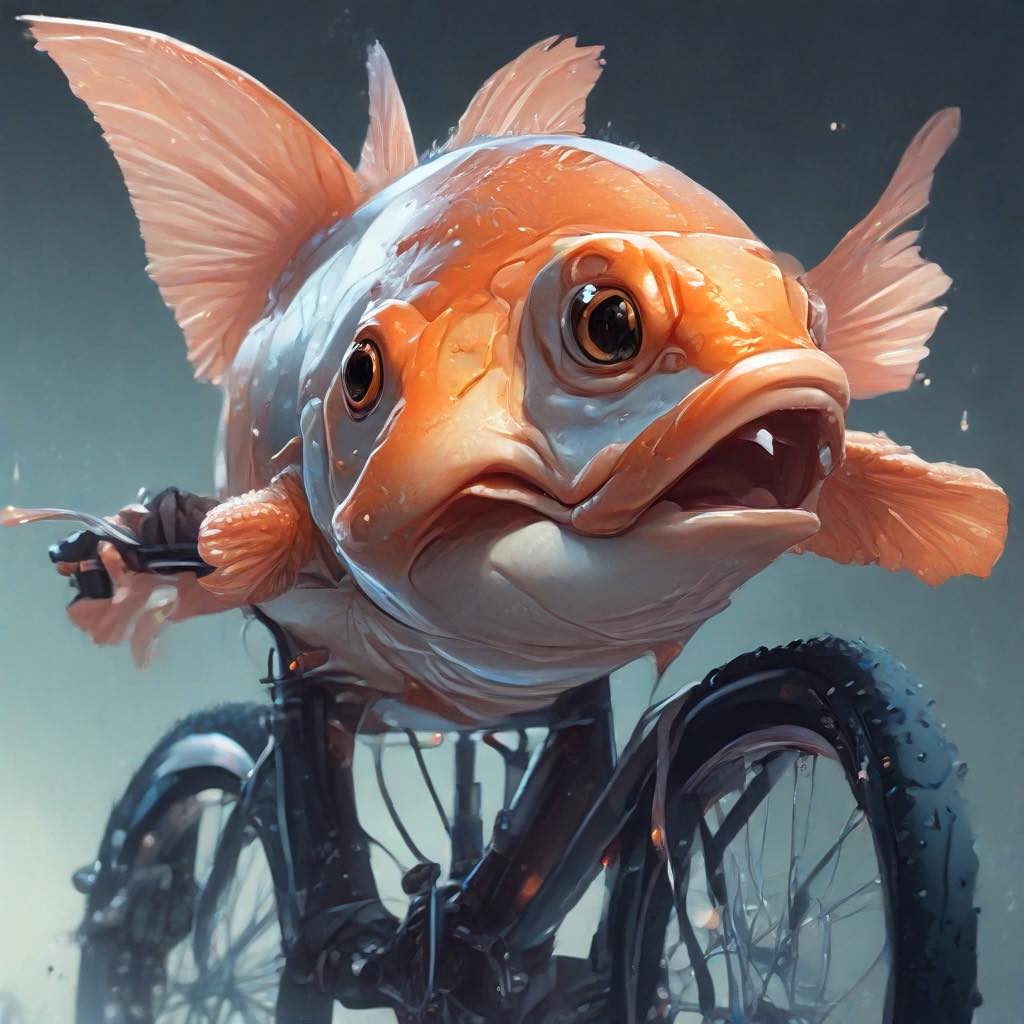}
    \end{tabular}
    \caption{``Janus'' artifacts appear when the student network does not have the capacity to match the teacher's sudden changes. This problem can be mitigated by relaxing the mode coverage requirement.}
    \label{fig:janus}
\end{figure}

The adversarial objective above encourages the prediction to be both sharp and flow-preserving, but this does not change the fact that the student does not have enough capacity to perfectly match the teacher as discussed in \Cref{sec:why-mse-fails}. With the MSE objective, it manifests blurry results. With the adversarial objective, it manifests the ``Janus'' artifacts.

As shown in \Cref{fig:janus}, the teacher model can sometimes generate drastic layout changes for adjacent noise inputs, but the student model does not have the same capacity to make such sharp changes. As a result, the adversarial loss sacrifices semantic correctness in need to preserve the sharpness and the layout, manifesting artifacts that feature conjoined heads and bodies.

Semantic correctness is more important than mode coverage by human preference. Therefore, after training with the original adversarial objective, we relax the flow preservation requirement. Specifically, we further finetune the model without the condition on $x_t$:

\begin{equation}
\begin{aligned}    
    & D'(x_{t-ns}, t-ns, c) \\
    & \quad \equiv \sigma\bigg(\mathrm{head}\Big( d(x_{t-ns}, t-ns, c)\Big)\bigg)
\end{aligned}
\end{equation}

We find that finetuning with this objective is effective in removing the ``Janus'' artifacts while still preserving the original flow to a great extent in practice. Therefore, at every stage of the progressive distillation, we first train with the conditional objective and then finetune with this unconditional objective. Since the unconditional objective only concerns per-sample quality, we use the skip-level teacher for distilling the one-step and two-step models to further retain quality and mitigate error accumulation.

\subsection{Fix the Schedule}

A prior work \cite{lin2023common} has shown that common diffusion schedules are flawed. Specifically, the schedule does not reach pure noise at $t=T$ during training, yet pure noise is given during inference, causing a discrepancy. Unfortunately, SDXL uses this flawed schedule. The effect is less obvious under a large number of inference steps but is particularly detrimental for few-step generations.

A hacky way to circumvent the problem is to hard swap pure noise $\epsilon$ as model input at $t=T$ during training. This way the model is trained to expect pure noise as input at $t=T$ and we still use \Cref{eq:conversion-eps-pred} with the old $\bar\alpha$ schedule at inference to avoid singularity. It incurs minimum changes to the sampling procedure with existing software ecosystems \cite{von-platen-etal-2022-diffusers}. This approach is also used by SDXL-Turbo \cite{sauer2023adversarial}.
\begin{equation}
\begin{aligned}
    &\mathbf{Forward}(x_0, \epsilon, t) \\
    &\quad = \begin{cases}
        \mathbf{forward}(x_0, \epsilon, t), &\text{when}\ t < T \\
        \epsilon, &\text{when}\ t=T
    \end{cases}
\end{aligned}
\end{equation}

\subsection{Distillation Procedure}

First, we perform distillation from 128 steps directly to 32 steps with MSE loss. We find MSE is sufficient for the early stage. We also apply classifier-free guidance (CFG) \cite{ho2022classifierfree} only in this stage. We use a guidance scale of 6 without any negative prompts.

Then, we switch to using adversarial loss to distill the step count in this order: $32 \rightarrow 8 \rightarrow 4 \rightarrow 2 \rightarrow 1$. At each stage, we first train with the conditional objective as in \Cref{sec:adversarial-objective} to preserve the probability flow, and then train with the unconditional objective as in \Cref{sec:mode-relax} to relax the mode coverage.

At each stage, we first train with LoRA using the two objectives, then we merge the LoRA and train the whole UNet further with the unconditional objective. We find finetuning the whole UNet can achieve even better performance, while the LoRA module can be used on other base models. Our LoRA settings are the same as LCM-LoRA \cite{luo2023lcmlora}, which uses rank 64 on all the convolution and linear weights except the input and output convolutions and the shared time embedding linear layers. We do not use LoRA on the discriminator. We re-initialize the discriminator at each stage.

We distill our models on a subset of LAION \cite{schuhmann2022laion5b} and COYO \cite{kakaobrain2022coyo-700m} dataset. We select images to be greater than 1024px and LAION images with aesthetic scores above 5.5. We additionally filter images by sharpness using a Laplacian filter and clean up the text prompts. The distillation is conducted on a square aspect ratio, but we find it generalizes well to other aspect ratios at inference time.

We use batch size 512 across 64 A100 80G GPUs. For the first $128 \rightarrow 32$ stage with MSE loss, we use learning rate 1e-5 with Adam $\beta_1=0.9,\beta_2=0.999$. For the remaining stages with adversarial loss, we use learning rates 1e-6 with LoRA and 5e-7 without LoRA for both the student and the discriminator networks. The Adam optimizer \cite{kingma2017adam} uses $\beta_1=0, \beta_2=0.99$ following prior works \cite{karras2020analyzing,kang2023scaling} without weight decay \cite{loshchilov2019decoupled}. We use gradient accumulation, VAE slicing, BF16 mixed precision \cite{micikevicius2018mixed}, flash attention \cite{dao2022flashattention,dao2023flashattention2}, and zero redundancy optimizer \cite{rajbhandari2020zero} to reduce the memory footprint.

\subsection{Stable Training Techniques}
\label{sec:stable-techniques}

For one-step and two-step distillations, we employ additional techniques to stabilize the training.

\subsubsection{Train Student Networks at Multiple Timesteps}

While we only need to train the one-step model at timestep \{1000\}, and the two-step model at timesteps \{500, 1000\} for complete image generation, we find training on more timesteps \{250, 500, 750, 1000\} improves stability. As an additional benefit, this allows our models to support SDEdit \cite{meng2022sdedit} at different timesteps as illustrated below:
\begin{center}
\includegraphics[width=\linewidth]{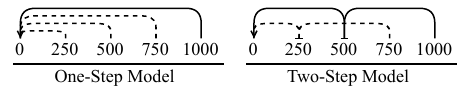}
\end{center}

\subsubsection{Train Discriminator at Multiple Timesteps}

We find training the one-step model using the above discriminator formulation very unstable. We find the root reason is that our discriminator uses the pre-trained diffusion UNet encoder as the backbone, yet the diffusion encoder is trained to only focus on high-frequency details at lower timesteps and low-frequency structures at higher timesteps. For one-step generations, the student network directly predicts $\hat{x}_0$. If we pass $\hat{x}_0$ and $t=0$ to the discriminator backbone, it is not able to critique the image structure. This leads to images with bad shapes and even divergence.

Our solution is to add noise to both teacher-predicted $x_0$ and student-predicted $\hat{x}_0$ to timesteps: \{10, 250, 500, 750\} randomly. This way the discriminator can critique the prediction on both high-frequency details and low-frequency structures. 

Specifically, we first draw $t* \leftarrow \{10, 250, 500, 750\}$ with uniform weighting 1:1:1:1 and sample a new noise $\epsilon* \sim \mathcal{N}(0, \mathbf{I})$. Then we apply the noise before passing it through the conditional and unconditional discriminators:
\begin{align}
    & D(x_t, \mathbf{forward}(\hat{x}_0, \epsilon*, t*), t, t*, c) \\
    & D'(\mathbf{forward}(\hat{x}_0, \epsilon*, t*), t*, c)
\end{align}

After the model is trained stable, we change the timesteps weighting to 5:1:1:1. This further improves details and removes noisy artifacts.

Note that this stabilization technique can also be viewed from the lens of bridging the distribution gap \cite{mescheder2018training}, discriminator augmentation \cite{karras2020training}, and multi-scale discriminator \cite{karnewar2020msggan}.

\subsubsection{Switch to $x_0$ Prediction}
\label{sec:switch_to_x0}

We find the one-step model with $\epsilon$-prediction formulation tends to generate noise artifacts likely due to numerical instability. We change the one-step model to $x_0$-prediction and it resolves the issue.

Specifically, we copy the network and convert the predicted $\hat\epsilon$ to $\hat{x}_0$ through conversion function $\mathbf{x}$ defined in \Cref{eq:conversion-eps-pred}. We use MSE to gradually guide the online model to $x_0$-prediction.
\begin{align}
    \hat{\epsilon} &= f_{\mathrm{frozen}}(x_t, t, c) \\
    \hat{x}_0 &= f_{\mathrm{online}}(x_t, t, c) \\
    \mathcal{L}_{\mathrm{convert}} &= \| \hat{x}_0 - \mathbf{x}(x_t, \hat{\epsilon}, t) \|_2^2
\end{align}

As discussed in \Cref{sec:why-mse-fails}, MSE loss cannot convert our model perfectly. The converted model generates blurry results, but this will be fixed by the adversarial objectives.

After the conversion, the one-step model is trained with adversarial objectives in $x_0$-prediction formulation, while the teacher model still operates in $\epsilon$-prediction formulation. Due to the substantial formulation change, we do not provide LoRA for one-step generation.

\section{Evaluation}

\subsection{Specification Comparison}

\Cref{tab:spec} shows the specification of our distilled models compared to others.

\begin{table}[h]
    \centering
    \setlength\tabcolsep{4pt}
    \begin{tabularx}{\linewidth}{Xcccc}
        \toprule
        \multirow{2}{*}{Method} & Steps & \multirow{2}{*}{Resolution} & CFG & Offer \\
         & Needed & & Free & LoRA \\
        \midrule
        SDXL \cite{podell2023sdxl} & 25+ & 1024px & No & - \\
        LCM \cite{luo2023latent,luo2023lcmlora} & 4+ & 1024px & Yes\&No & Yes \\
        Turbo \cite{sauer2023adversarial} & 1+ & 512px & Yes & No \\
        \textbf{Ours} & \textbf{1+} & \textbf{1024px} & \textbf{Yes} & \textbf{Yes} \\
        \bottomrule
    \end{tabularx}
    \vspace{-6pt}
    \caption{Model specifications. Our method requires the fewest amount of steps to produce high-quality samples.}
    \label{tab:spec}
\end{table}

\begin{figure*}[t]
    \centering
    \captionsetup{justification=raggedright,singlelinecheck=false}
    \small
    \setlength\tabcolsep{4pt}
    \begin{tabularx}{\textwidth}{|X|X|X|X|X|X|X|X|X|}
        \normalsize{\textbf{SDXL}} \cite{podell2023sdxl} & \multicolumn{4}{l|}{\normalsize{\textbf{Ours}}} & \multicolumn{2}{l|}{\normalsize{\textbf{Turbo}} \cite{sauer2023adversarial}} & \multicolumn{2}{l|}{\normalsize{\textbf{LCM}} \cite{luo2023latent}} \\

        \footnotesize{64NFE, CFG6} & \multicolumn{4}{l|}{\footnotesize{No CFG}} & \multicolumn{2}{l|}{\footnotesize{512px, No CFG}} & \footnotesize{8NFE, CFG8} & \footnotesize{No CFG} \\
        
        32 Steps & 8 Steps & 4 Steps & 2 Steps & 1 Step & 4 Steps & 1 Step & 4 Steps & 1 Step \\
    \end{tabularx}
    
    \begin{subfigure}[b]{\textwidth}
        \centering
        \setlength\tabcolsep{1pt}
        \begin{tabularx}{\textwidth}{@{}XX@{}X@{}X@{}XX@{}XX@{}X@{}}
            \includegraphics[width=\linewidth]{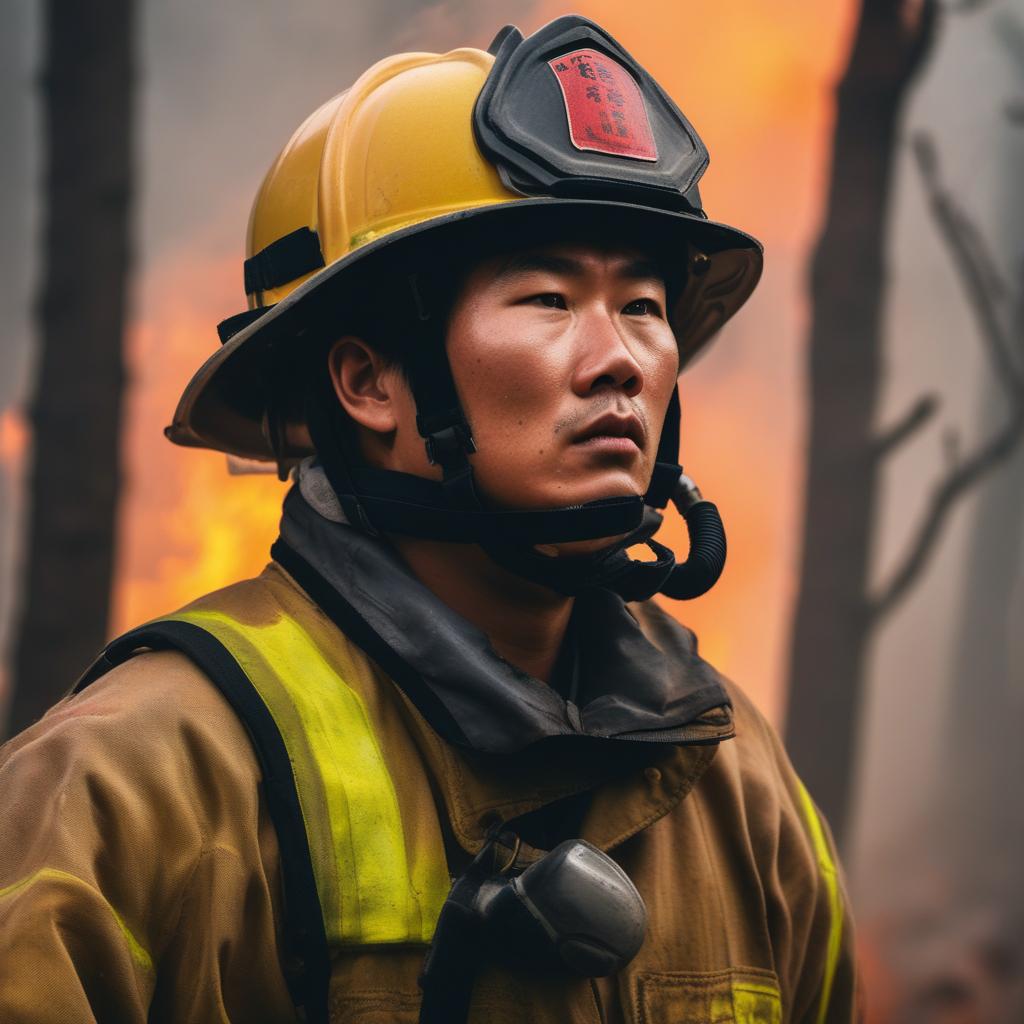} &
            \includegraphics[width=\linewidth]{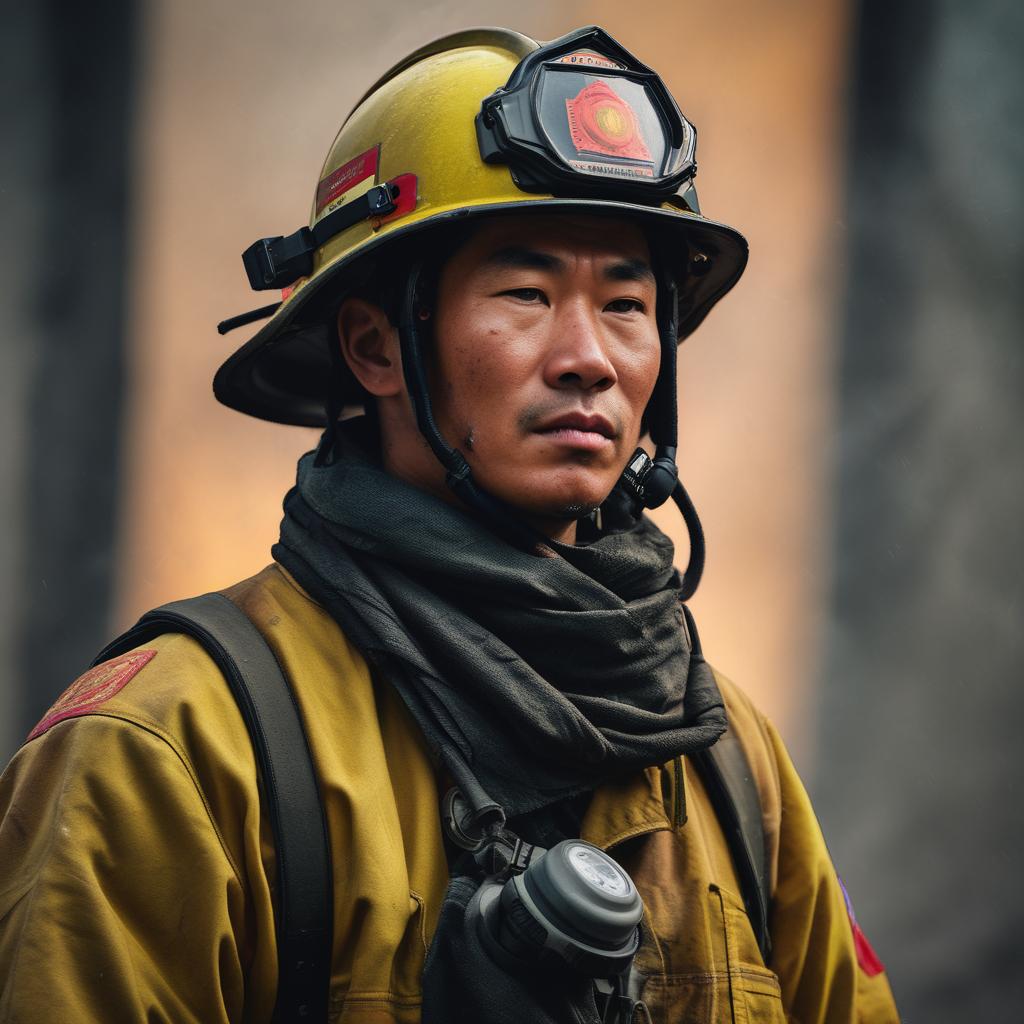} &
            \includegraphics[width=\linewidth]{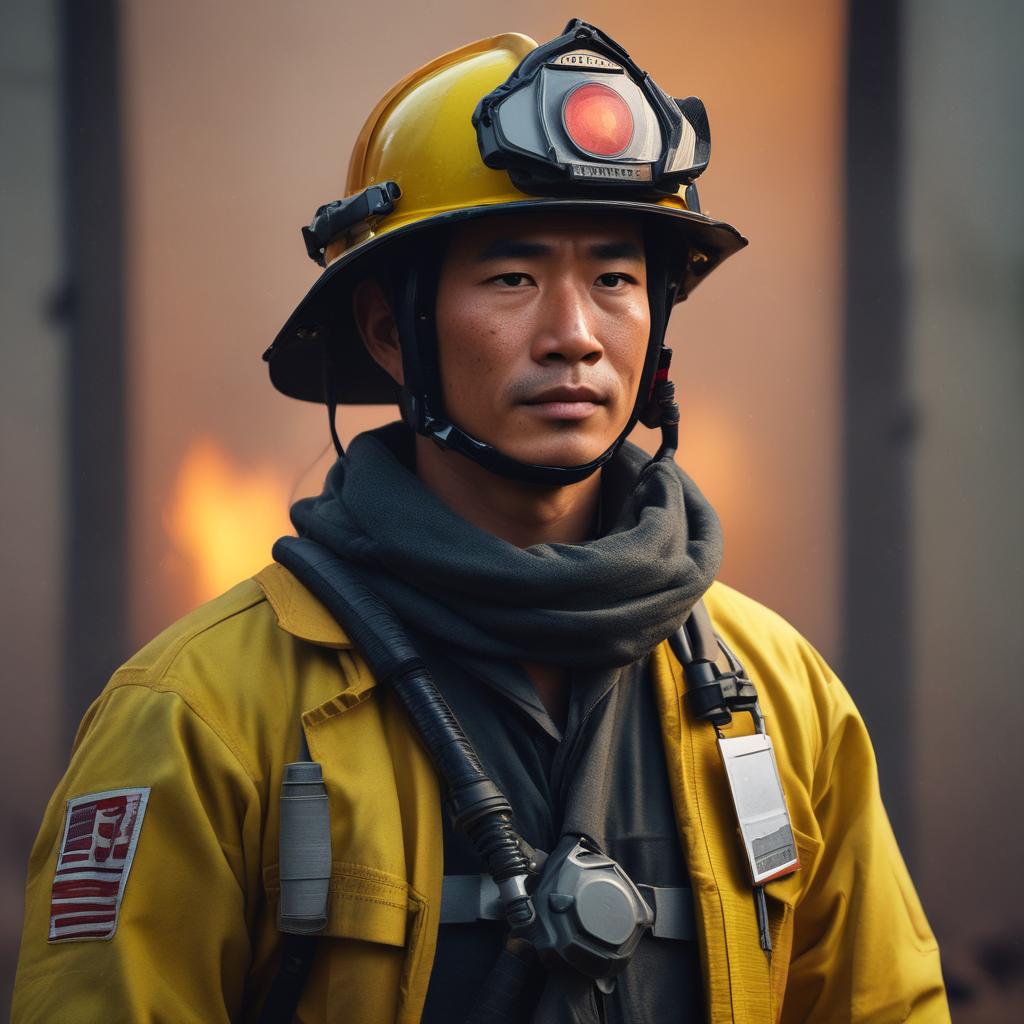} &
            \includegraphics[width=\linewidth]{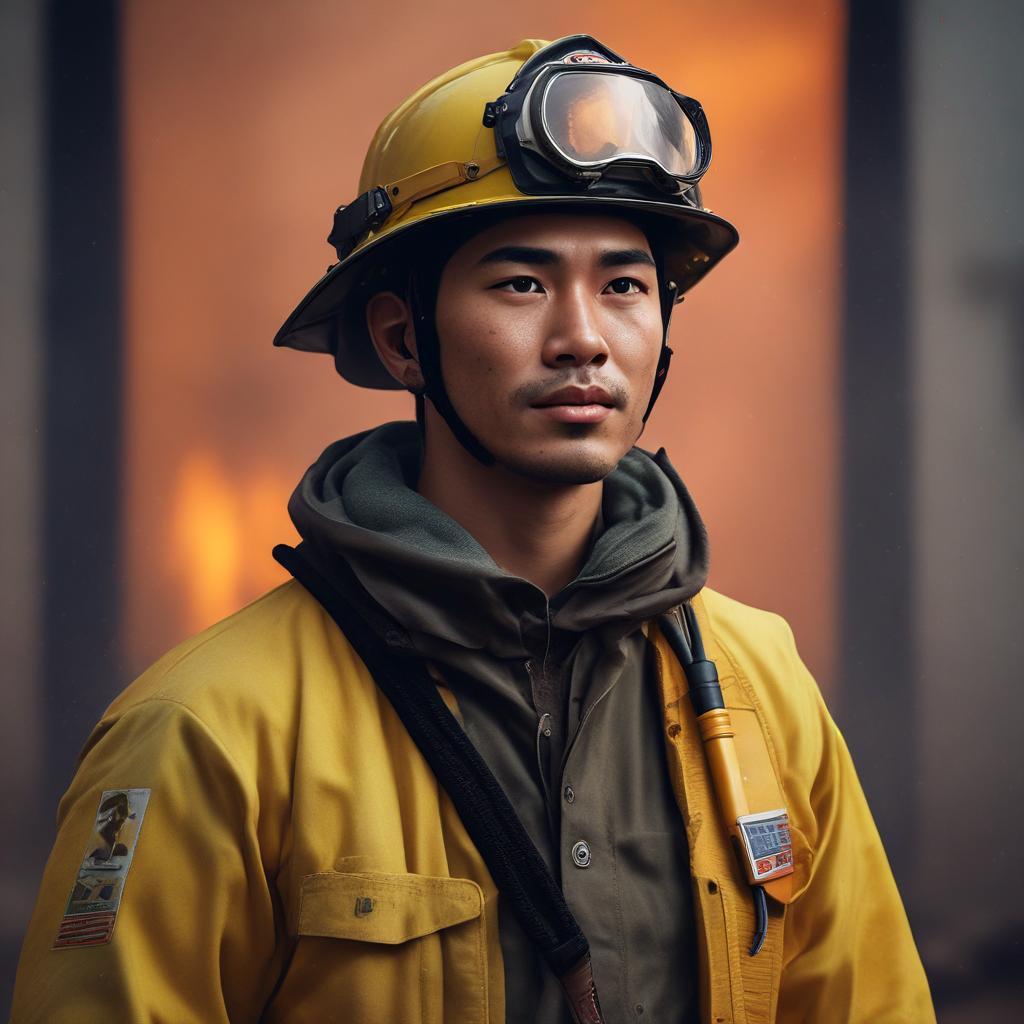} &
            \includegraphics[width=\linewidth]{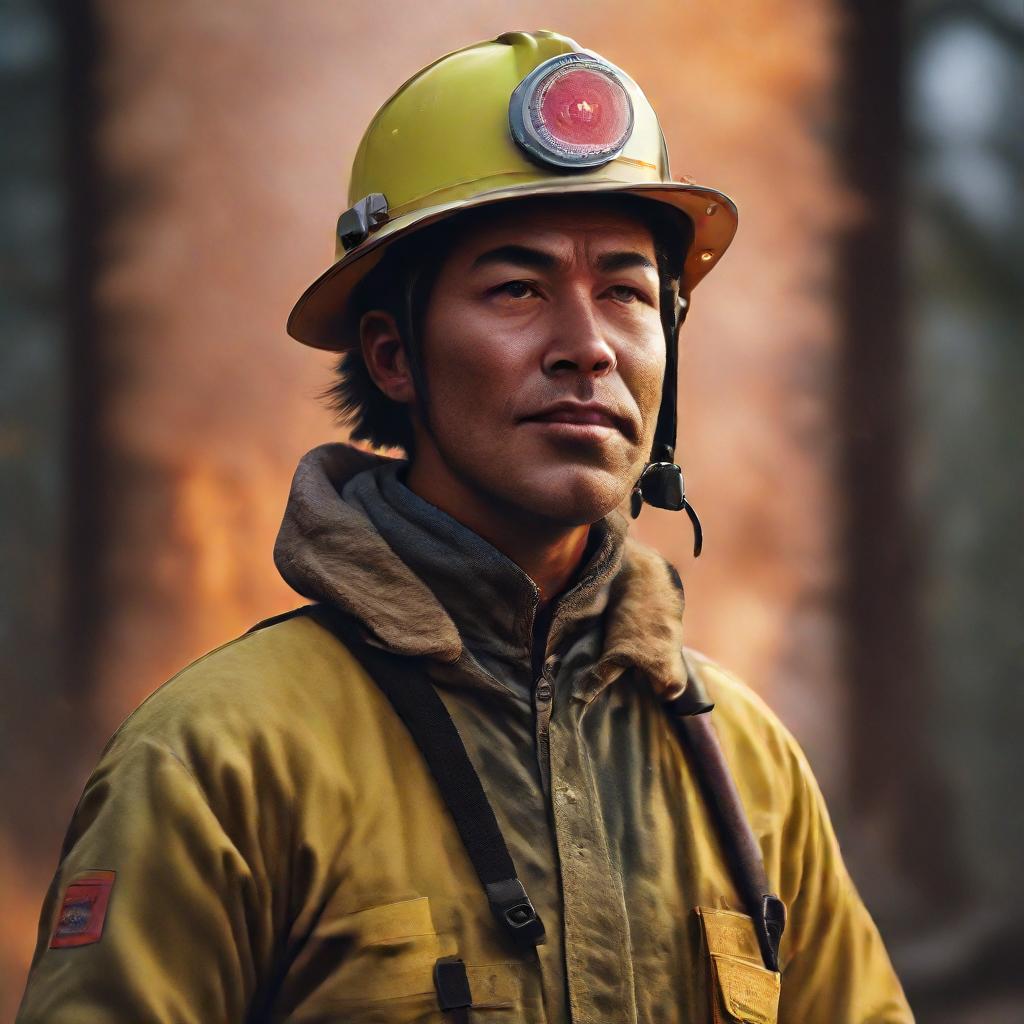} &
            \includegraphics[width=\linewidth]{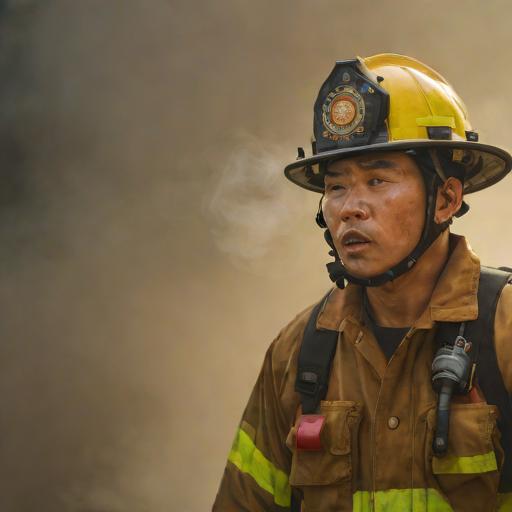} &
            \includegraphics[width=\linewidth]{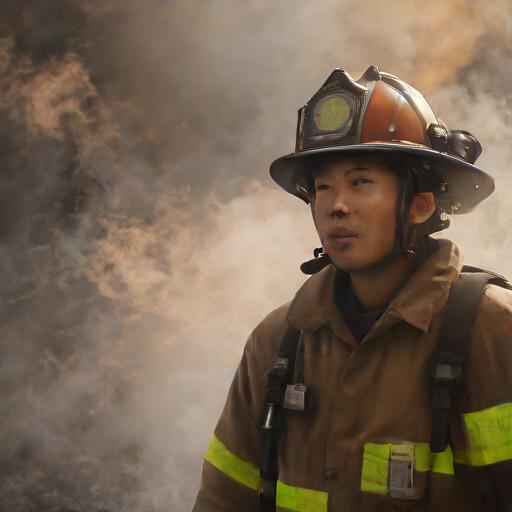} &
            \includegraphics[width=\linewidth]{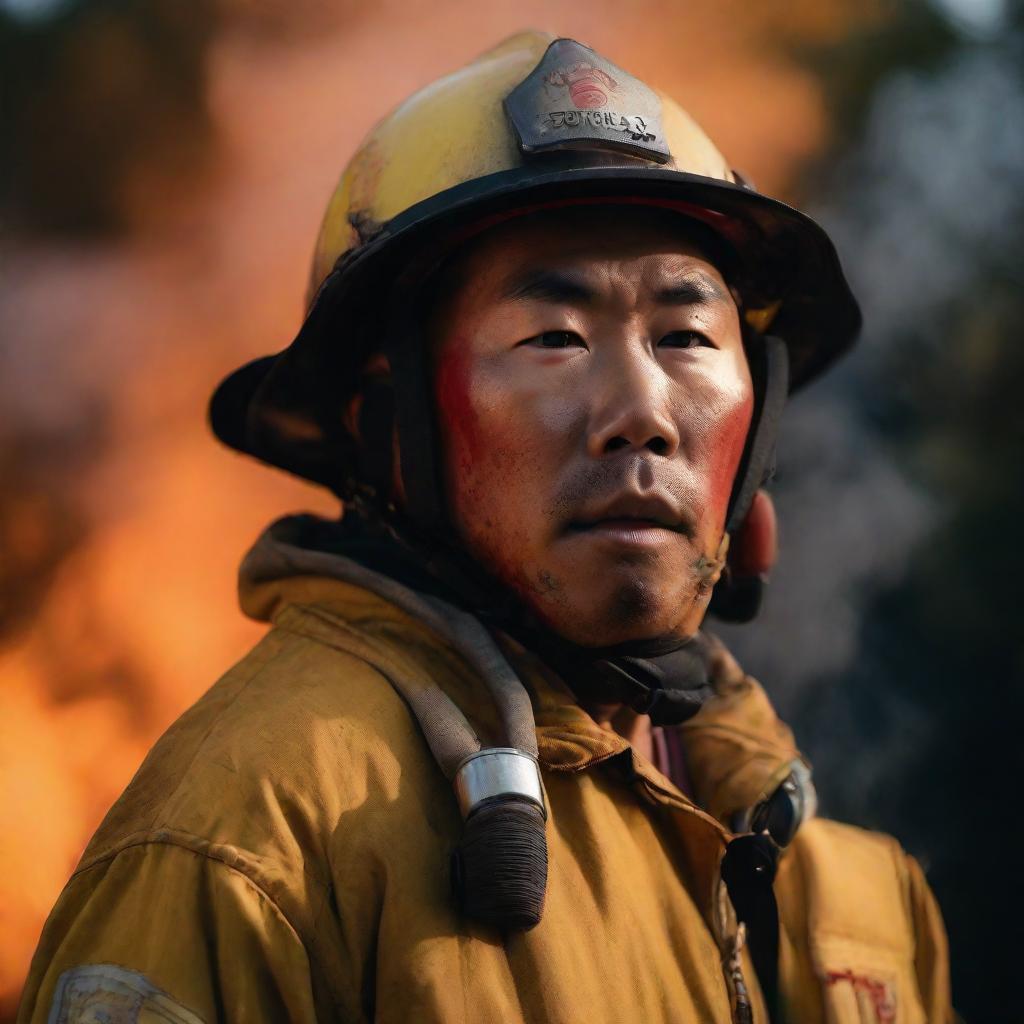} &
            \includegraphics[width=\linewidth]{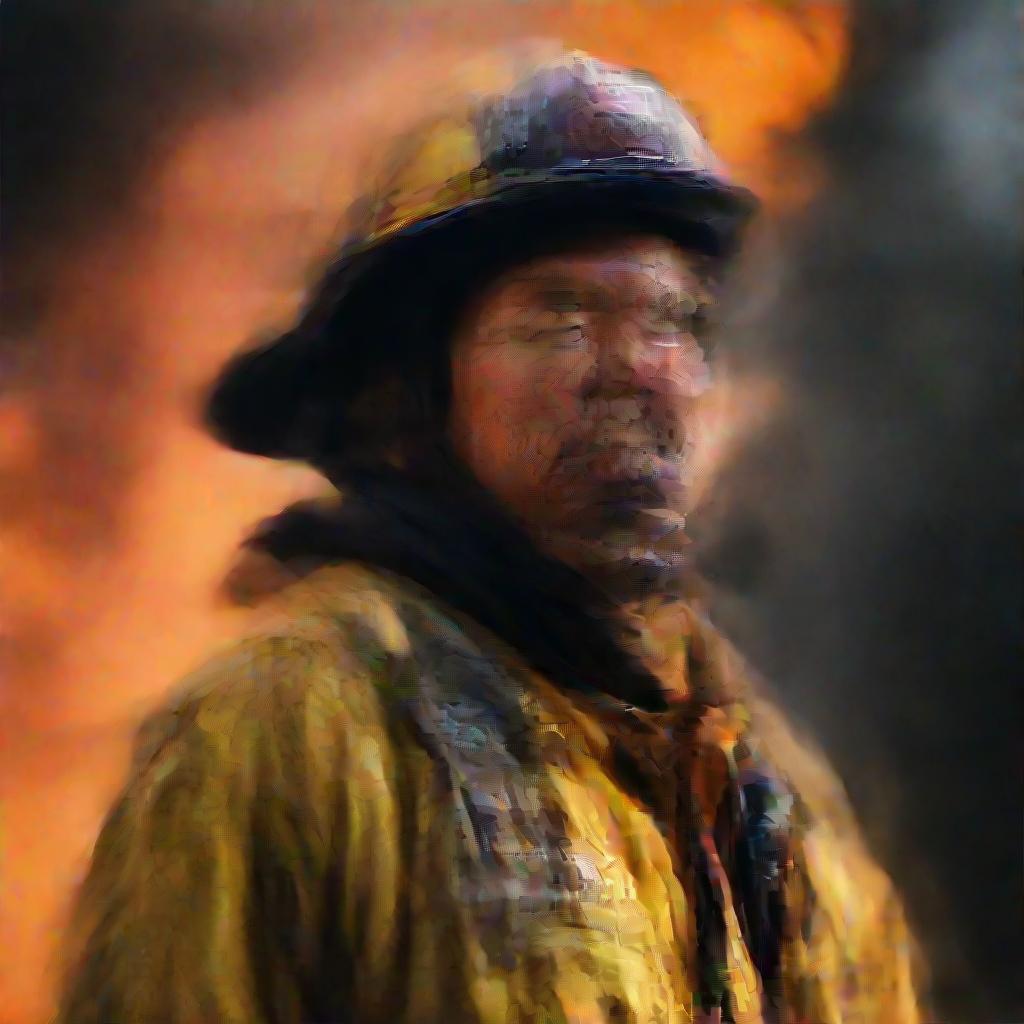}
        \end{tabularx}
        \vspace{-6pt}
        \caption{An Asian firefighter with a rugged jawline rushes through the billowing smoke of an autumn blaze.}
    \end{subfigure}

    \begin{subfigure}[b]{\textwidth}
        \centering
        \setlength\tabcolsep{1pt}
        \begin{tabularx}{\textwidth}{@{}XX@{}X@{}X@{}XX@{}XX@{}X@{}}
            \includegraphics[width=\linewidth]{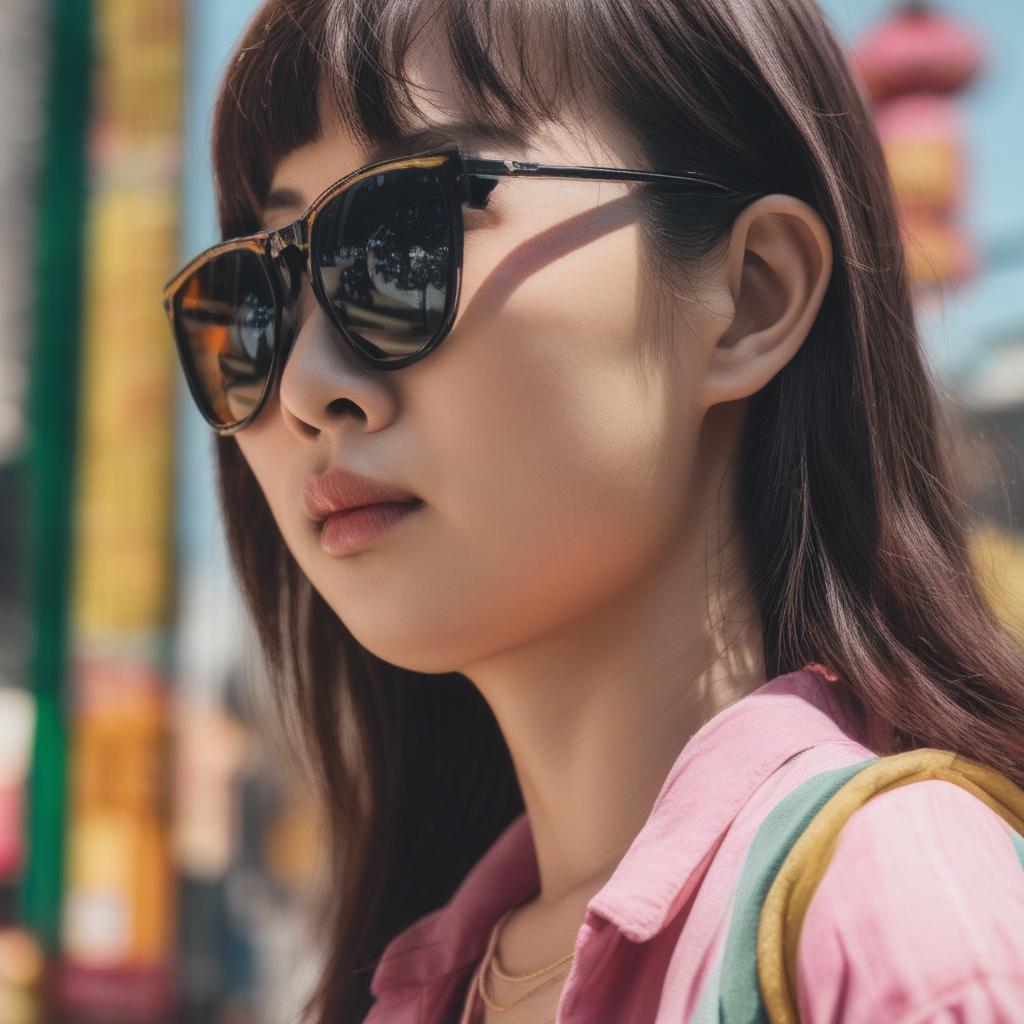} &
            \includegraphics[width=\linewidth]{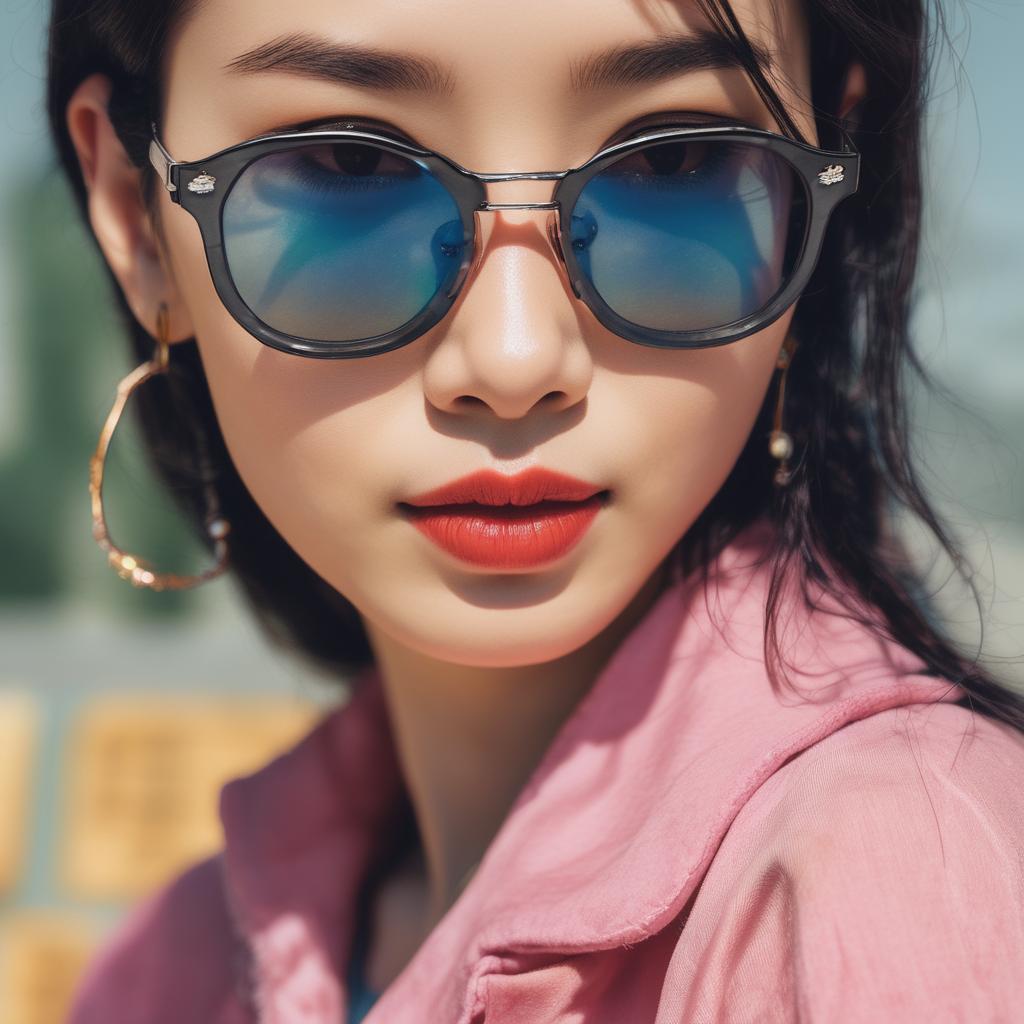} &
            \includegraphics[width=\linewidth]{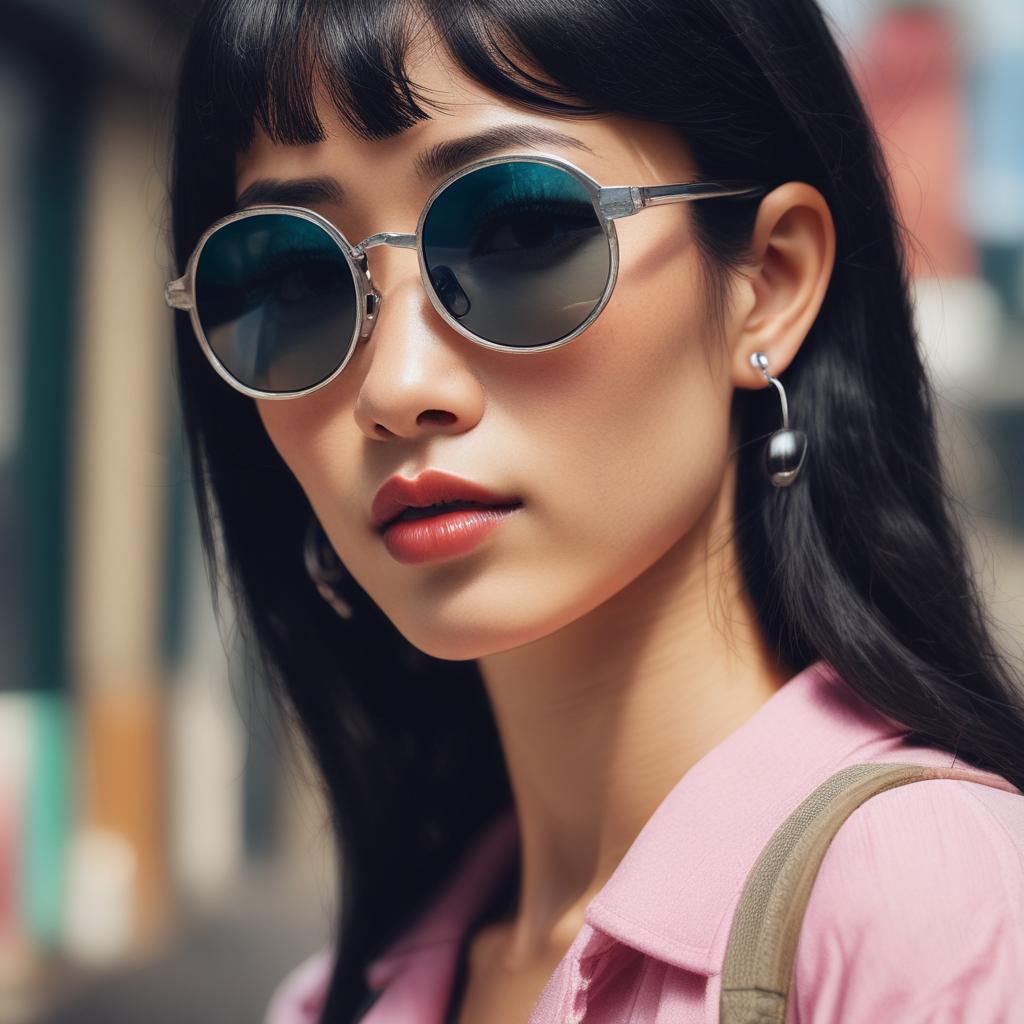} &
            \includegraphics[width=\linewidth]{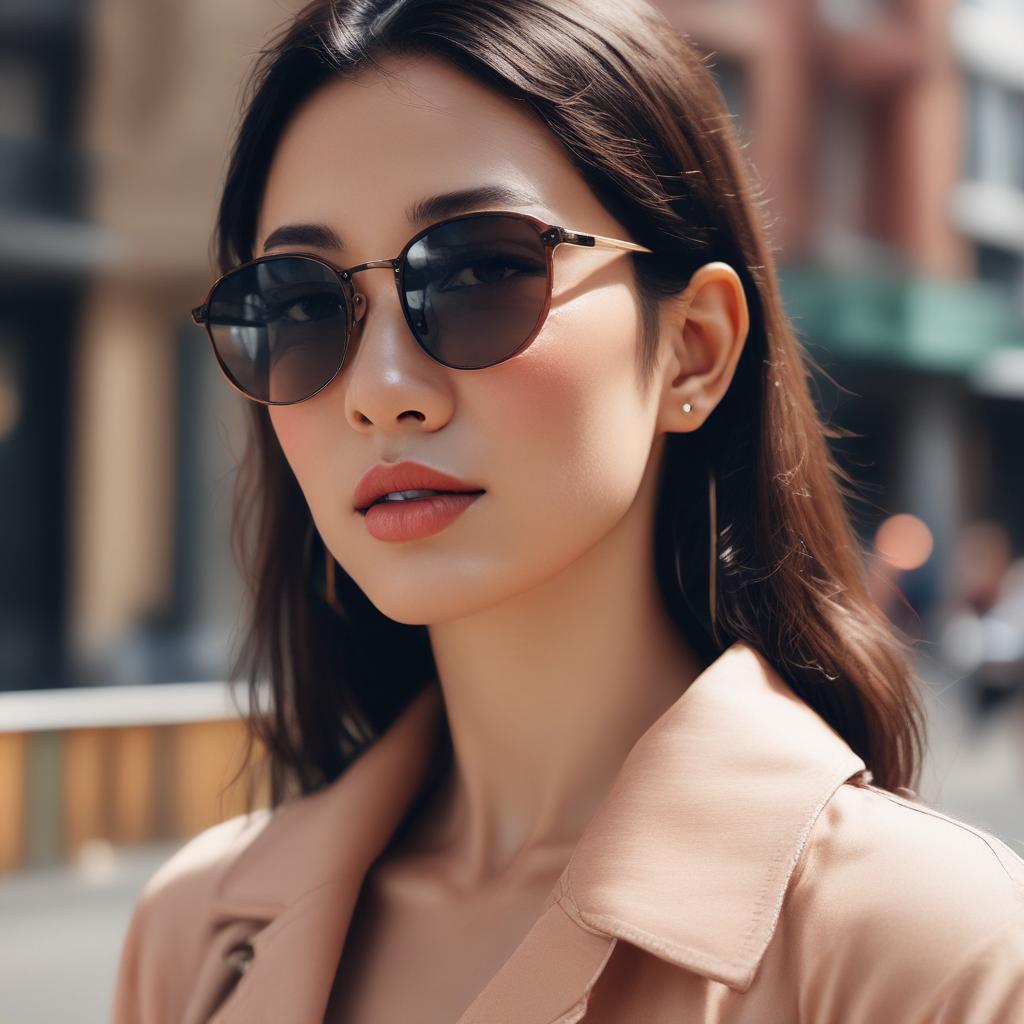} &
            \includegraphics[width=\linewidth]{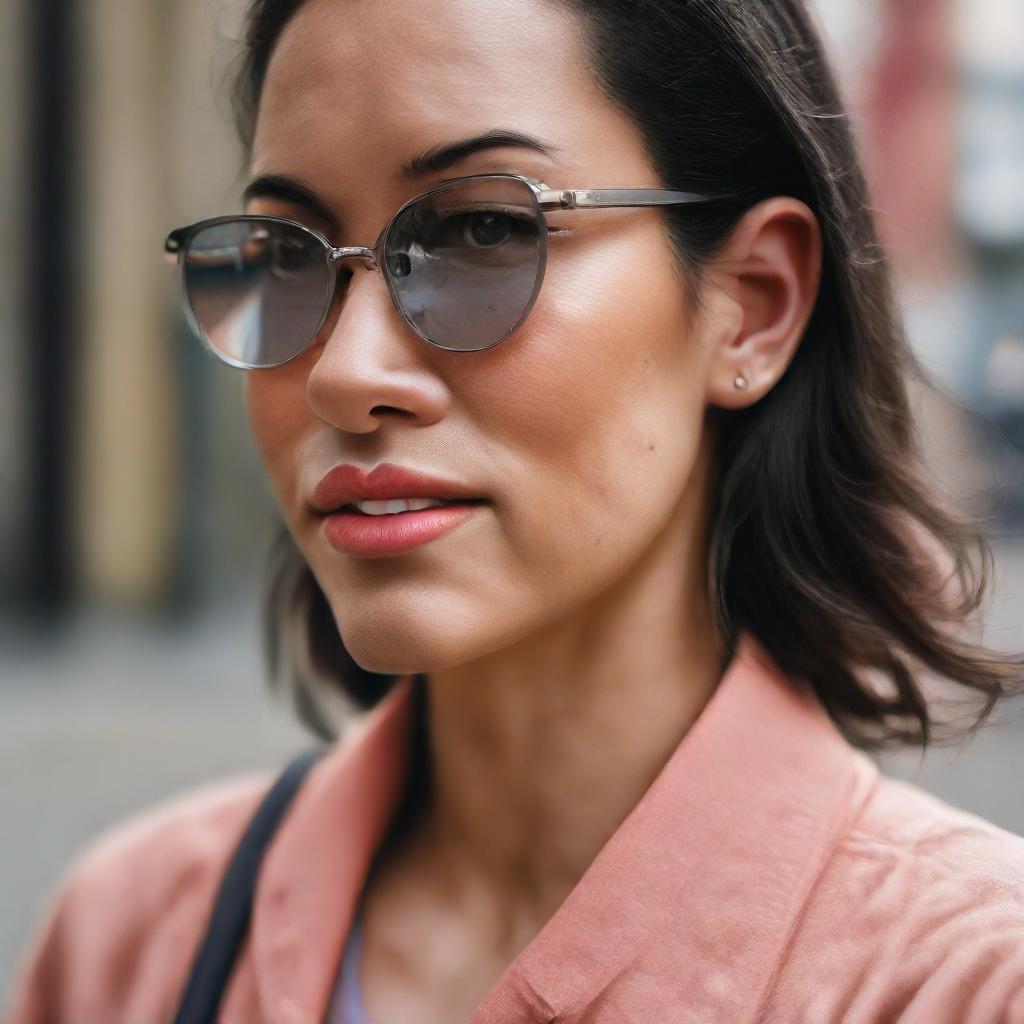} &
            \includegraphics[width=\linewidth]{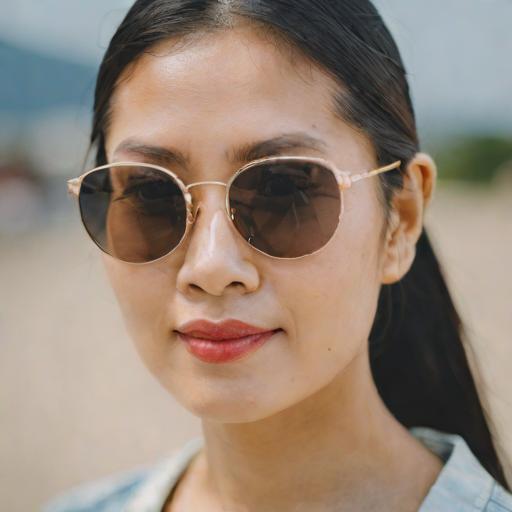} &
            \includegraphics[width=\linewidth]{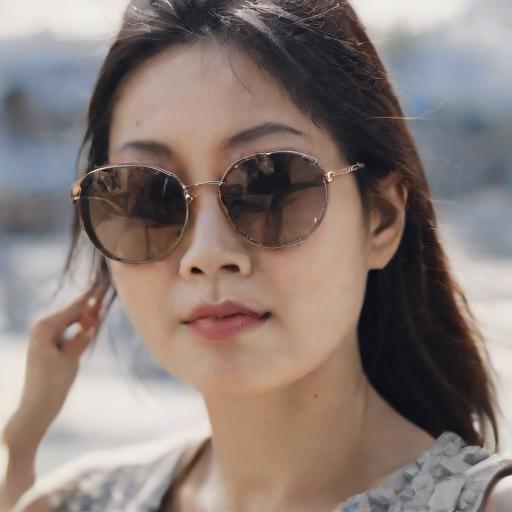} &
            \includegraphics[width=\linewidth]{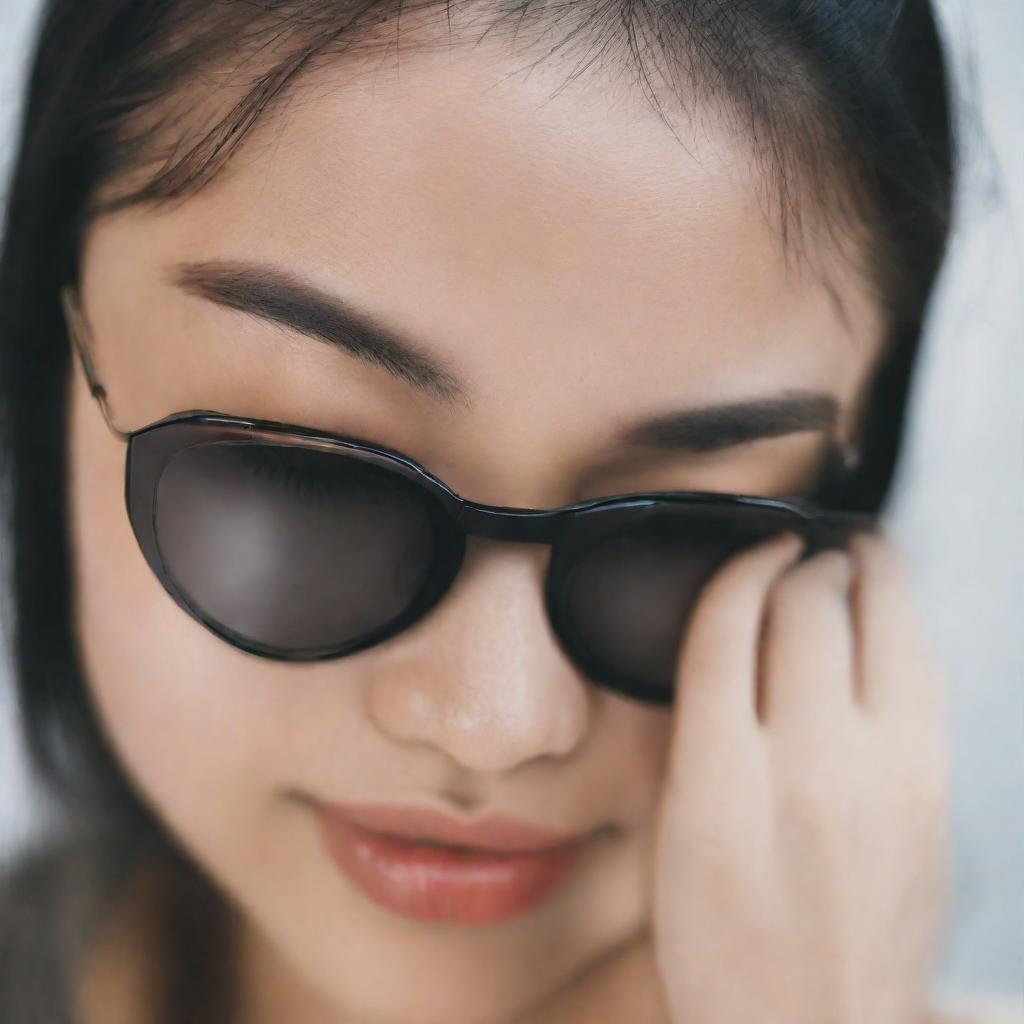} &
            \includegraphics[width=\linewidth]{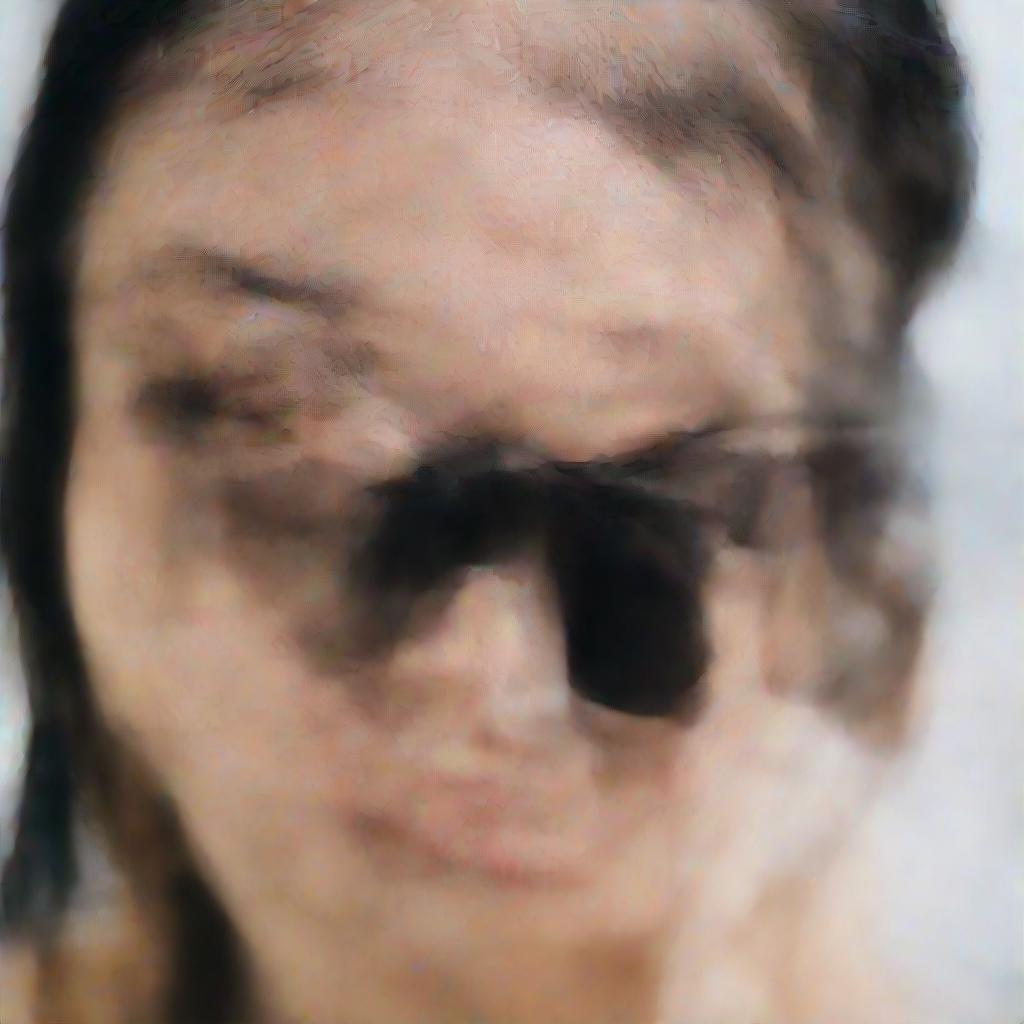}
        \end{tabularx}
        \vspace{-6pt}
        \caption{A close-up of an Asian lady with sunglasses.}
    \end{subfigure}

    \begin{subfigure}[b]{\textwidth}
        \centering
        \setlength\tabcolsep{1pt}
        \begin{tabularx}{\textwidth}{@{}XX@{}X@{}X@{}XX@{}XX@{}X@{}}
            \includegraphics[width=\linewidth]{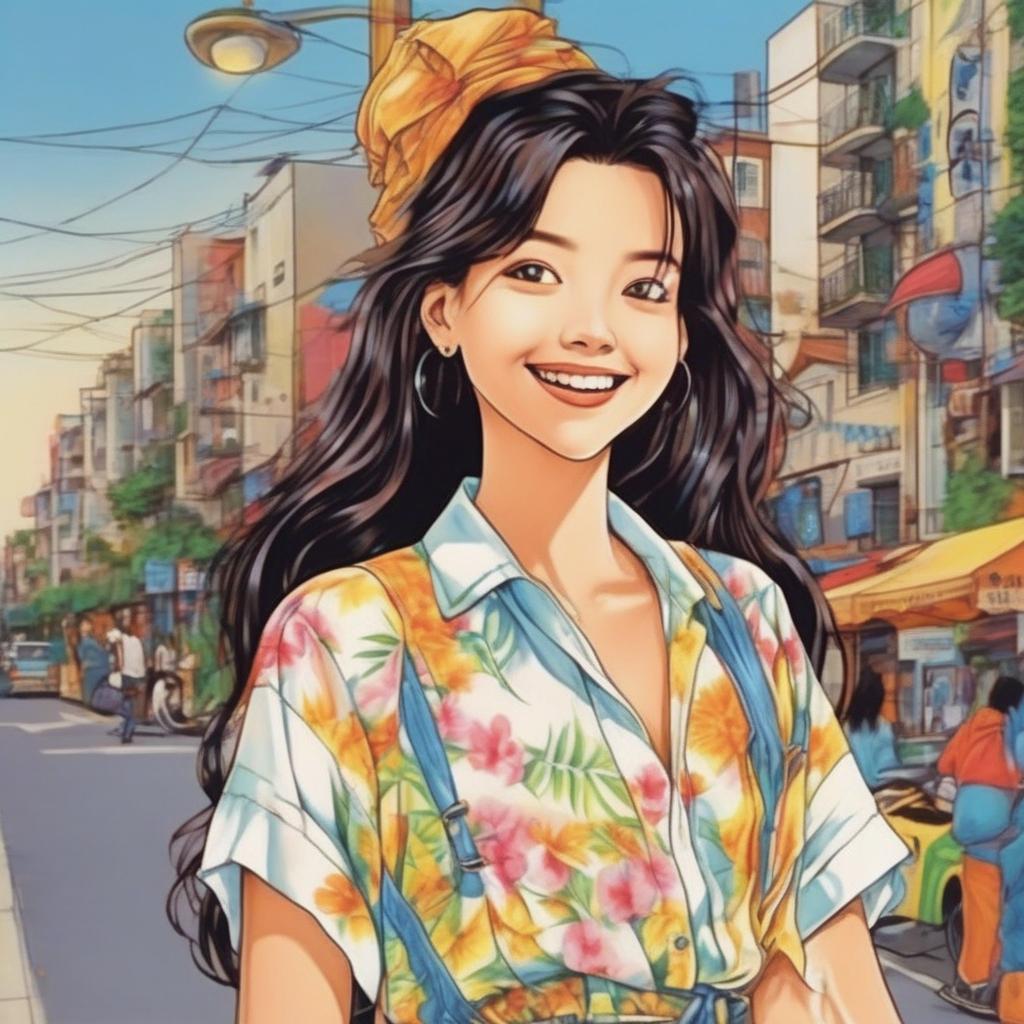} &
            \includegraphics[width=\linewidth]{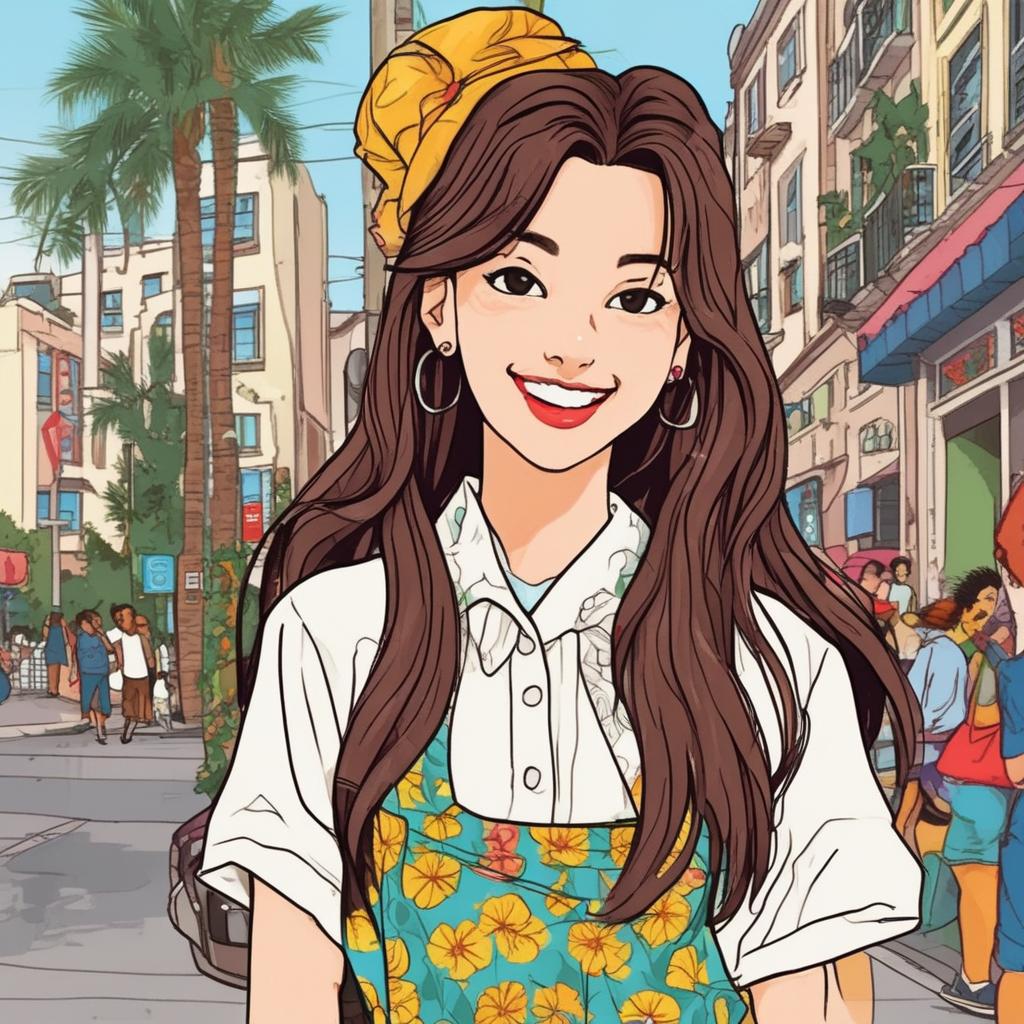} &
            \includegraphics[width=\linewidth]{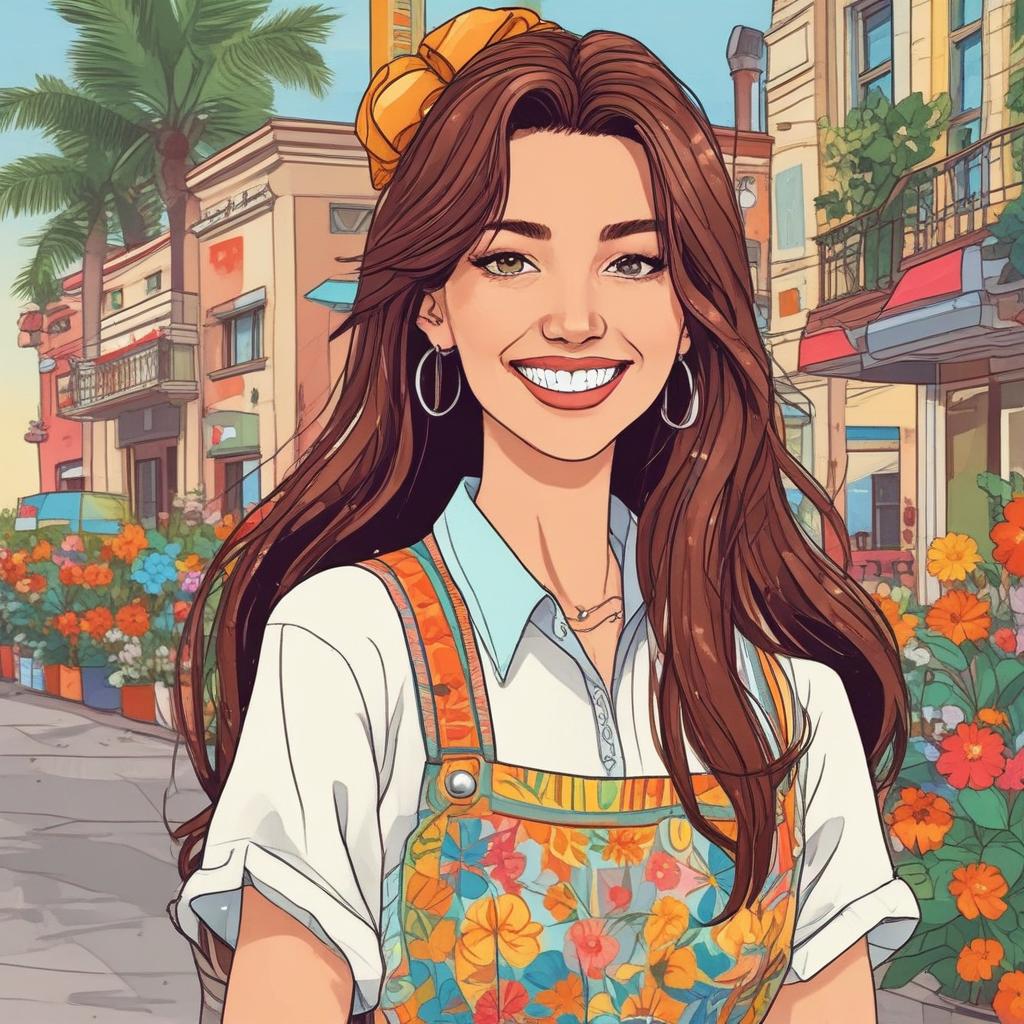} &
            \includegraphics[width=\linewidth]{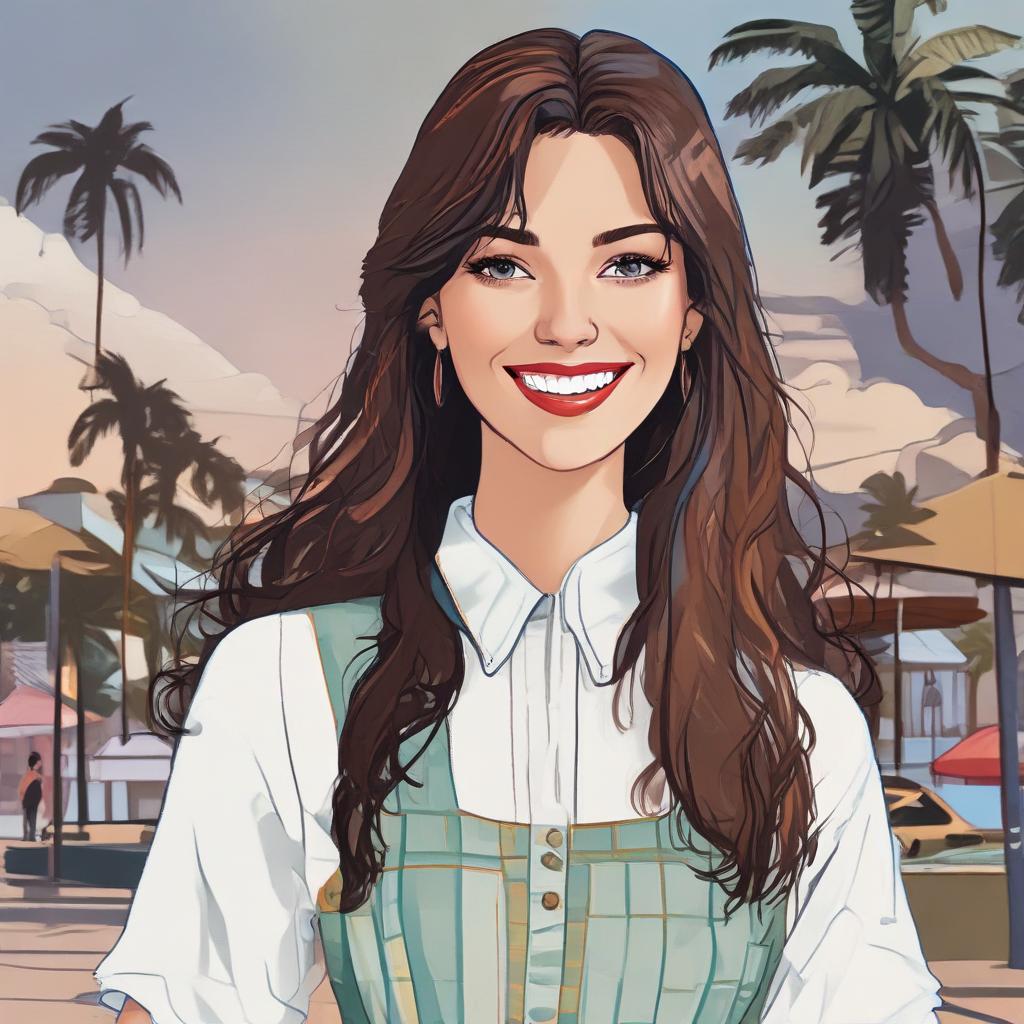} &
            \includegraphics[width=\linewidth]{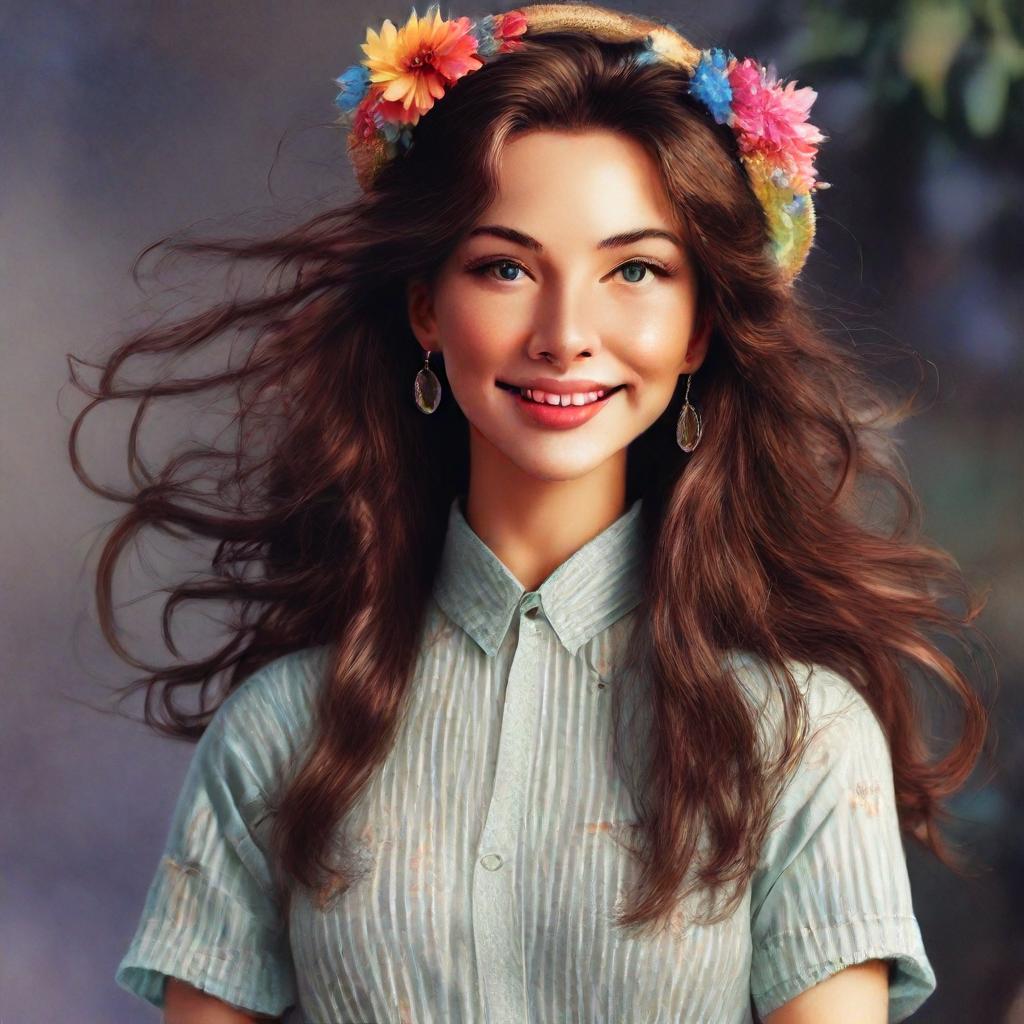} &
            \includegraphics[width=\linewidth]{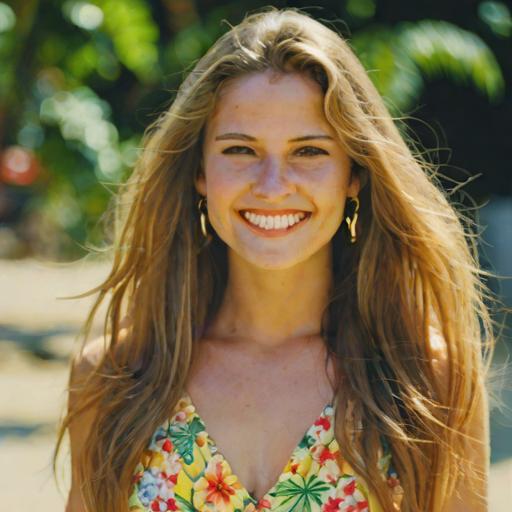} &
            \includegraphics[width=\linewidth]{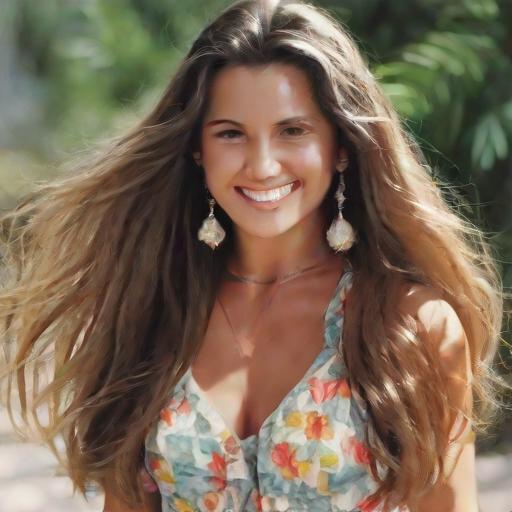} &
            \includegraphics[width=\linewidth]{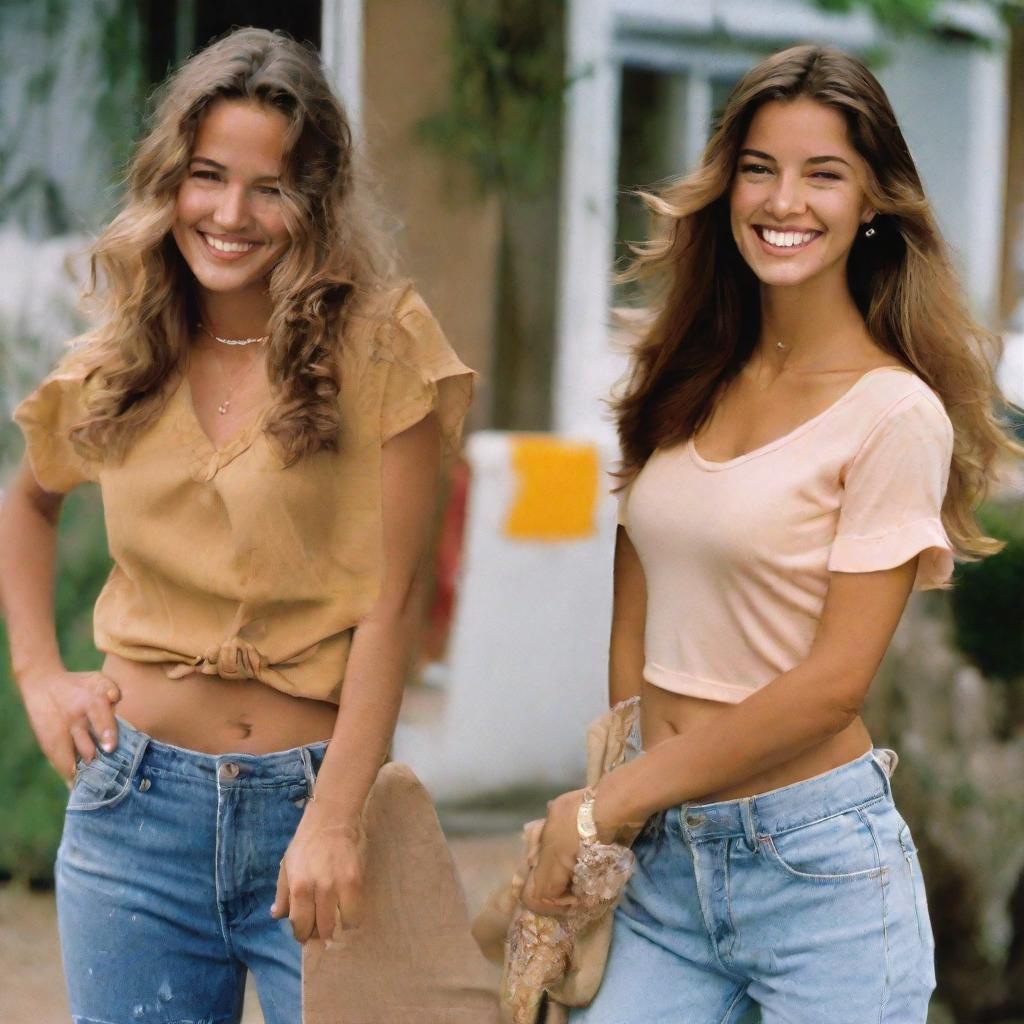} &
            \includegraphics[width=\linewidth]{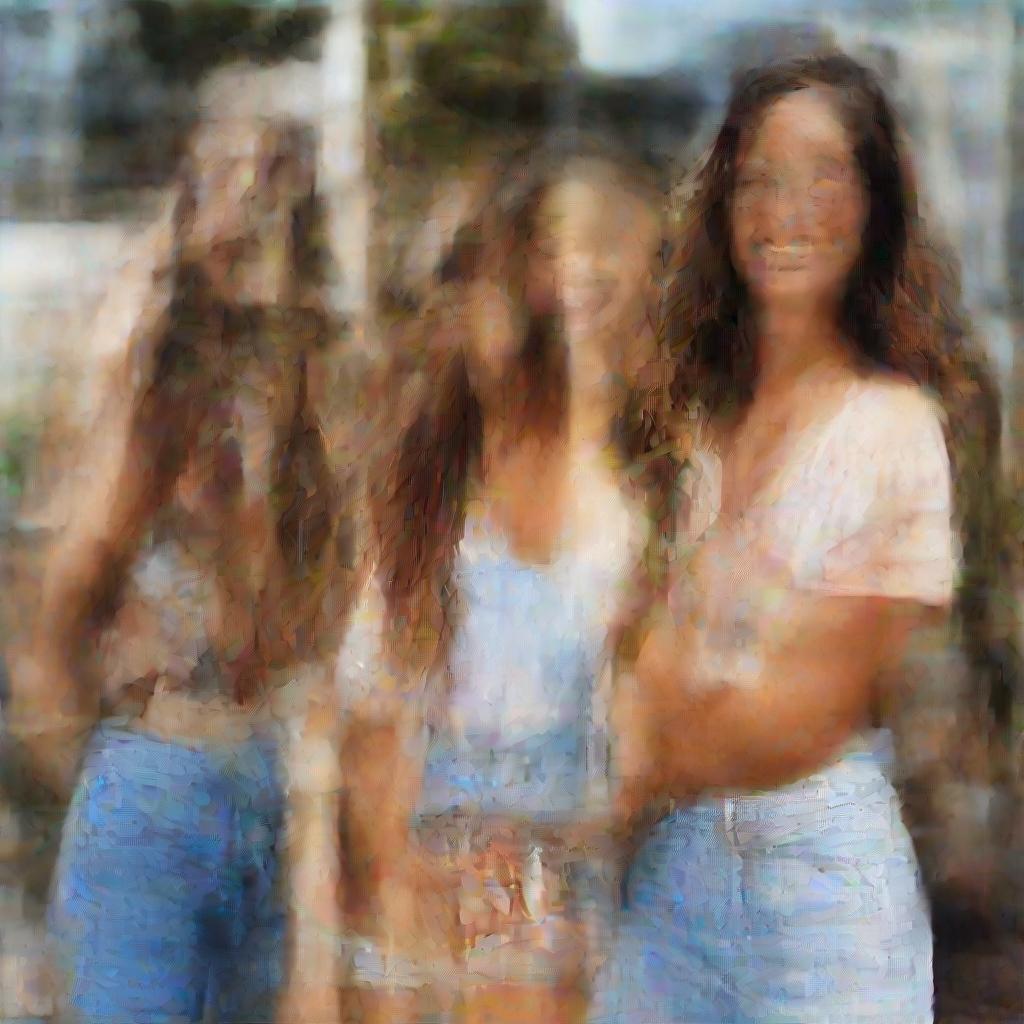}
        \end{tabularx}
        \vspace{-6pt}
        \caption{The 90s, a beautiful woman with a radiant smile and long hair, dressed in summer attire.}
    \end{subfigure}

    \begin{subfigure}[b]{\textwidth}
        \centering
        \setlength\tabcolsep{1pt}
        \begin{tabularx}{\textwidth}{@{}XX@{}X@{}X@{}XX@{}XX@{}X@{}}
            \includegraphics[width=\linewidth]{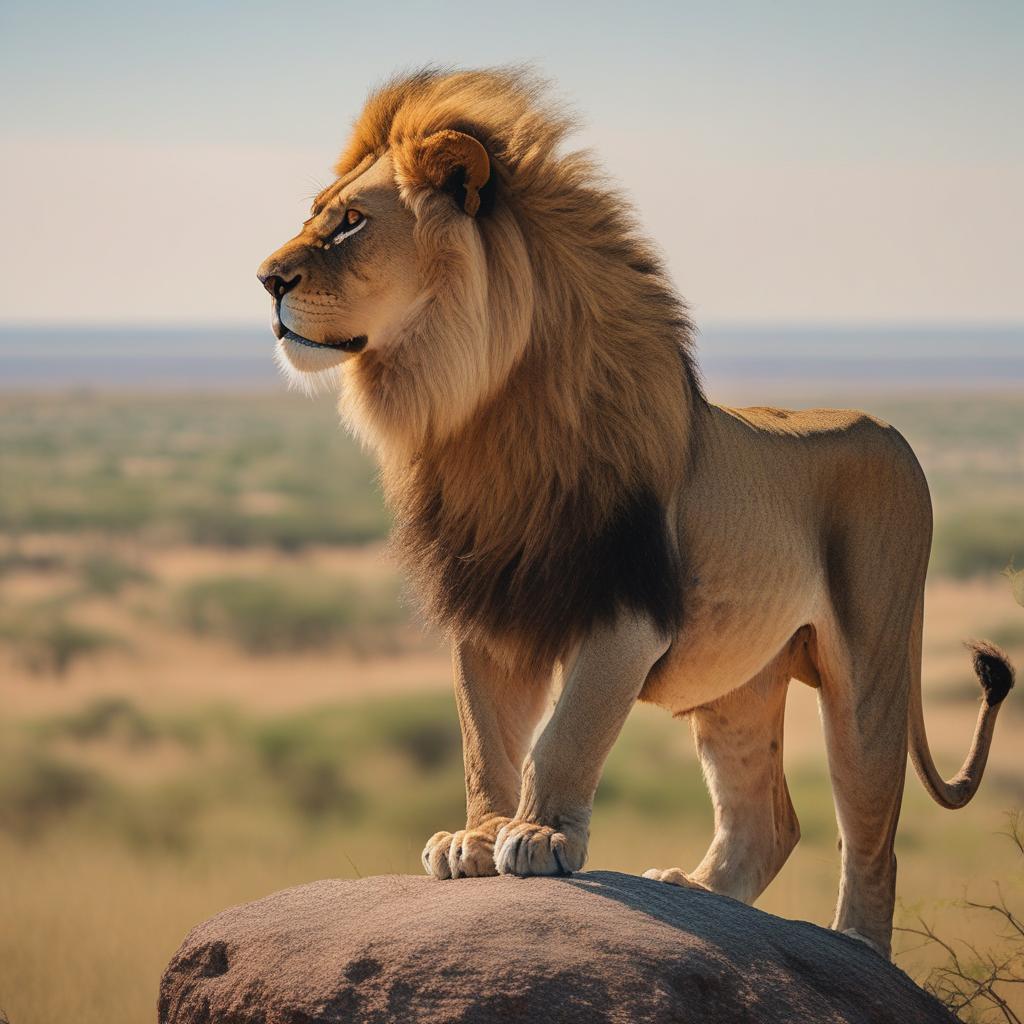} &
            \includegraphics[width=\linewidth]{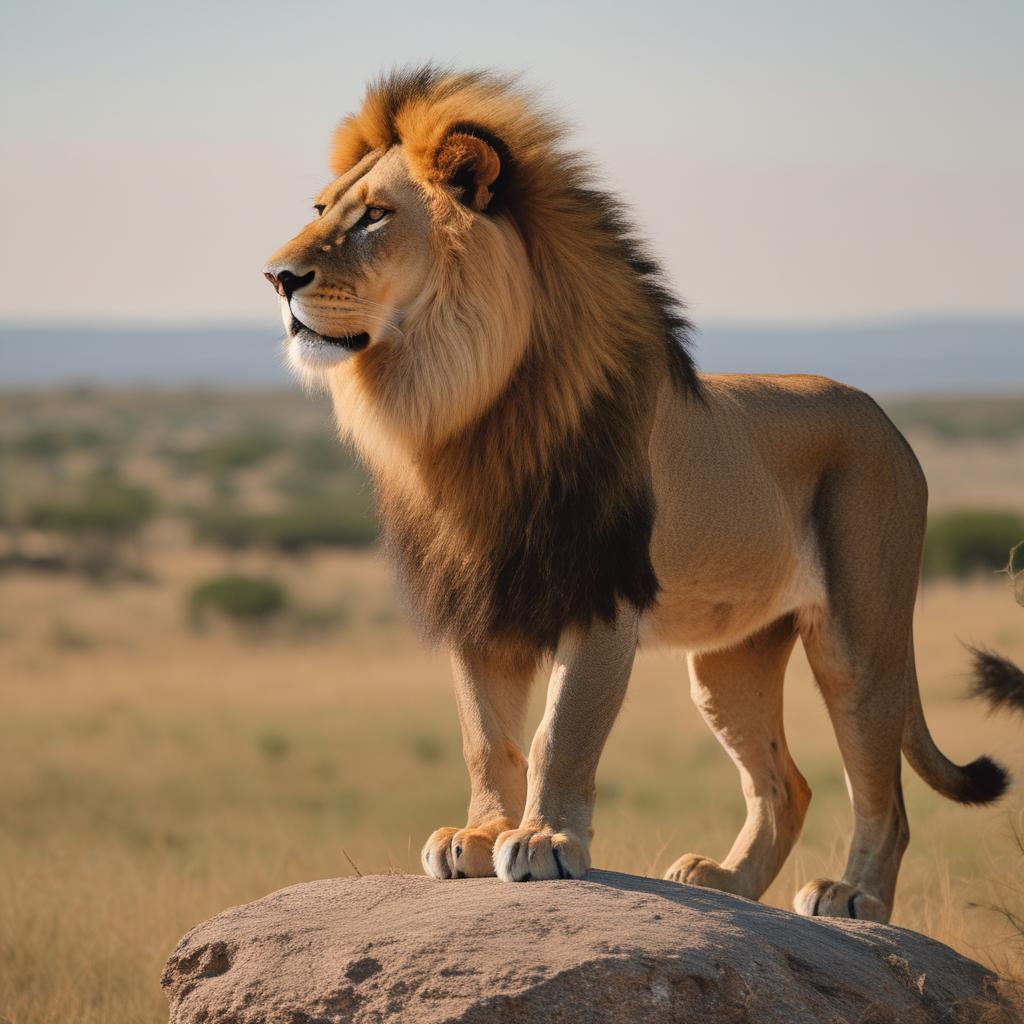} &
            \includegraphics[width=\linewidth]{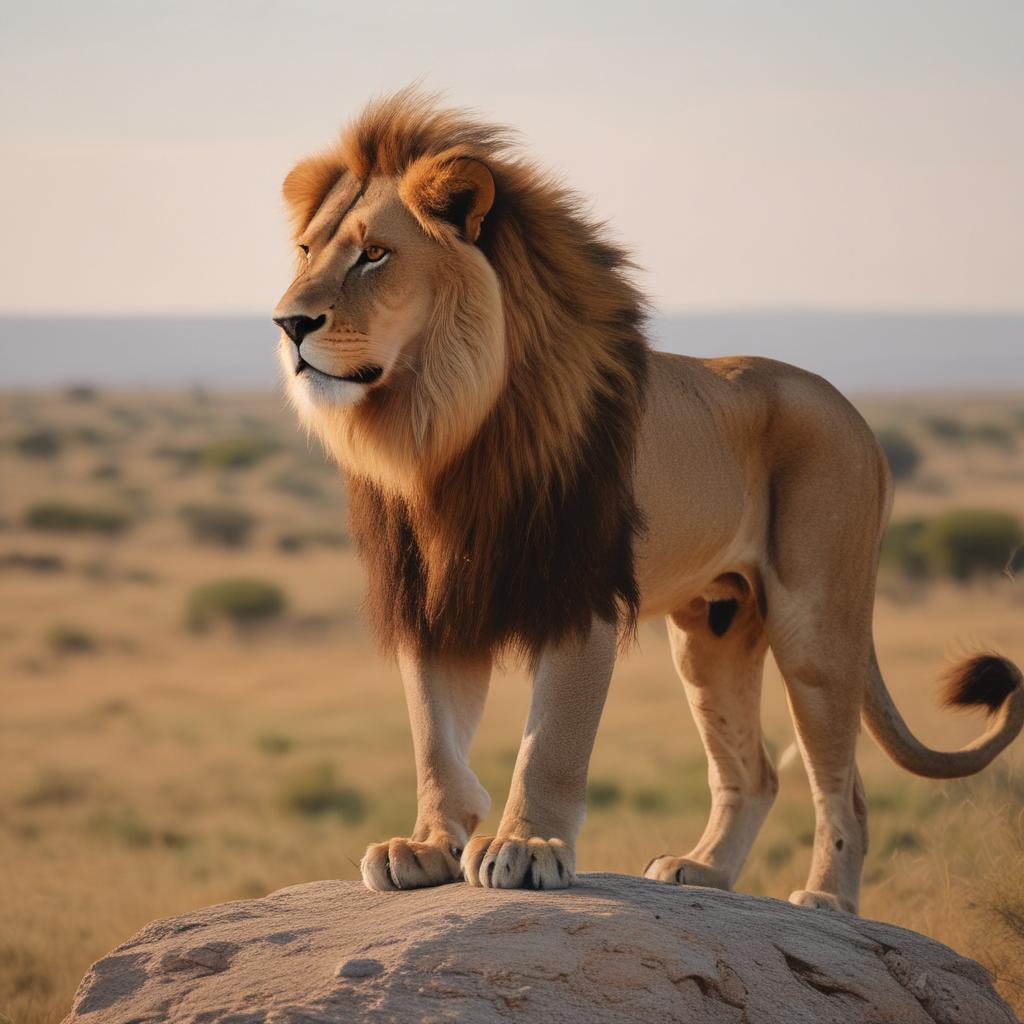} &
            \includegraphics[width=\linewidth]{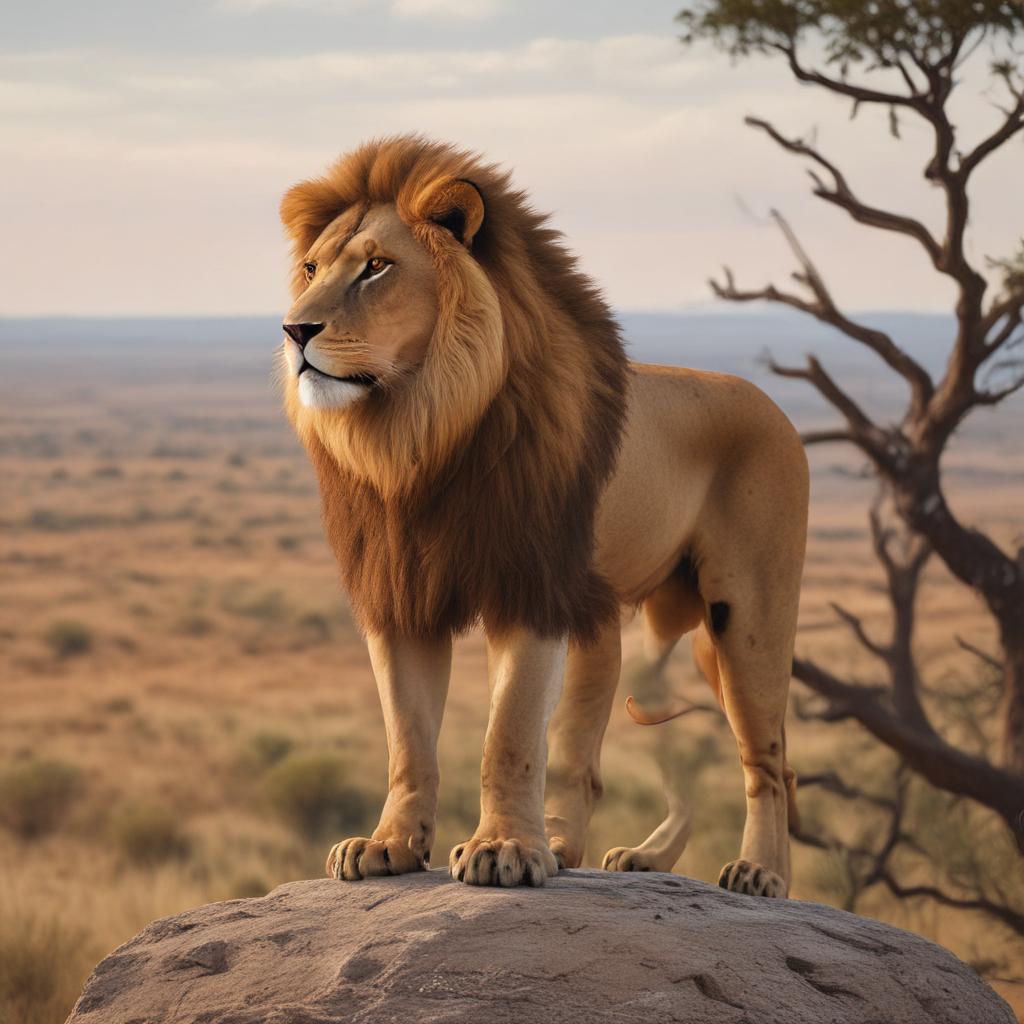} &
            \includegraphics[width=\linewidth]{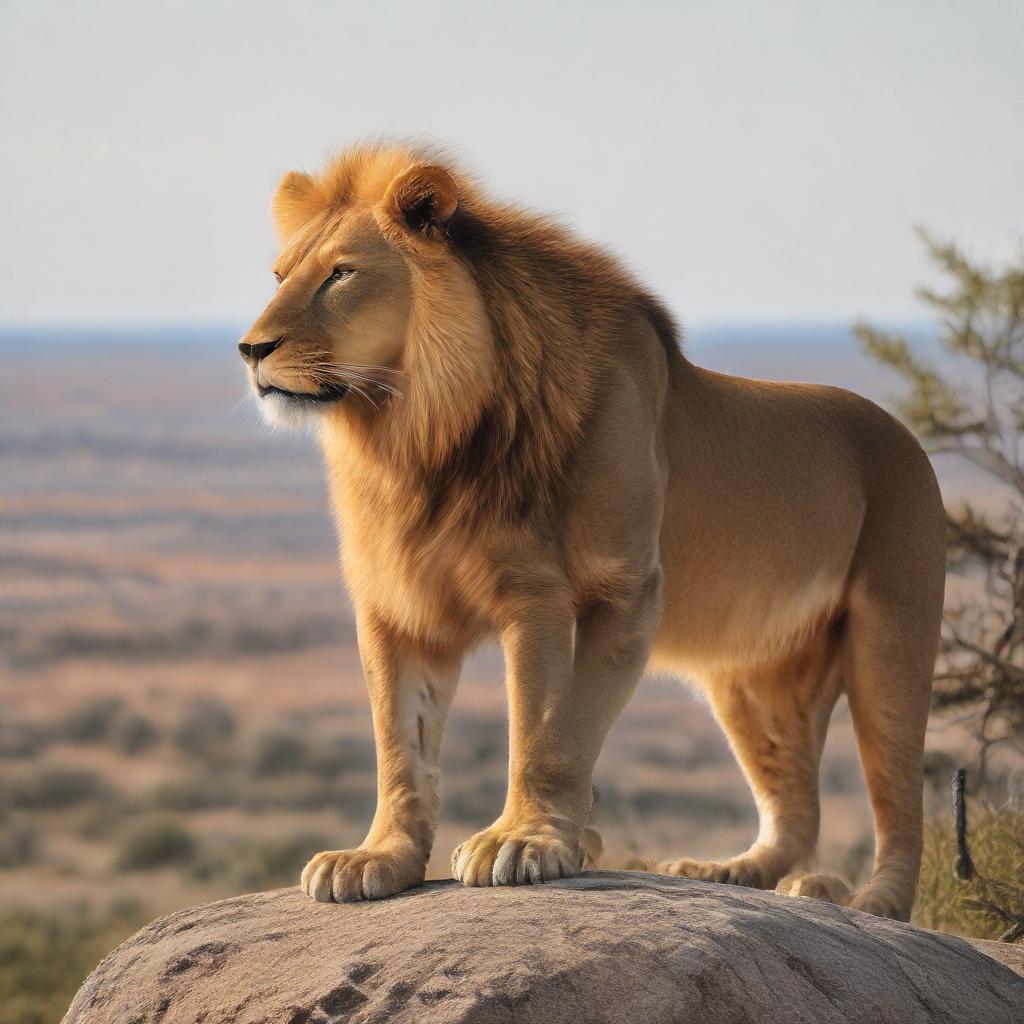} &
            \includegraphics[width=\linewidth]{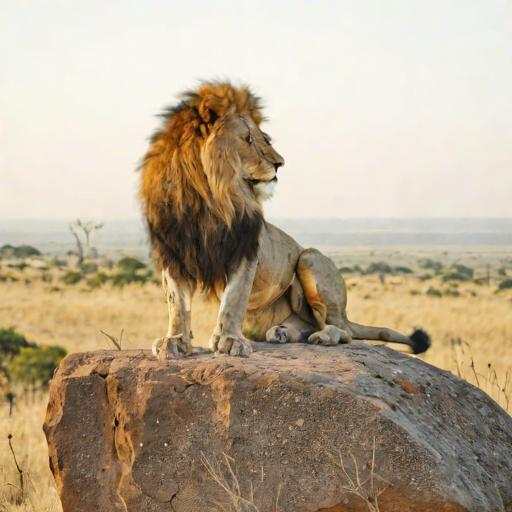} &
            \includegraphics[width=\linewidth]{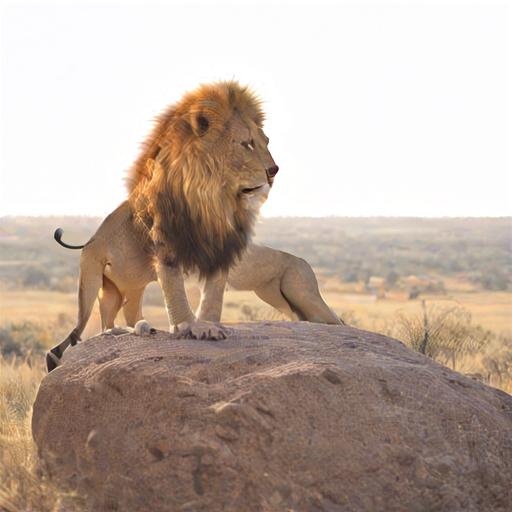} &
            \includegraphics[width=\linewidth]{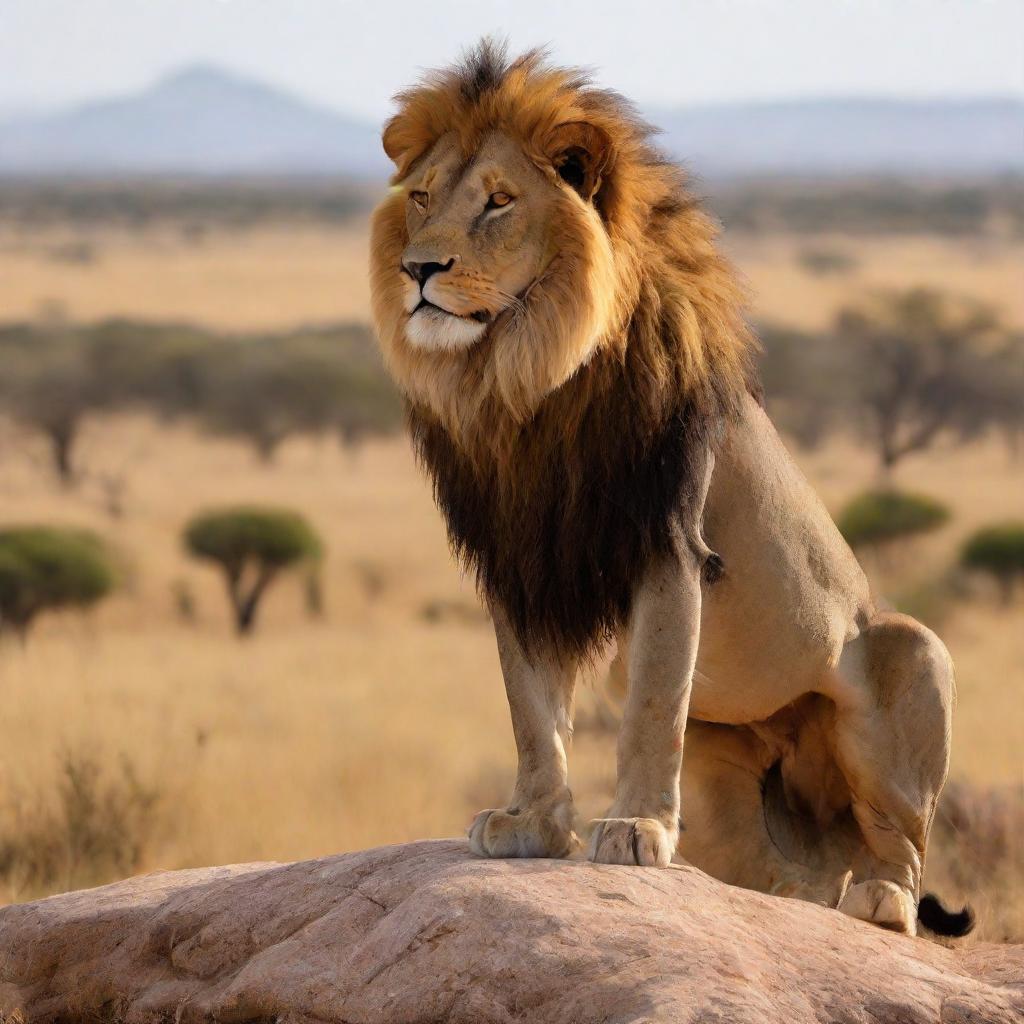} &
            \includegraphics[width=\linewidth]{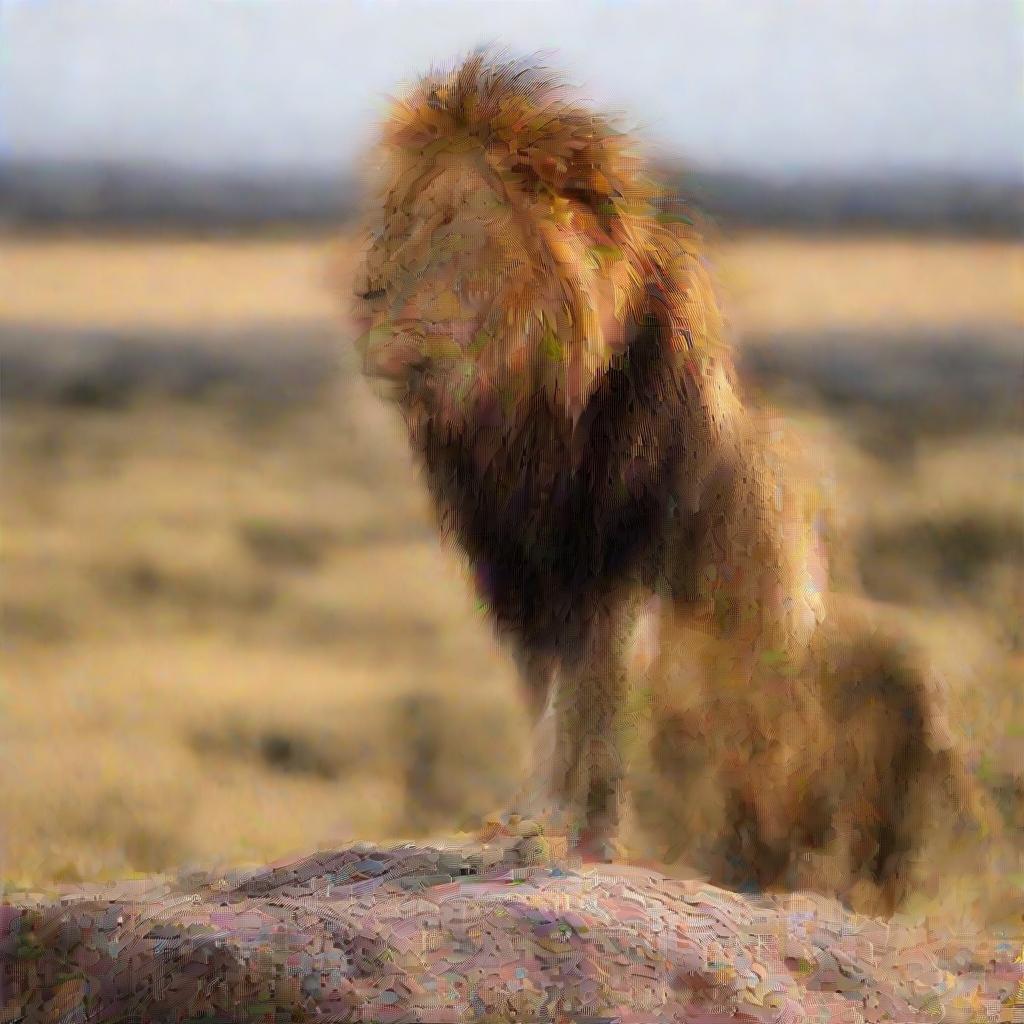}
        \end{tabularx}
        \vspace{-6pt}
        \caption{A majestic lion stands proudly on a rock, overlooking the vast African savannah.}
    \end{subfigure}

    \begin{subfigure}[b]{\textwidth}
        \centering
        \setlength\tabcolsep{1pt}
        \begin{tabularx}{\textwidth}{@{}XX@{}X@{}X@{}XX@{}XX@{}X@{}}
            \includegraphics[width=\linewidth]{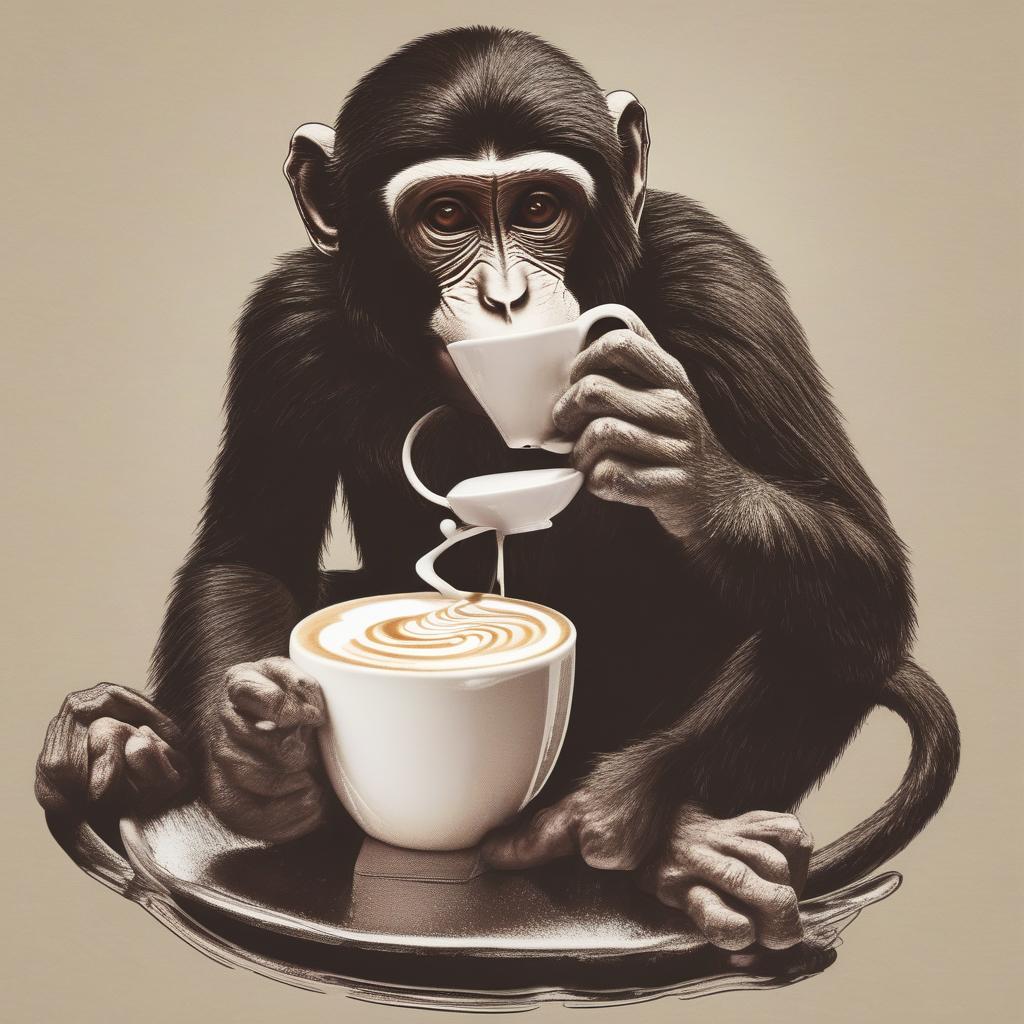} &
            \includegraphics[width=\linewidth]{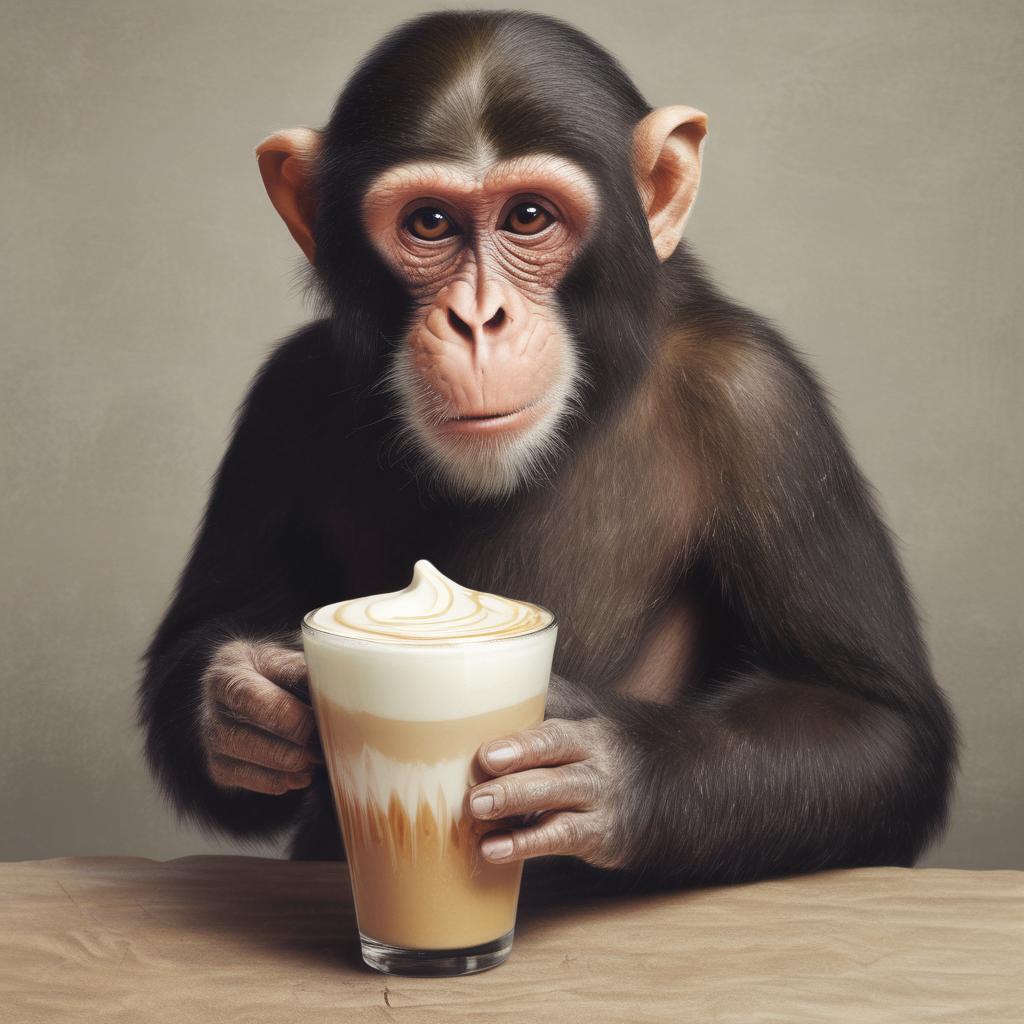} &
            \includegraphics[width=\linewidth]{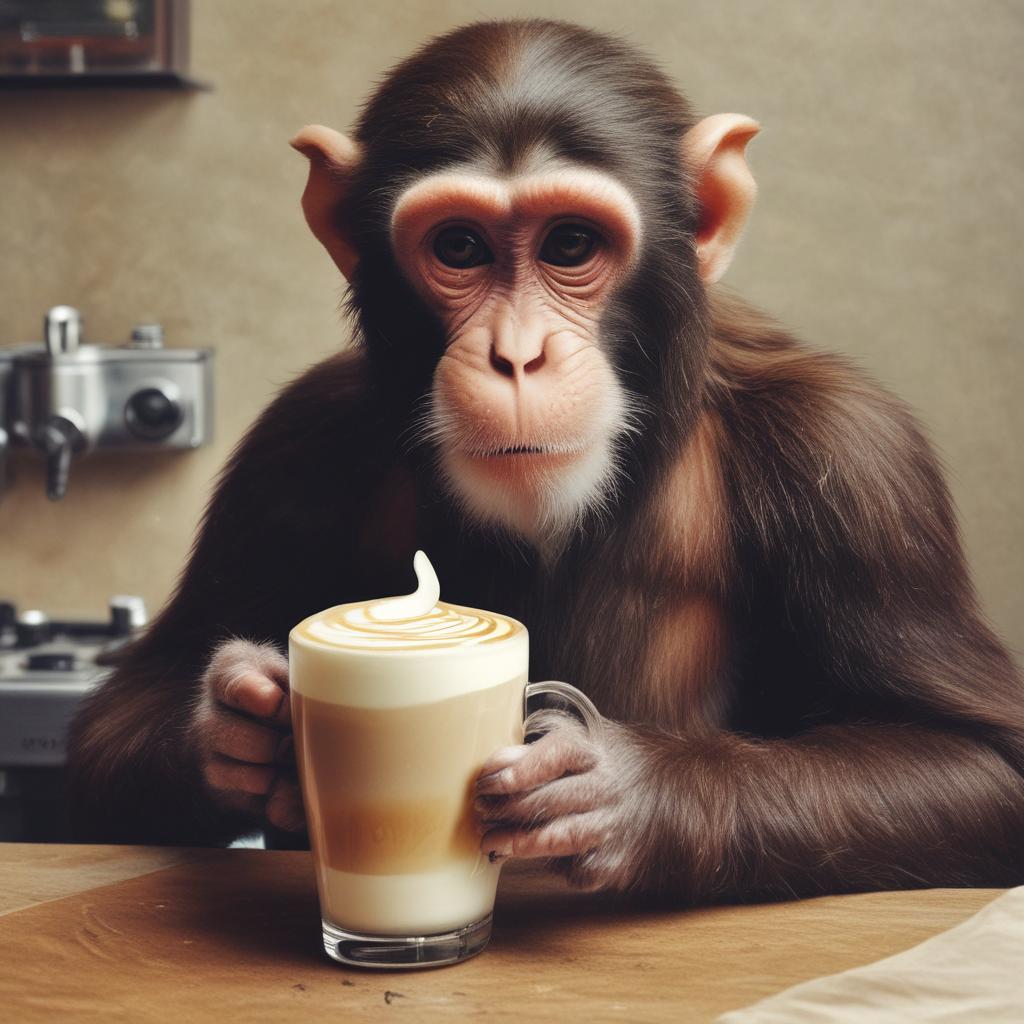} &
            \includegraphics[width=\linewidth]{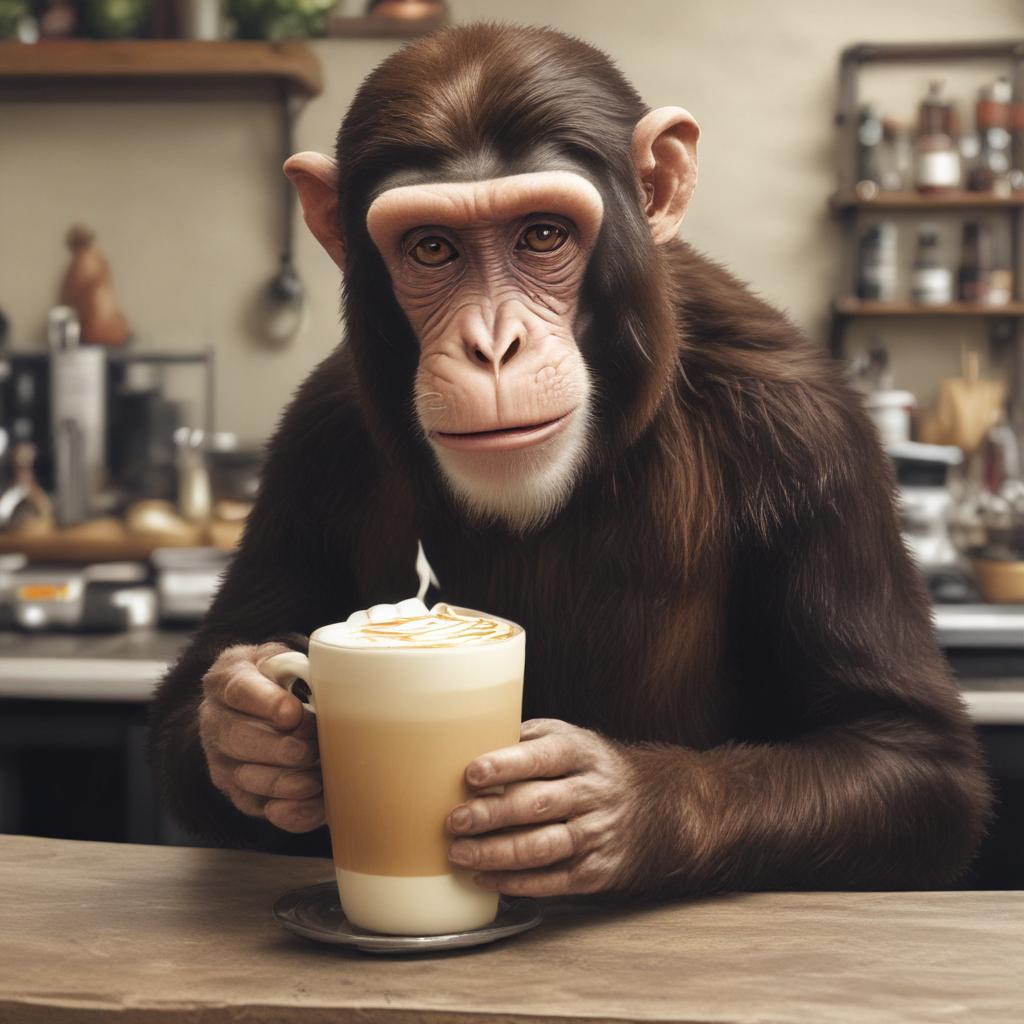} &
            \includegraphics[width=\linewidth]{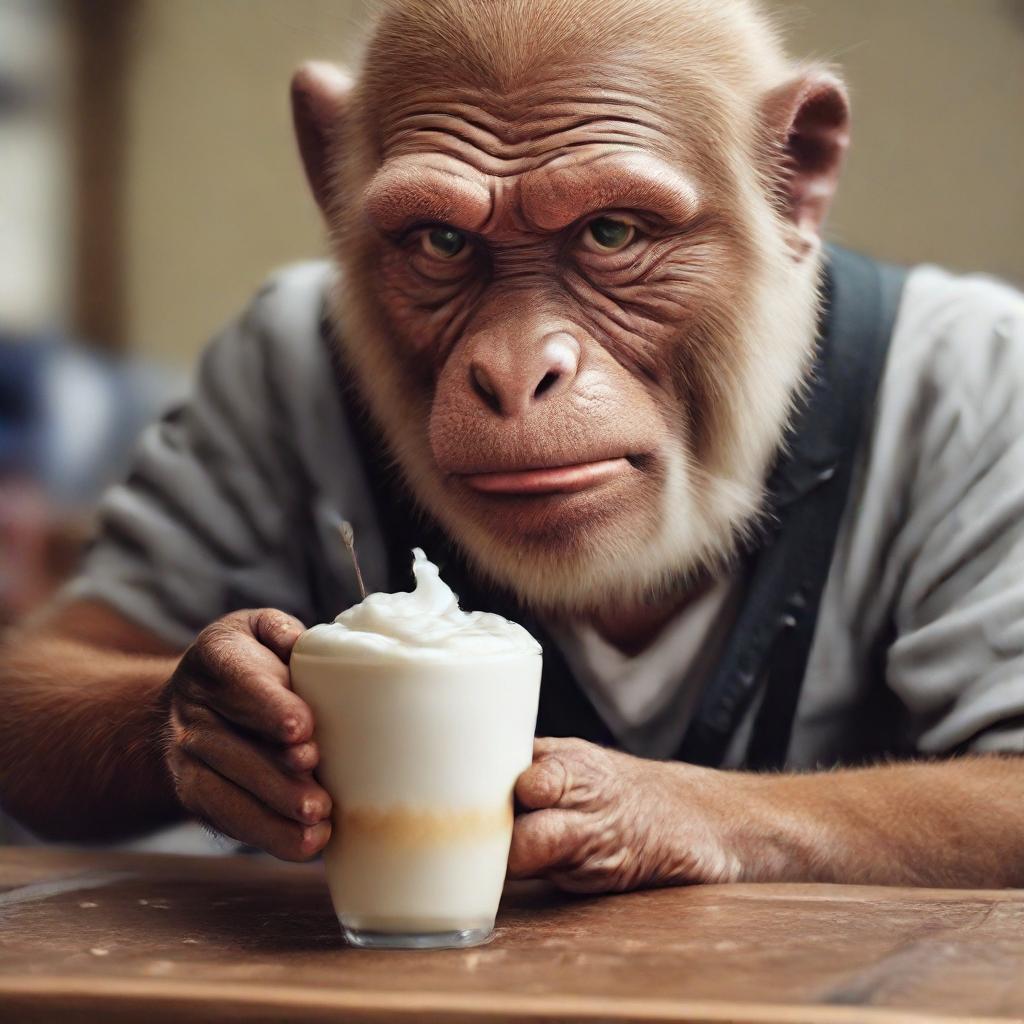} &
            \includegraphics[width=\linewidth]{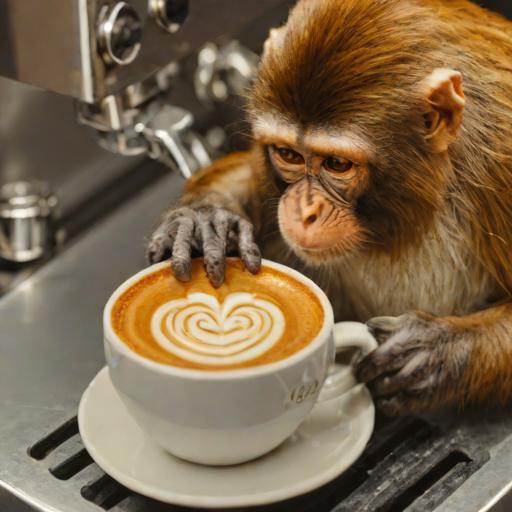} &
            \includegraphics[width=\linewidth]{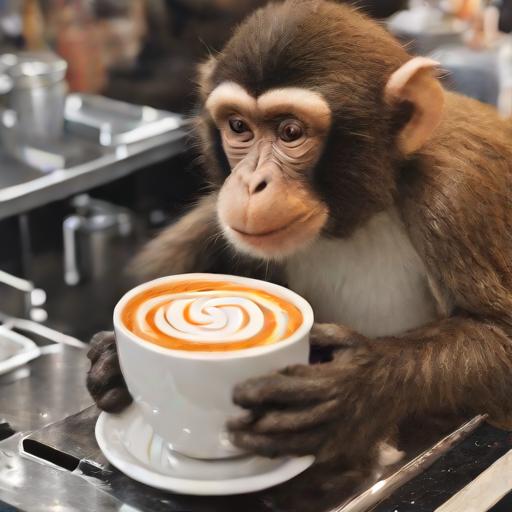} &
            \includegraphics[width=\linewidth]{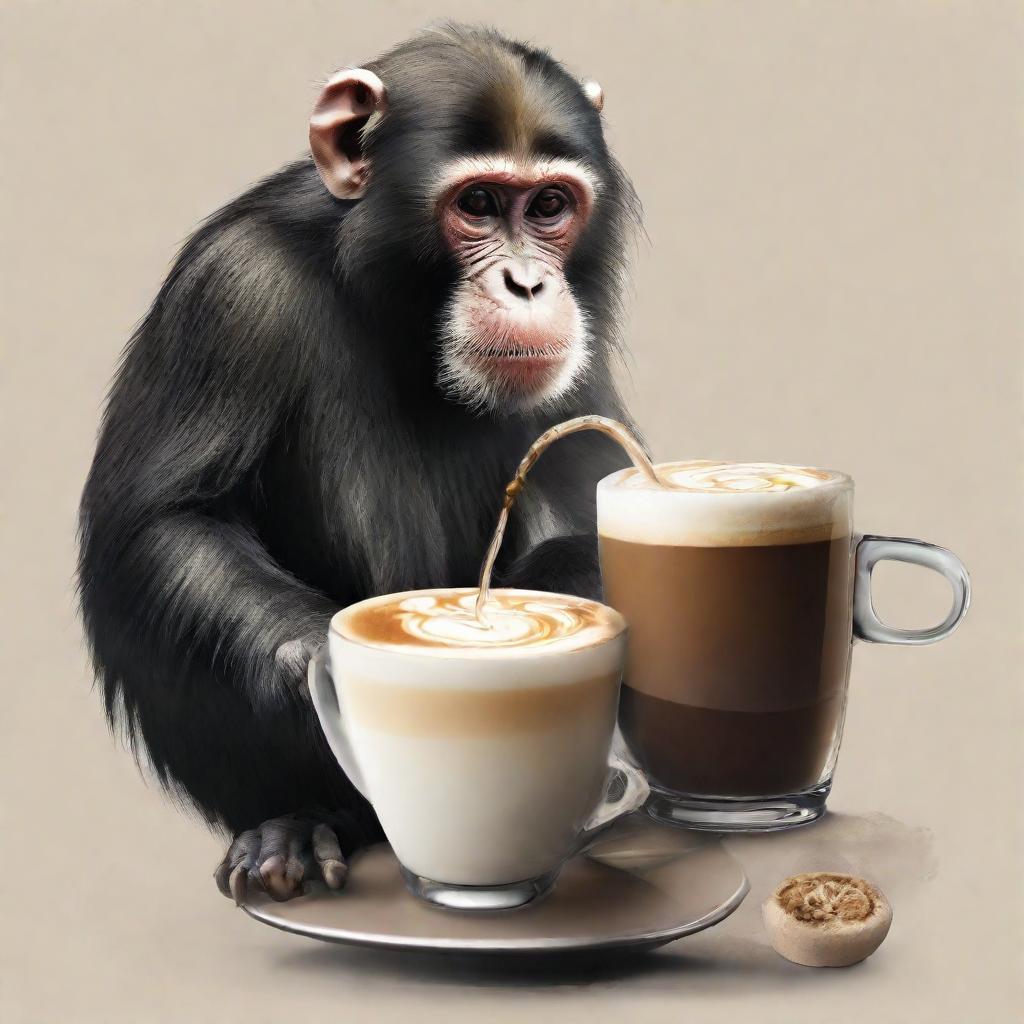} &
            \includegraphics[width=\linewidth]{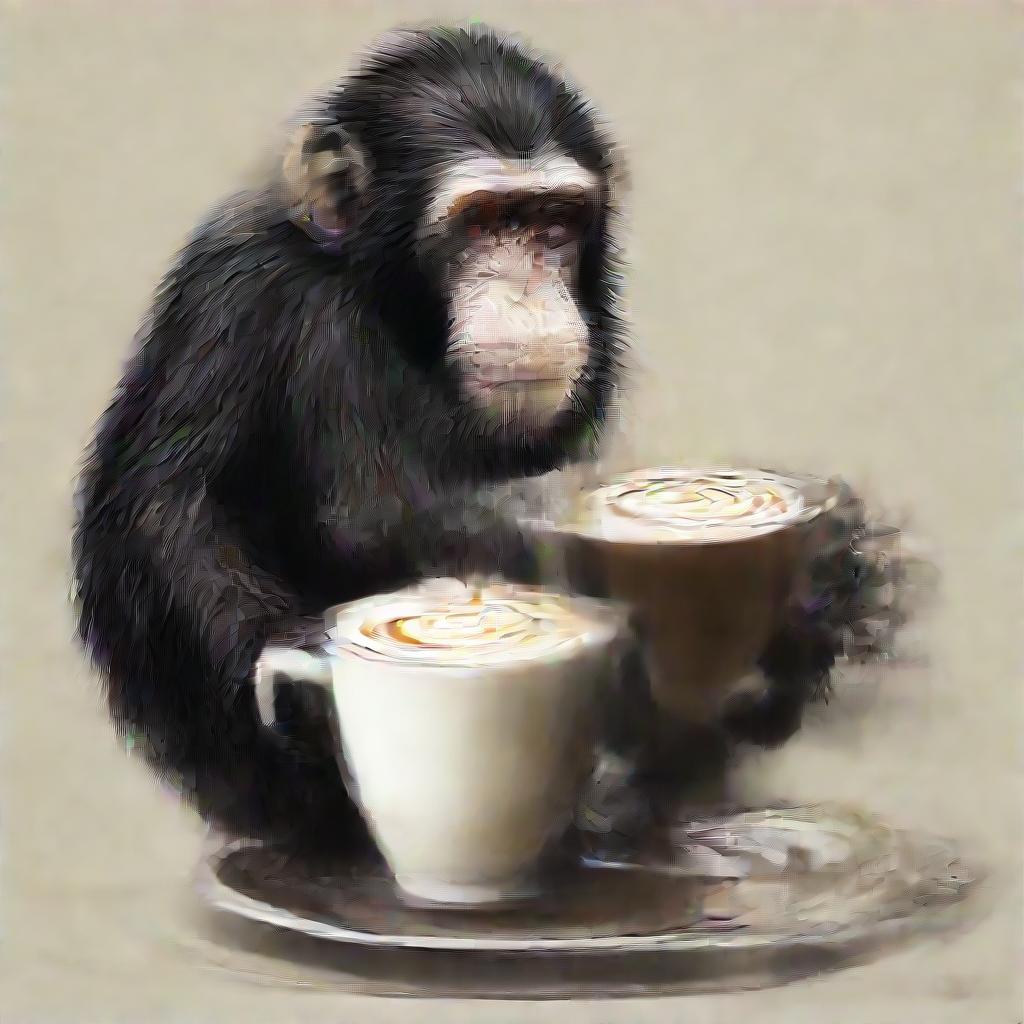}
        \end{tabularx}
        \vspace{-6pt}
        \caption{A monkey making latte art.}
    \end{subfigure}

    \begin{subfigure}[b]{\textwidth}
        \centering
        \setlength\tabcolsep{1pt}
        \begin{tabularx}{\textwidth}{@{}XX@{}X@{}X@{}XX@{}XX@{}X@{}}
            \includegraphics[width=\linewidth]{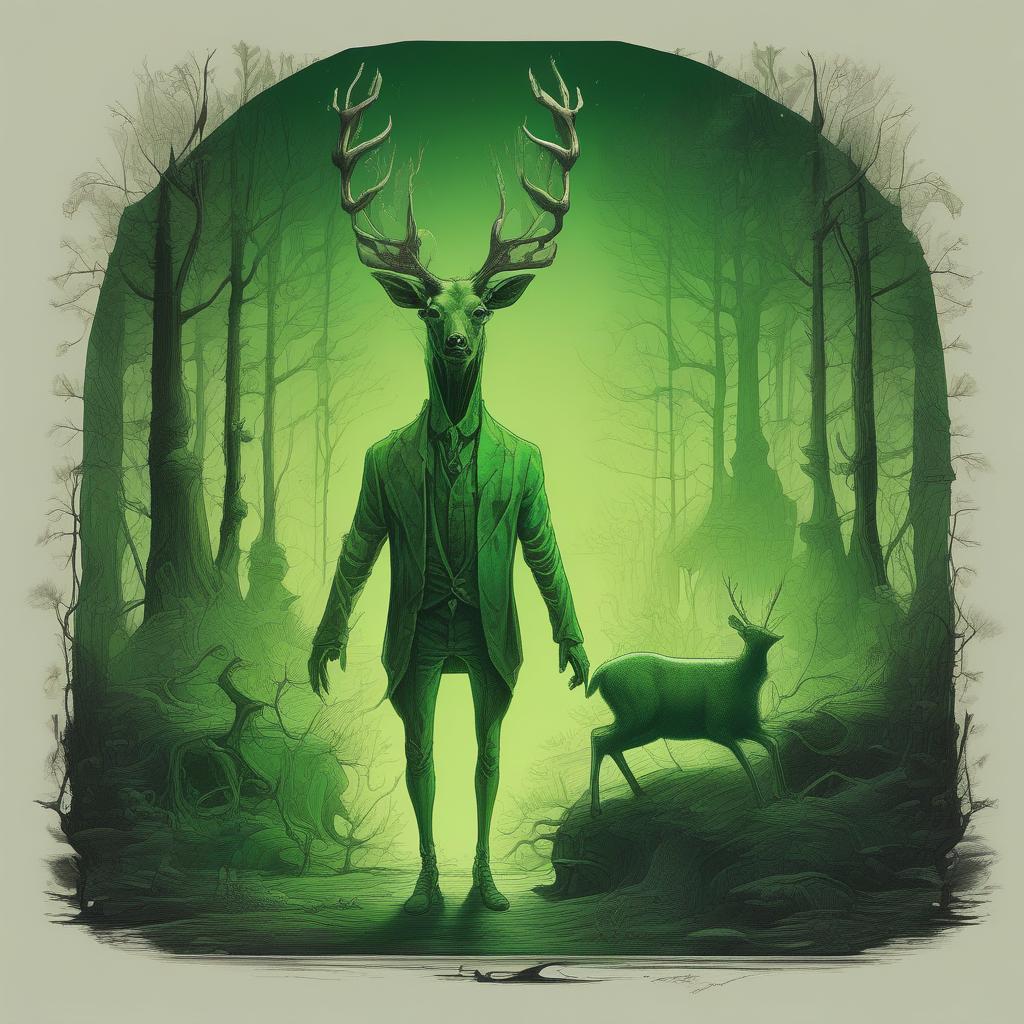} &
            \includegraphics[width=\linewidth]{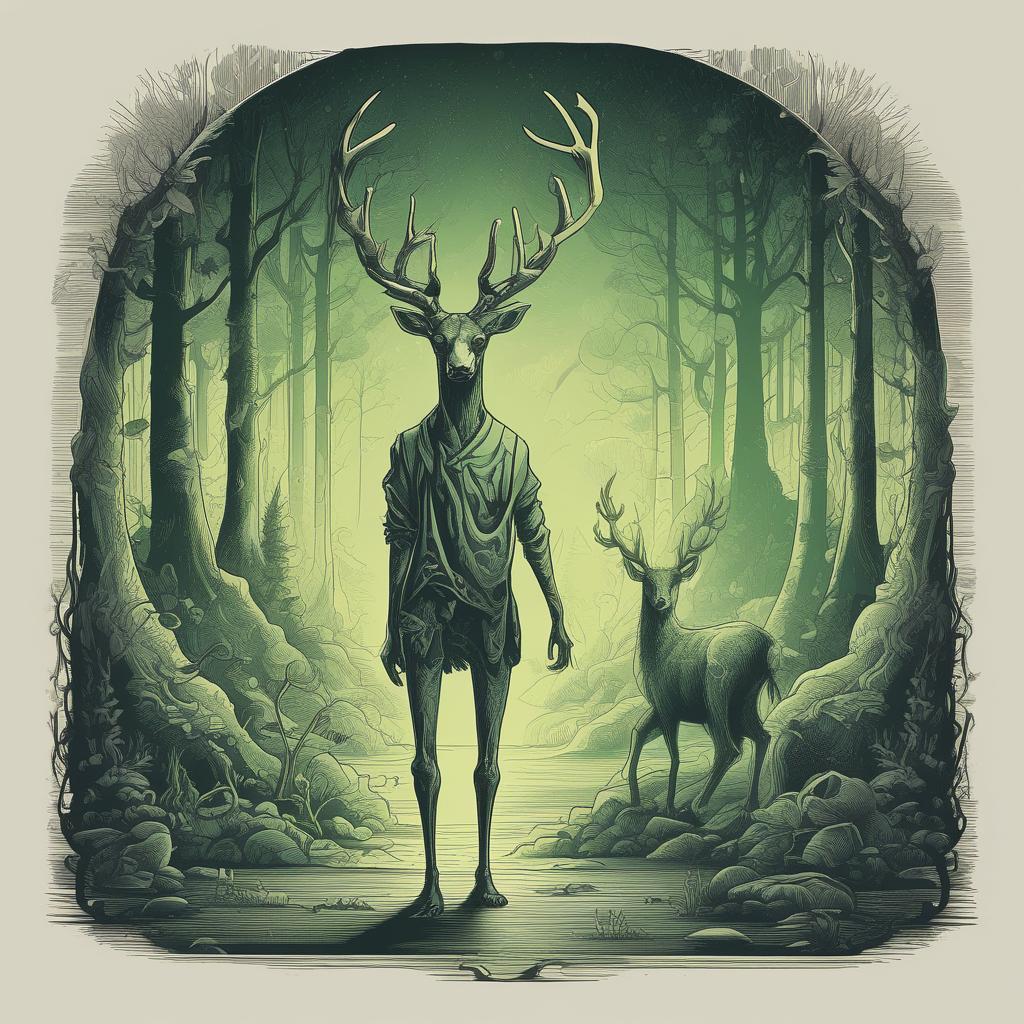} &
            \includegraphics[width=\linewidth]{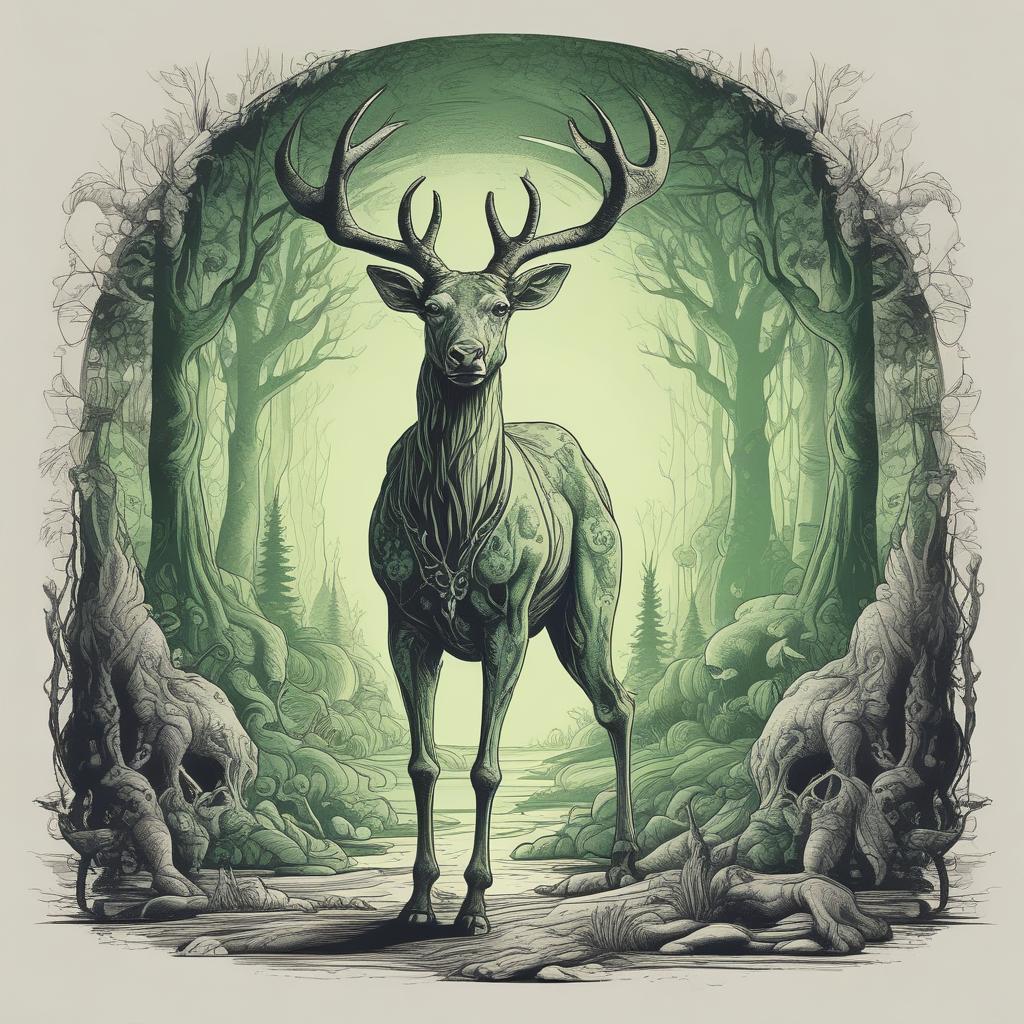} &
            \includegraphics[width=\linewidth]{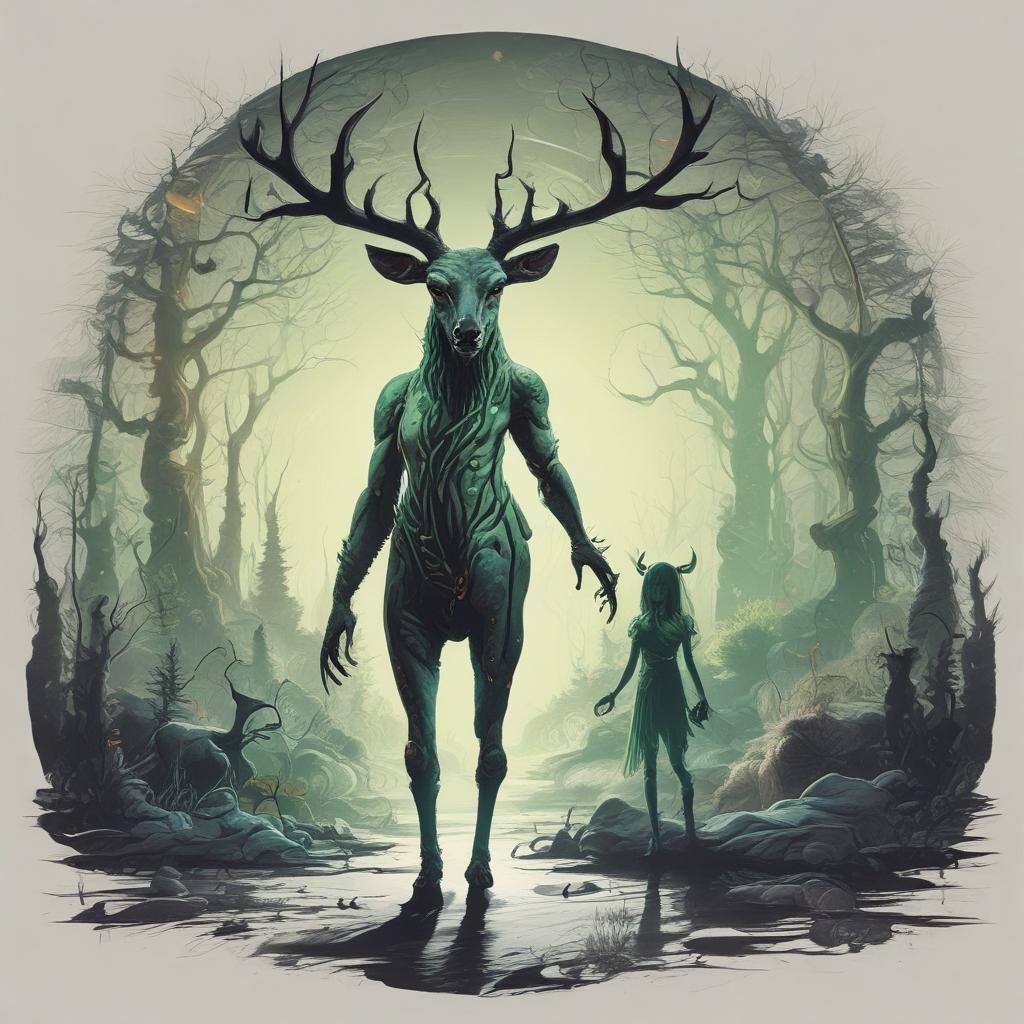} &
            \includegraphics[width=\linewidth]{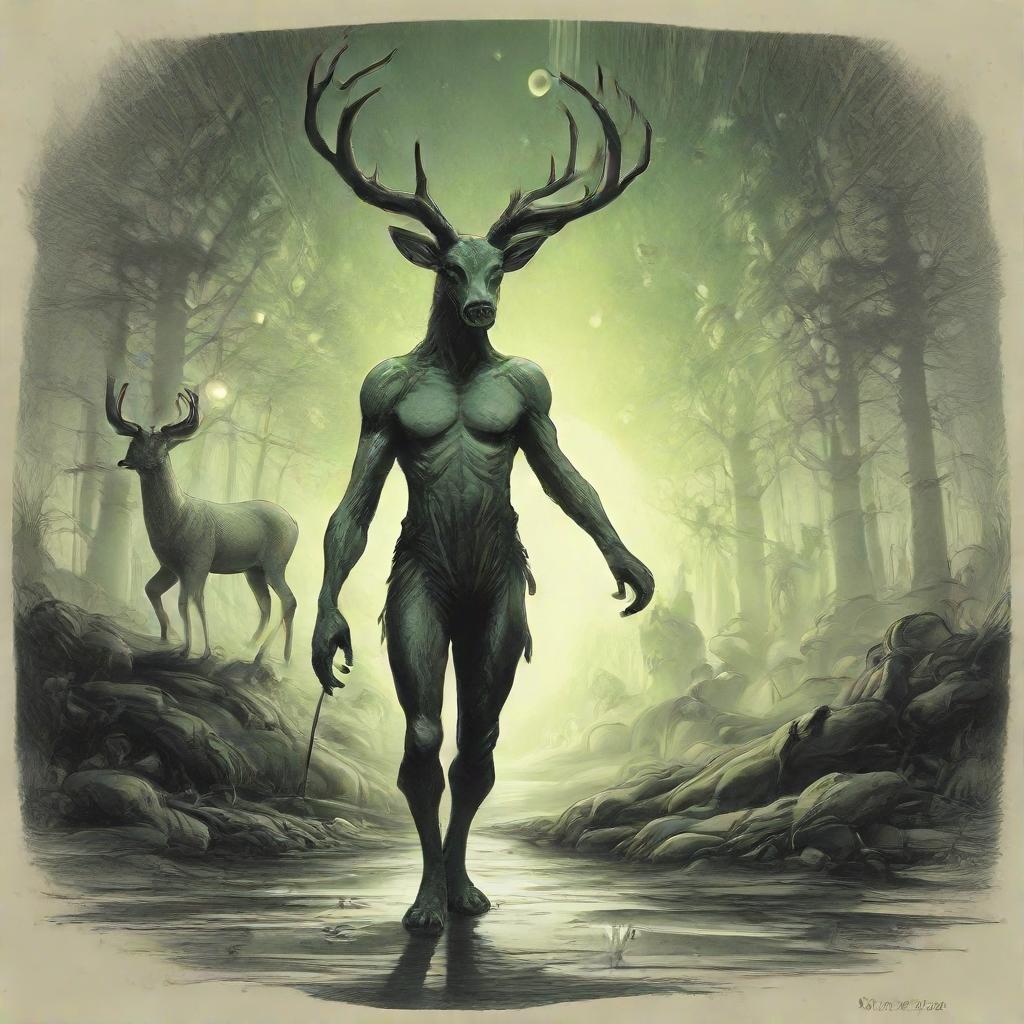} &
            \includegraphics[width=\linewidth]{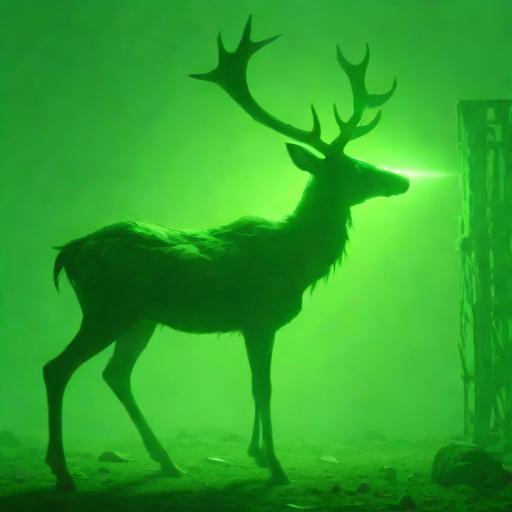} &
            \includegraphics[width=\linewidth]{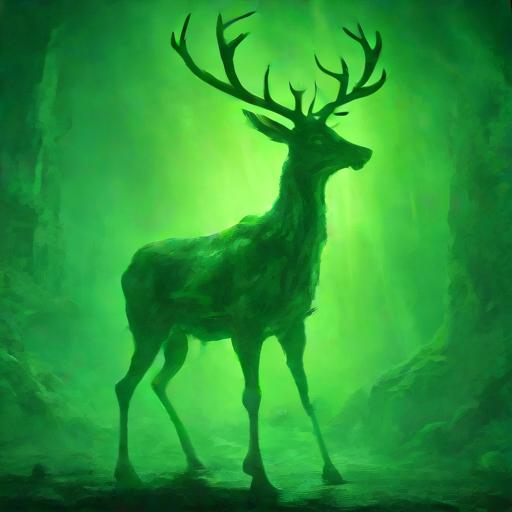} &
            \includegraphics[width=\linewidth]{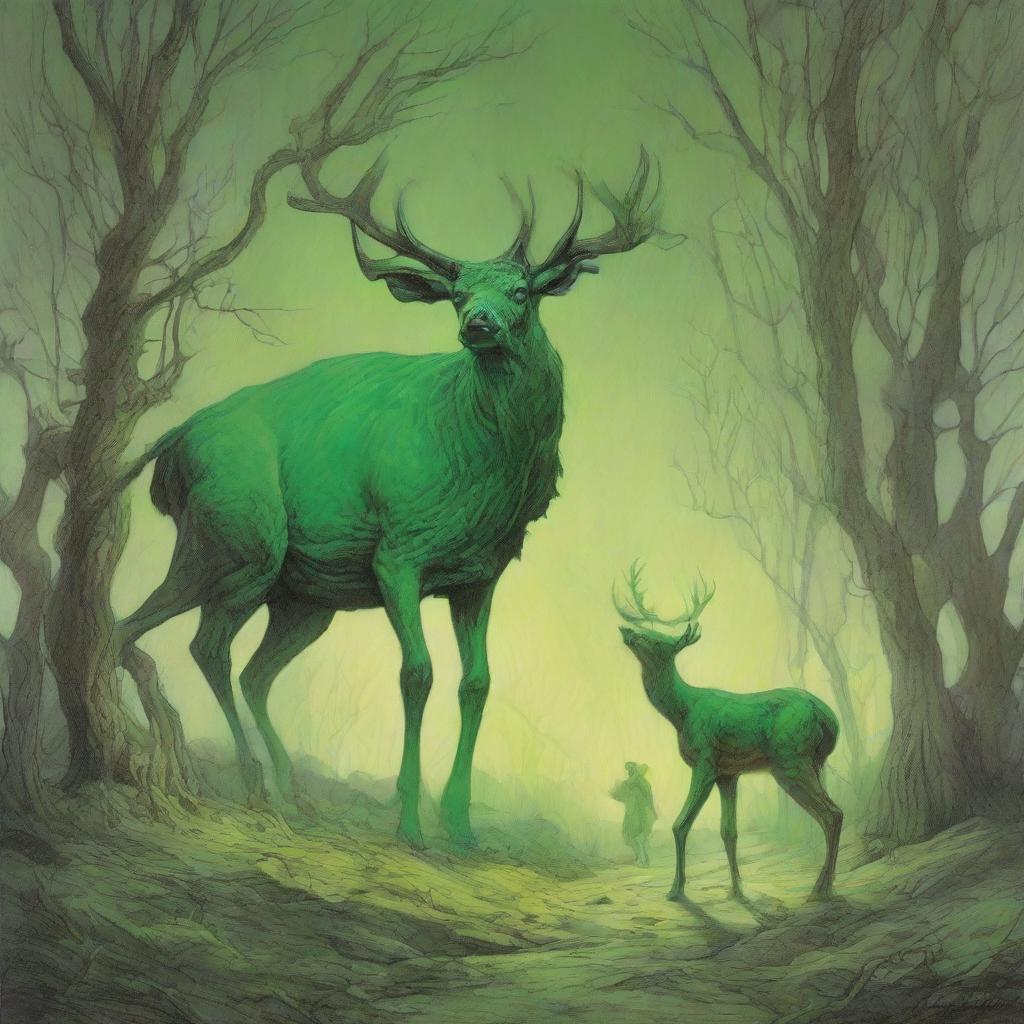} &
            \includegraphics[width=\linewidth]{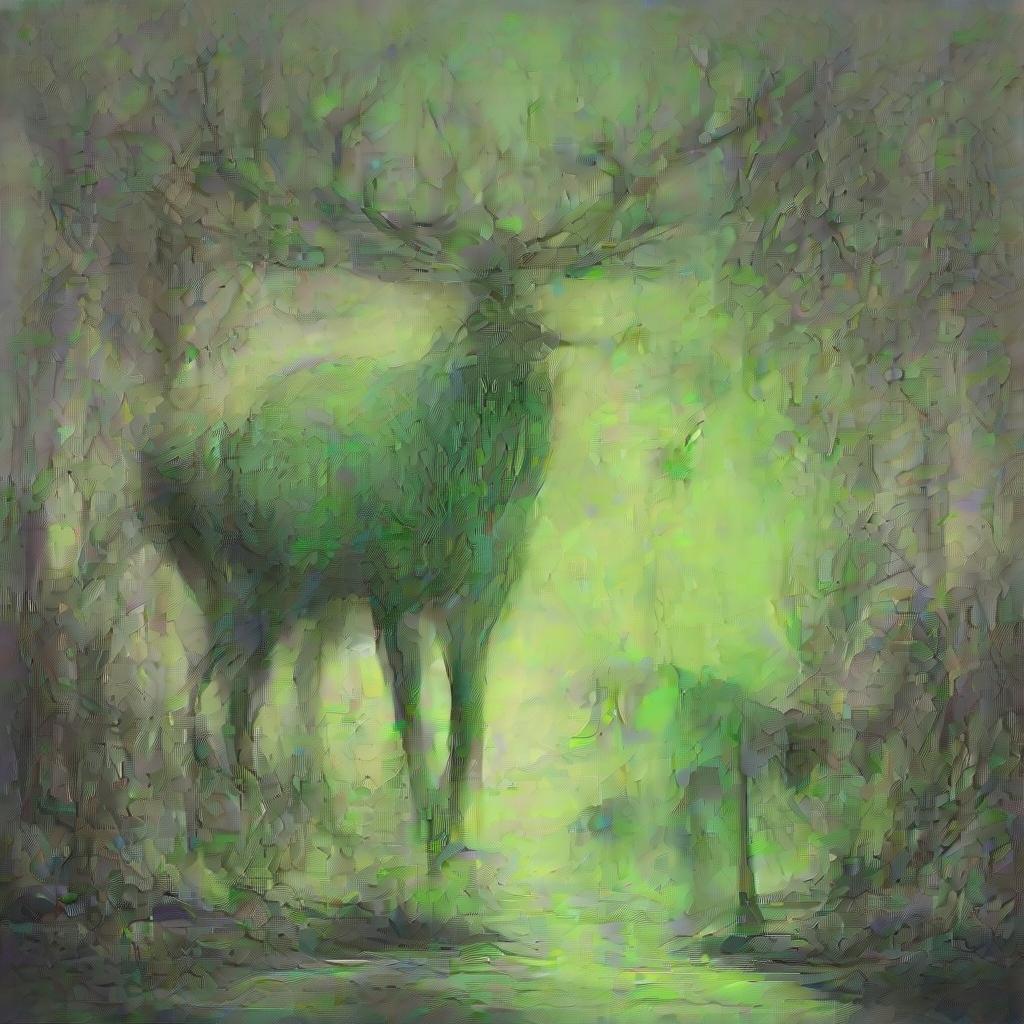}
        \end{tabularx}
        \vspace{-6pt}
        \caption{In a fantastical scene, a creature with a human head and deer body emanates a green light.}
    \end{subfigure}

    \begin{subfigure}[b]{\textwidth}
        \centering
        \setlength\tabcolsep{1pt}
        \begin{tabularx}{\textwidth}{@{}XX@{}X@{}X@{}XX@{}XX@{}X@{}}
            \includegraphics[width=\linewidth]{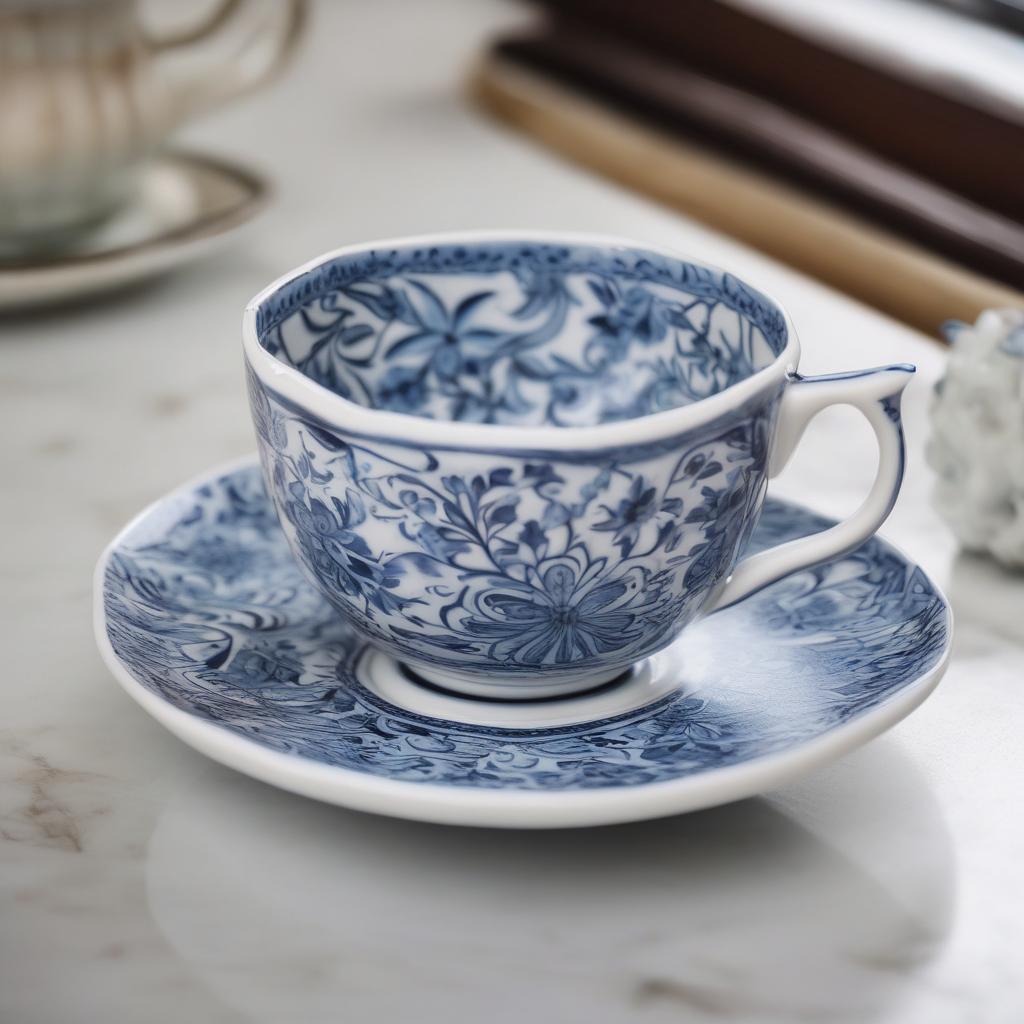} &
            \includegraphics[width=\linewidth]{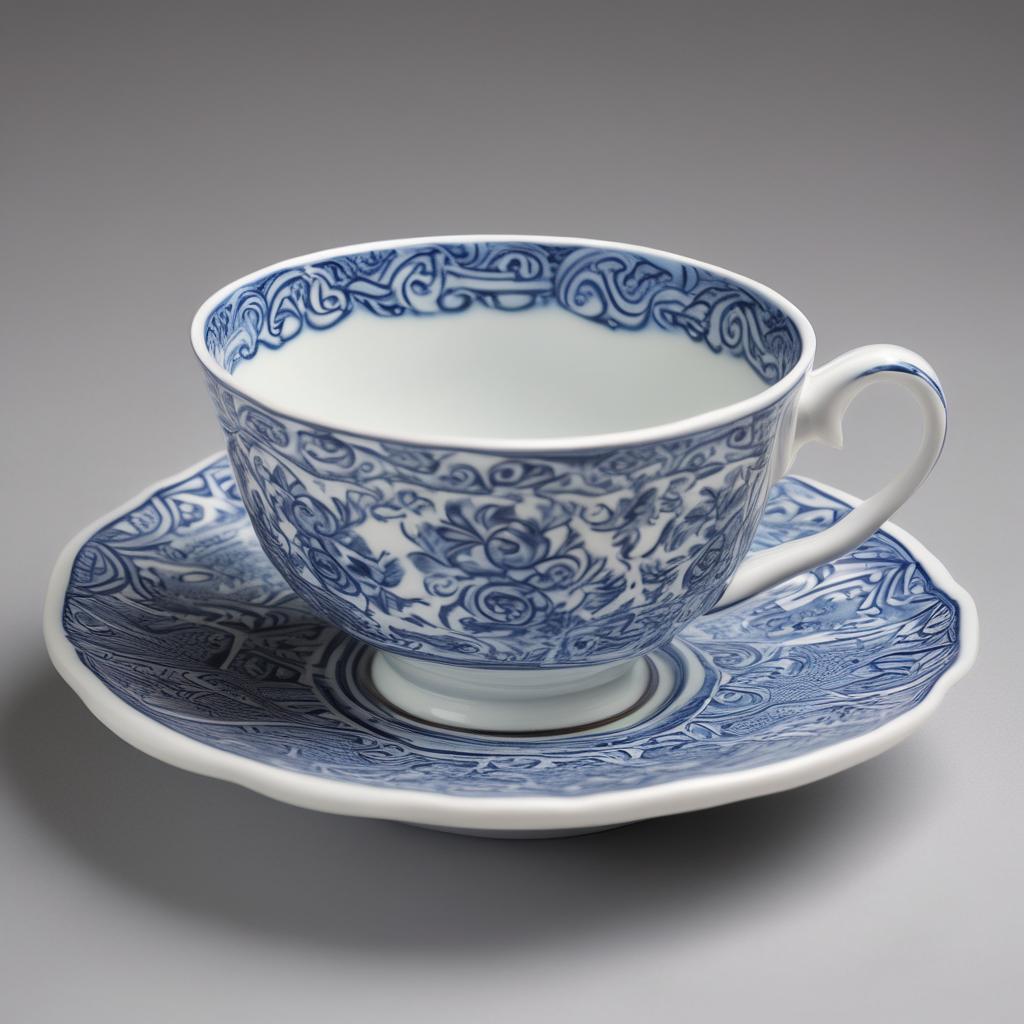} &
            \includegraphics[width=\linewidth]{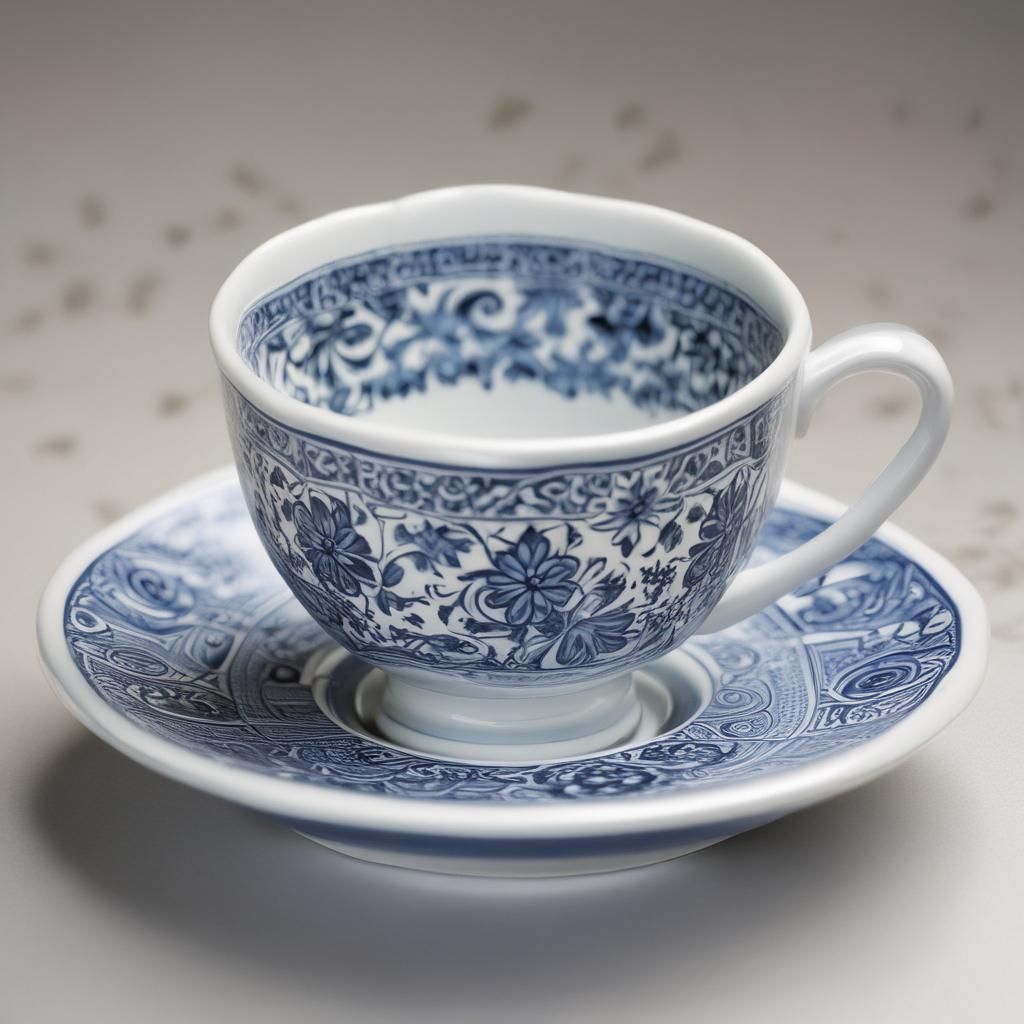} &
            \includegraphics[width=\linewidth]{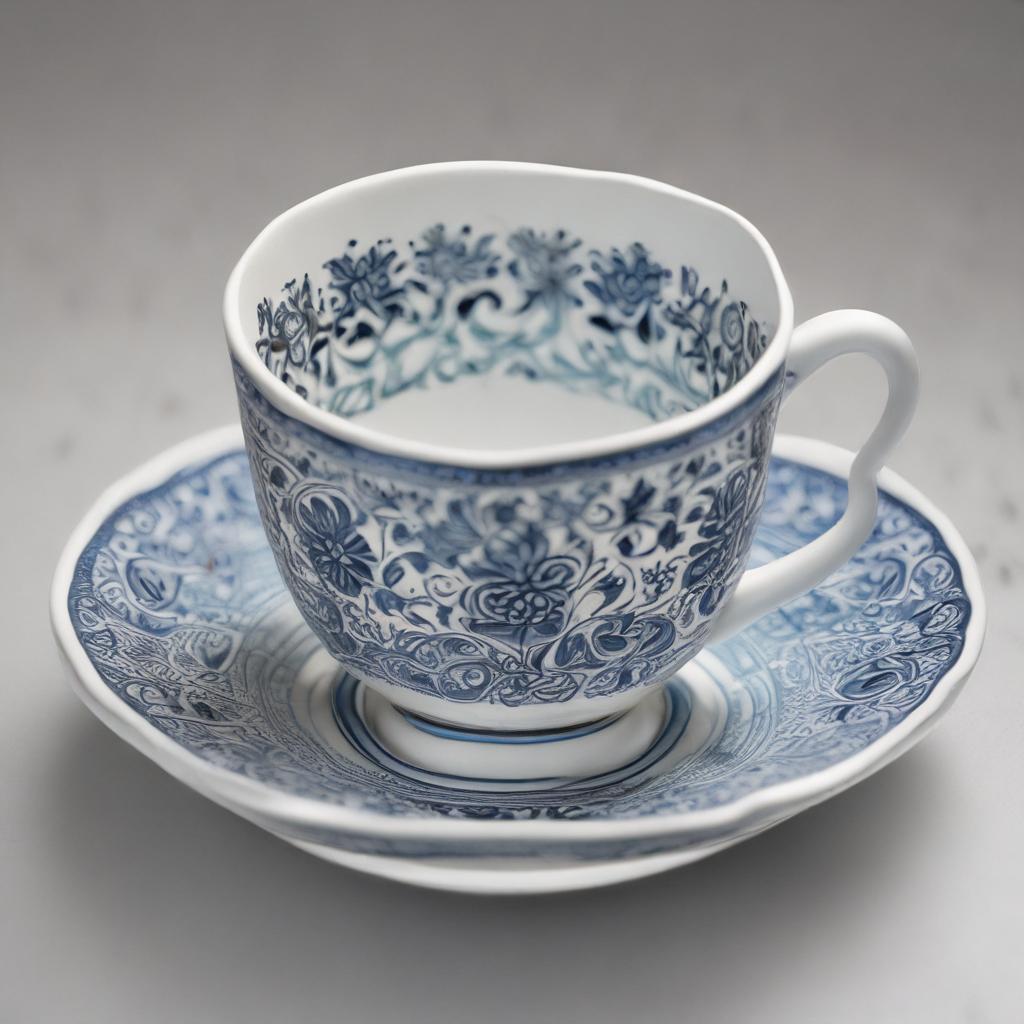} &
            \includegraphics[width=\linewidth]{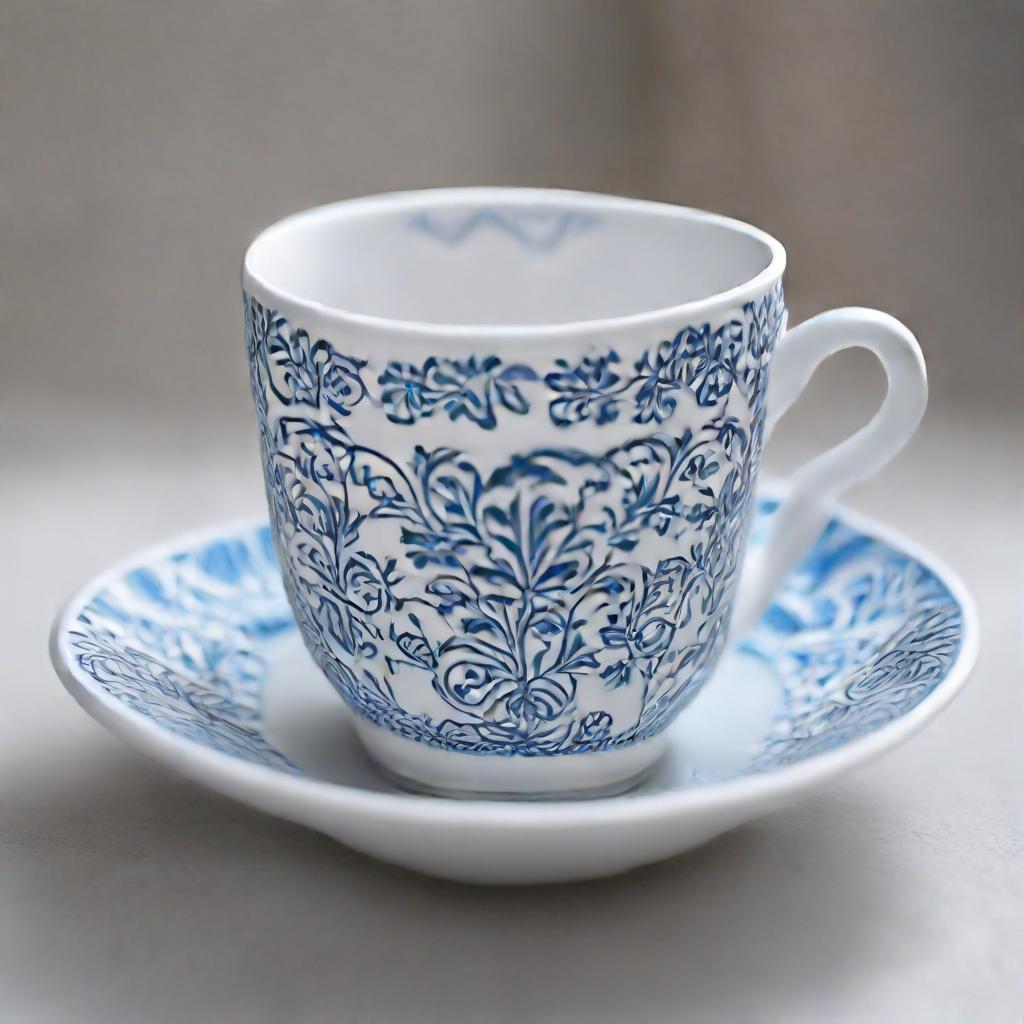} &
            \includegraphics[width=\linewidth]{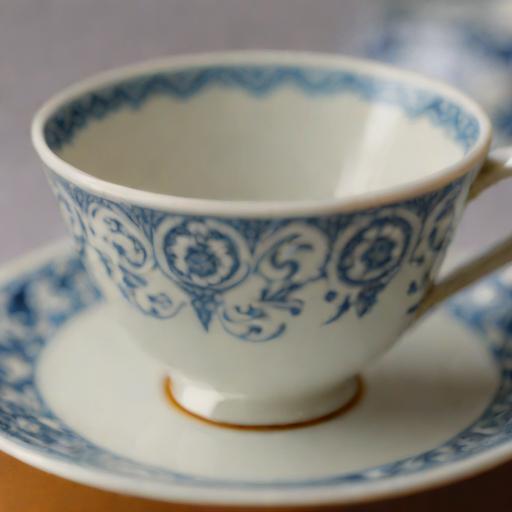} &
            \includegraphics[width=\linewidth]{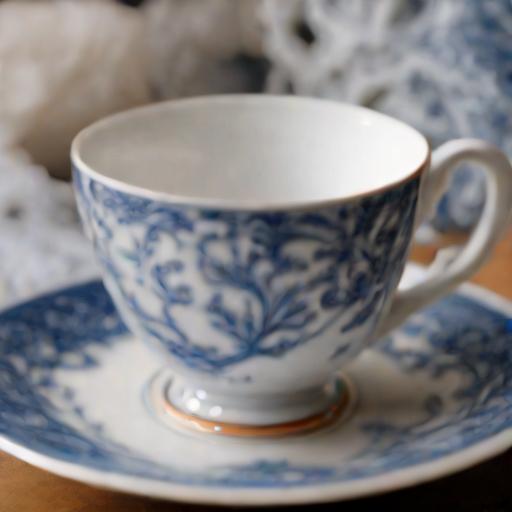} &
            \includegraphics[width=\linewidth]{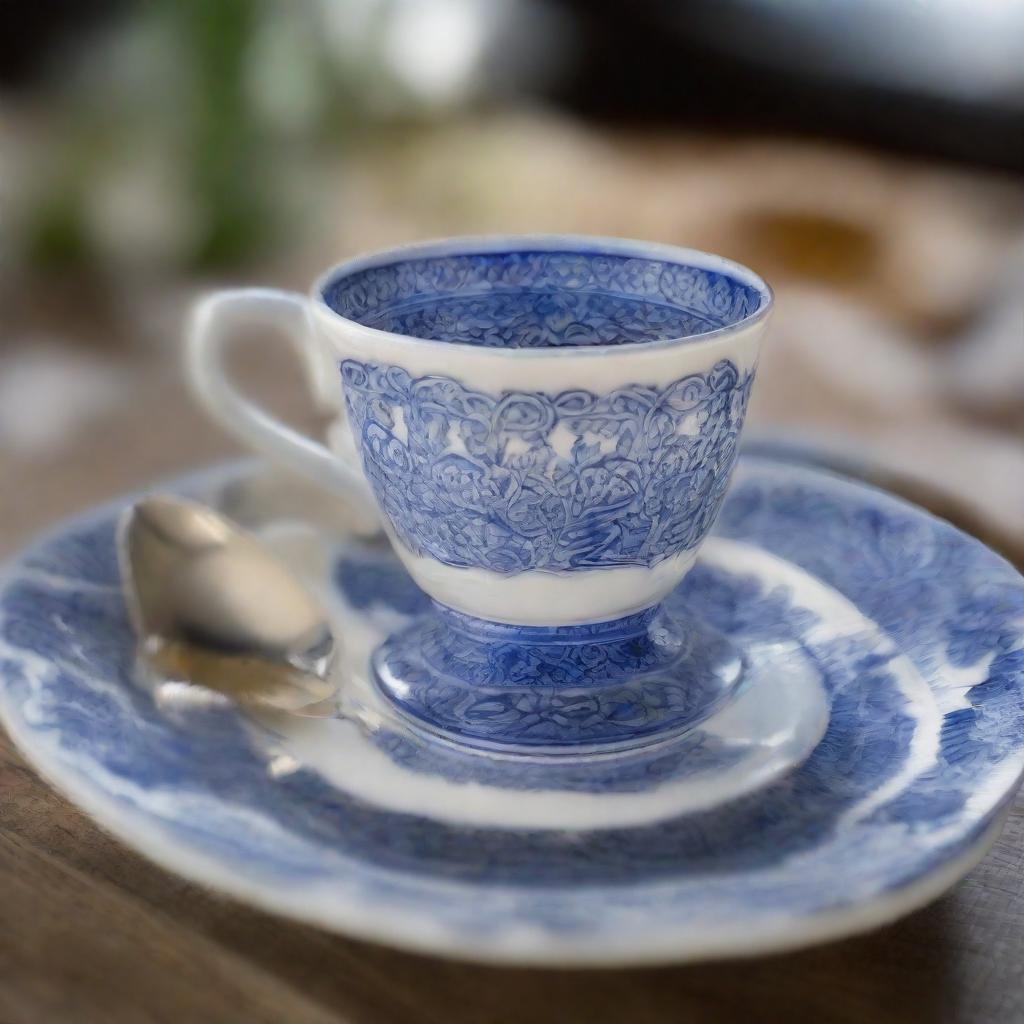} &
            \includegraphics[width=\linewidth]{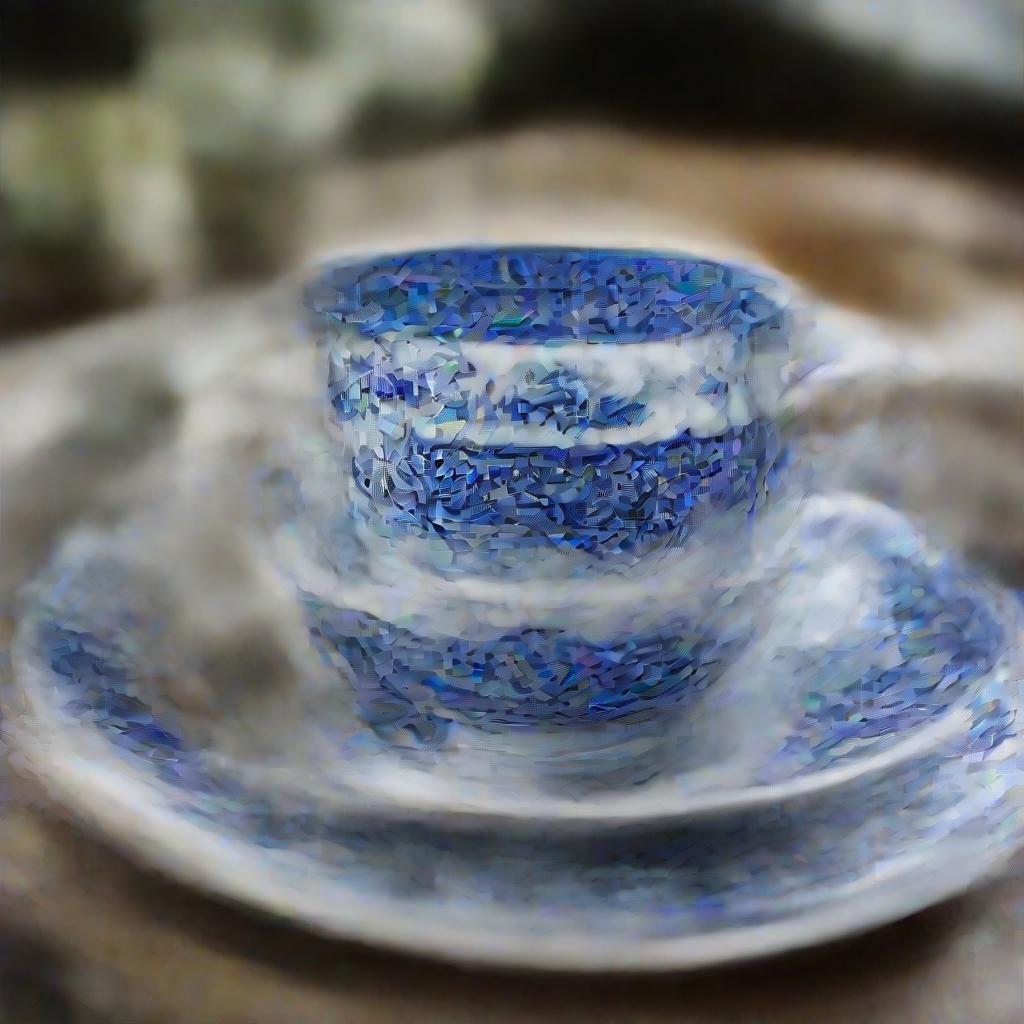}
        \end{tabularx}
        \vspace{-6pt}
        \caption{A delicate porcelain teacup sits on a saucer, its surface adorned with intricate blue patterns.}
    \end{subfigure}
    
    \begin{subfigure}[b]{\textwidth}
        \centering
        \setlength\tabcolsep{1pt}
        \begin{tabularx}{\textwidth}{@{}XX@{}X@{}X@{}XX@{}XX@{}X@{}}
            \includegraphics[width=\linewidth]{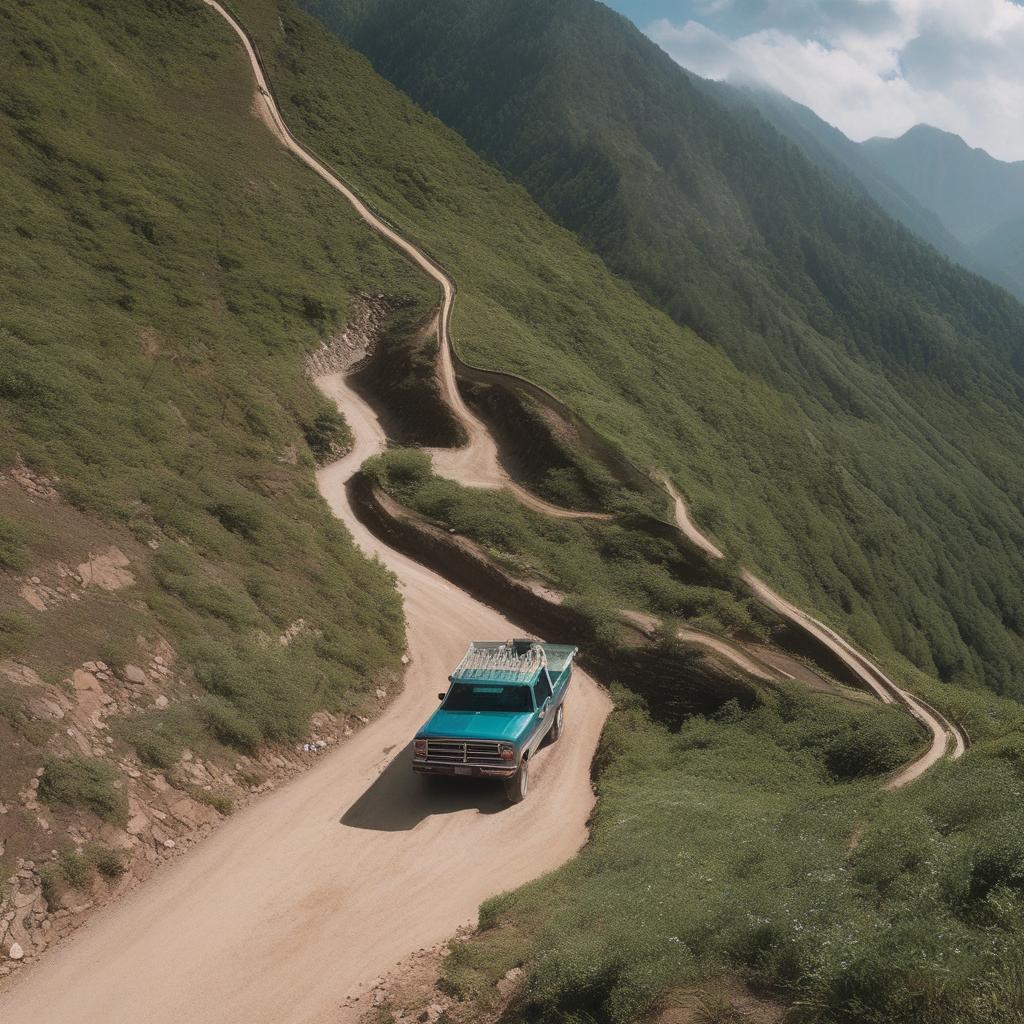} &
            \includegraphics[width=\linewidth]{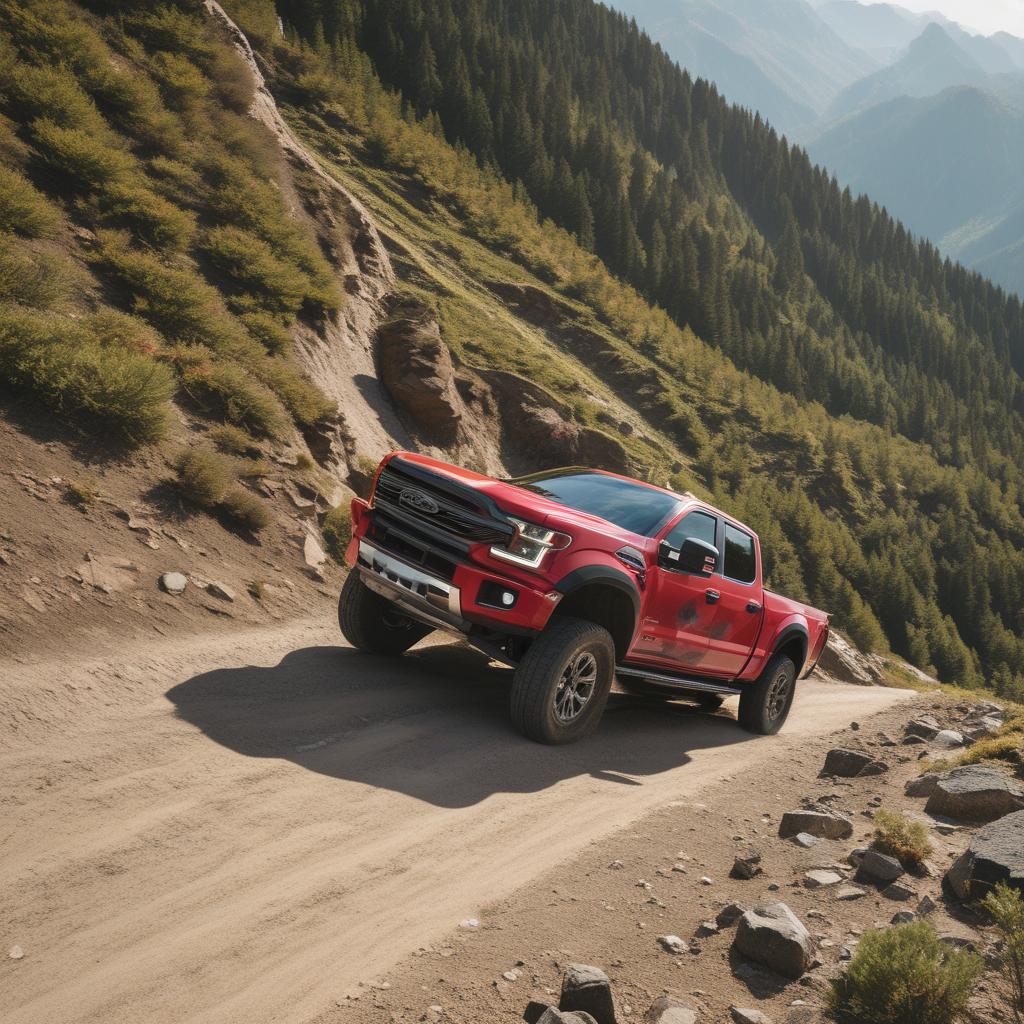} &
            \includegraphics[width=\linewidth]{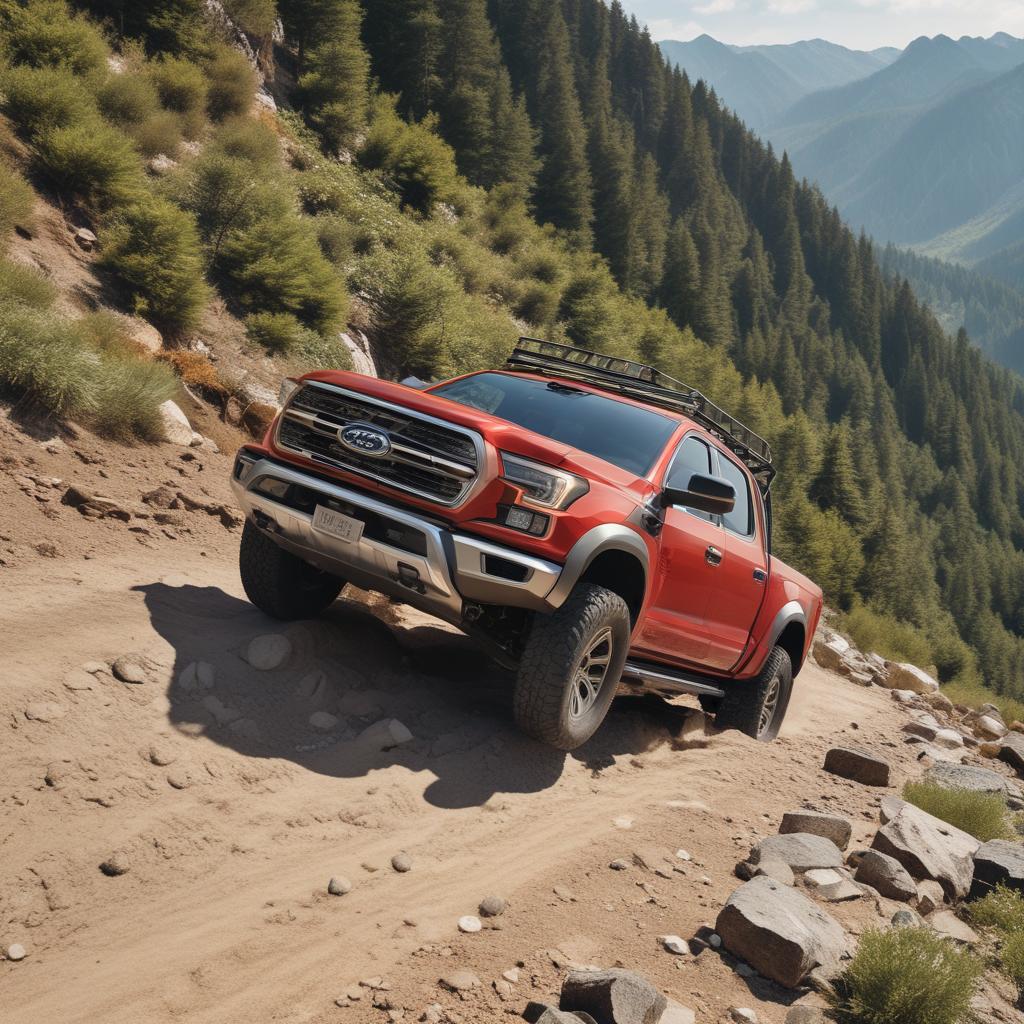} &
            \includegraphics[width=\linewidth]{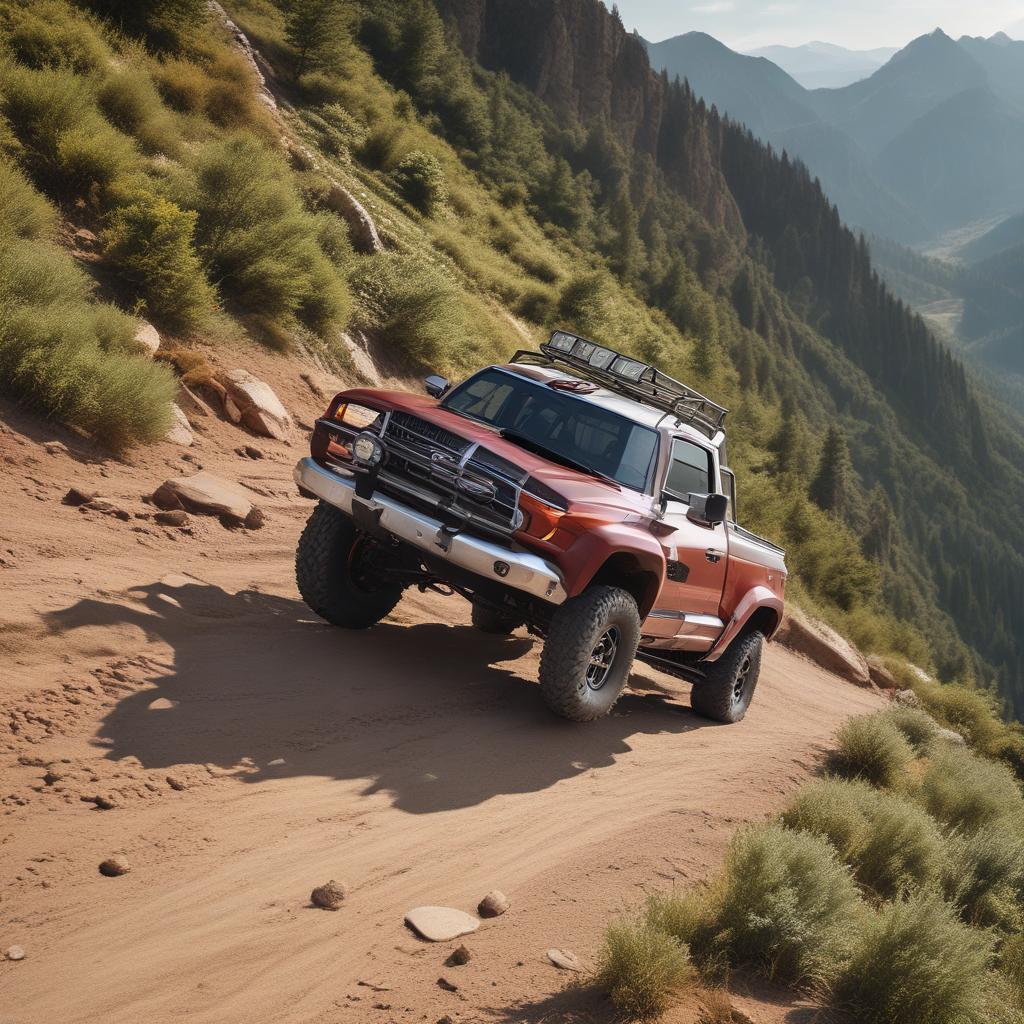} &
            \includegraphics[width=\linewidth]{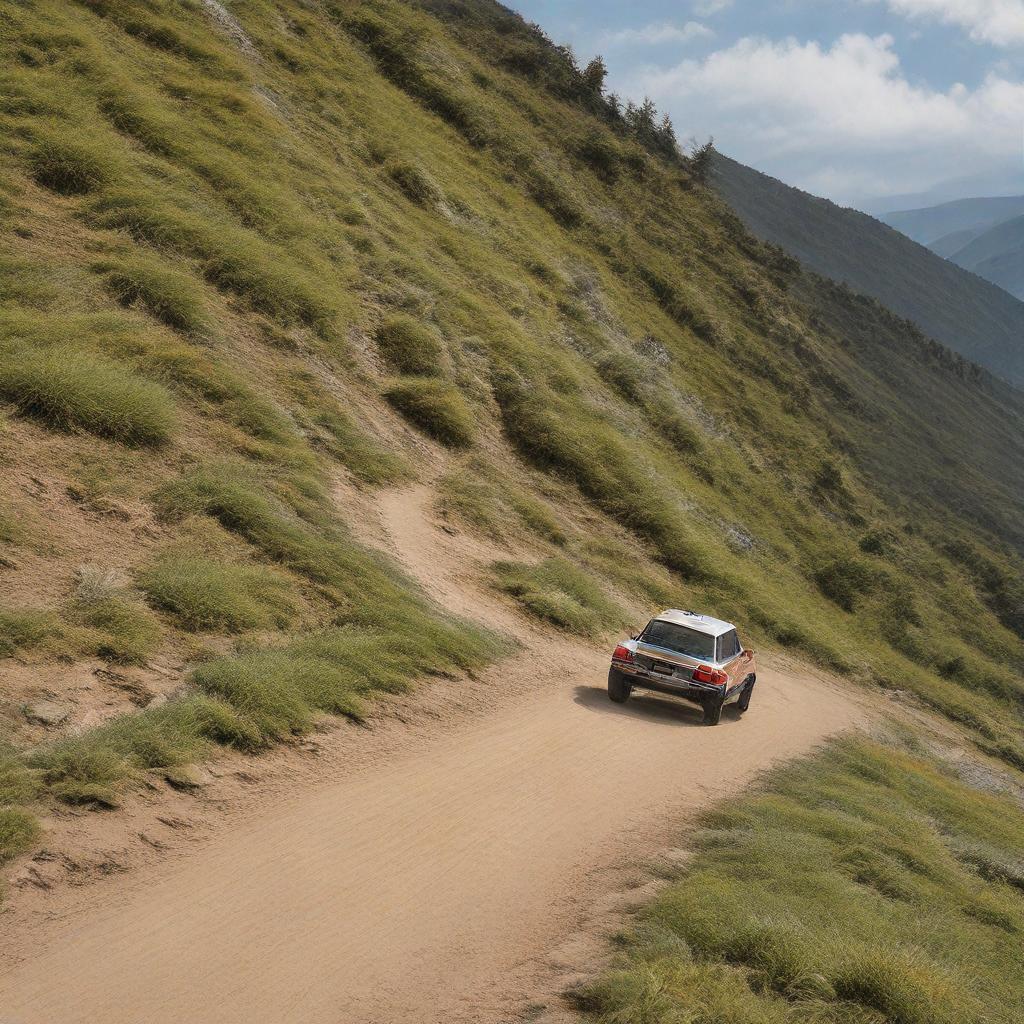} &
            \includegraphics[width=\linewidth]{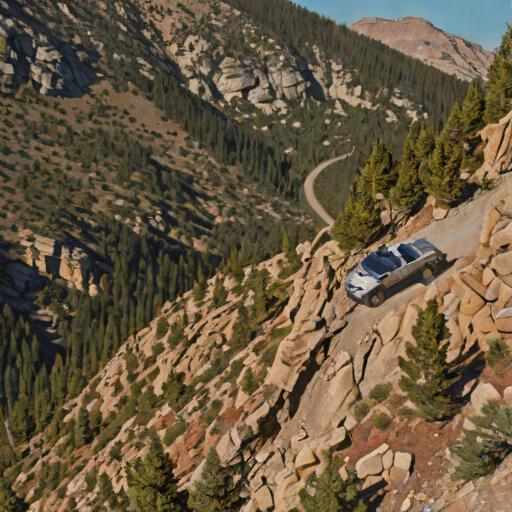} &
            \includegraphics[width=\linewidth]{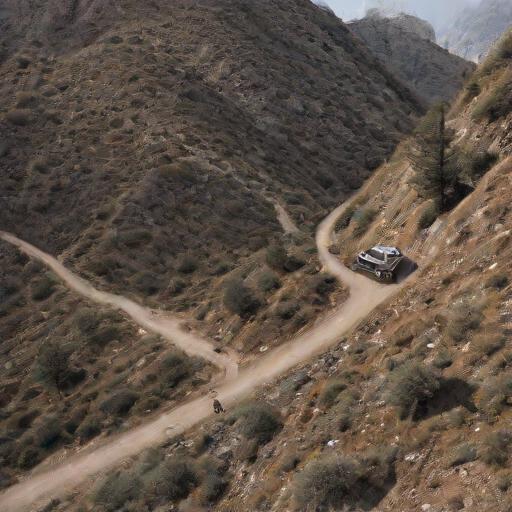} &
            \includegraphics[width=\linewidth]{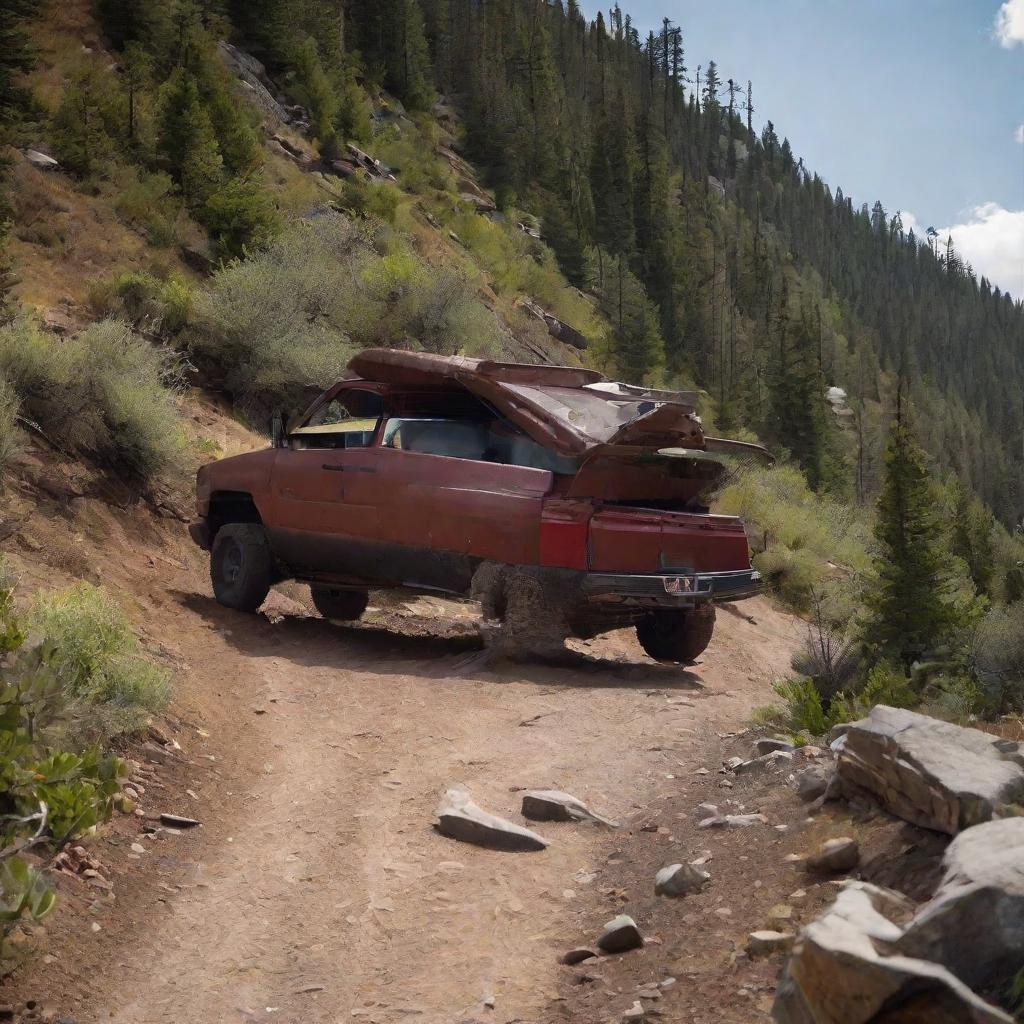} &
            \includegraphics[width=\linewidth]{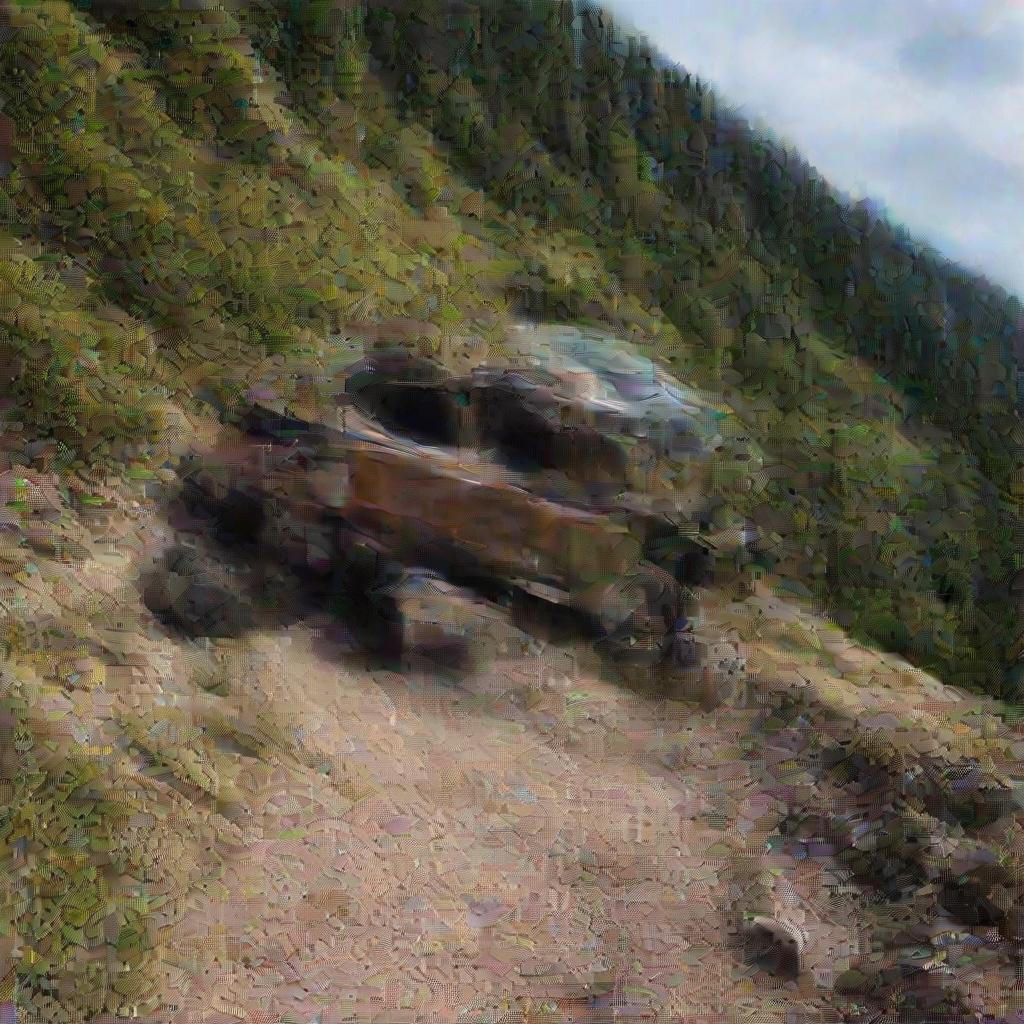}
        \end{tabularx}
        \vspace{-6pt}
        \caption{A pickup truck going up a mountain switchback.}
    \end{subfigure}
    \caption{Qualitative comparison. Our models here are fully trained instead of LoRA. Models are given the same initial noise for each prompt, except SDXL-Turbo since it only supports 512px resolution. Our model produces the best results in all prompts, and it best preserves the style and layout of the original model. Note 1: Original SDXL and ours use Euler sampler while other methods use their default samplers. Note 2: some methods require classifier-free guidance (CFG) at inference, which doubles the number of function evaluations (NFE) and doubles the computation. (Please zoom in to view at the full resolution.)}
    \label{fig:qualitative}
\end{figure*}

\begin{figure*}[t]
    \centering
    \captionsetup{justification=raggedright,singlelinecheck=false}
    \small
    \setlength\tabcolsep{4pt}
    \begin{tabularx}{\textwidth}{|X|X|X|X|X|X|X|X|X|}
        \normalsize{\textbf{SDXL}} \cite{podell2023sdxl} & \multicolumn{6}{l|}{\normalsize{\textbf{Ours}}} & \multicolumn{2}{l|}{\normalsize{\textbf{LCM}} \cite{luo2023latent,luo2023lcmlora}} \\

        \footnotesize{64NFE, CFG6} & \multicolumn{6}{l|}{\footnotesize{No CFG}} & \footnotesize{8NFE, CFG8} & \footnotesize{No CFG} \\
        
        32 Steps & \multicolumn{2}{l|}{8 Steps} & \multicolumn{2}{l|}{4 Steps} & \multicolumn{2}{l|}{2 Steps} & \multicolumn{2}{l|}{4 Steps} \\
        
         & Full & LoRA & Full & LoRA & Full & LoRA & Full & LoRA
    \end{tabularx}

    \begin{subfigure}[b]{\textwidth}
        \centering
        \setlength\tabcolsep{1pt}
        \begin{tabularx}{\textwidth}{@{}XX@{}X@{}X@{}X@{}X@{}XX@{}X@{}}
            \includegraphics[width=\linewidth]{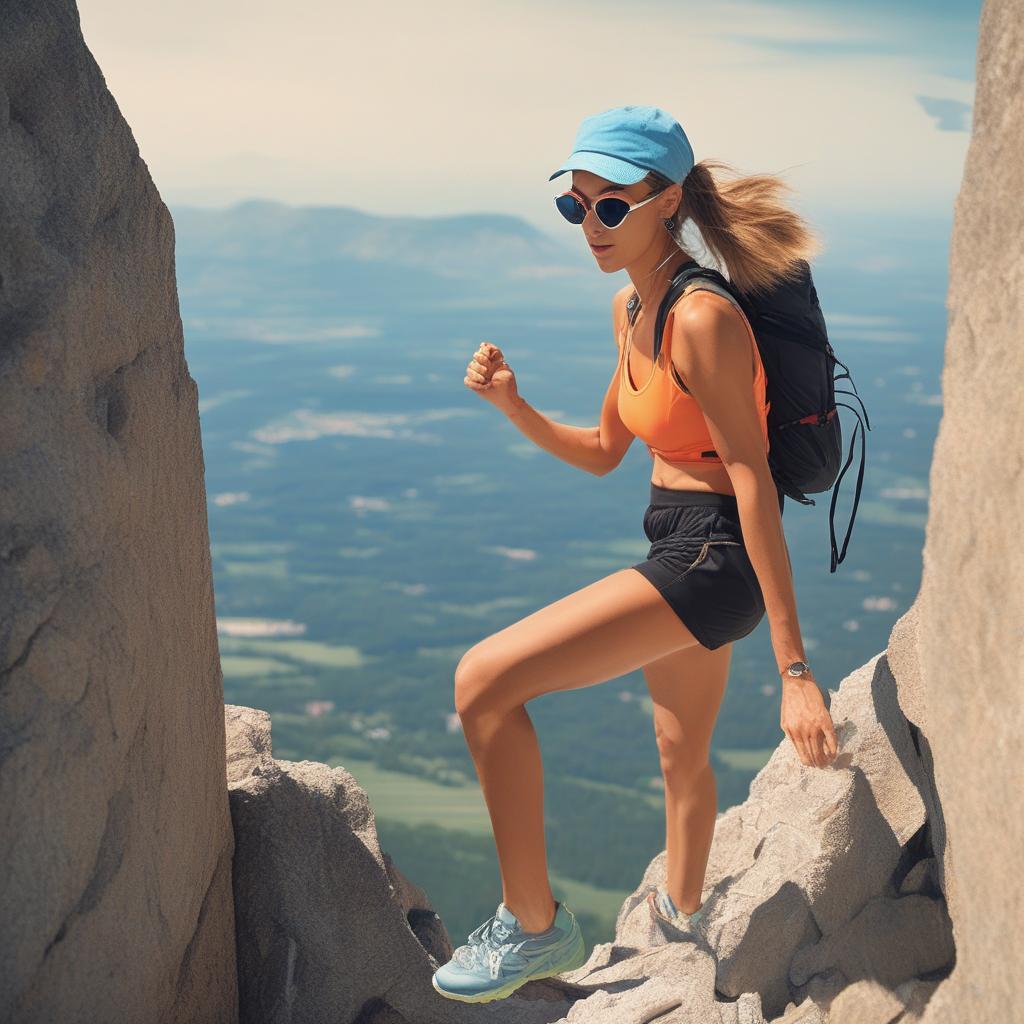} &
            \includegraphics[width=\linewidth]{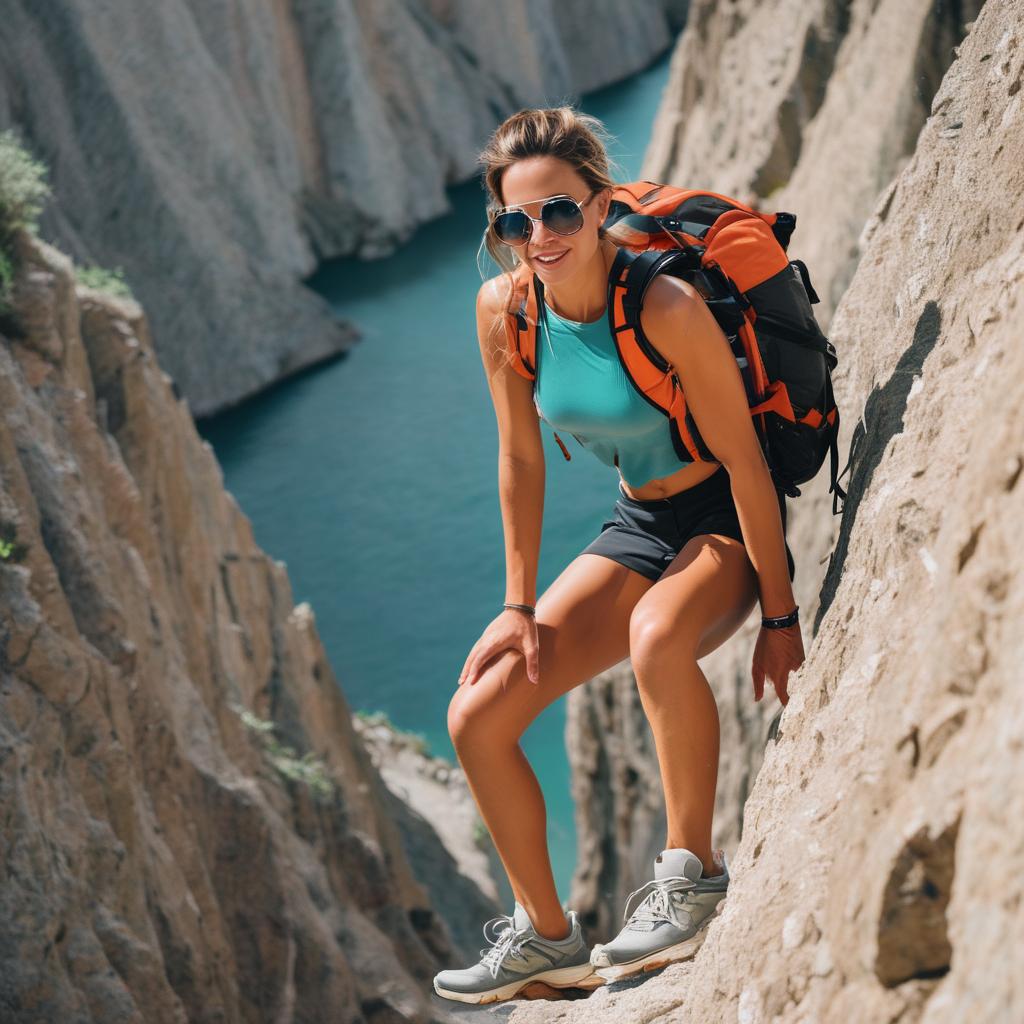} &
            \includegraphics[width=\linewidth]{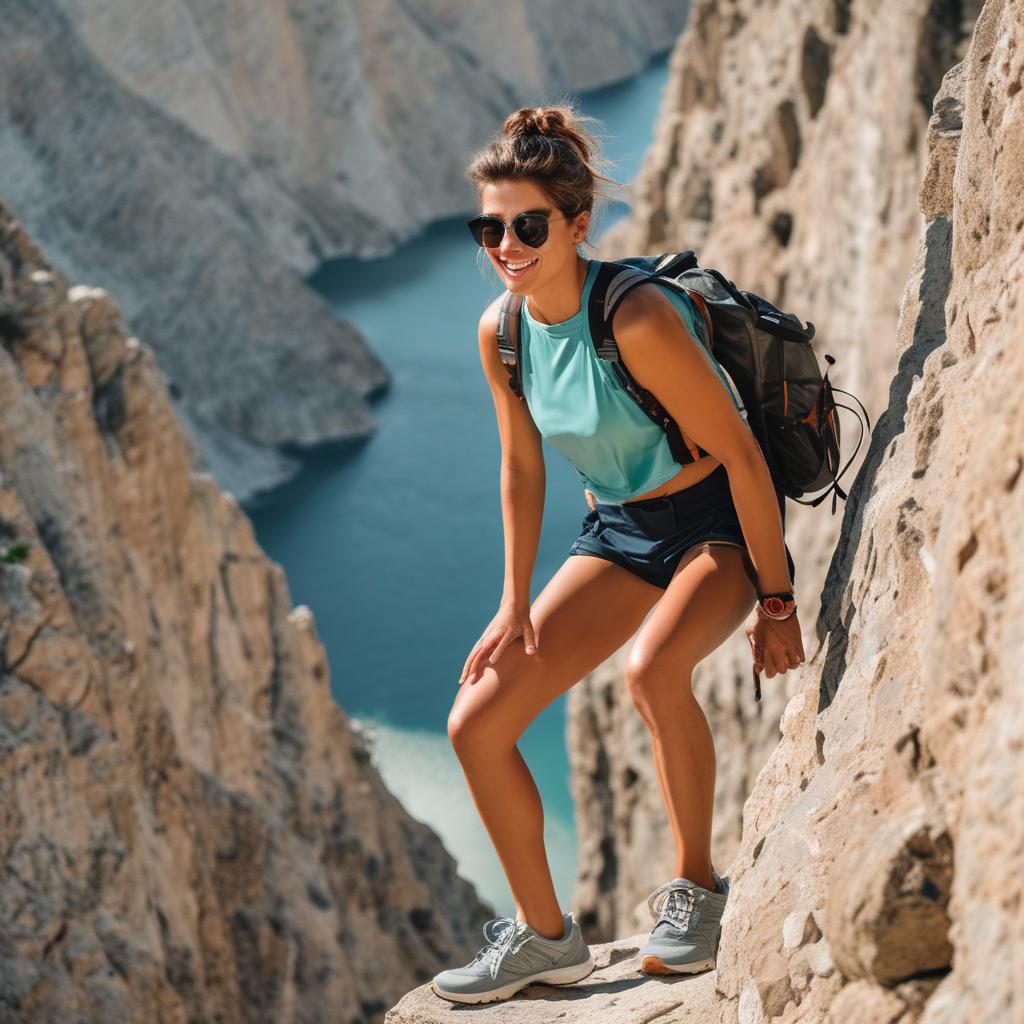} &
            \includegraphics[width=\linewidth]{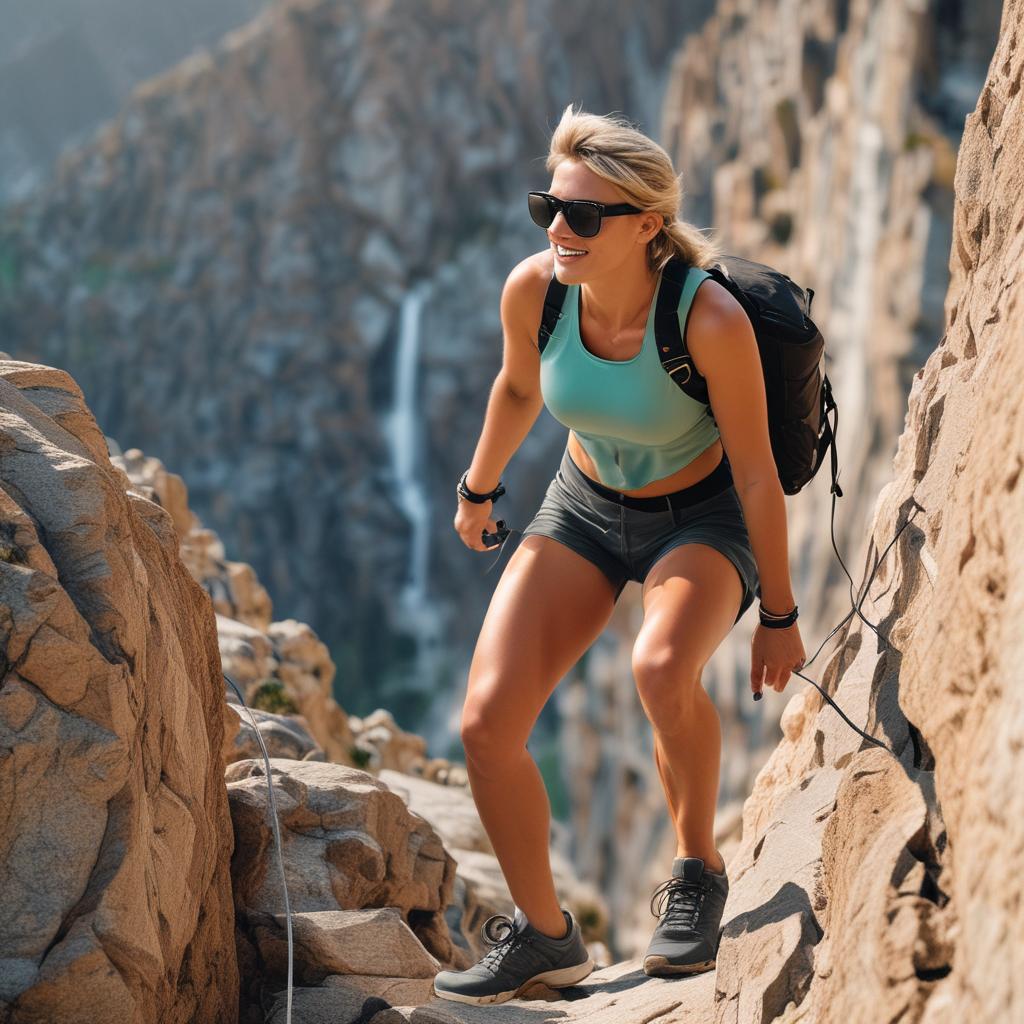} &
            \includegraphics[width=\linewidth]{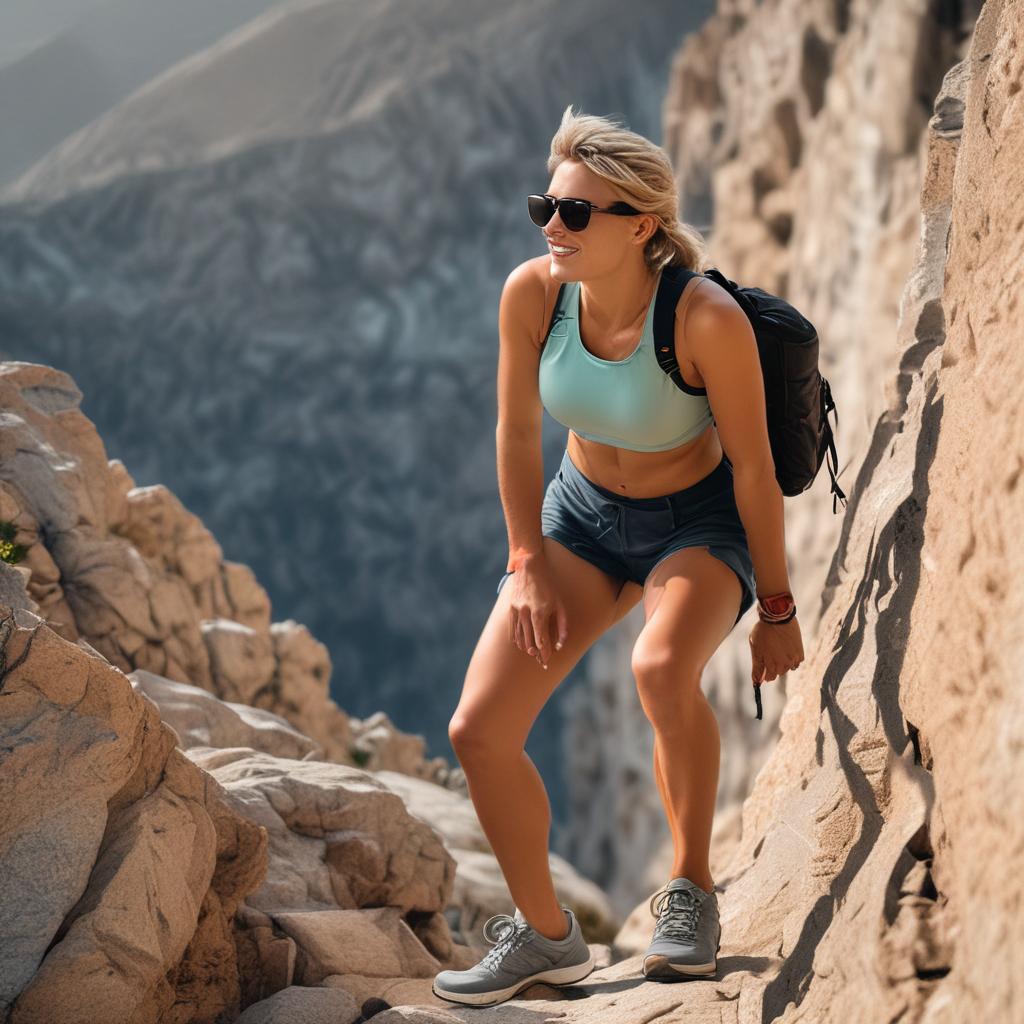} &
            \includegraphics[width=\linewidth]{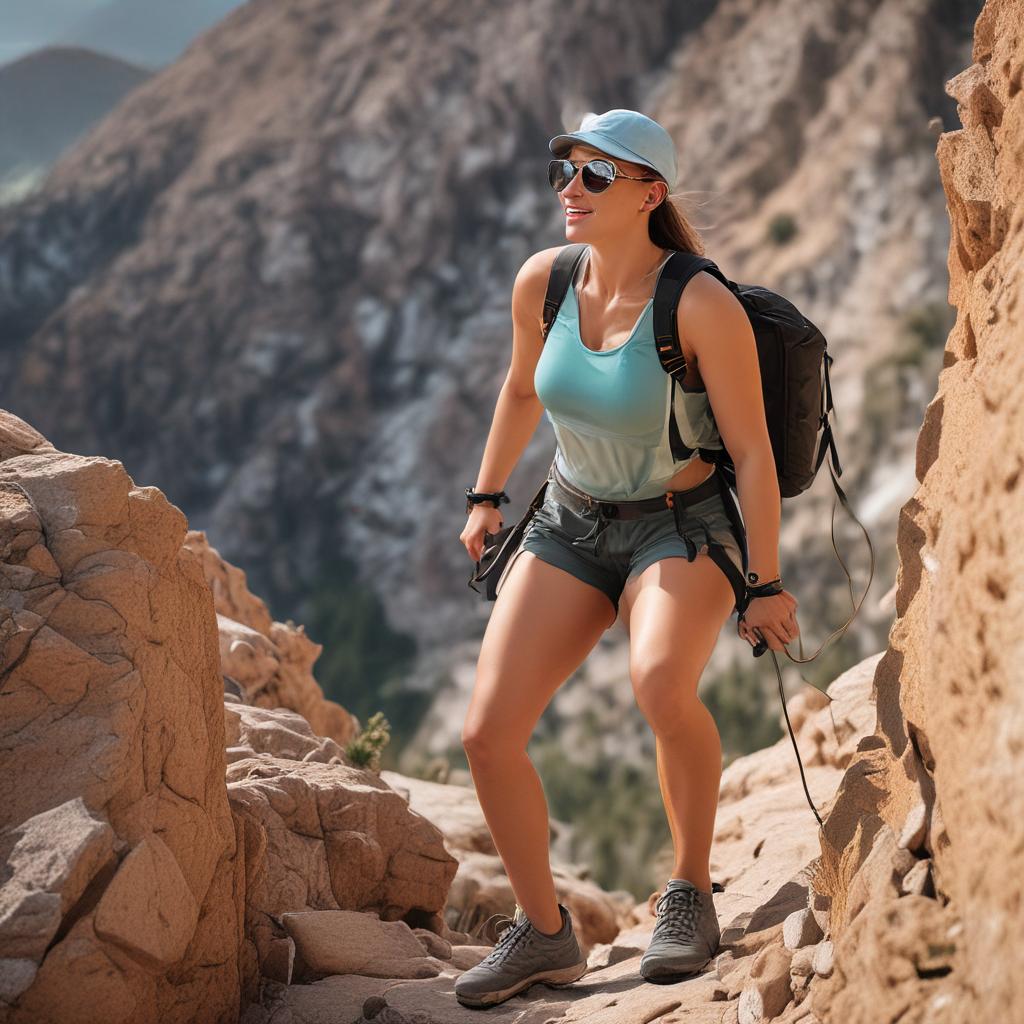} &
            \includegraphics[width=\linewidth]{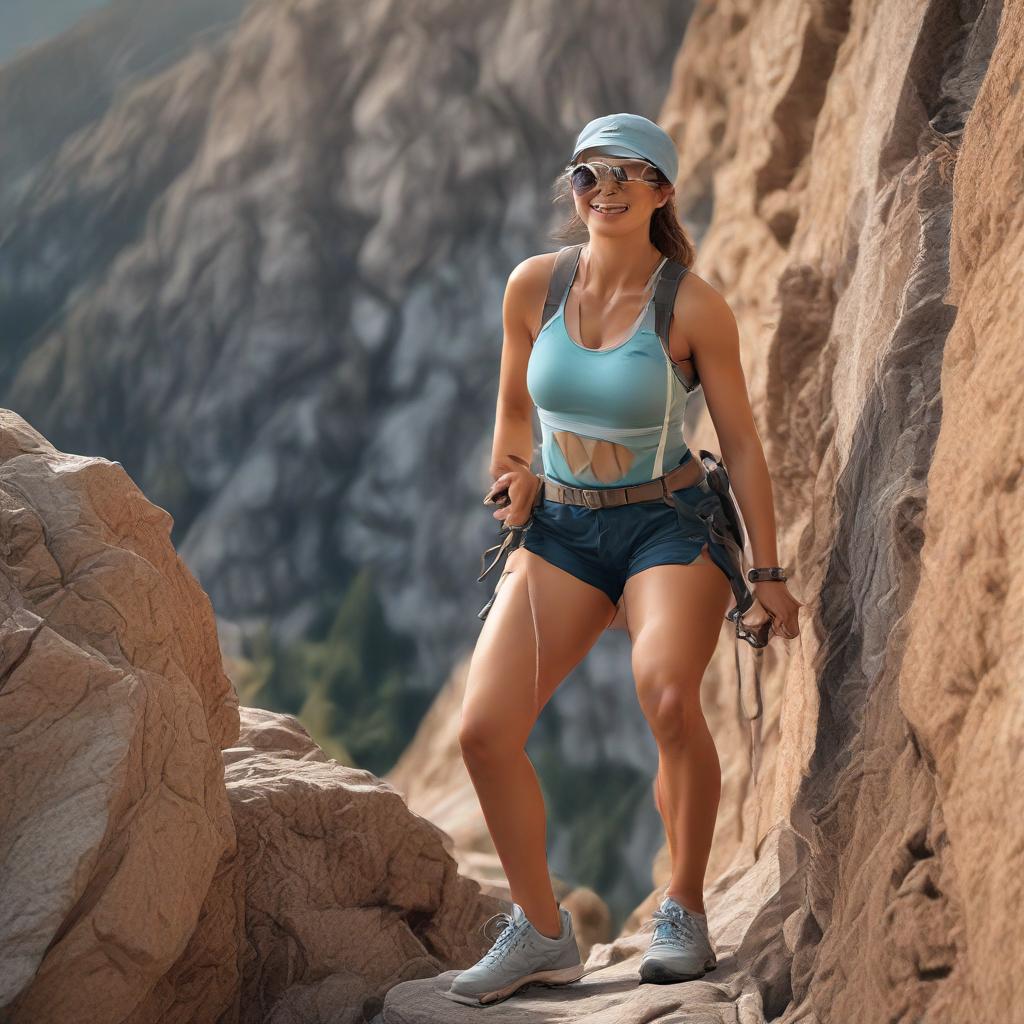} &
            \includegraphics[width=\linewidth]{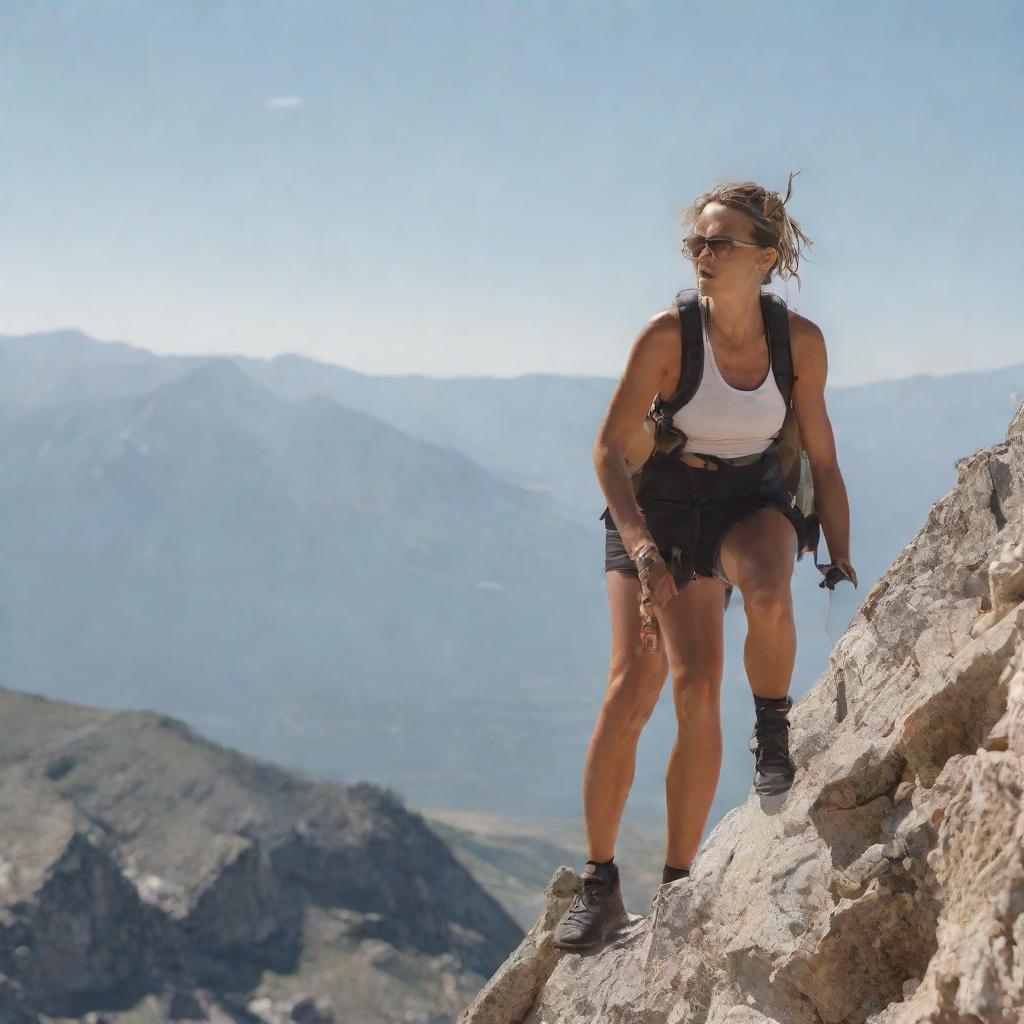} &
            \includegraphics[width=\linewidth]{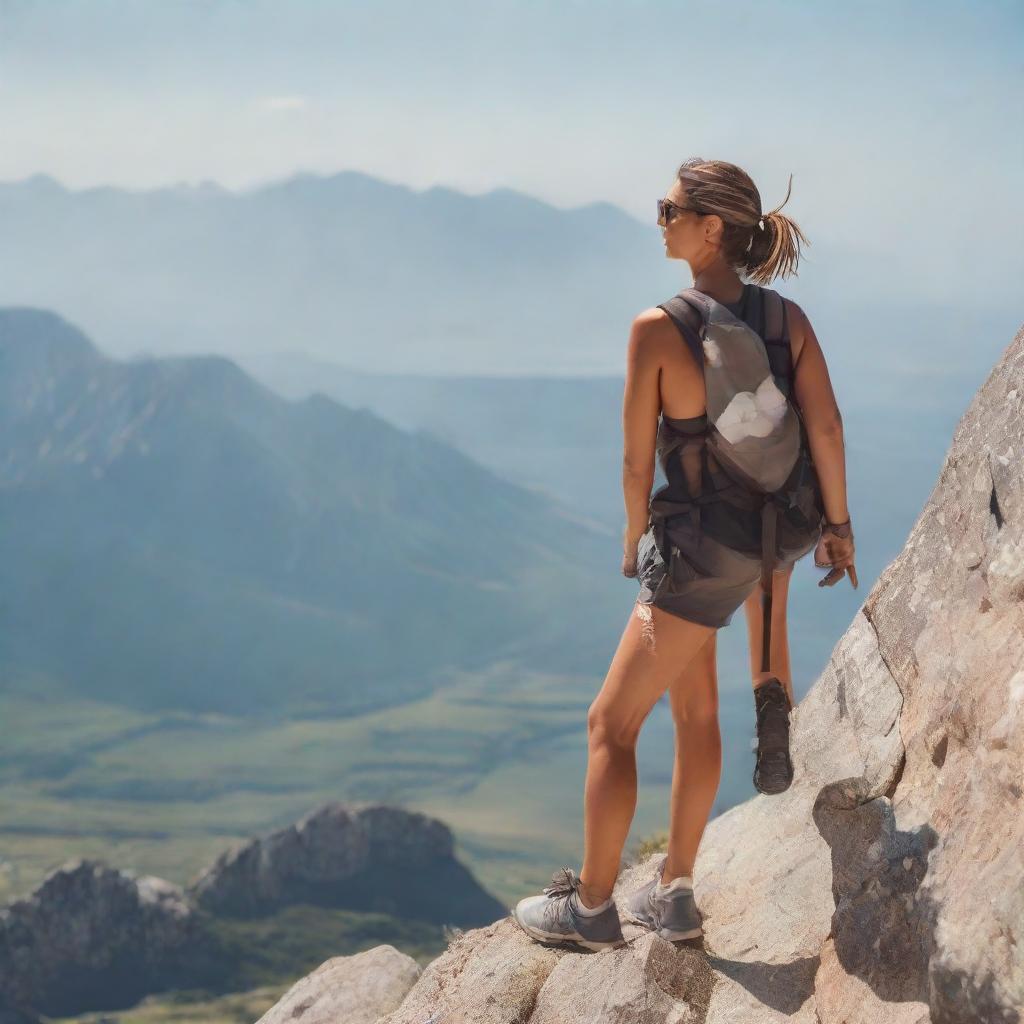}
        \end{tabularx}
        \vspace{-6pt}
        \caption{A tanned woman, dressed in sportswear and sunglasses, climbing a peak with a group during the summer.}
    \end{subfigure}
    
    \begin{subfigure}[b]{\textwidth}
        \centering
        \setlength\tabcolsep{1pt}
        \begin{tabularx}{\textwidth}{@{}XX@{}X@{}X@{}X@{}X@{}XX@{}X@{}}
            \includegraphics[width=\linewidth]{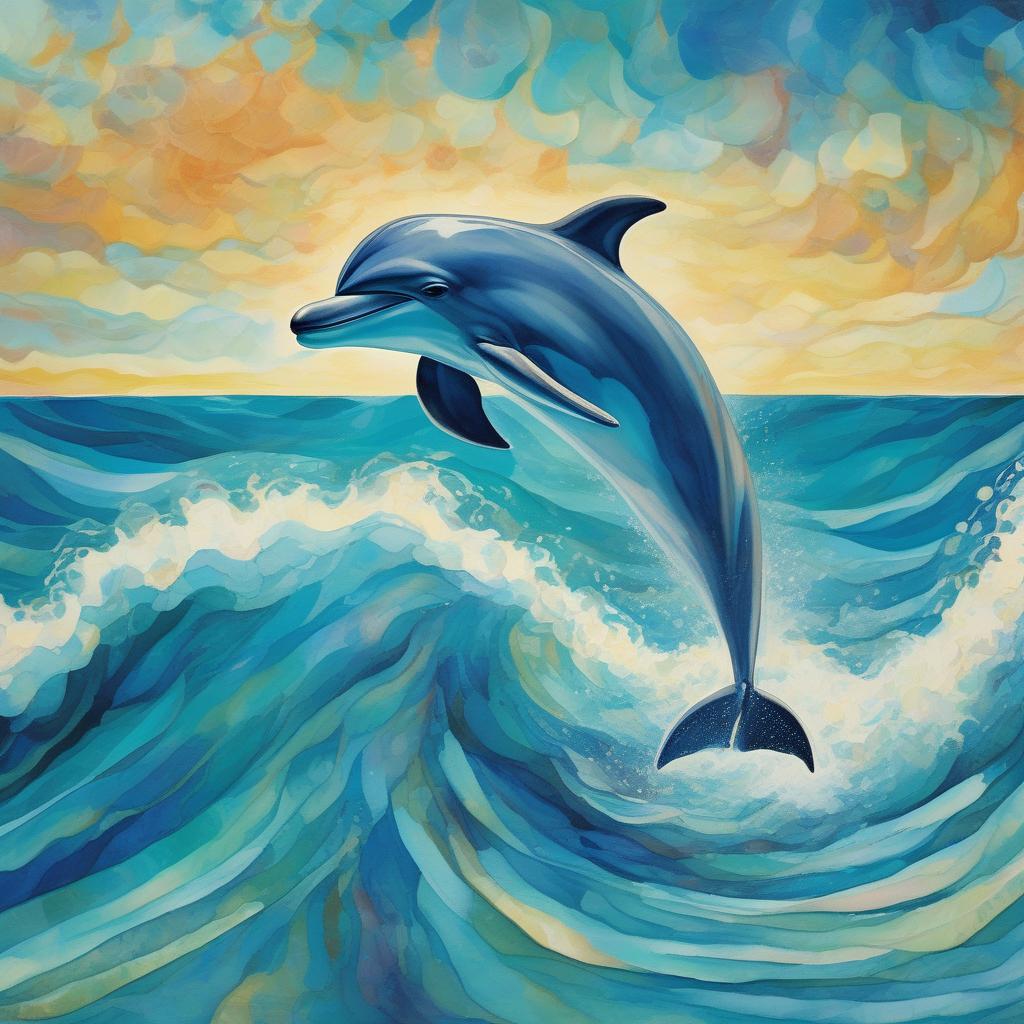} &
            \includegraphics[width=\linewidth]{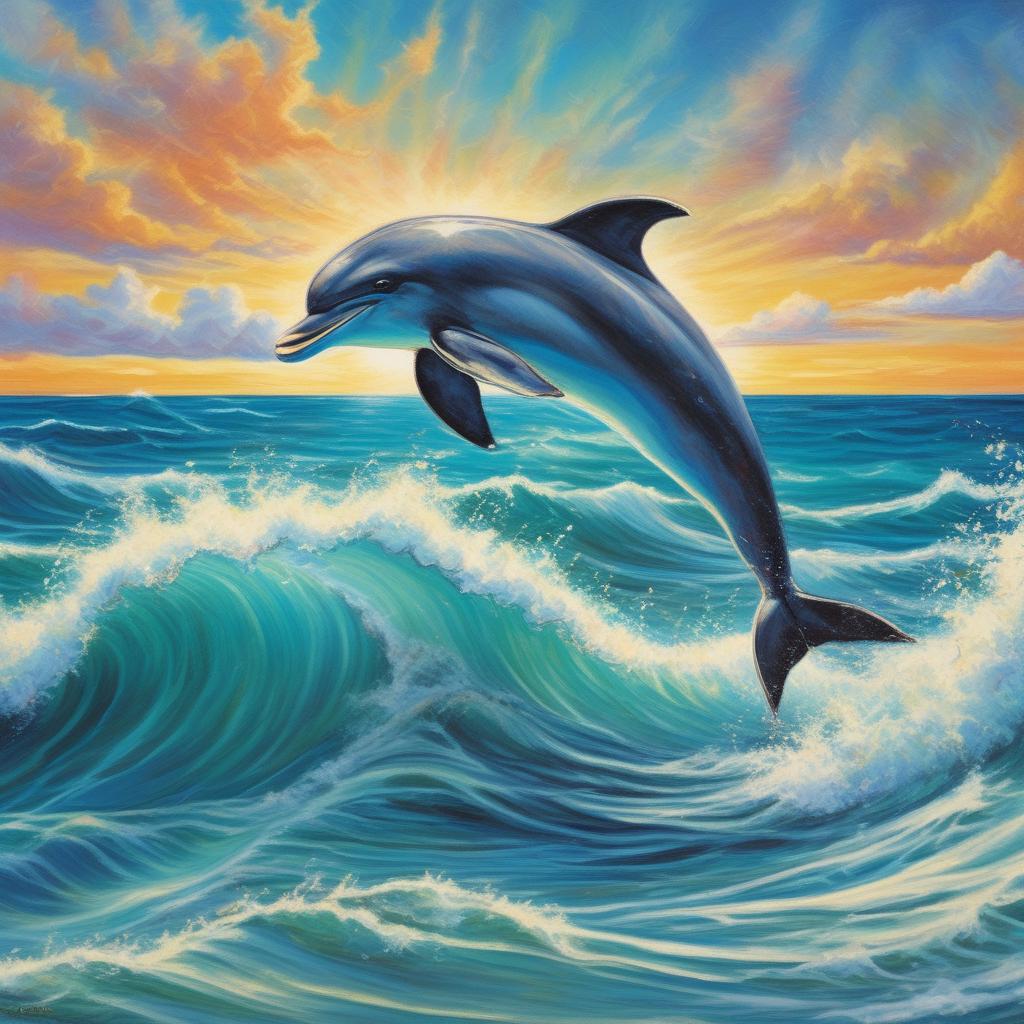} &
            \includegraphics[width=\linewidth]{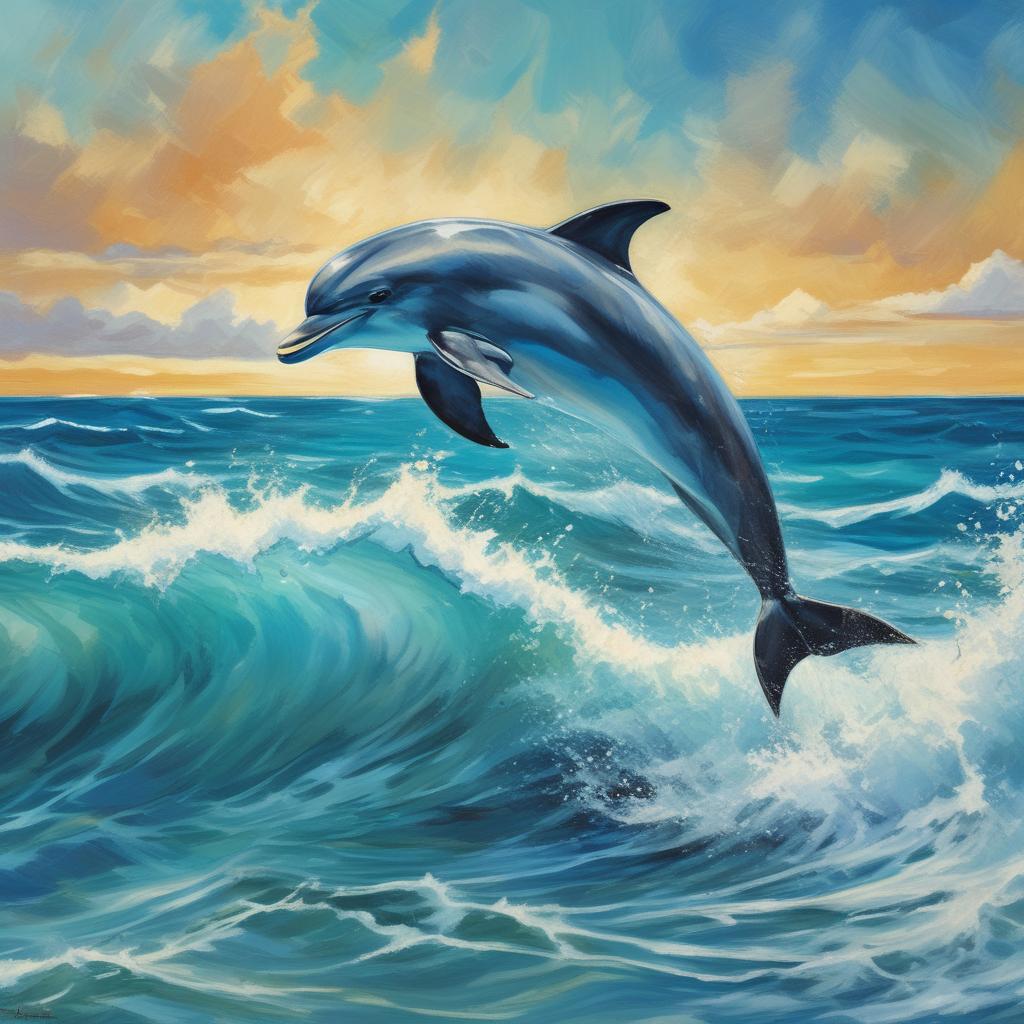} &
            \includegraphics[width=\linewidth]{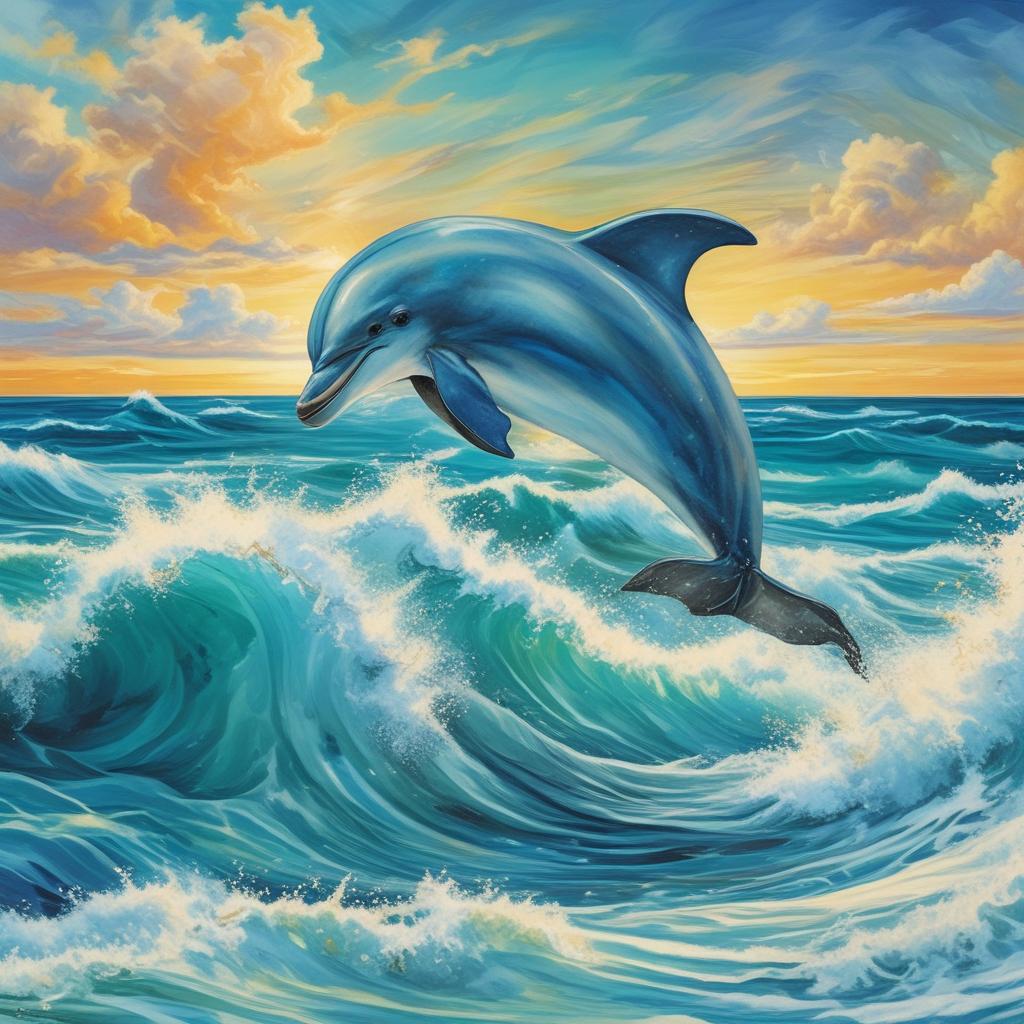} &
            \includegraphics[width=\linewidth]{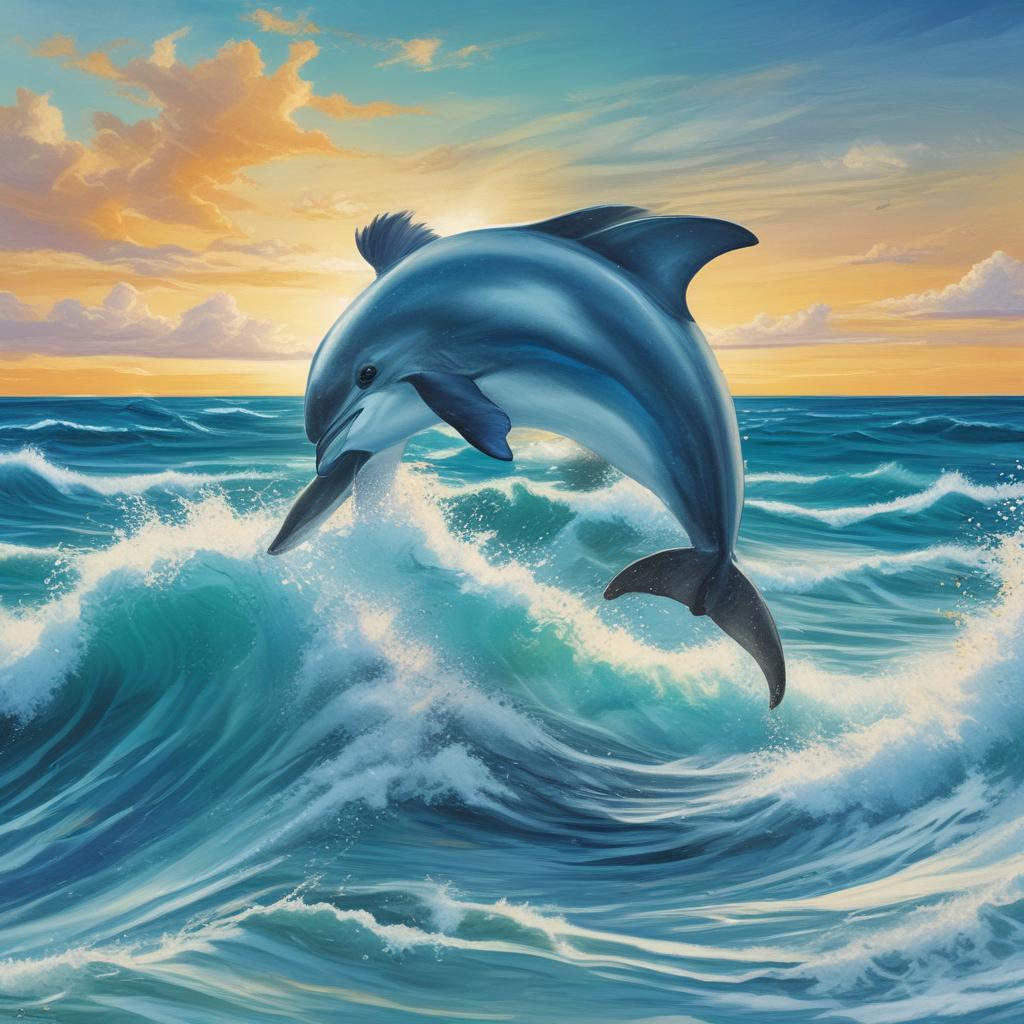} &
            \includegraphics[width=\linewidth]{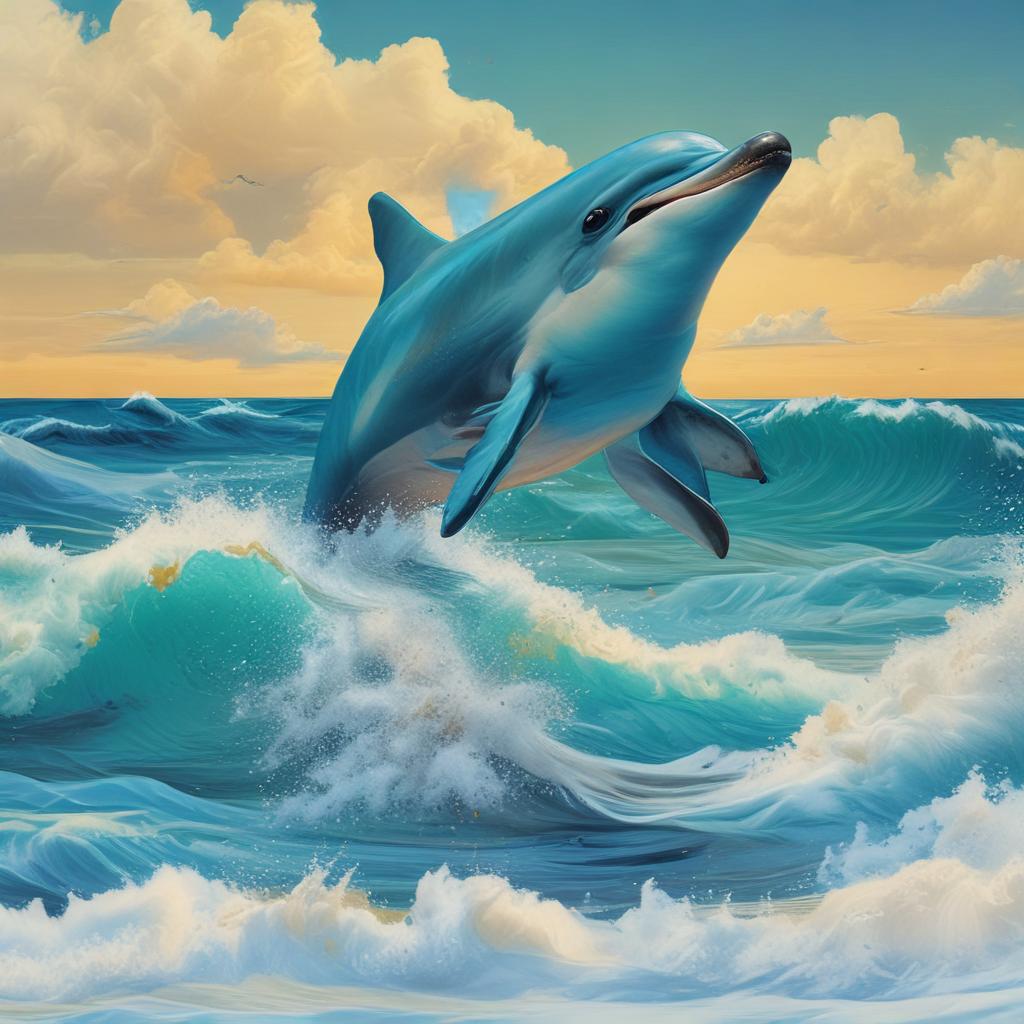} &
            \includegraphics[width=\linewidth]{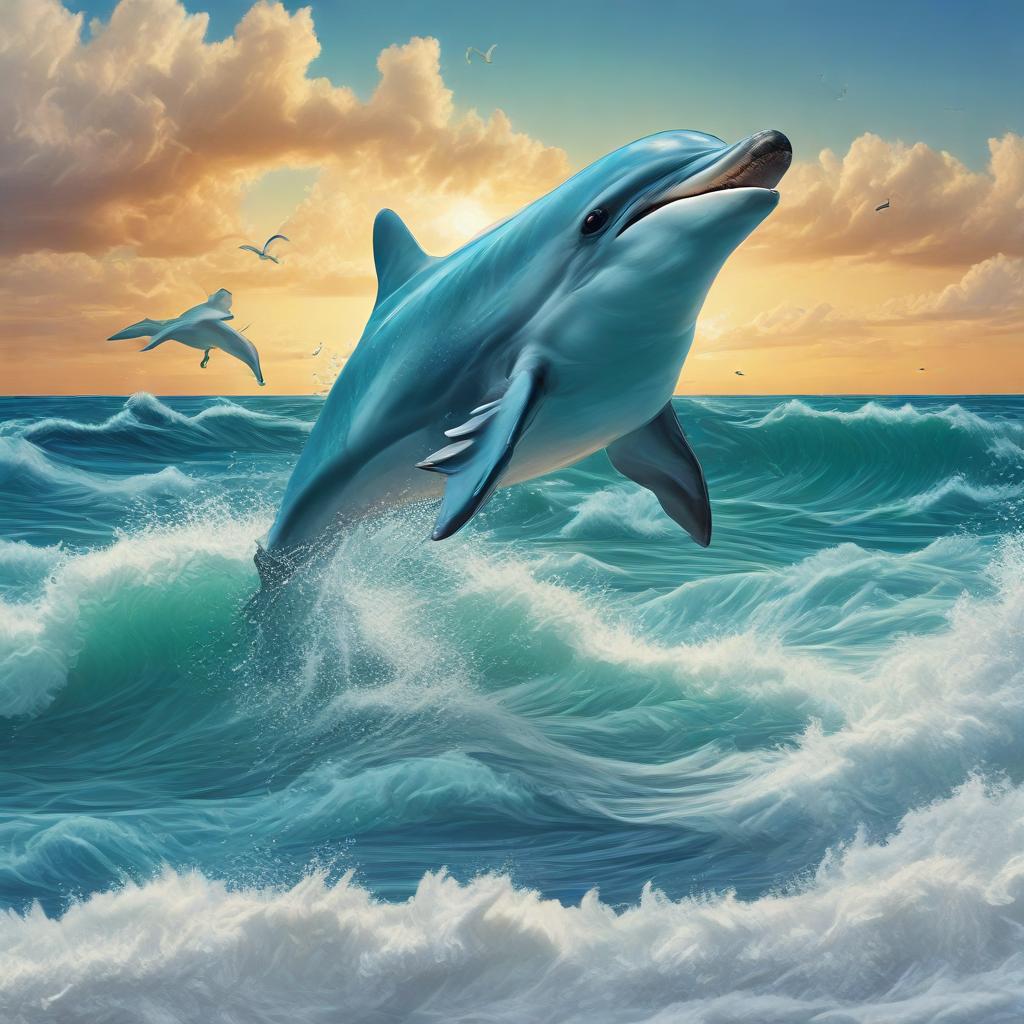} &
            \includegraphics[width=\linewidth]{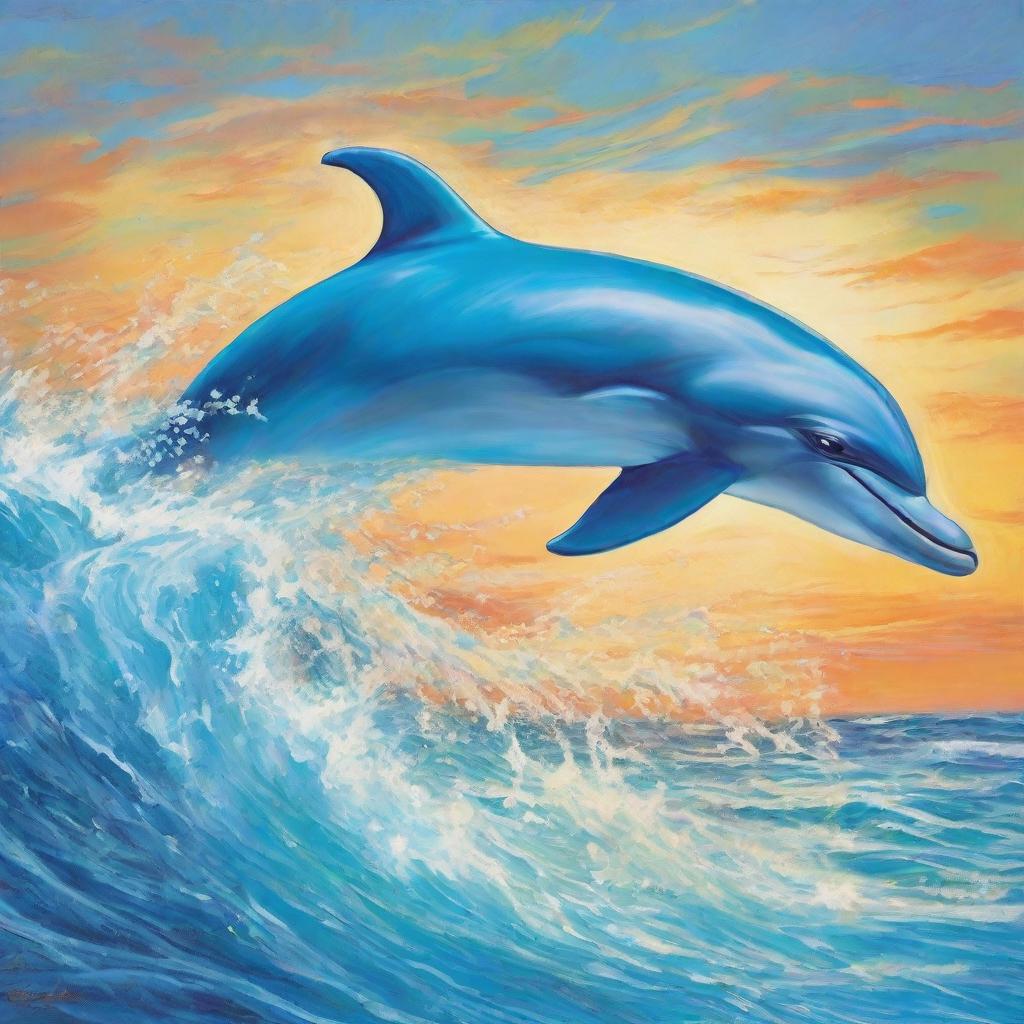} &
            \includegraphics[width=\linewidth]{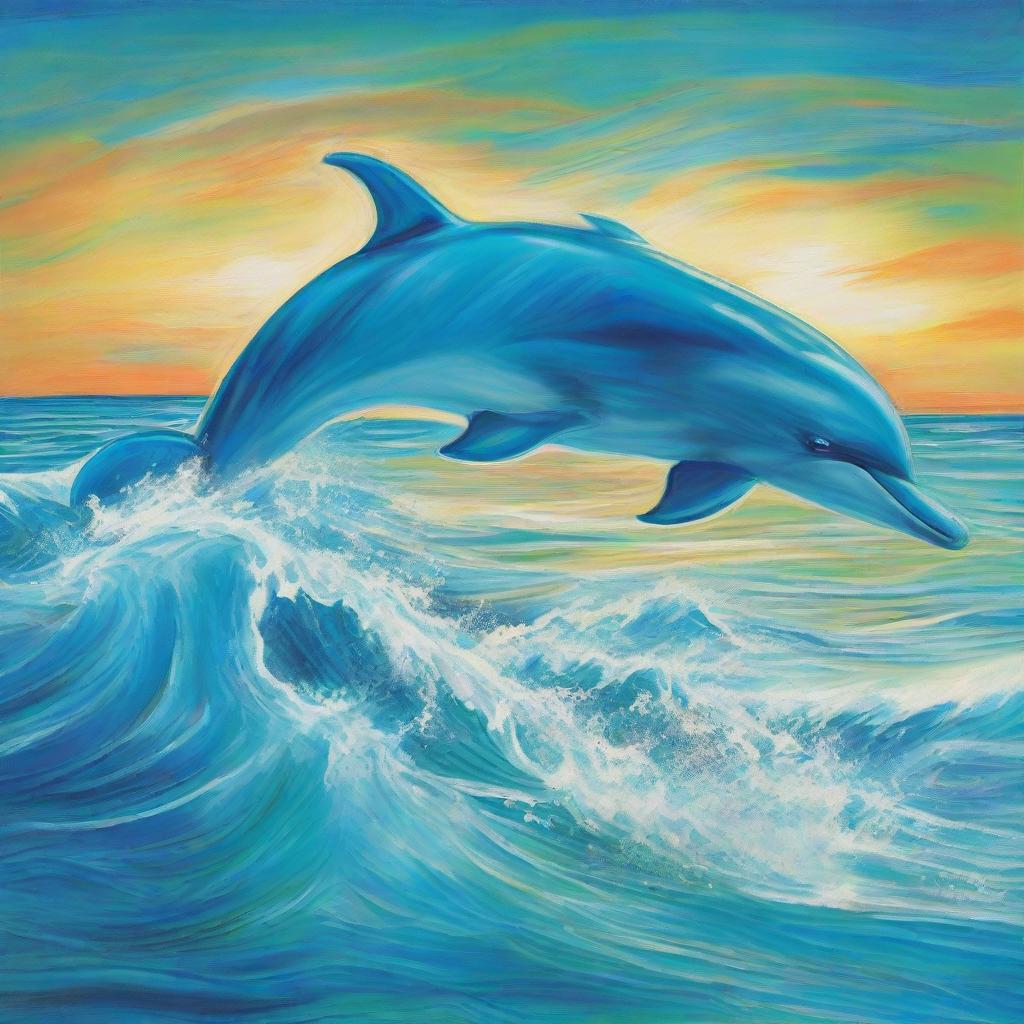}
        \end{tabularx}
        \vspace{-6pt}
        \caption{A dolphin leaps through the waves, set against a backdrop of bright blues and teal hues.}
    \end{subfigure}

    \begin{subfigure}[b]{\textwidth}
        \centering
        \setlength\tabcolsep{1pt}
        \begin{tabularx}{\textwidth}{@{}XX@{}X@{}X@{}X@{}X@{}XX@{}X@{}}
            \includegraphics[width=\linewidth]{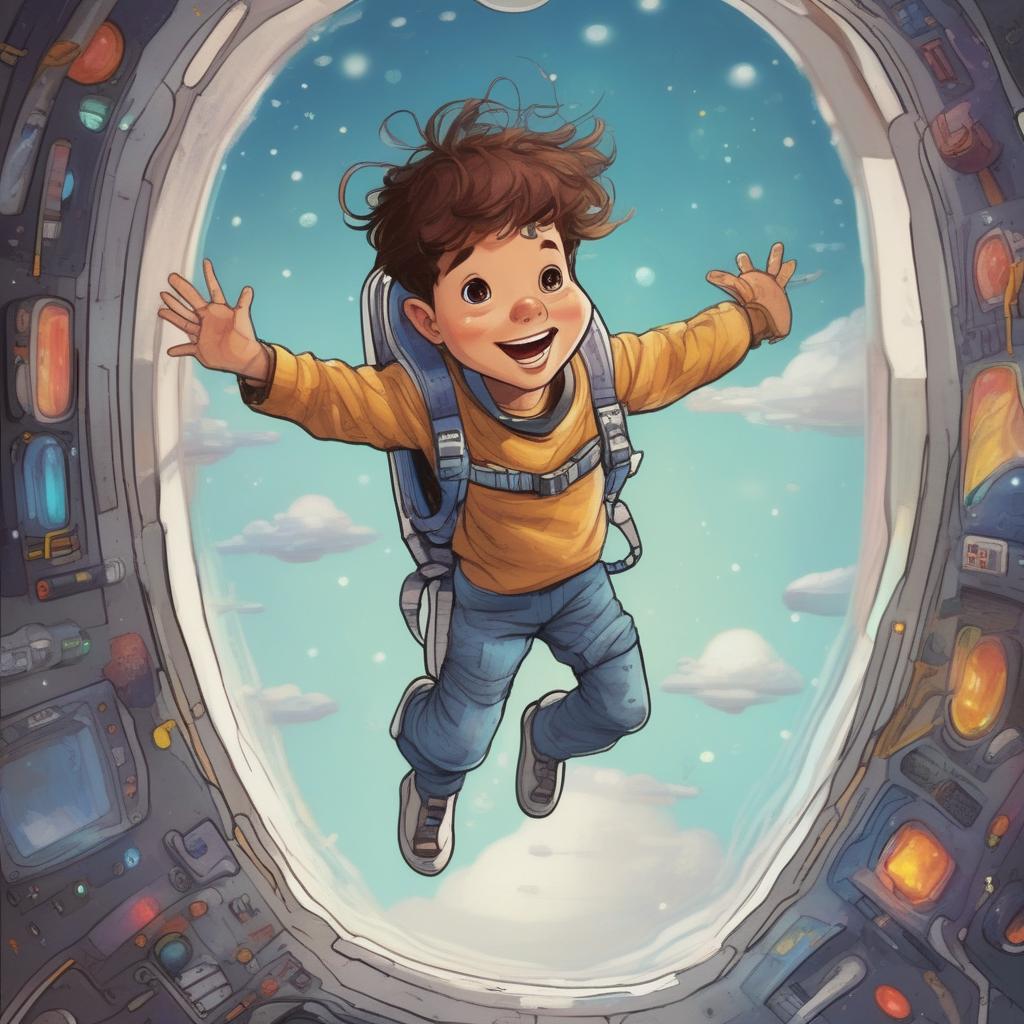} &
            \includegraphics[width=\linewidth]{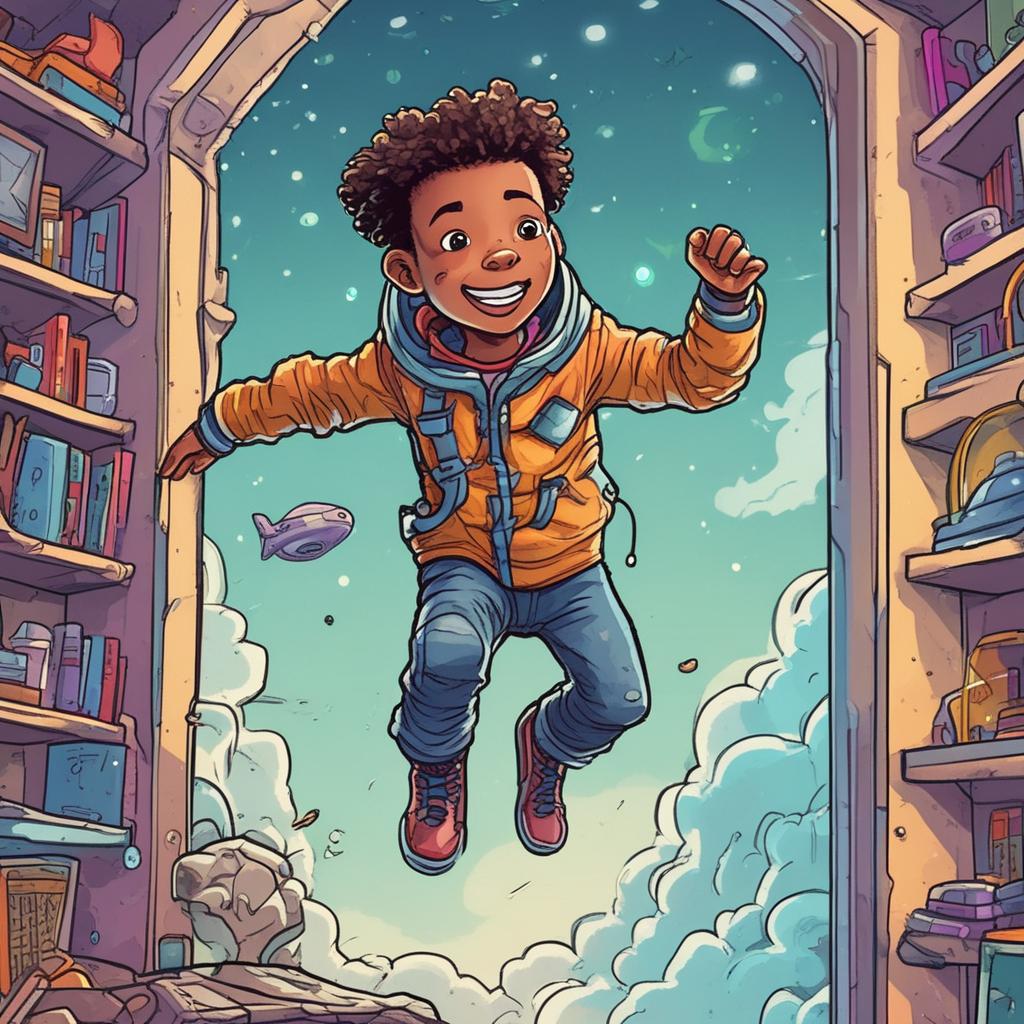} &
            \includegraphics[width=\linewidth]{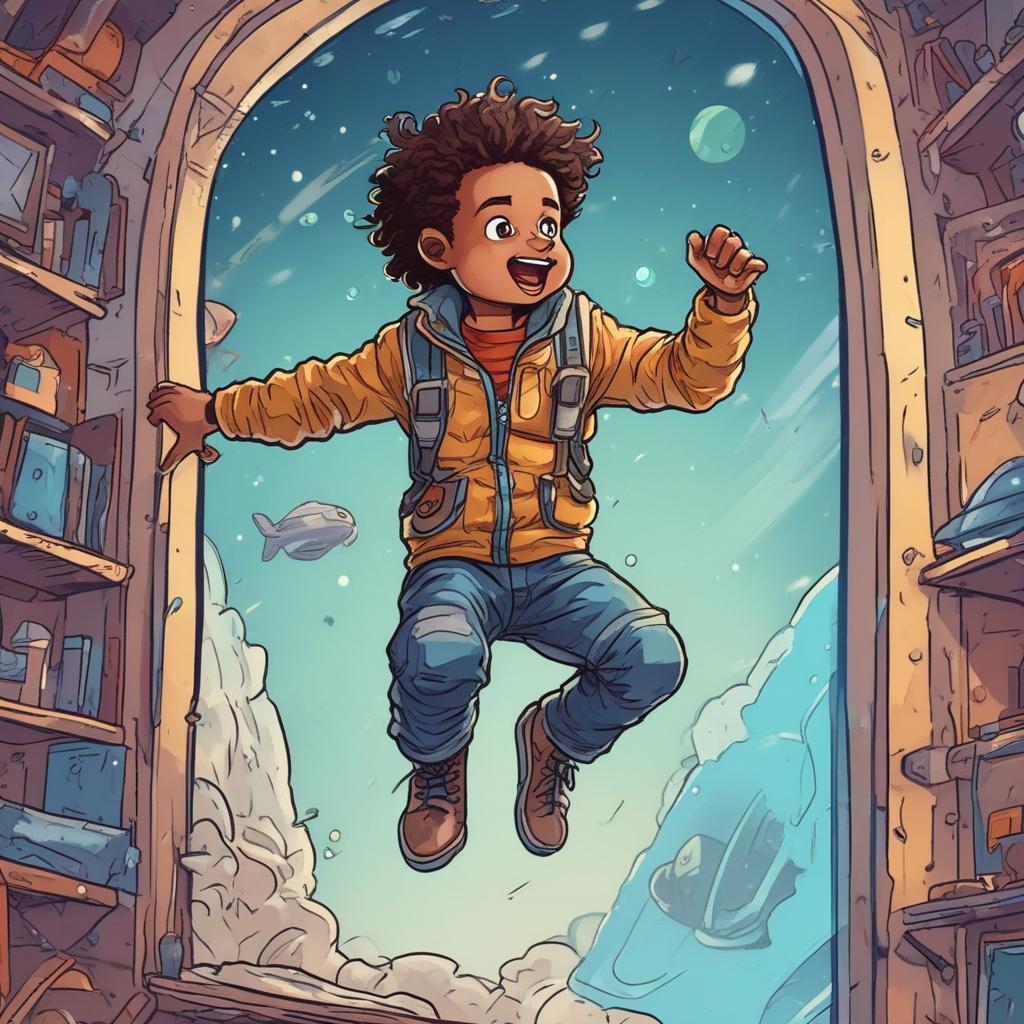} &
            \includegraphics[width=\linewidth]{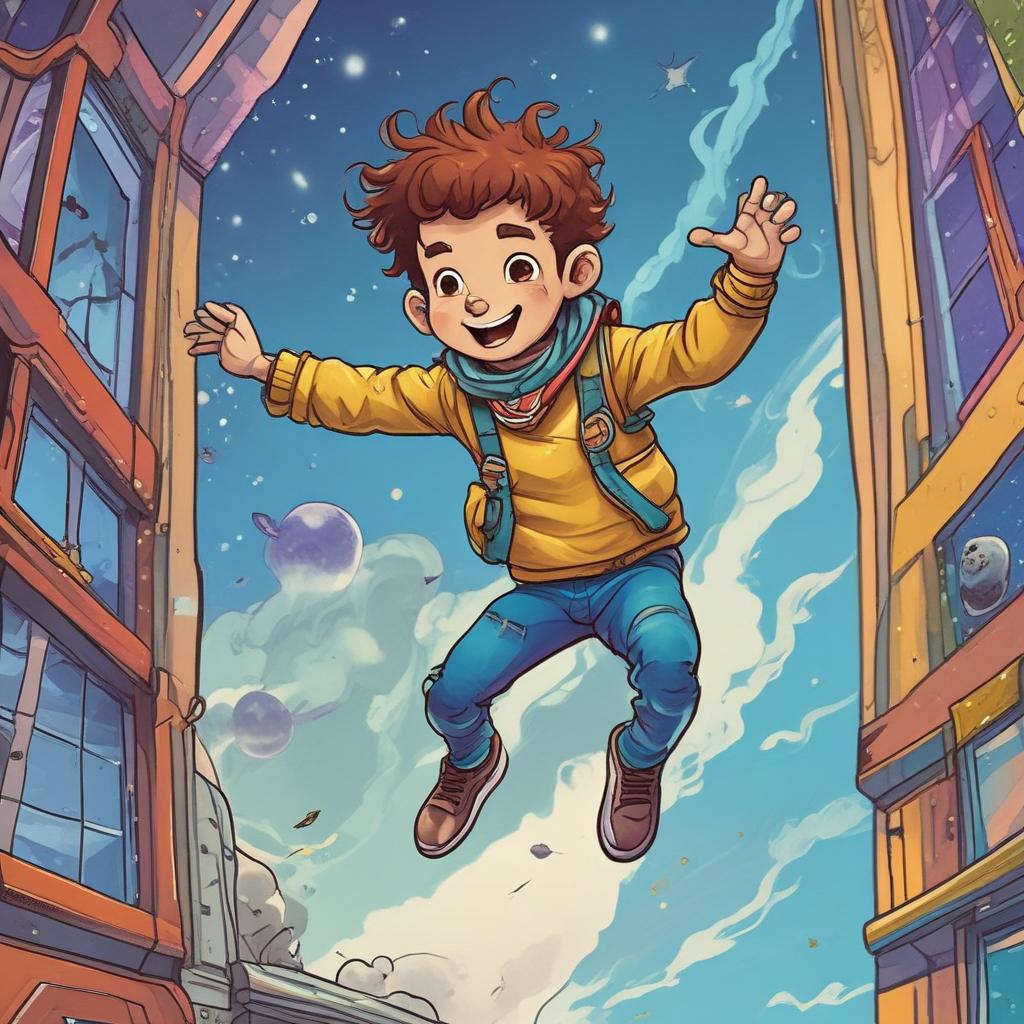} &
            \includegraphics[width=\linewidth]{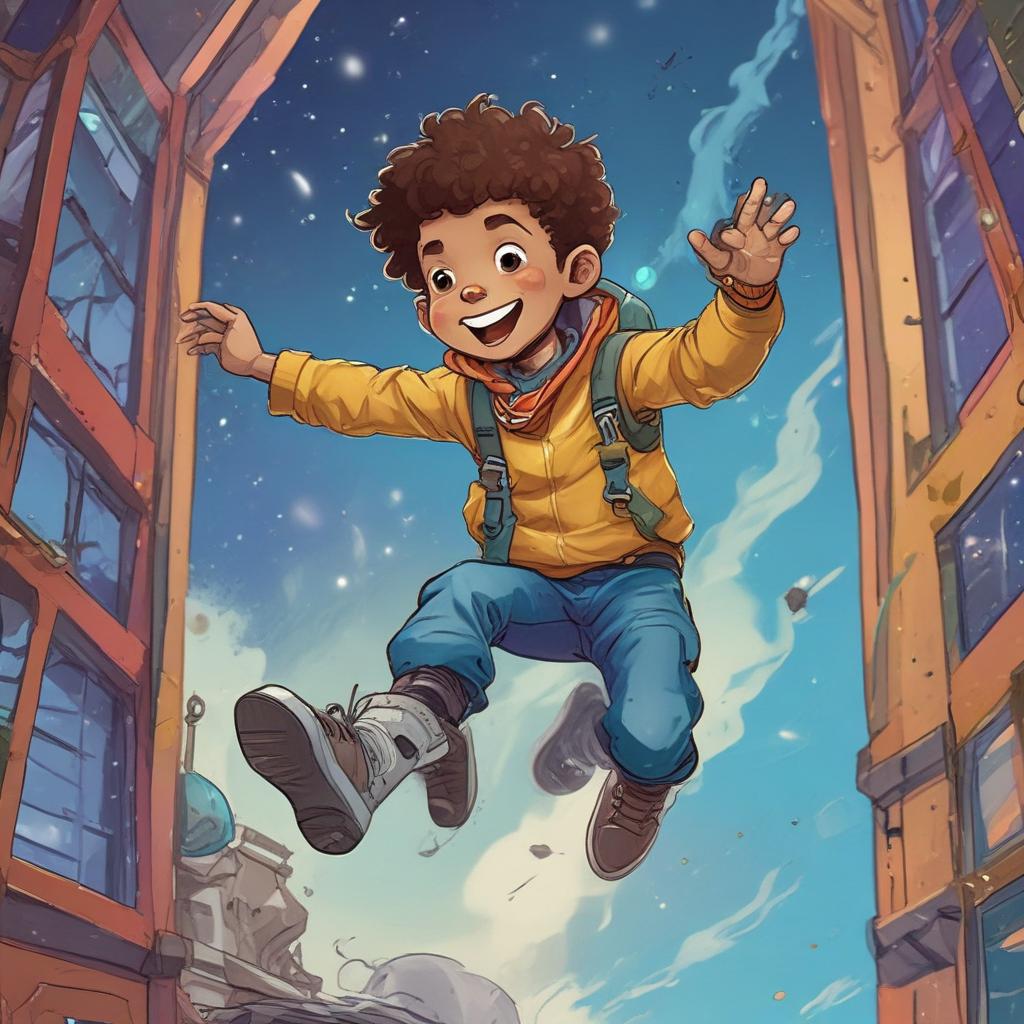} &
            \includegraphics[width=\linewidth]{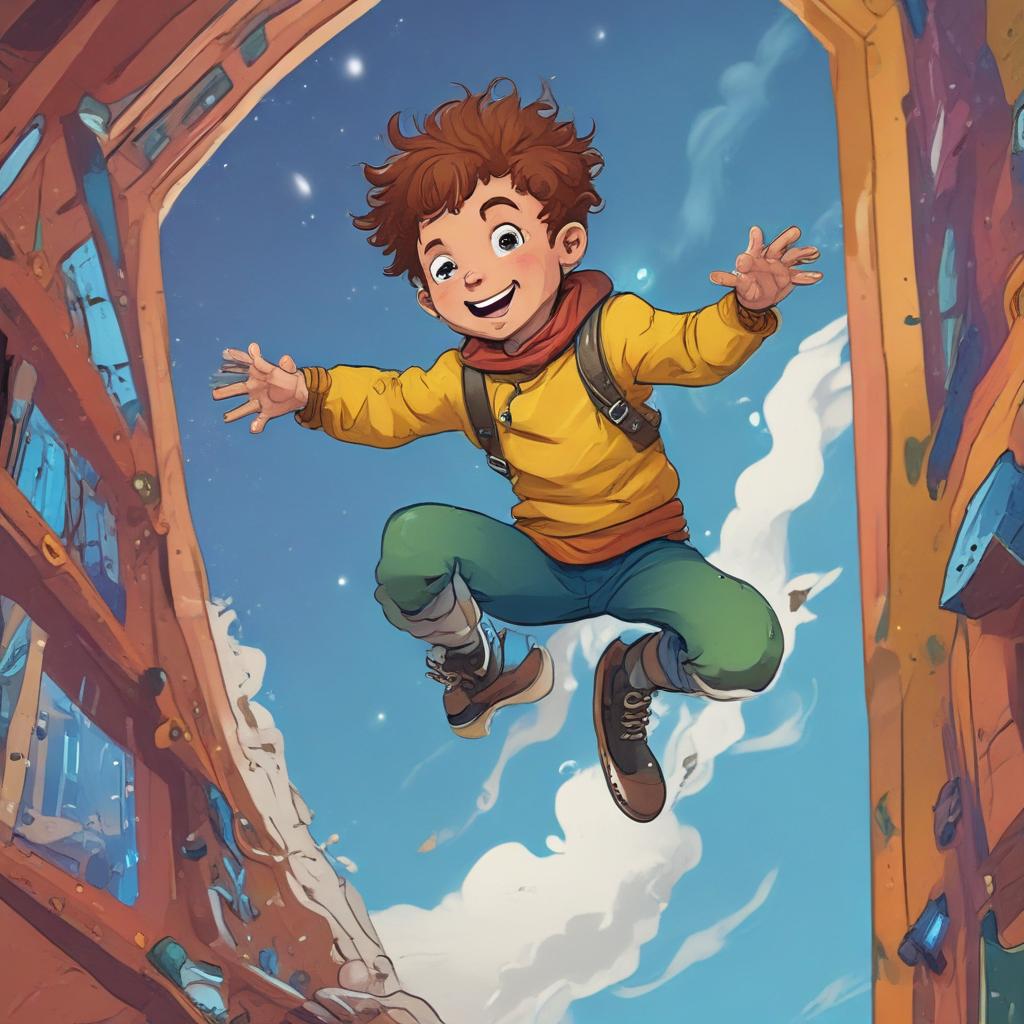} &
            \includegraphics[width=\linewidth]{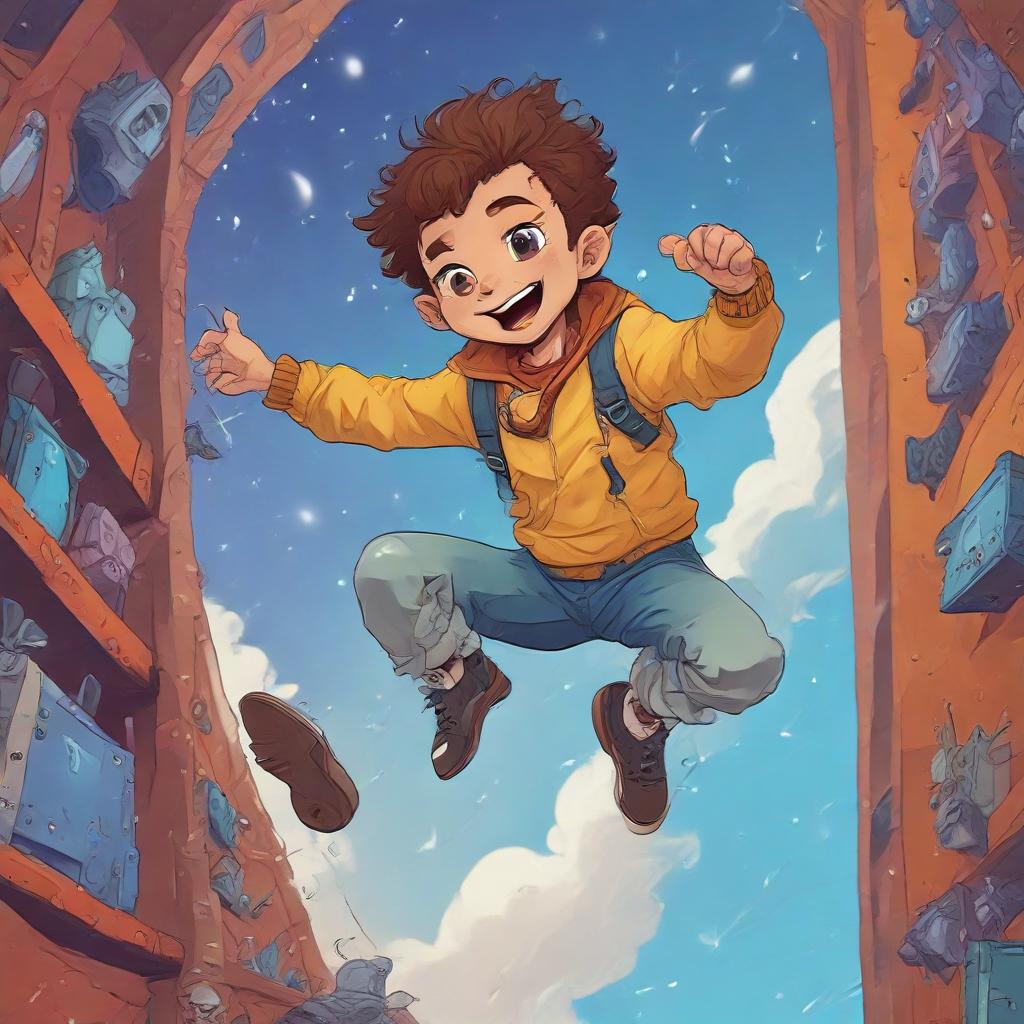} &
            \includegraphics[width=\linewidth]{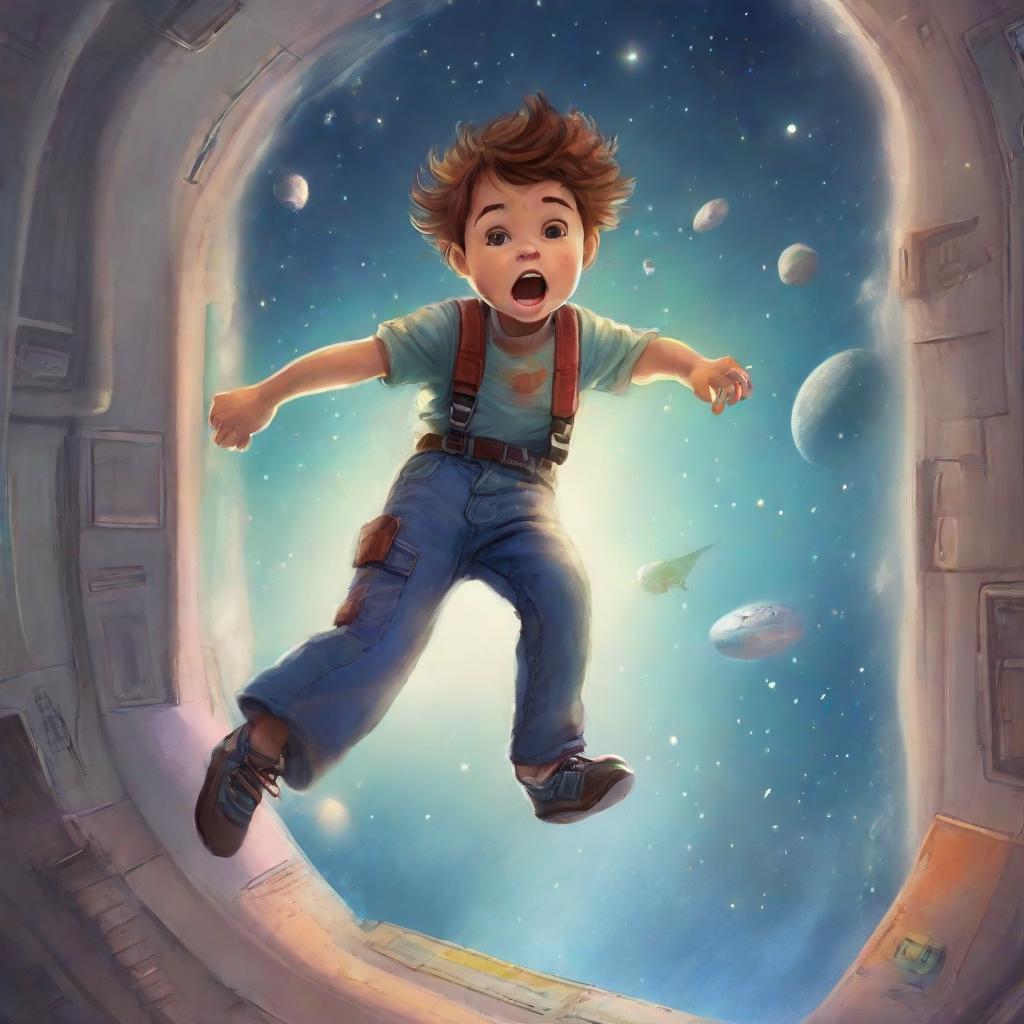} &
            \includegraphics[width=\linewidth]{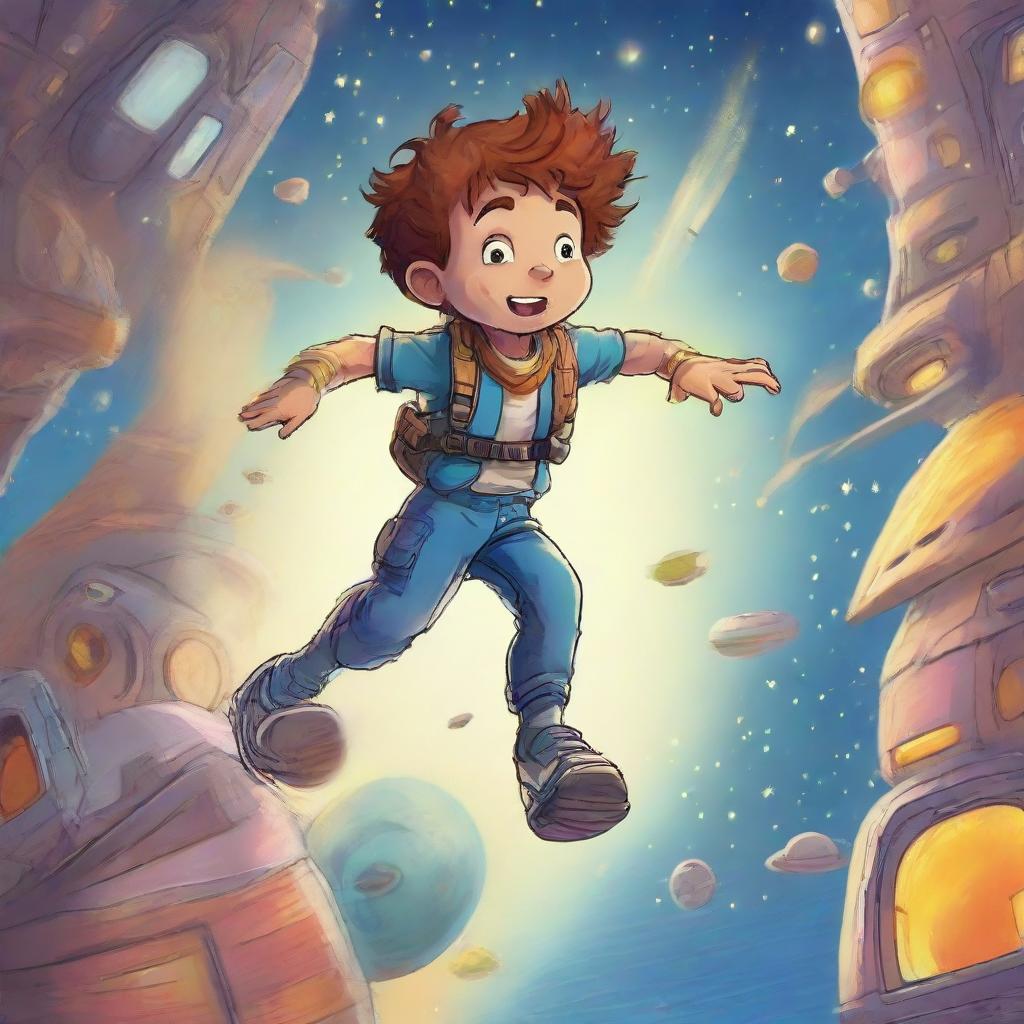}
        \end{tabularx}
        \vspace{-6pt}
        \caption{A boy jumping off a spaceship.}
    \end{subfigure}
    \vspace{-12pt}
    \caption{Comparison between fully-trained and LoRA-trained models. Training the full UNet has slightly better details, but our LoRA models are also very high quality. Note we do not provide 1-step LoRA. (Please zoom in to view at the full resolution.)}
    \label{fig:qualitative-lora}
    \vspace{-6pt}
\end{figure*}

\subsection{Qualitative Comparison}

\Cref{fig:qualitative} compares our method against other open-source distillation models: SDXL-Turbo \cite{sauer2023adversarial} and LCM \cite{luo2023latent}. Our method is substantially better in overall quality and details. Our method is also substantially better in the preservation of the style and layout of the original model. Furthermore, we find our 4-step and 8-step model can often outperform the original SDXL model for 32 steps. This is because our progressive distillation starts all the way from 128 steps.

\Cref{fig:qualitative-lora} compares our LoRA models against the fully trained models. We find that fully trained models have better structures and details. This is less noticeable on 8-step models, but more observable on 2-step models.

\subsection{Quantitative Comparison}

\begin{table}[t]
    \centering
    \setlength\tabcolsep{3pt}
    \begin{tabularx}{\linewidth}{Xcrrr}
        \toprule
        Method & Steps & \makecell{FID $\downarrow$ \\ \footnotesize{(Whole)}} & \makecell{FID $\downarrow$ \\ \footnotesize{(Patch)}} & CLIP $\uparrow$ \\
        \midrule
        SDXL \cite{podell2023sdxl} & 32 & 18.49 & 35.89 & 26.48 \\
        \midrule
        LCM \cite{luo2023latent} & 1 & 80.01 & 158.90 & 23.65 \\
        LCM \cite{luo2023latent} & 4 & 21.85 & 42.53 & 26.09\\
        LCM-LoRA \cite{luo2023lcmlora} & 4 & 21.50 & 40.38 & 26.18 \\
        \midrule
        \multirow{2}{*}{Turbo \cite{sauer2023adversarial}} & 1 & 23.71 & 43.69 & 26.36 \\
         & 4 & 22.58 & 42.65 & 26.18 \\
        \midrule
        \multirow{4}{*}{Ours} & 1 & 22.61 & \textbf{41.53} & 26.02 \\
        & 2 & 23.11 & \textbf{35.12} & 25.98 \\
        & 4 & 22.30 & \textbf{33.52} & 26.07 \\
        & 8 & 21.43 & \textbf{33.55} & 25.86 \\
        \midrule
        \multirow{3}{*}{Ours-LoRA} & 2 & 23.39 & \textbf{40.54} & 26.18 \\
        & 4 & 23.01 & \textbf{34.10} & 26.04 \\
        & 8 & 22.30 & \textbf{33.92} & 25.77 \\
        \bottomrule
    \end{tabularx}
    \vspace{-6pt}
    \caption{Quantitative comparison. FID-Whole reflects high-level diversity and quality. FID-Patch reflects high-resolution details. CLIP score reflects text-alignment. Our models have significantly better high-resolution details while retaining similar performance in diversity and text alignment.}
    \label{tab:quantitative}
    \vspace{-10pt}
\end{table}

\Cref{tab:quantitative} shows Fréchet Inception Distance (FID) \cite{heusel2018gans,parmar2022aliased} and CLIP score \cite{radford2021learning}. Following the convention, we generate images using the first 10K prompts from the COCO \cite{lin2015microsoft} validation dataset. FID metric is computed against the corresponding ground truth images from COCO.

FID is normally computed by resizing the whole image to 299px for the InceptionV3 network \cite{szegedy2015rethinking}. This only assesses the high-level sample quality and diversity. The metric shows that our model achieves similar performance as other distillation techniques. All distillation methods have worse FID compared to the original SDXL likely due to the reduction in diversity.

We additionally propose to calculate FID on patches of images to assess high-resolution details. Specifically, we calculate FID on the 299px center-cropped patch of every image. For Turbo, we resize the 512px to 1024px before the crop for a fair comparison. The metric shows that our models have much better high-resolution details compared to other methods. Additionally, the metric shows that our model has better high-resolution details compared to original SDXL models for 32 steps because our distillation starts from 128 steps. It also shows that the quality degrades as the number of inference steps decreases.

CLIP score shows that our method achieves similar text-alignment performance compared to other methods.

\section{Ablation}

\subsection{Apply LoRA on Other Base Models}

\begin{figure}[h]
    \centering
    \footnotesize
    \setlength\tabcolsep{4pt}
    \begin{tabularx}{\linewidth}{|X|X|X|X|X|}
        SDXL & New Base & \multicolumn{3}{l|}{New Base + Our LoRA} \\
        32 Steps & 32 Steps &  8 Steps & 4 Steps & 2 Steps
    \end{tabularx}
    \setlength\tabcolsep{1pt}
    \begin{tabularx}{\linewidth}{@{}XX@{}X@{}X@{}X@{}}
        \includegraphics[width=\linewidth]{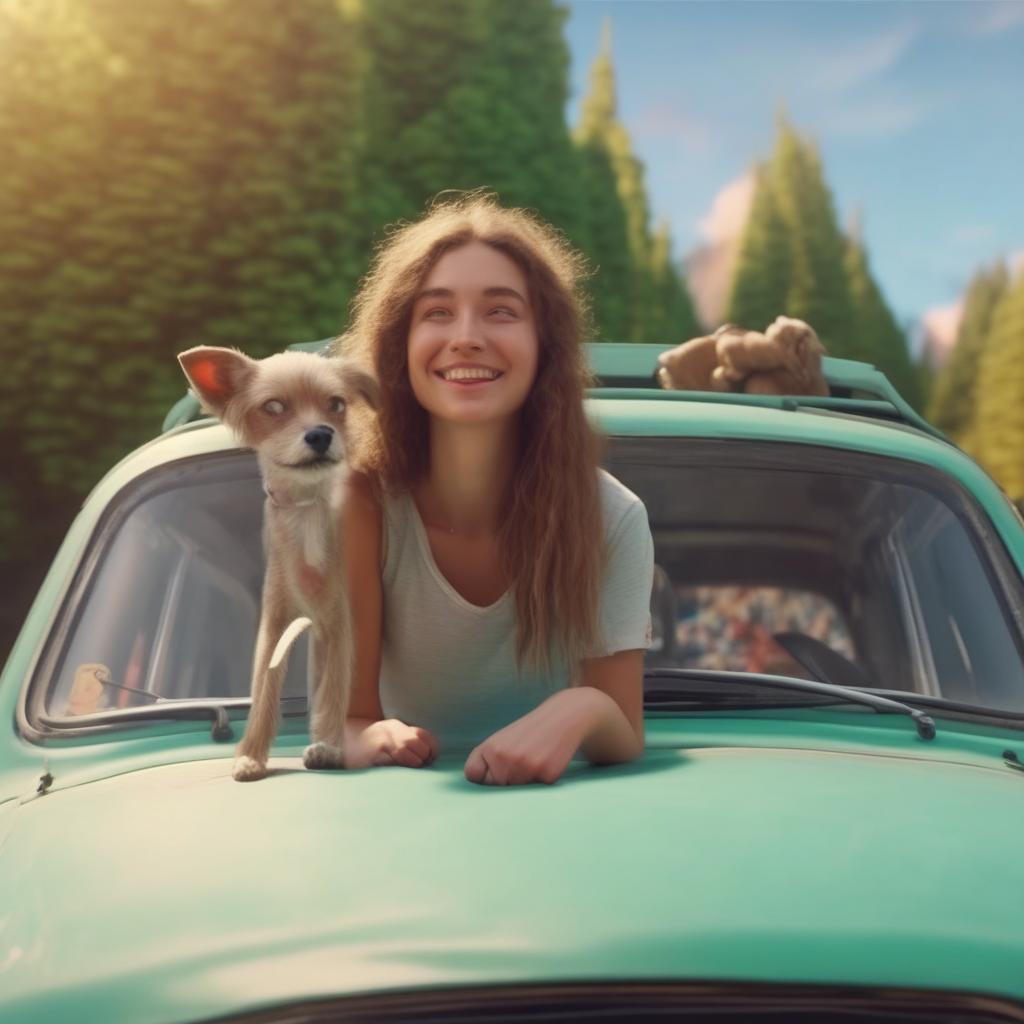} &
        \includegraphics[width=\linewidth]{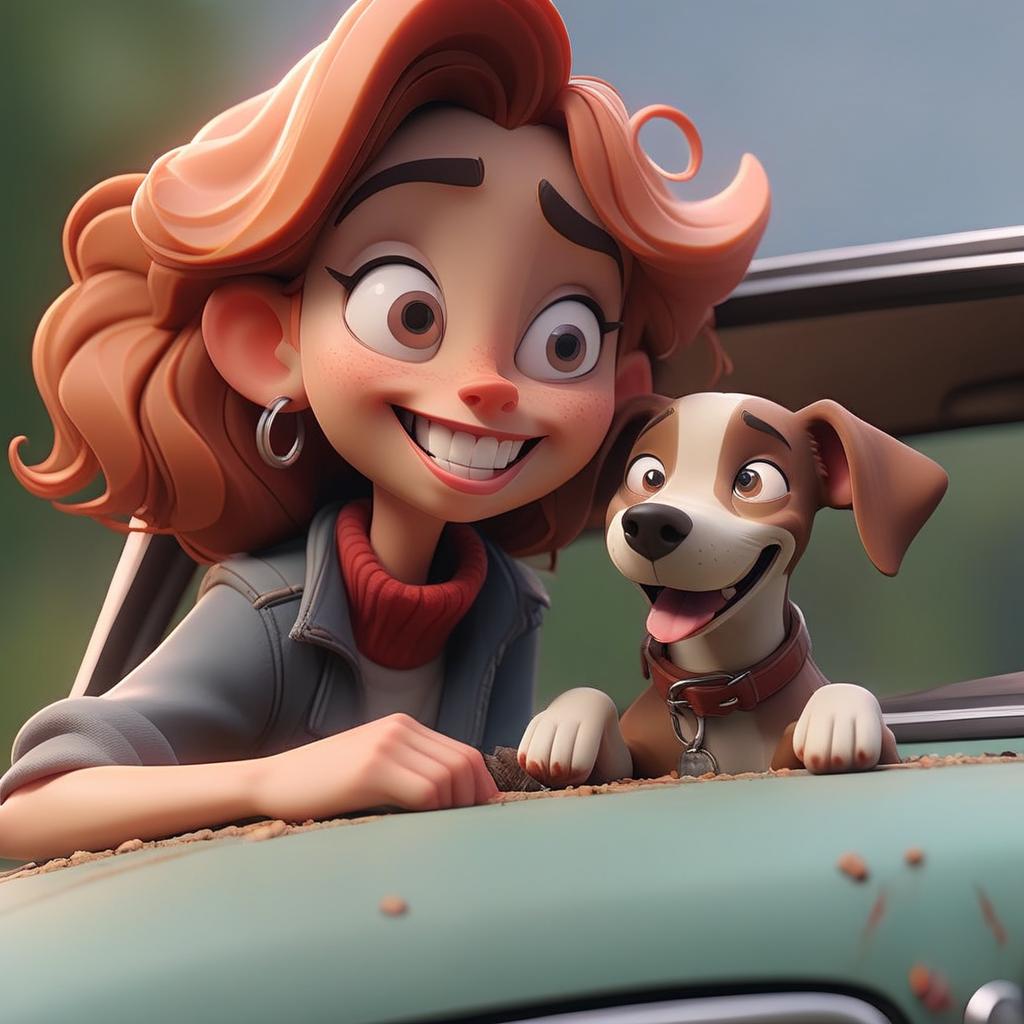} &
        \includegraphics[width=\linewidth]{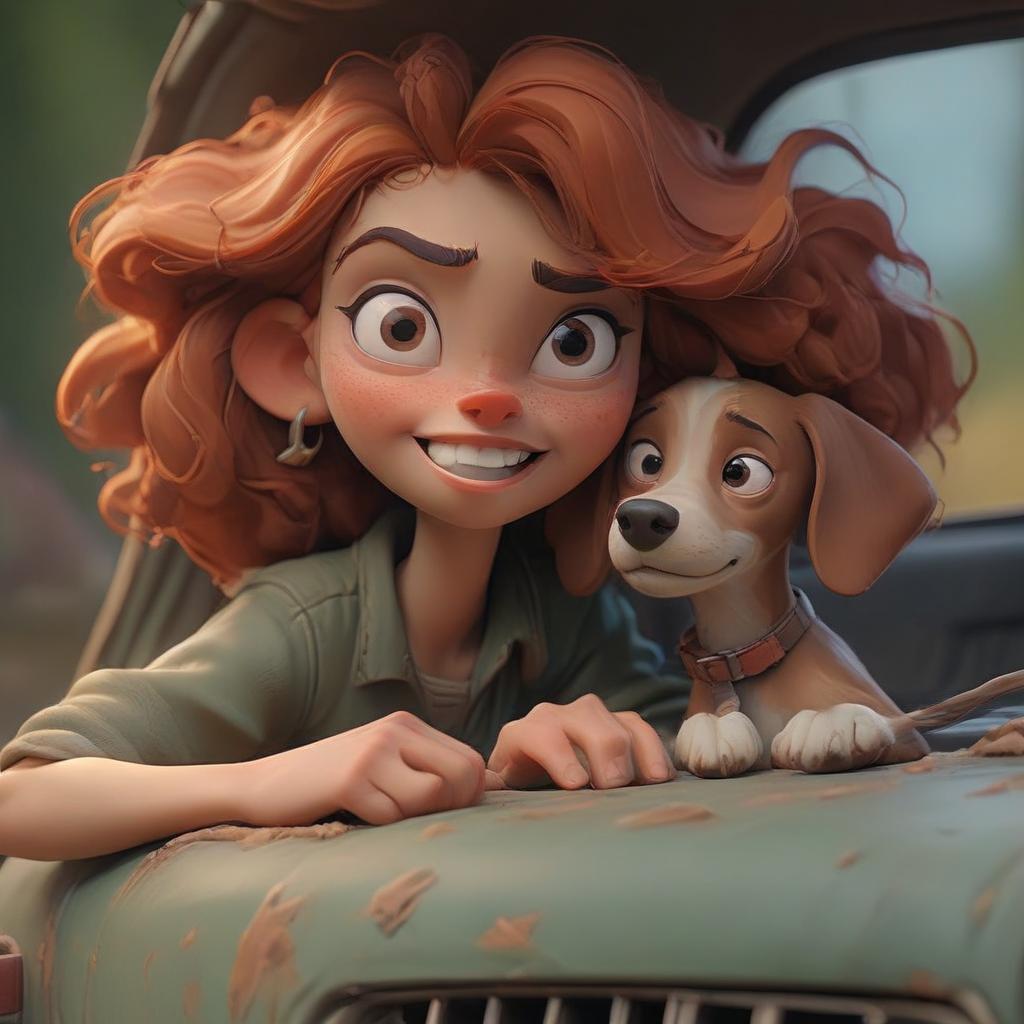} &
        \includegraphics[width=\linewidth]{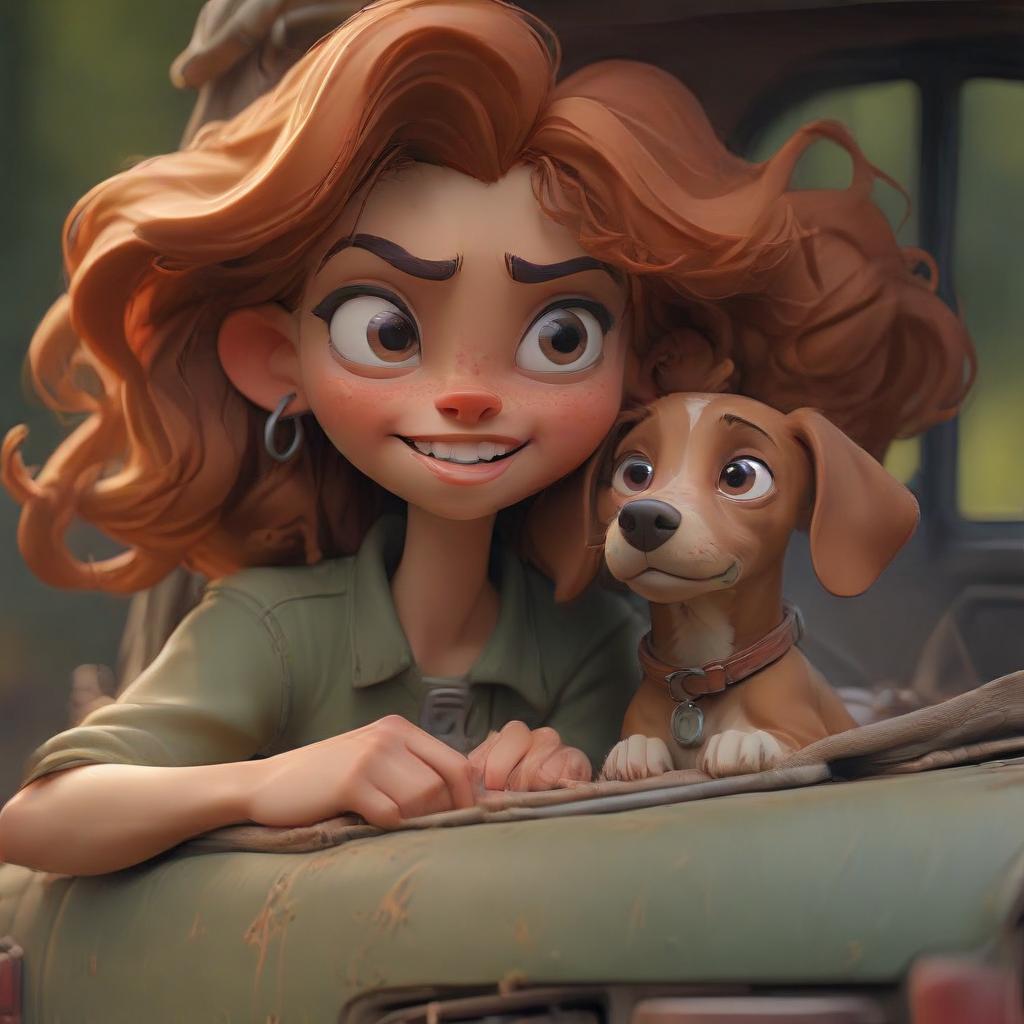} &
        \includegraphics[width=\linewidth]{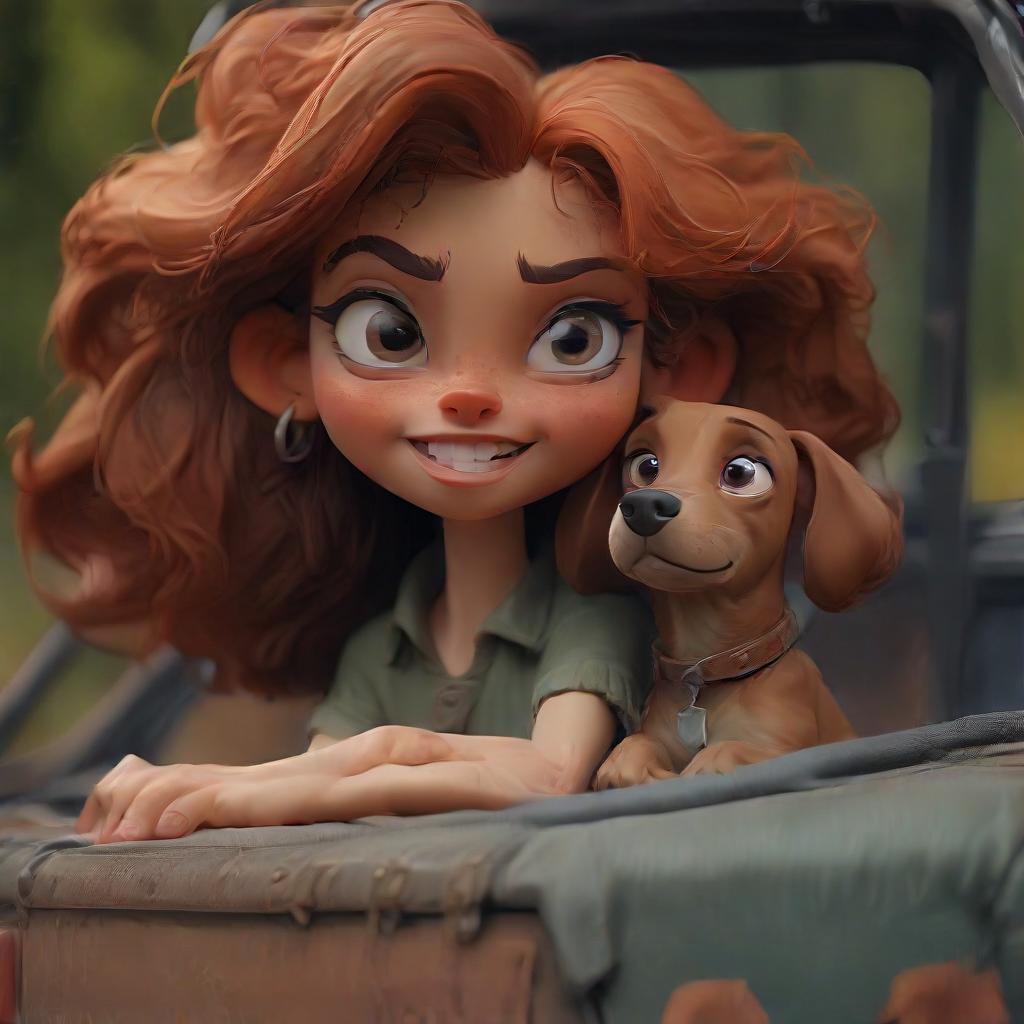} \\
        \includegraphics[width=\linewidth]{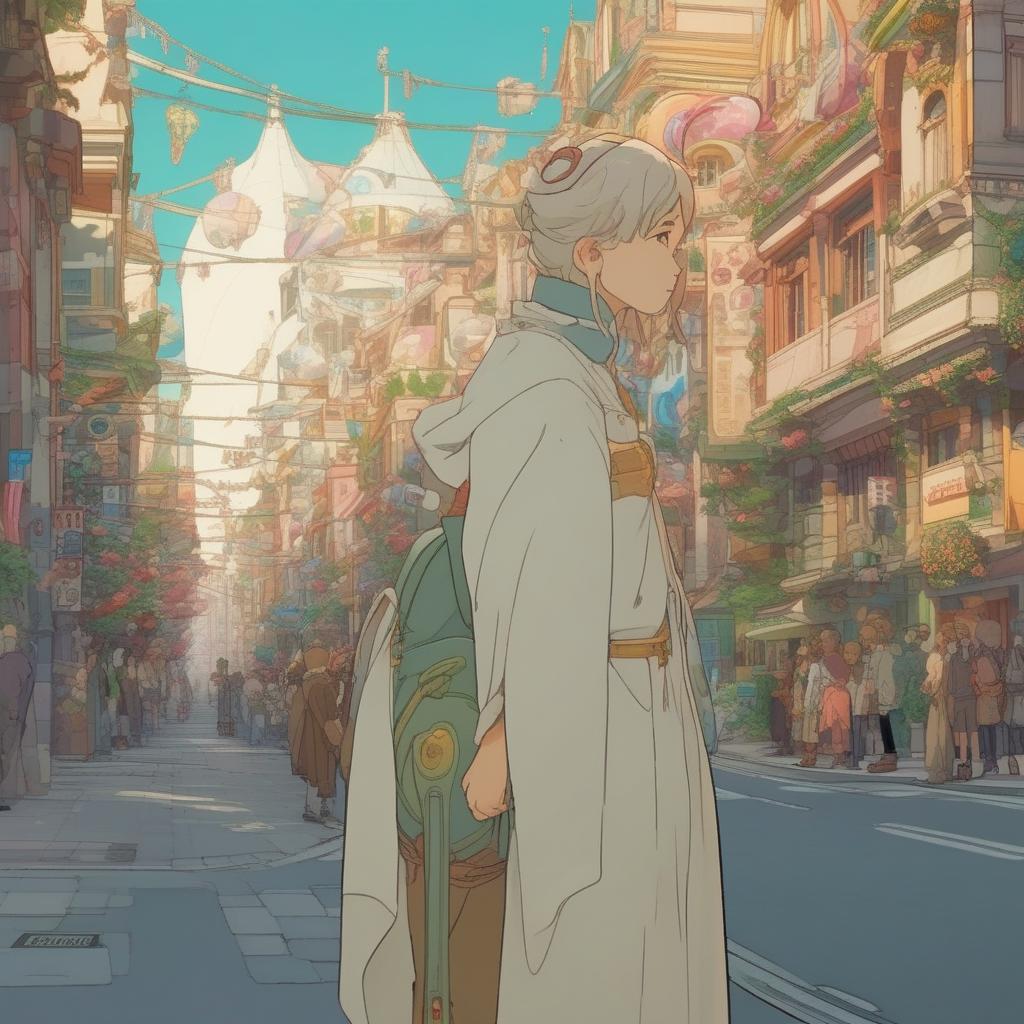} &
        \includegraphics[width=\linewidth]{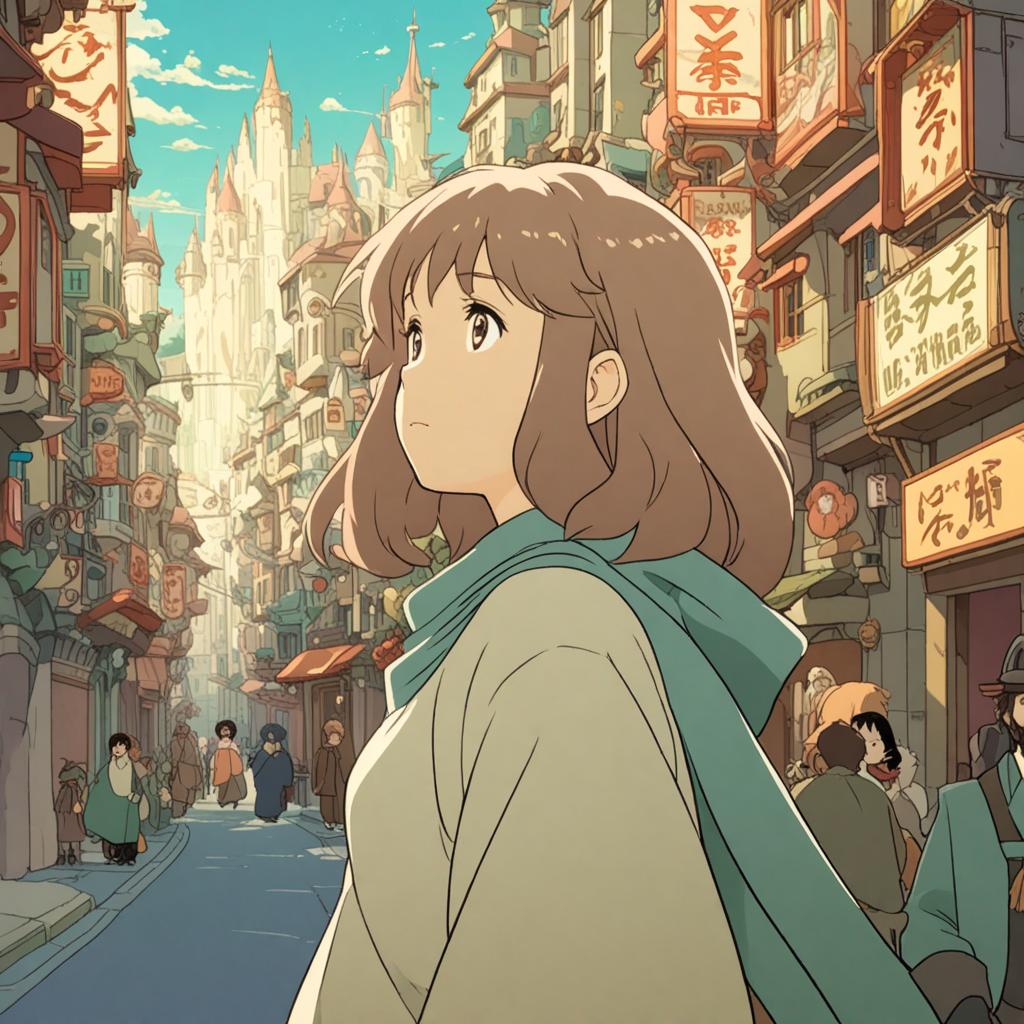} &
        \includegraphics[width=\linewidth]{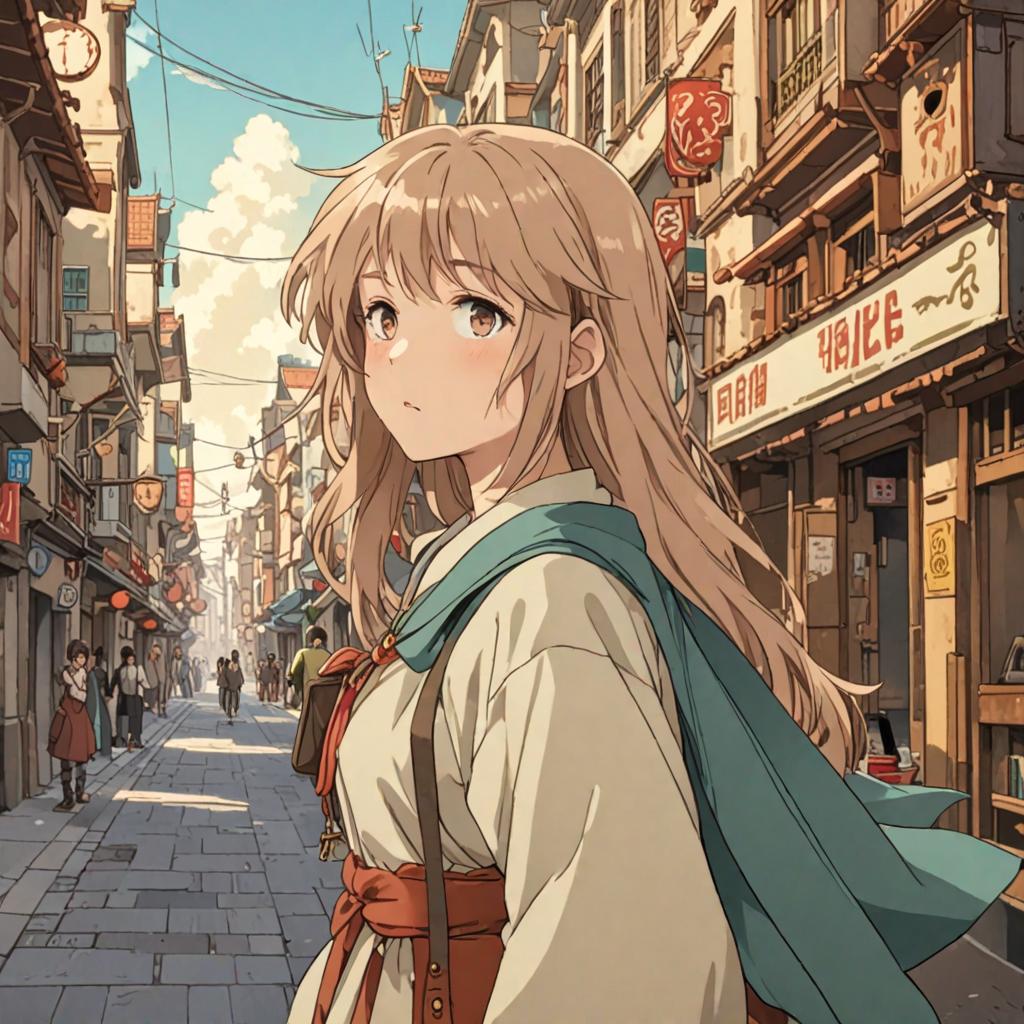} &
        \includegraphics[width=\linewidth]{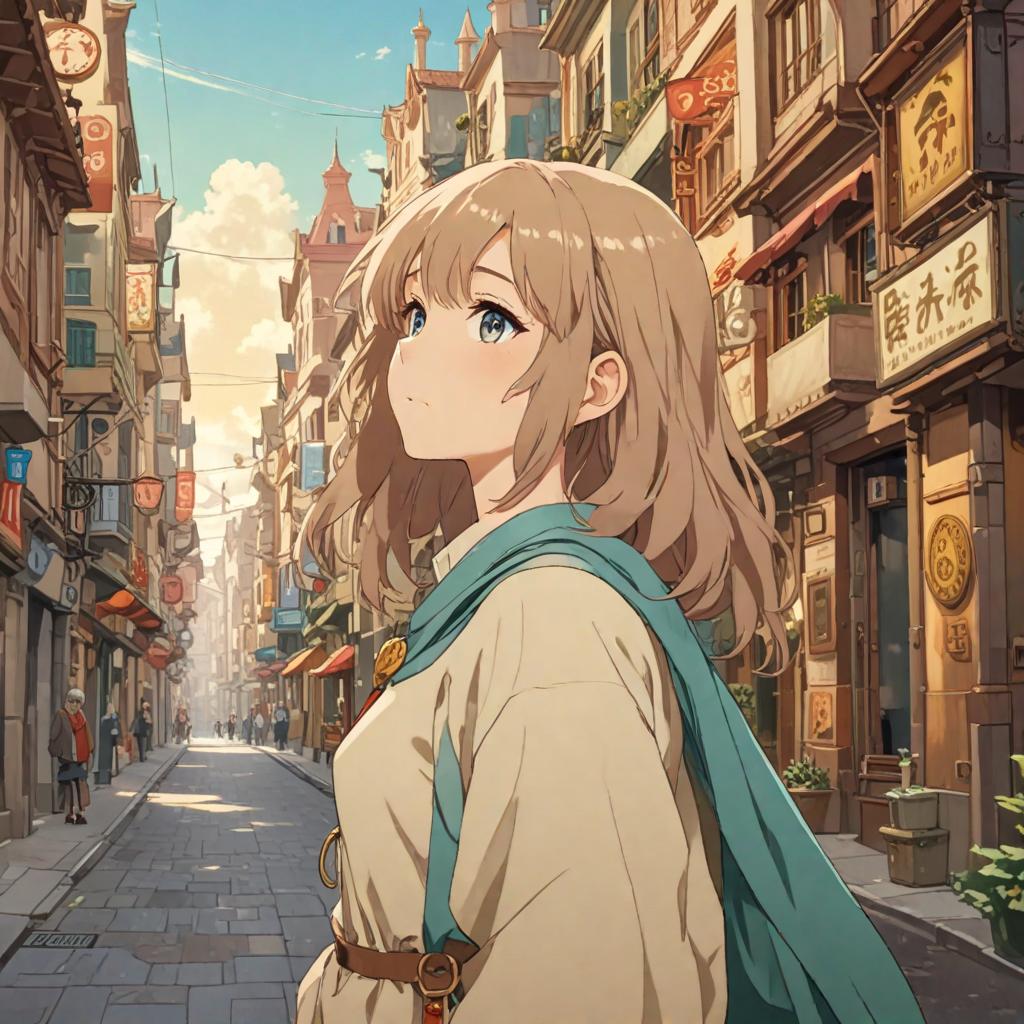} &
        \includegraphics[width=\linewidth]{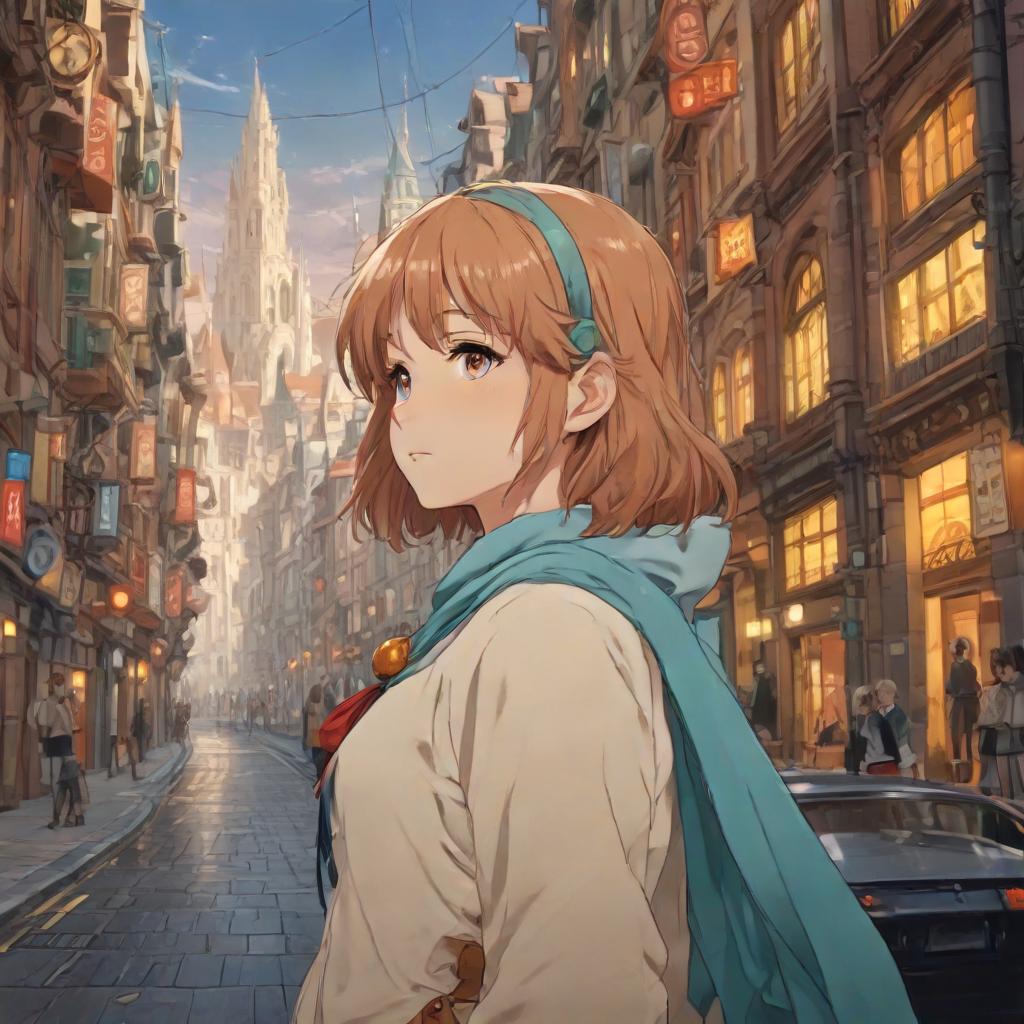} \\
        \includegraphics[width=\linewidth]{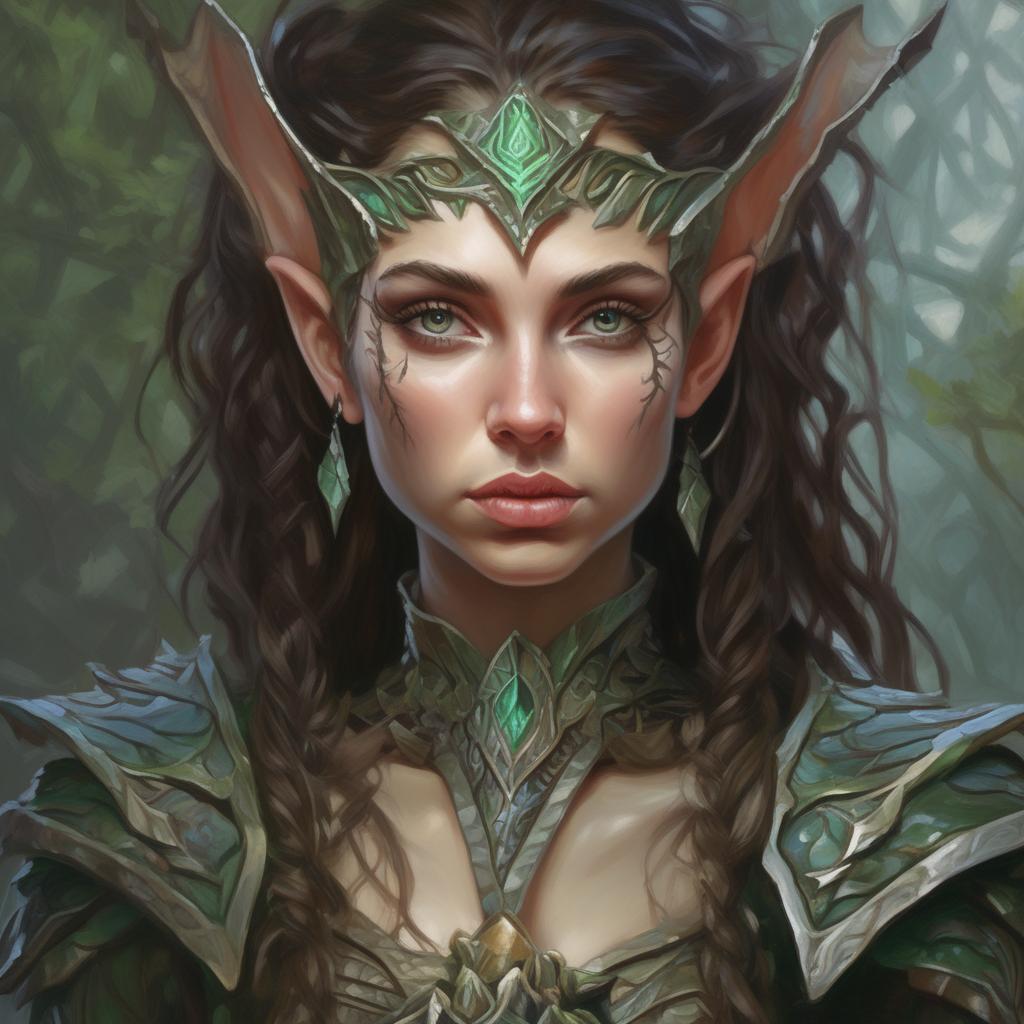} &
        \includegraphics[width=\linewidth]{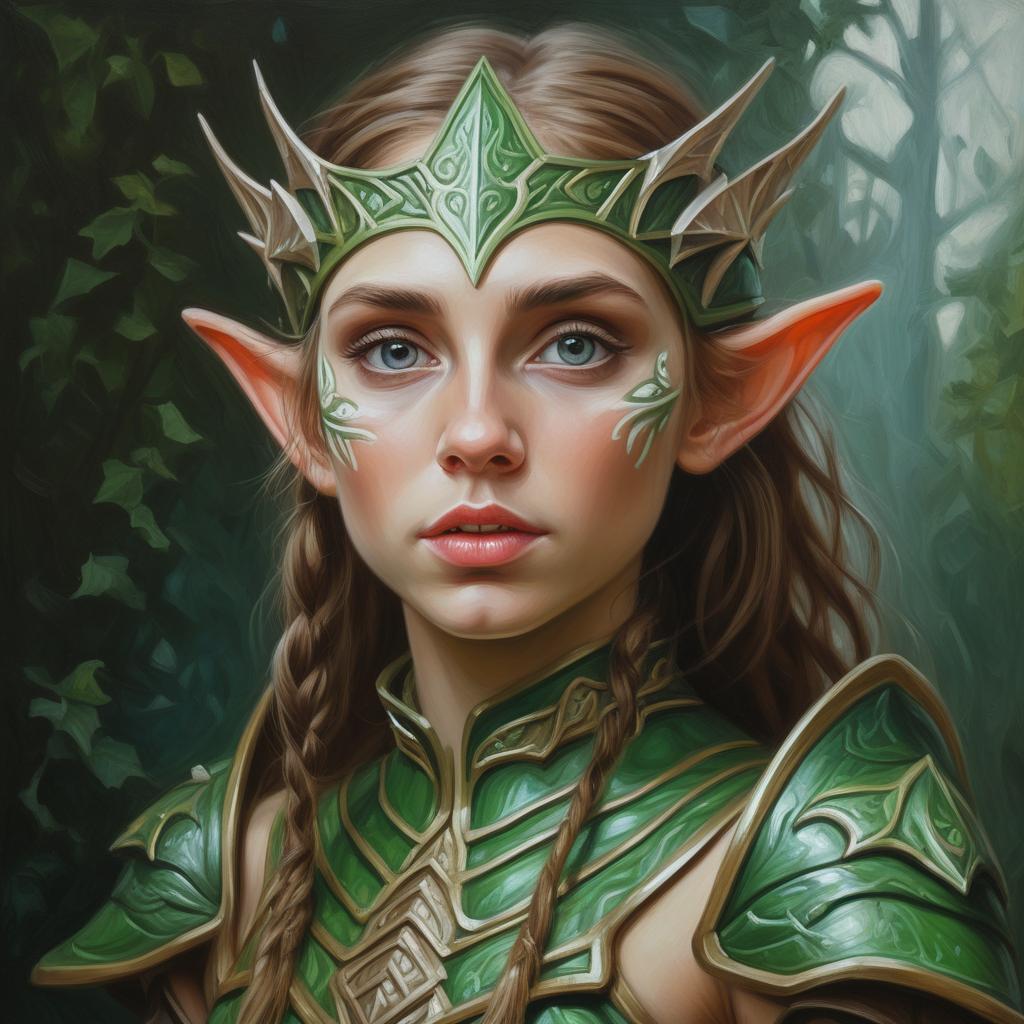} &
        \includegraphics[width=\linewidth]{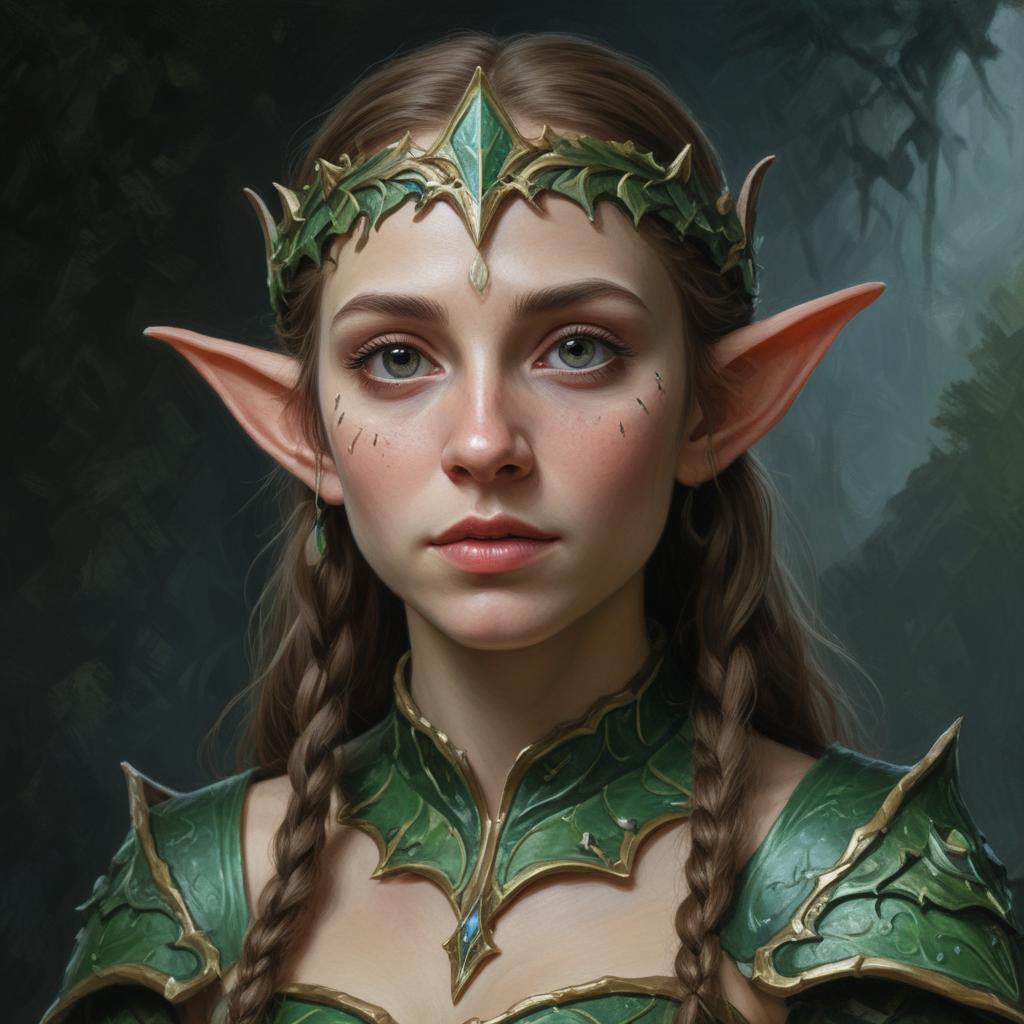} &
        \includegraphics[width=\linewidth]{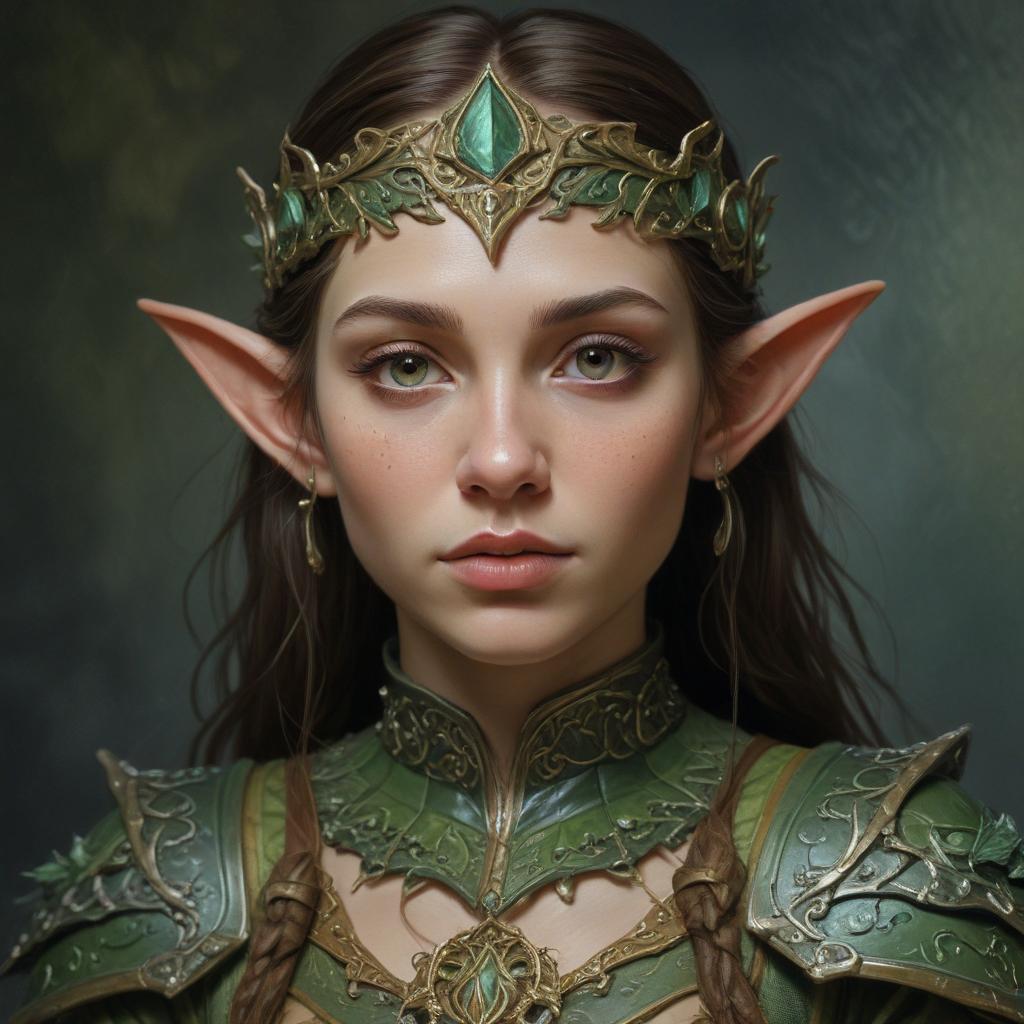} &
        \includegraphics[width=\linewidth]{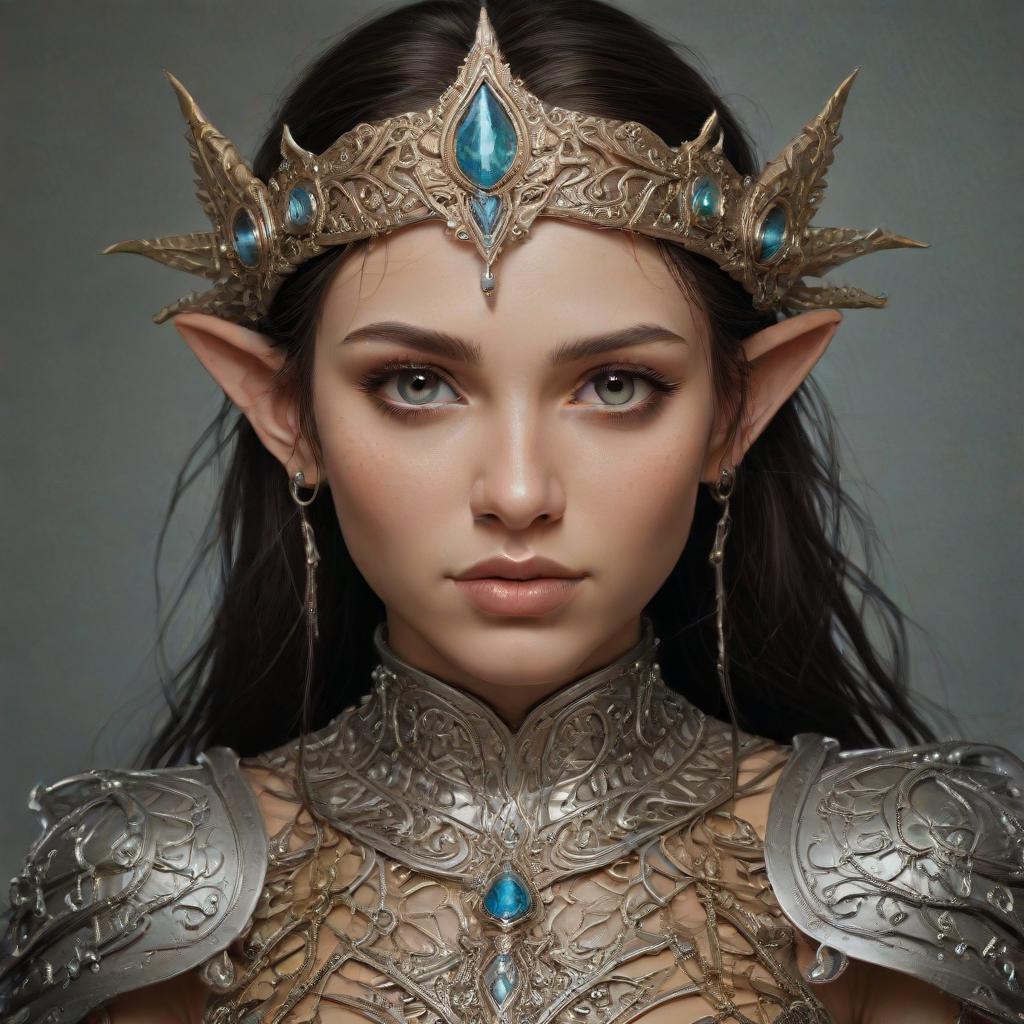}
    \end{tabularx}
    \vspace{-6pt}
    \caption{Our distillation LoRA can be applied to other base models, \eg cartoon \cite{samaritan}, anime \cite{aamxl}, and realistic \cite{realvisxl} base models.}
    \label{fig:lora-base}
\end{figure}

\Cref{fig:lora-base} shows that our distillation LoRA model can be applied to different base models. Specifically, we test it on third-party cartoon \cite{samaritan}, anime \cite{aamxl}, and realistic \cite{realvisxl} base models. Our distillation LoRAs are able to keep the style and layout of the new base model to a great extent.

\subsection{Inference with Different Aspect Ratios}

\begin{figure}[h]
    \centering
    \setlength\tabcolsep{1pt}
    \begin{tabularx}{\linewidth}{XXXXXXX}
        \includegraphics[width=\linewidth]{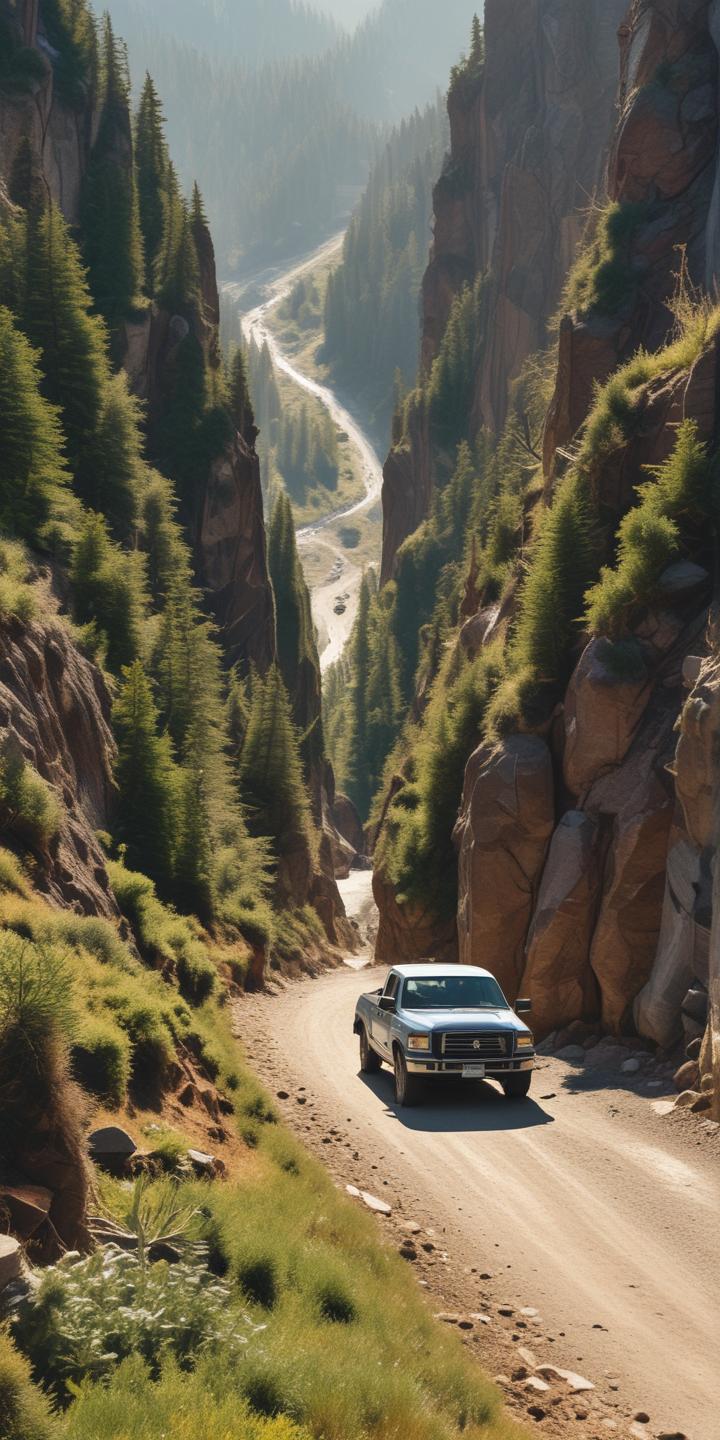} &
        \includegraphics[width=\linewidth]{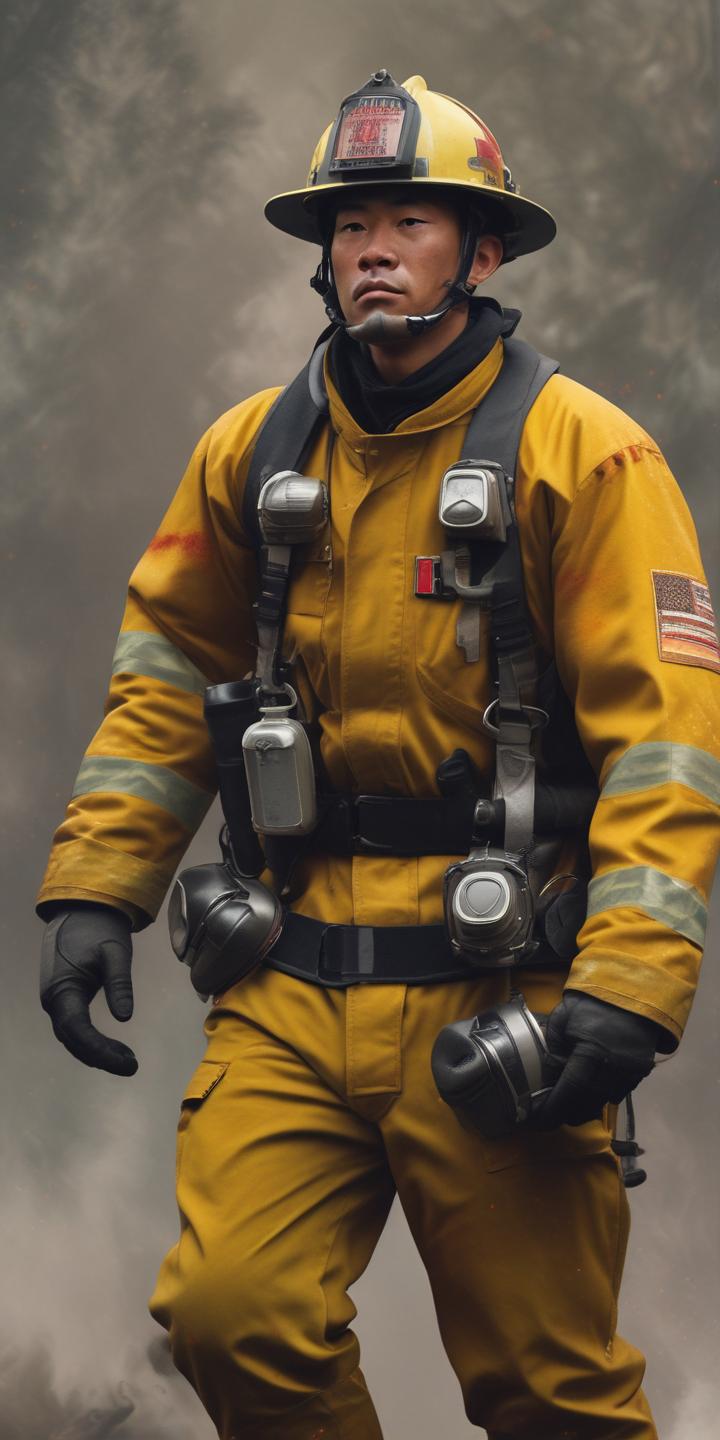} &
        \includegraphics[width=\linewidth]{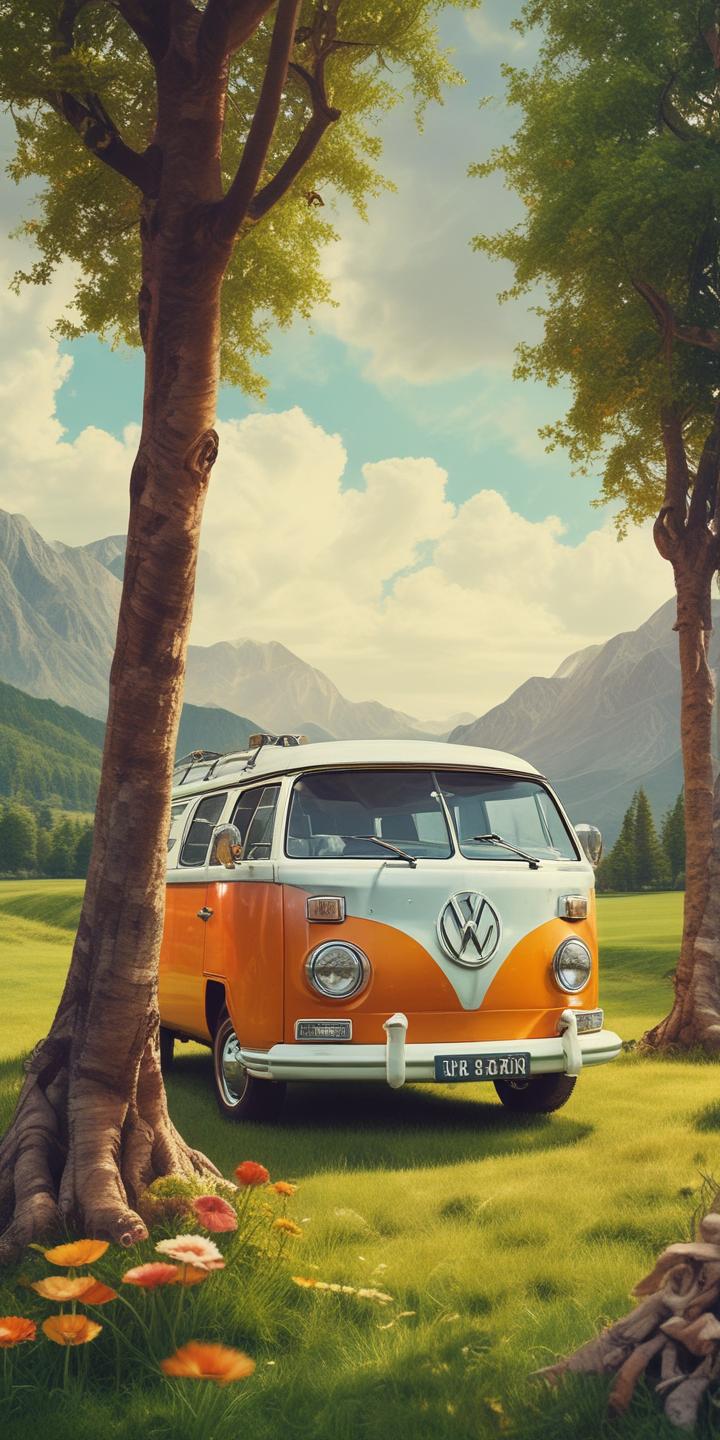} &
        \includegraphics[width=\linewidth]{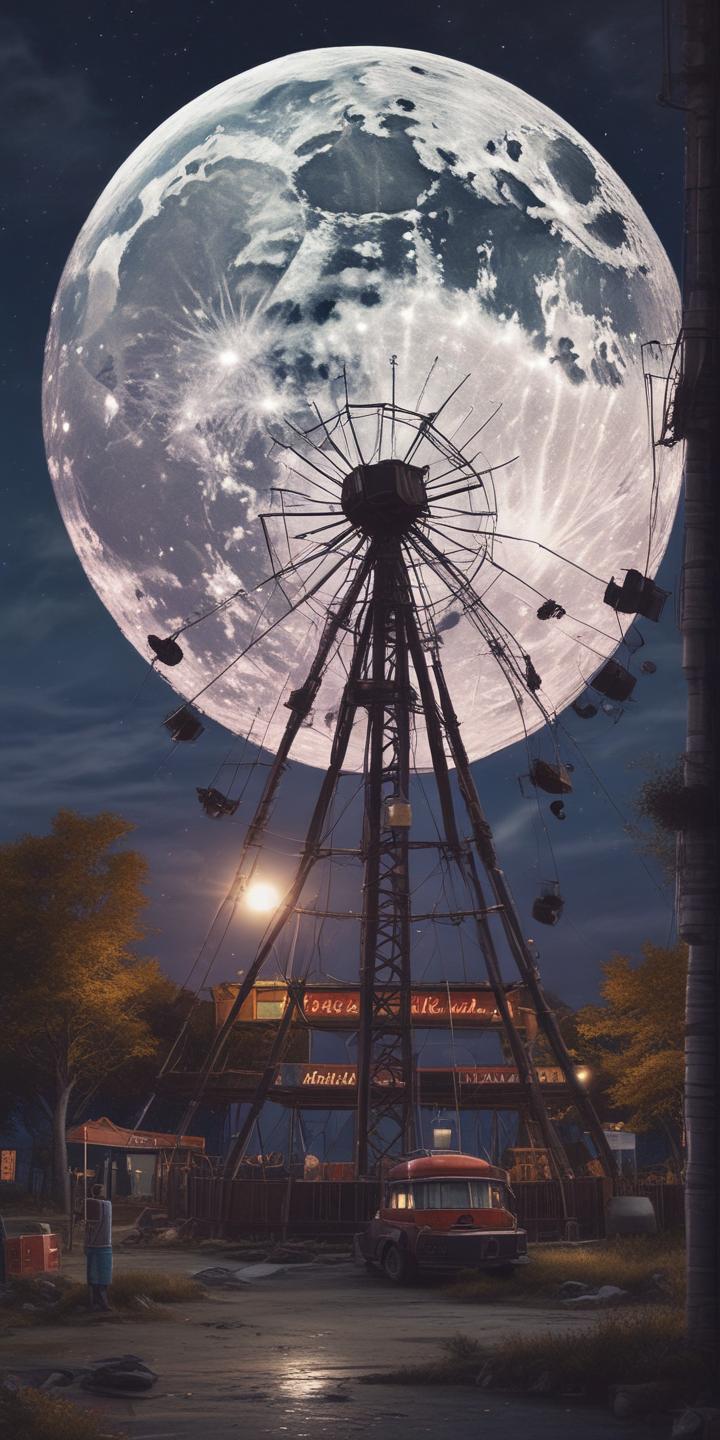} &
        \includegraphics[width=\linewidth]{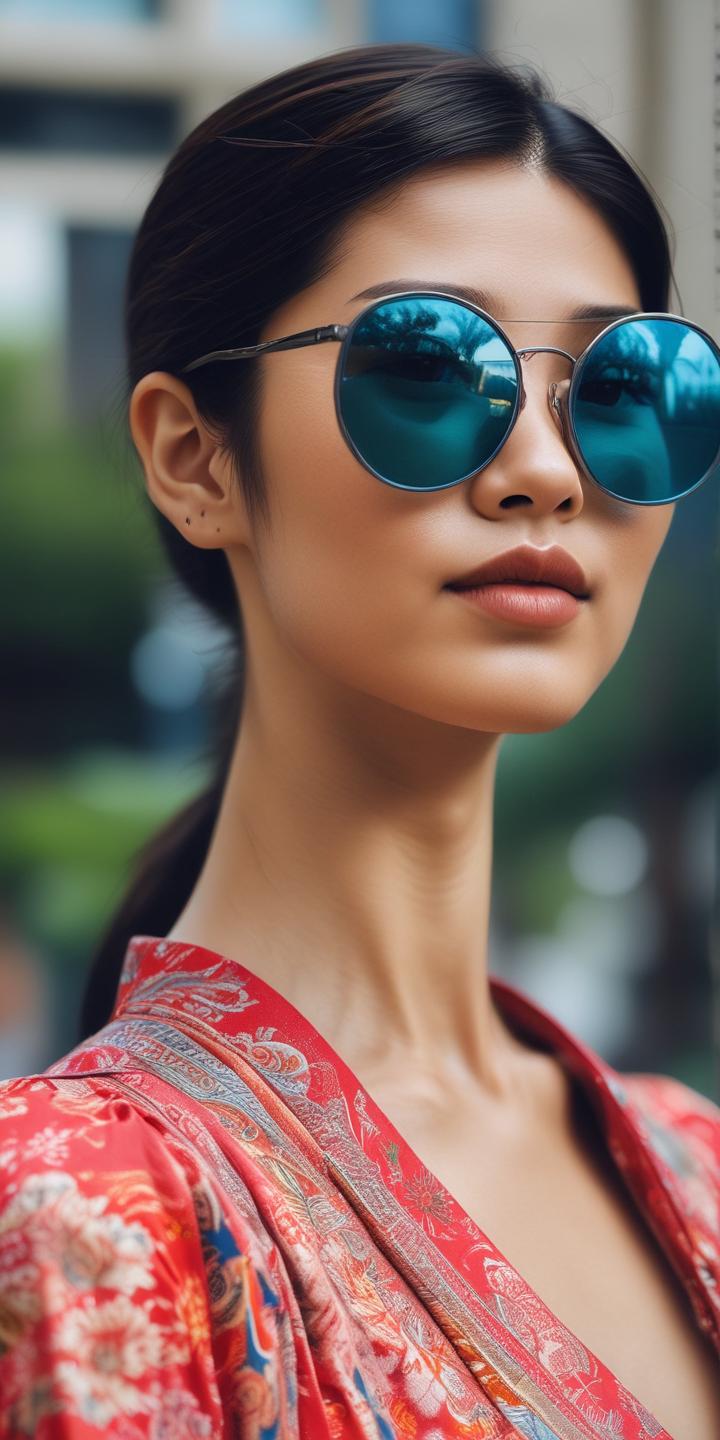} &
        \includegraphics[width=\linewidth]{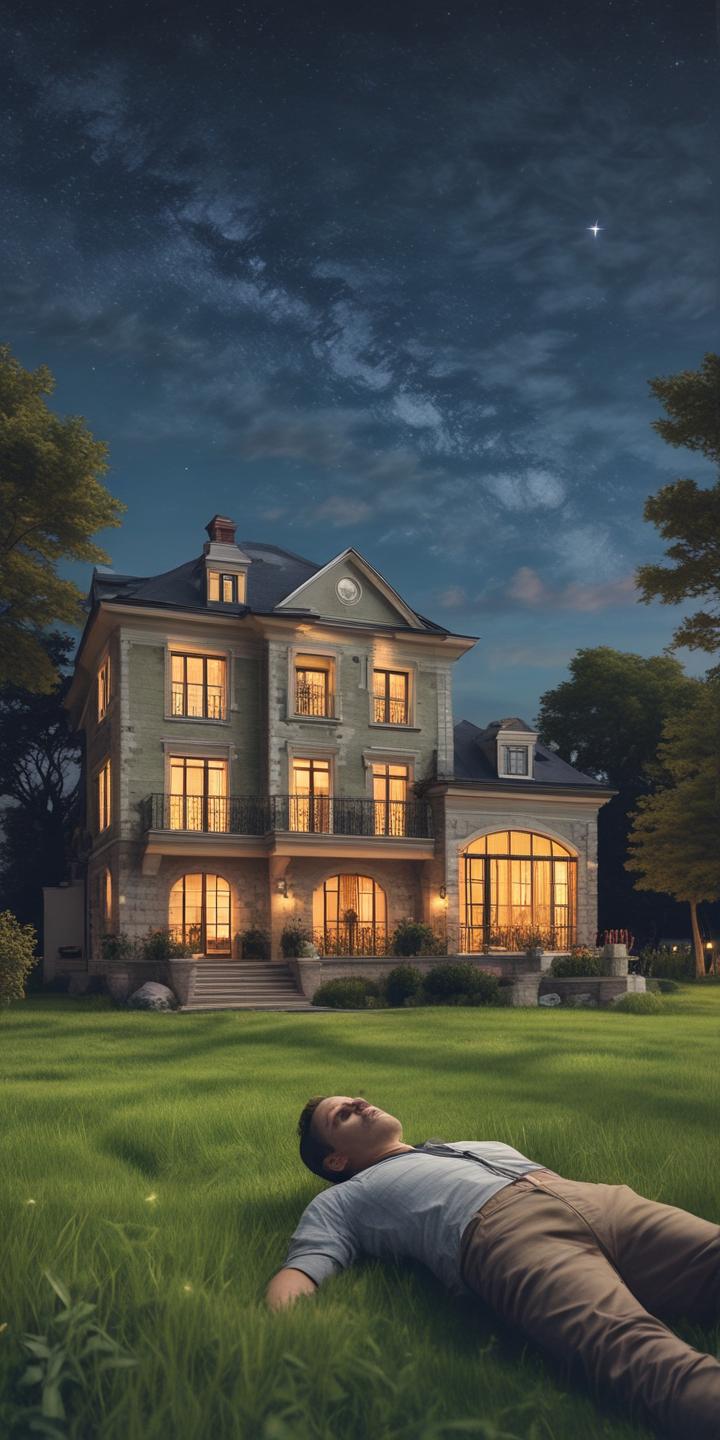}
    \end{tabularx}
    \vspace{-6pt}
    \caption{Our model is trained only on square images but still can generate different aspect ratios. The example images are 1:2 aspect ratio, 720$\times$1440px, generated by our 4-step model.}
    \vspace{-15pt}
    \label{fig:aspect-ratio}
\end{figure}

\Cref{fig:aspect-ratio} shows that are models can mostly retain the ability to infer at different resolutions and aspect ratios despite the distillation is only performed on square images. However, we do notice an increasing amount of bad cases when performing 1-step and 2-step generations. This can be improved by distilling with multiple aspect ratios, which we leave for future improvements.

\subsection{Compatibility with ControlNet}

\begin{figure}[h]
    \centering
    \footnotesize
    \setlength\tabcolsep{4pt}
    \begin{tabularx}{\linewidth}{|X|X|X|X|X|X|}
        Control & SDXL & \multicolumn{4}{l|}{Ours} \\
         & 32 Steps & 8 Steps & 4 Steps & 2 Steps & 1 Step
    \end{tabularx}
    \setlength\tabcolsep{1pt}
    \begin{tabularx}{\linewidth}{@{}XXX@{}X@{}X@{}X@{}}
        \includegraphics[width=\linewidth]{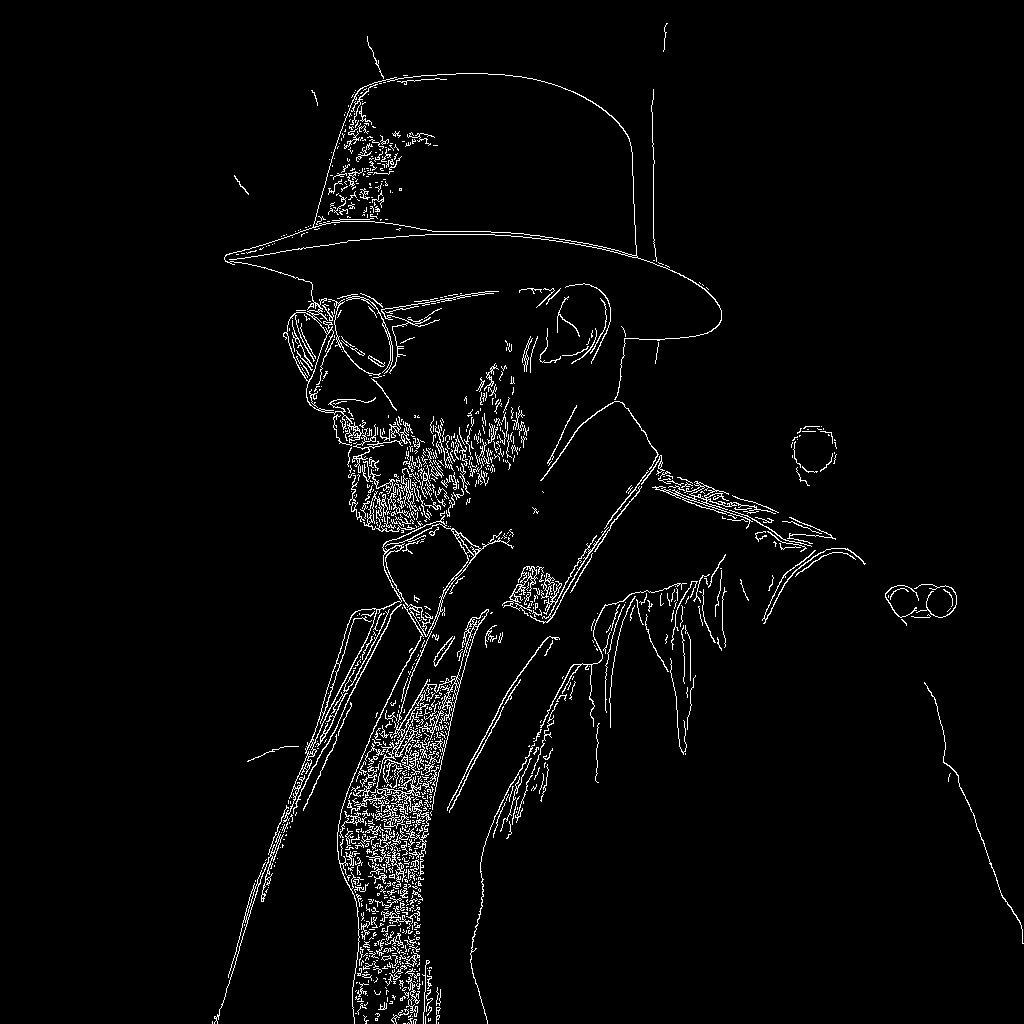} &
        \includegraphics[width=\linewidth]{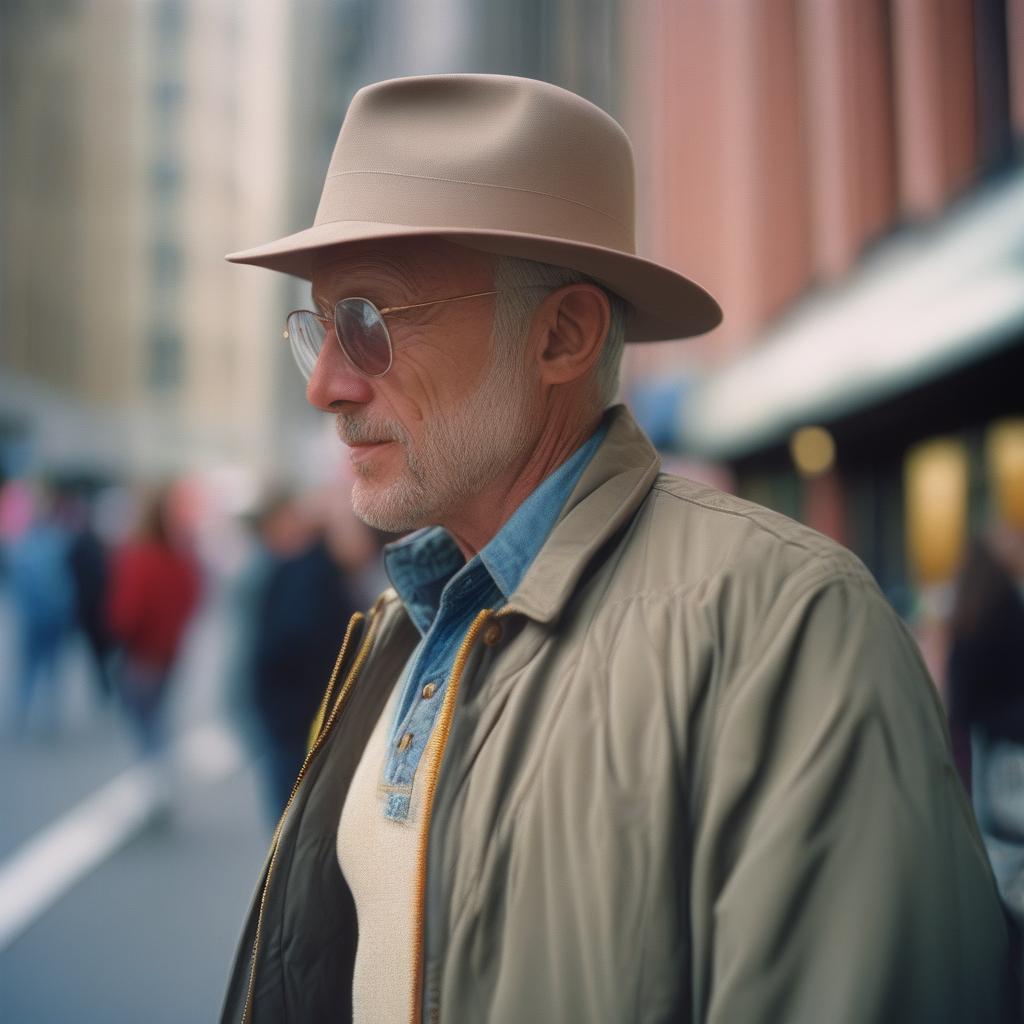} &
        \includegraphics[width=\linewidth]{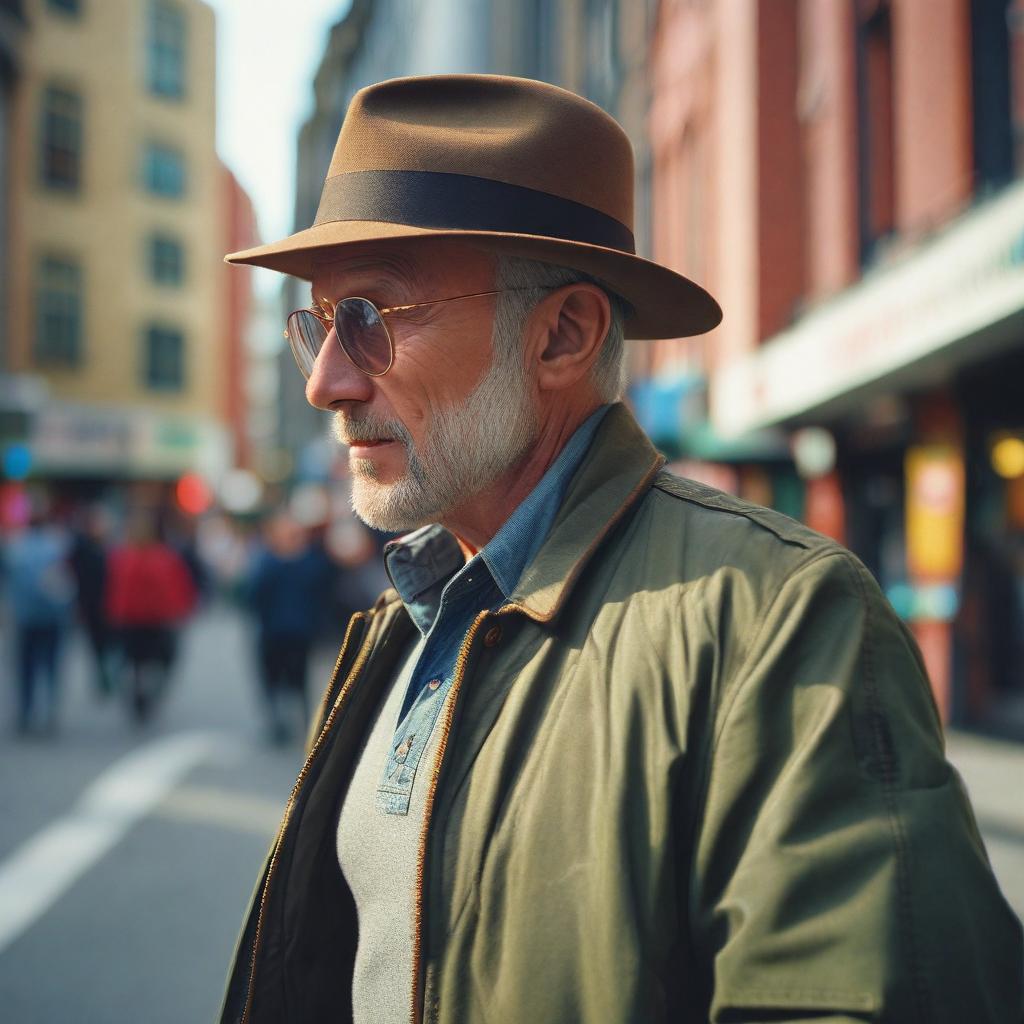} &
        \includegraphics[width=\linewidth]{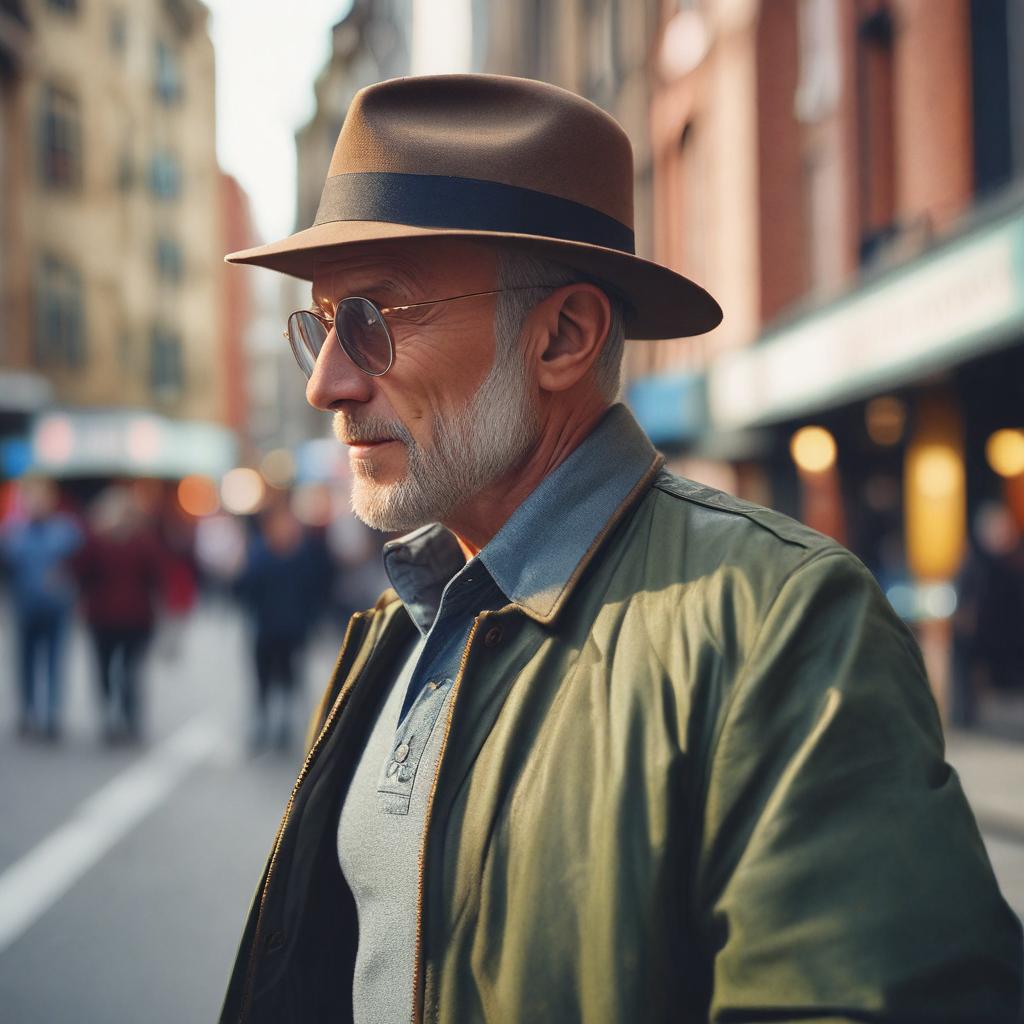} &
        \includegraphics[width=\linewidth]{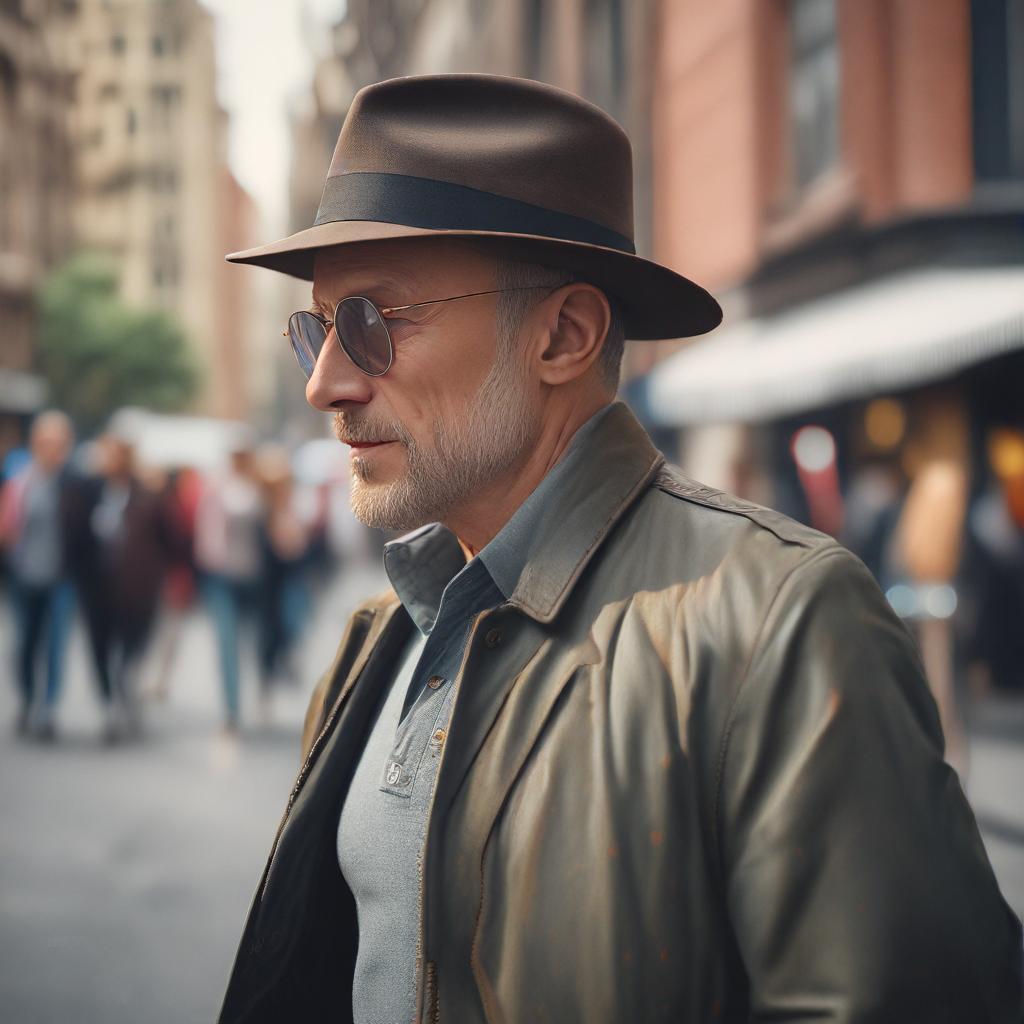} &
        \includegraphics[width=\linewidth]{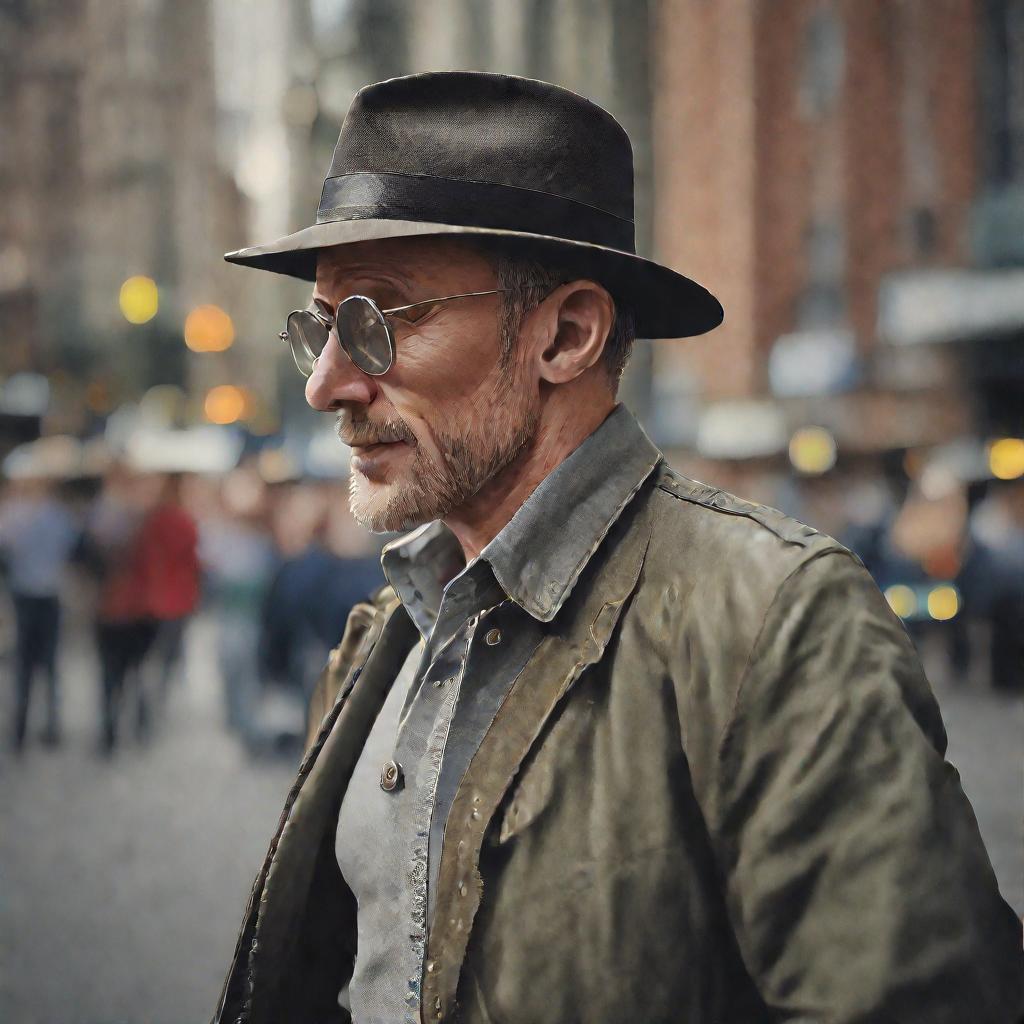} \\

        \includegraphics[width=\linewidth]{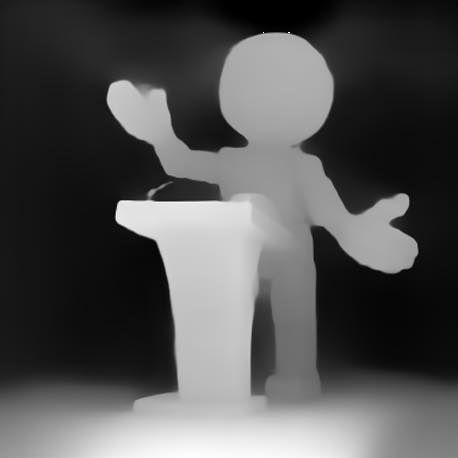} &
        \includegraphics[width=\linewidth]{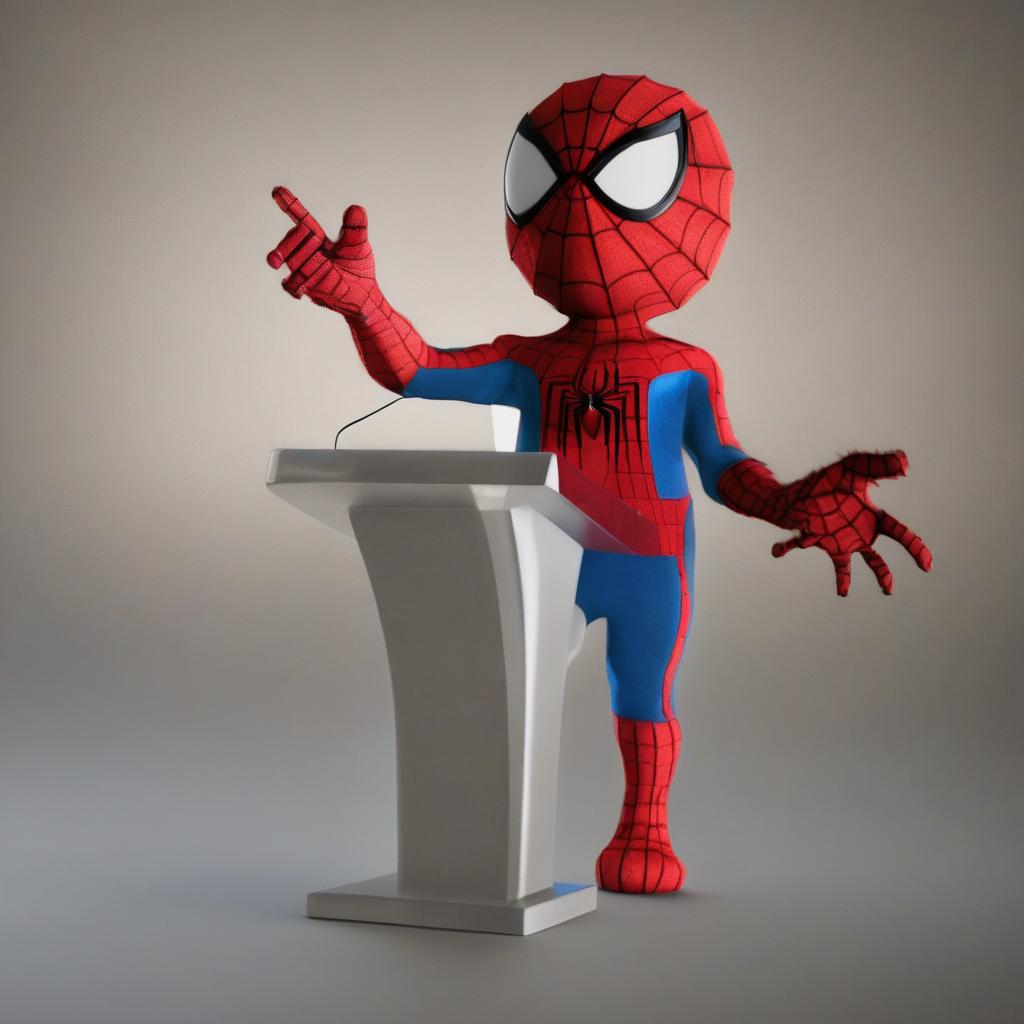} &
        \includegraphics[width=\linewidth]{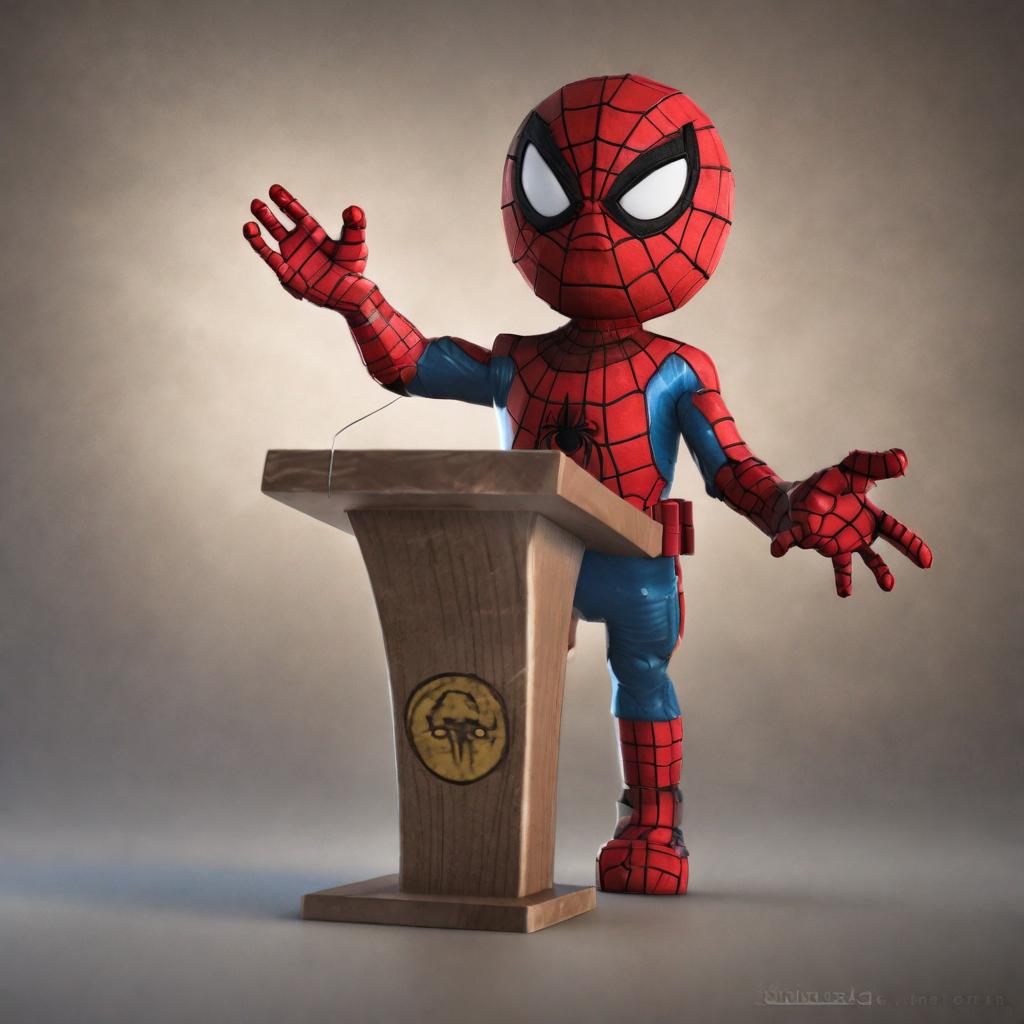} &
        \includegraphics[width=\linewidth]{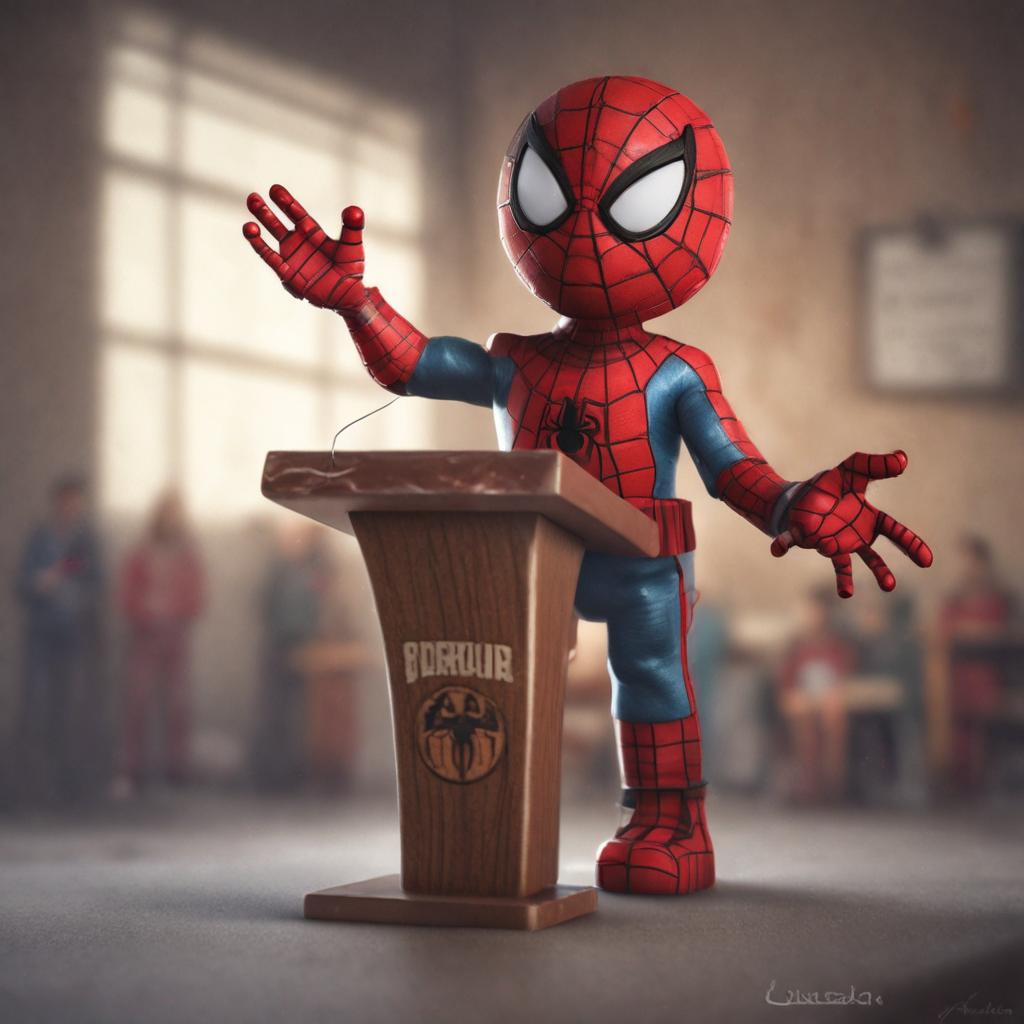} &
        \includegraphics[width=\linewidth]{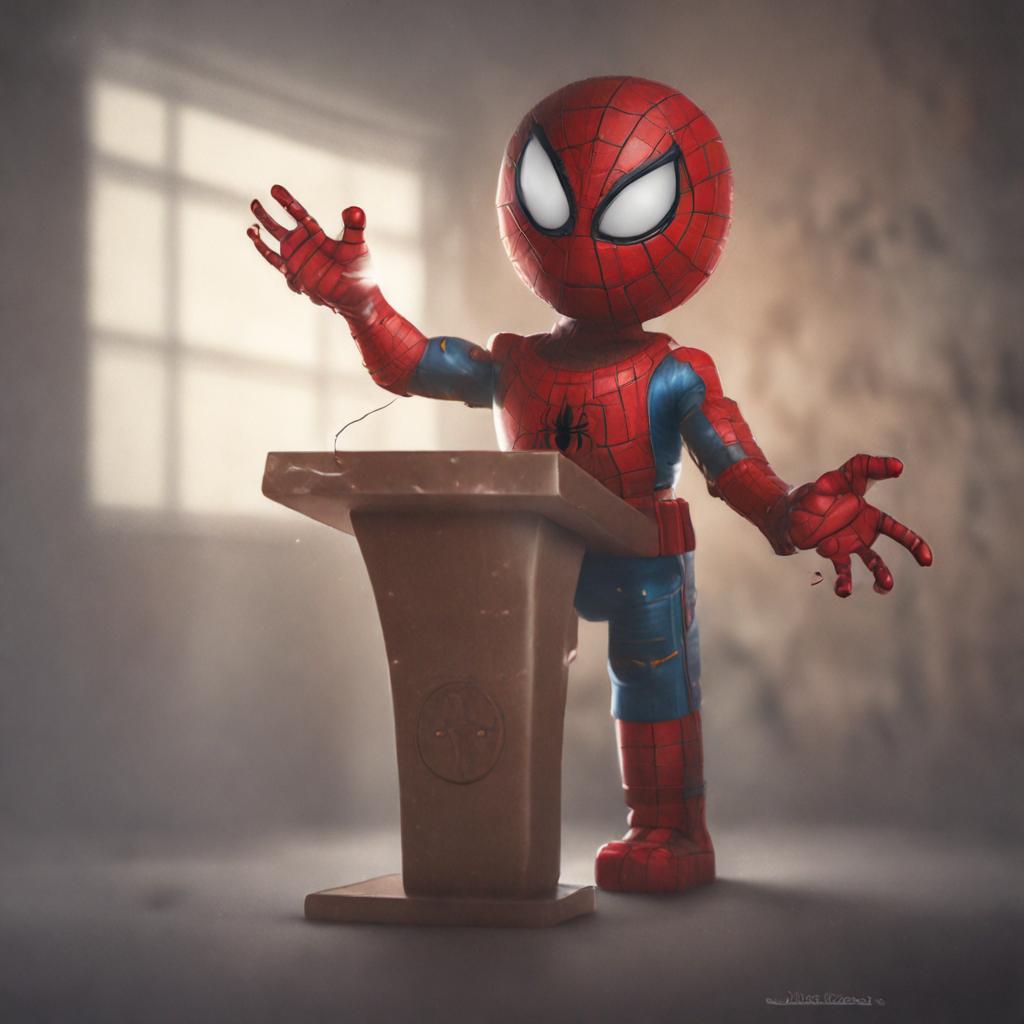} &
        \includegraphics[width=\linewidth]{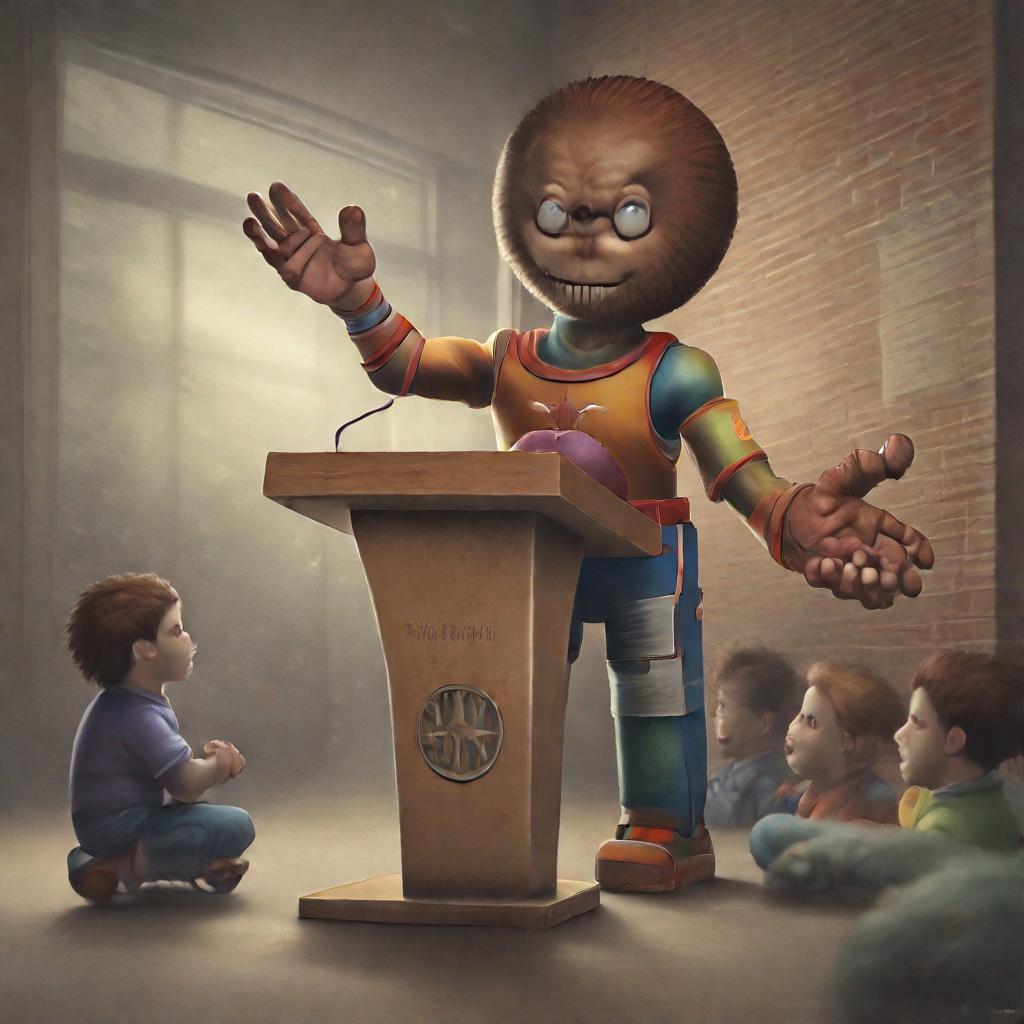}
    \end{tabularx}
    \vspace{-6pt}
    \caption{Our models are compatible with ControlNet \cite{zhang2023adding}. Examples shown are generation conditioned on canny edge and depth.}
    \label{fig:controlnet}
    \vspace{-12pt}
\end{figure}

\Cref{fig:controlnet} shows that our models are compatible with ControlNet \cite{zhang2023adding}. We test it on the canny edge \cite{controlcanny} and depth \cite{controldepth} ControlNet. We observe that our models follow the condition correctly, with some quality degradation as the number of inference steps decreases.

\section{Limitation}

Unlike other methods \cite{sauer2023adversarial,luo2023latent,luo2023lcmlora} having a single distilled checkpoint that supports multiple inference step settings, our method produces separate checkpoints for each corresponding inference step setting. This is usually not an issue in production when the number of inference steps is fixed. In case the number of inference steps must be flexible, our LoRA modules can mitigate the checkpoint switching issue.

Our method produces distilled student models with the same architecture as the teacher model. However, we believe that the UNet architecture is not optimal for one-step generation. We inspect the feature maps at each UNet layer and find that most of the generation is carried out by the decoder. We leave this problem to future improvements.

\section{Conclusion}

To sum up, we have presented SDXL-Lightning, our state-of-the-art one-step/few-step text-to-image generative models resulting from our novel progressive adversarial diffusion distillation method. In our evaluation, we have found that our models produce superior image quality compared to prior works. We are open-sourcing our models to advance the research in generative AI.

\clearpage

%%%%%%%%% REFERENCES
{\small
\bibliographystyle{ieee_fullname}
\bibliography{main}
}

\end{document}